\documentclass[a4paper, oneside, 12pt, authoryear, index, custommargin]{classes/PhDThesisPSnPDF}

\input{preamble/preamble}

\title{On Zero-Shot \\ Reinforcement Learning}
\author{Scott Jeen}

\dept{Department of Engineering}

\university{University of Cambridge}
\crest{\includegraphics[width=0.2\textwidth]{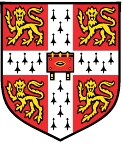}}
\degreetitle{Doctor of Philosophy}

\college{Jesus College}

\degreedate{October 2024} 

\subject{PhD Thesis} \keywords{Artificial Intelligence, Machine Learning, Reinforcement Learning, Zero-Shot Reinforcement Learning, University of Cambridge}

\ifdefineAbstract
\fi

\ifdefineChapter
\fi

\begin{document}

\includepdf[pages=1,pagecommand={}]{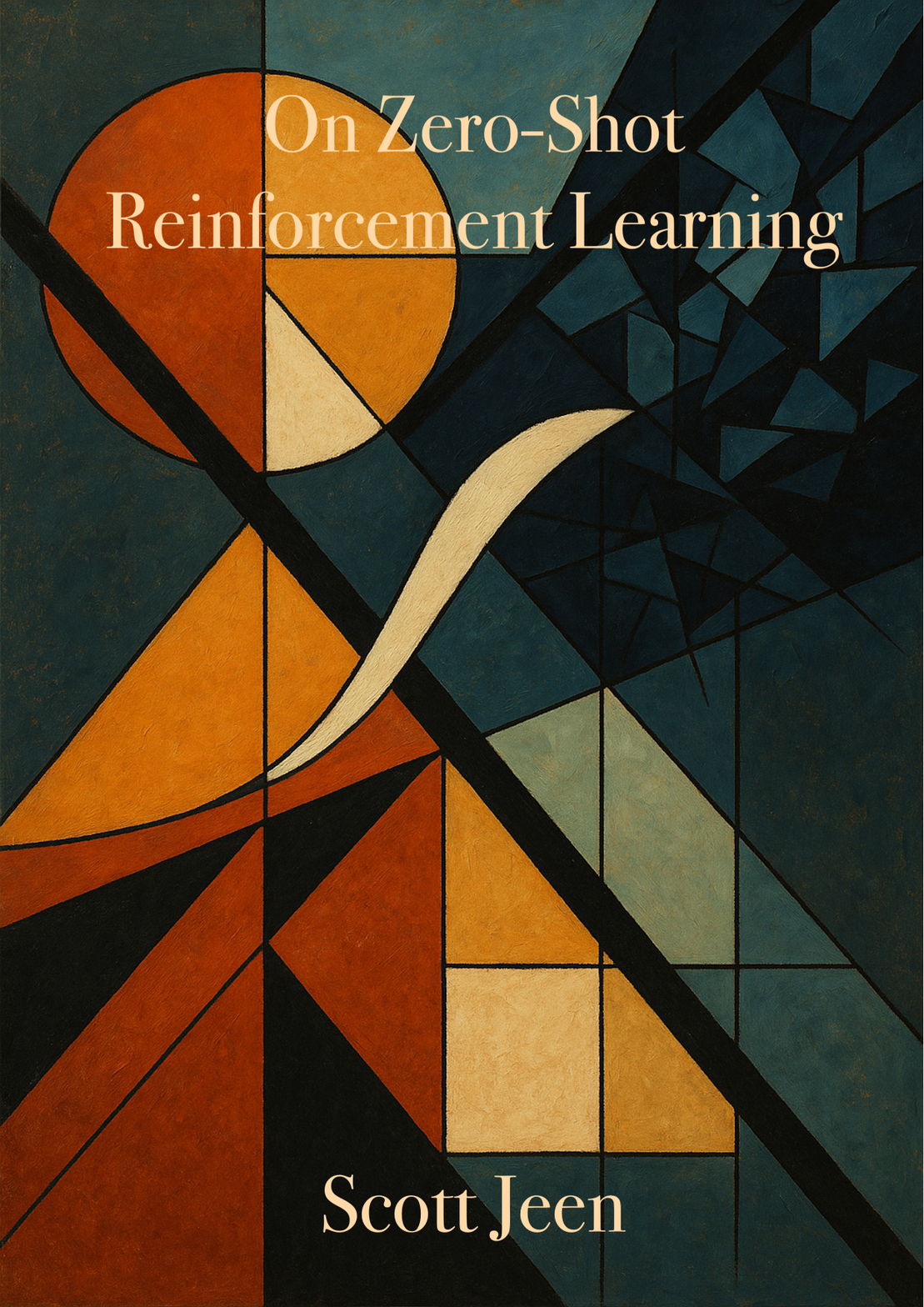}

% \thispagestyle{empty}
% \topskip0pt
% \vspace*{\fill}
% \begin{center}
% \textit{Cover generated by GPT-4o conditioned on this thesis' text.}
% \end{center}
% \vspace*{\fill}

\frontmatter

\maketitle

\cleardoublepage

\thispagestyle{empty}
\topskip0pt
\vspace*{\fill}
\begin{center}
\textit{For Barry and Helen Sealey.}
\end{center}
\vspace*{\fill}

\begin{declaration}
This thesis is the result of my own work and includes nothing which is the outcome of work done in collaboration except as declared in the preface and specified in the text. It is not substantially the same as any work that has already been submitted, or is being concurrently submitted, for any degree, diploma or other qualification at the University of Cambridge or any other University or similar institution except as declared in the preface and specified in the text. It does not exceed the prescribed word limit for the relevant Degree Committee.
\end{declaration}

\begin{abstract}
    Modern reinforcement learning (RL) systems capture deep truths about general, human problem-solving. In domains where new data can be simulated cheaply, these systems uncover sequential decision-making policies that far exceed the ability of any human. Society faces many problems whose solutions require this skill, but they are often in domains where new data cannot be cheaply simulated. In such scenarios, we can \textit{learn} simulators from existing data, but these will only ever be approximately correct, and can be pathologically incorrect when queried outside of their training distribution. As a result, a misalignment between the environments in which we train our agents and the real-world in which we wish to deploy our agents is inevitable. Dealing with this misalignment is the primary concern of \textit{zero-shot reinforcement learning}, a problem setting where the agent must generalise to a new task or domain with \textit{zero practice shots}. Whilst impressive progress has been made on methods that perform zero-shot RL in idealised settings, new work is needed if these results are to be replicated in real-world settings. In this thesis, we argue that doing so requires us to navigate (at least) three constraints. First, the \textit{data quality constraint}: real-world datasets are small and homogeneous. Second, the \textit{observability constraint}: states, dynamics and rewards in the real-world are often only \textit{partially} observed. And third, the \textit{data availability constraint}: \textit{a priori} access to data cannot always be assumed. This work proposes a suite of methods that perform zero-shot RL subject to these constraints. In a series of empirical studies we expose the failings of existing methods, and justify our techniques for remedying them. We believe these designs take us a step closer to RL methods that can be deployed to solve real-world problems.
\end{abstract}

\begin{acknowledgements}      
This thesis is the result of many privileged experiences, shaped by many special people. Of those people, I am especially indebted to four, without whom this work would not exist.  

The first two are Barry and Helen Sealey. In 2018 I’d been accepted to a Masters program at Cambridge, but couldn’t afford it. One evening, I told my friend, Charlie Cave, about my predicament, and he suggested that his grandparents, Barry and Helen, may be able to help. They had previously helped young people in similar situations, he said.  He encouraged me to write to them, and if they couldn’t help, maybe they’d point me in the direction of others who could. I did, and a week later I was in their home, in a new suit, for lunch. I nervously explained that I wanted to work hard to have a positive influence on the world, and that I felt the degree at Cambridge could help in that pursuit. Despite having little evidence to inform a decision, they chose to believe me, and funded me. That decision changed my life. It allowed me the opportunity to meet the people who have shaped me personally and academically, and it opened my eyes to a new world. For that privilege I am forever indebted to them, and I hope this thesis in some way vindicates their support. I dedicate it to them. 

One of the people Barry and Helen's generosity allowed me to meet, and the third person I wish to acknowledge, is Professor Julian Allwood. After a lecture of his, I asked if he would be willing to supervise my Masters thesis. Like Barry and Helen, his acceptance was conditioned on little evidence. In the months that followed, Julian convinced me that not only was it possible to make the world a better place, \textit{I} could (and should) make the world a better place. He was the first person I'd met who voiced this so emphatically. Though this thesis represents a significant departure from what we worked on together, to me it is entirely coherent with his message. I thank him for both believing in me and inspiring me.

The fourth person to acknowledge is my supervisor, and friend, Professor Jonathan Cullen. Jon is perhaps the only person on earth who would have let me write this thesis; I wonder if that makes him mad or brilliant (it's the latter). Jon is wonderfully open-minded, passionate, humble and intelligent. He has provided huge moral support throughout the PhD, and I thank him dearly for letting me follow my curiosity, even if it worried him at times. Thank you for trusting me Jon--I hope you're pleased with how this turned out. The world would be a better place with more people like you in it.

There are many others to thank. I first thank those that funded this work: the EPSRC, Emerson Electric, the Alan Turing Institute, the Cambridge Service for Data-Driven Discovery, and Jesus College. I thank Seb Hickman for his loyal friendship, wit, intellect and conversations about the future. I thank Tom Bewley for his friendship, and for his support throughout the projects that constitute Chapters \ref{chapter: from low quality data} and \ref{chapter: under partial observability}. I thank Tom Kirk, Ben Day, Conor Sheehan, Rachael Curzons, Logan Stewart, Harry Collard, Alex Atack, and Felix Pollock at Foresight Data Machines for inspiring me to work harder. I look forward to telling people I interned with them before they were a billion dollar company, and that Tom never actually let me visit a steel plant. I thank Hollie Berman for supporting me to pursue a PhD in 2020. I thank Professor Alessandro Abate for his guidance in the early days, and in particular for his help on the project that constitutes Chapter \ref{chapter: no prior data}. And I thank many of the amazing people I've met in Cambridge over the years for their friendship and inspiration, including, but by no means limited to: Hannes Gauch, Jack Lynch, Arduin Findeas, Luke Cullen, Grace Beaney-Colverd, Sam Stephenson, José Azevedo, Karla Cervantes Barrón, Sarah Nelson, Catherine Richards, Francis Heil, Simone Hamilton, Timo Herbertz, Josh Grantham, Eloise Ackman, Nick Kingsley, and Declan Marshall. 

Finally, I thank my family for their unending support. In particular, I thank my parents Shirley and Ian. I would say that I hope this makes you proud, but I know that however I chose to live my life you'd have been proud, and that is the greatest privilege. Thank you too, to my sister Susan, brother-in-law Ross, and step-mum Angela for always being there for me. And very finally, my biggest thanks go to my partner Cat Chapman. Thank you for indulging me in daily discussions about what's wrong with the world and how we're going to fix it. Thank you for helping me through the toughest periods. I couldn't have done it without you, and I endeavour to provide you the support you provide me. I love you all dearly.  
\end{acknowledgements}

% *********************** Adding TOC and List of Figures ***********************

\tableofcontents

\listoffigures

\listoftables

% \printnomenclature[space] space can be set as 2em between symbol and description
%\printnomenclature[3em]

\printnomenclature[8em]

\nomenclature[n]{$s_t \in \mathcal{S}$}{State at timestep $t$ and state space.}
\nomenclature[n]{$a_t \in \mathcal{A}$}{Action at timestep $t$ and action space.}
\nomenclature[n]{$r_t \in \mathbb{R}$}{Reward at timestep $t$.}
\nomenclature[n]{$o_t \in \mathcal{O}$}{Observation at timestep $t$ and observation space.}
\nomenclature[n]{$R(s_{t+1})$}{Reward function.}
\nomenclature[n]{$\rho_0$}{Initial state distribution.}
\nomenclature[n]{$\pi \in \Pi$}{Agent's policy and policy space.}
\nomenclature[n]{$V^\pi(s)$}{State value function for policy $\pi$.}
\nomenclature[n]{$Q^\pi(s, a)$}{State-action value function for policy $\pi$.}
\nomenclature[n]{$T$}{Terminal timestep of an episodic Markov Decision Process.}
\nomenclature[n]{$\alpha$}{Learning rate.}
\nomenclature[n]{$\gamma$}{Reward discount factor.}
\nomenclature[n]{$\mathcal{L}(\cdot)$}{Loss function.}
\nomenclature[n]{$L$}{Context length.}
\nomenclature[n]{$\tau_{t}^L \in \mathcal{T}$}{Trajectory at timestep $t$ of length $L$ and trajectory space.}
\nomenclature[n]{$\mathcal{N}(\cdot)$}{Gaussian distribution.}
\nomenclature[n]{$\text{Unif.}(\cdot)$}{Uniform distribution.}
\nomenclature[n]{$\varphi(s_{t})$}{A map from states to features.}
\nomenclature[n]{$z \in \mathcal{Z}$}{Task representation and task space.}
\nomenclature[n]{$\psi^\pi(s, a)$}{The successor features of $(s,a)$ subject to policy $\pi$.}
\nomenclature[n]{$P(s_{t+1} \vert s_t, a_t)$}{Markovian transition function.}
\nomenclature[n]{$\psi(s, a, z)$}{The universal successor features of $(s,a)$ for task $z$.}
\nomenclature[n]{$G_t$}{Discounted return from timestep $t$ to terminal timestep $T$.}
\nomenclature[n]{$\mathcal{D}$}{Dataset of (reward-labelled) transitions.}
\nomenclature[n]{$\Delta(X)$}{The set of possible distributions over $X$.}

\nomenclature[a]{\textbf{SGD}}{Stochastic gradient descent.}
\nomenclature[a]{\textbf{FB}}{Forward-Backward representation.}
\nomenclature[a]{\textbf{USF}}{Universal successor features.}
\nomenclature[a]{\textbf{CQL}}{Conservative $Q$-Learning.}
\nomenclature[a]{\textbf{GC-IQL}}{Goal Conditioned-Implicit $Q$-Learning.}
\nomenclature[a]{\textbf{SAC}}{Soft Actor Critic.}
\nomenclature[a]{\textbf{TD3}}{Twin Delayed Deep Deterministic policy gradients}
\nomenclature[a]{\textbf{SR}}{Successor representation.}
\nomenclature[a]{\textbf{SF}}{Successor features.}
\nomenclature[a]{\textbf{SM}}{Successor measures.}
\nomenclature[a]{\textbf{MPC}}{Model predictive control.}
\nomenclature[a]{\textbf{MPPI}}{Model predictive path integral.}
\nomenclature[a]{\textbf{RL}}{Reinforcement learning.}
\nomenclature[a]{\textbf{MBRL}}{Model-based reinforcement learning.}
\nomenclature[a]{\textbf{DQN}}{Deep $Q$ Networks.}
\nomenclature[a]{\textbf{MDP}}{Markov decision process.}
\nomenclature[a]{\textbf{POMDP}}{Partially-observable Markov Decision Process.}
\nomenclature[a]{\textbf{BC}}{Behaviour cloning.}
\nomenclature[a]{\textbf{GCRL}}{Goal-conditioned reinforcement learning.}
\nomenclature[a]{\textbf{RNN}}{Recurrent Neural Network.}
\nomenclature[a]{\textbf{ReLU}}{Rectified linear unit.}
\nomenclature[a]{\textbf{OOD}}{Out of distribution.}
\nomenclature[a]{\textbf{GRU}}{Gated recurrent unit.}
\nomenclature[a]{\textbf{PPO}}{Proximal Policy Optimisation.}
\nomenclature[a]{\textbf{TD}}{Temporal difference.}
\nomenclature[a]{\textbf{IQM}}{Interquartile mean.}
\nomenclature[a]{\textbf{PEARL}}{Probabilistic Emission-Abating Reinforcement Learning.}
\nomenclature[a]{\textbf{RS}}{Random shooting.}
\nomenclature[a]{\textbf{FB-M}}{Forward Backward representations with memory.}
\nomenclature[a]{\textbf{HILP}}{Hilbert foundation policy.}
\nomenclature[a]{\textbf{VC-FB}}{Value-Conservative Forward Backward representations.}
\nomenclature[a]{\textbf{MC-FB}}{Measure-Conservative Forward Backward representations.}
\nomenclature[a]{\textbf{ExORL}}{Exploratory Data for Offline Reinforcement Learning.}

% ******************************** Main Matter *********************************
\mainmatter

\chapter{Introduction}\label{chapter: introduction}
\textit{``Like every other destruction of optimism, whether in a whole civilization or in a single individual, there must have been unspeakable catastrophes for those who dared to expect progress. But we should feel more than sympathy for those people. We should take it personally. For if any of those earlier experiments in optimism had succeeded, our species would be exploring the stars by now, and you and I would be immortal.''}
\begin{flushright}
    —David Deutsch
\end{flushright}
\section{Automated Problem Solving}

Society makes progress by solving problems. The human brain exhibits a problem solving ability that is uniquely general \citep{deutsch1998fabric, deutsch2011beginning}, and so we are uniquely capable of rapid progress \citep{pinker2018enlightenment}. Our problem-solving ability is so valuable that we go to great lengths to build tools that \textit{automatically} solve problems for us, so we can focus this ability elsewhere \citep{asimov1989}. The Water Clocks of Ancient Egypt were an early example. A container is filled with water, and a small hole is opened to allow water to exit at constant flow rate. The changing water level, as measured by a scale internal to the container, indicates the passage of time. Automating timekeeping no doubt allowed the Egyptians to focus their valuable attention on other problems whose solutions further catalysed progress. Many others have followed since. Gutenberg's printing press automated scribing, Babbage's analytical engine (would have) automated arithmetic, Ford's assembly line automated manufacturing, Shockley's transistor automated electronics, and Berners Lee's World Wide Web automated information distribution. Each tool liberating humanity from problems of ever-increasing scope. The natural next step in this chronology is toward a meta-problem solver, a tool that solves not just one problem, but \textit{many} problems on humanity's behalf.

Designing such a tool has been the long-standing goal of artificial intelligence (AI) research \citep{mccarthy2006proposal}. After early efforts with \textit{symbolic} AI\footnote{or \textit{good old fashioned} AI (GOFAI) \citep{haugeland1989artificial}}\citep{newell1956logic, newell2007computer, russell2016artificial} and \textit{machine learning} \citep{hinton1986learning, cortes1995support, mackay2003information, williams2006, breiman2001random, bishop2006pattern, murphy2012machine}, \textit{deep learning} has emerged as the dominant paradigm \citep{goodfellow2016deep, lecun2015deep}. The typical \textit{supervised} learning setup trains a deep neural network (DNN) to predict the labels of  datapoints in a dataset. Training involves updating the weights of the network with backpropogation \citep{rumelhart1986} and stochastic gradient descent \citep{robbins1951stochastic, leon1998online} to minimise the prediction errors computed by a loss function. This recipe revolutionised image recognition \citep{lecun1989backpropagation, lecun1998gradient, krizhevsky2012imagenet, he2016deep, dosovitskiy2020image}, machine translation \citep{sutskever2014sequence, bahdanau2014neural}, speech recognition \citep{hinton2012deep, van2016wavenet}, and natural language processing \citep{mikolov2013distributed, vaswani2017attention, devlin2018bert, brown2020}, and has led to the creation of fluent chat-bots \citep{achiam2023gpt, team2023gemini, team2024gemma} and image generators \citep{ho2020denoising, rombach2022high, ramesh2021zero} that were unimaginable before their release. 

Each of these tools return a \textit{single} action in response to an input. When presented with a picture of a spotty animal with a long neck, the image recognition system returns the word ``giraffe''. When asked to translate ``où est la gare?'' to English, the translation system returns the phrase ``where is the train station?''. And when asked for the meaning of life, the language system returns the number ``42'' \citep{adams1995hitch}. Yet, the hardest problems we face require a temporally-extended \textit{sequence} of actions. Consider the problem a physicist faces when trying to explain unexpected experimental results. They may first check their apparatus was setup as expected, and isn't malfunctioning. If the checks are passed, they may spend a few hours inspecting and rewriting the code that converts their measurements to human-legible results. If their results remain repeatable, they may spend a few days re-reading the literature that motivated the experiments, and discussing the results with colleagues. If no explanation is found, they may conclude that the existing theory is inadequate, and spend the next months, or even years, developing new theory to explain the results. In principle, a model could be trained with supervised learning to mimic this sequence of actions\footnote{Indeed, this idea has received enthusiastic recent support \citep{janner2021offline, chen2021decision, yang2023foundation}}. But, to succeed, the model needs access to a broad dataset of actions, ideally cataloguing all possible decisions the physicist could have made, with the quality of each decision ranked by how helpful it was in achieving the goal of explaining the results. Not only would curating such a dataset be impractical, it is notoriously difficult to manually assign credit for a successful outcome to the correct decision, or decisions, in a temporally extended sequence \citep{minsky1961steps, sutton1984temporal, richards2019dendritic}. 

The alternative is to let the model interact with the world to uncover useful sequences of decisions autonomously, which is the concern of \textit{reinforcement learning} (RL) \citep{waltz1965heuristic, mendel19708, Sutton2018, bertsekas2019reinforcement}. Here, a decision-making \textit{agent} interacts sequentially with a problem-solving \textit{environment}. The agent is communicated a high-level objective via \textit{rewards}, and its task is to learn the sequence of actions that maximise their sum. \textit{Deep} RL—the modelling of the traditional components of an RL agent with DNNs \citep{mnih2013, arulkumaran2017deep, li2017deep}—has helped create tools that exhibit remarkable sequential decision-making skill. These systems have proven to master perfect information games like chess, go, and shogi  \citep{silver2016mastering, silver2017, silver2018general, schrittwieser2020}, beat expert humans at imperfect information games like poker, Starcraft, and Diplomacy \citep{vinyals2019grandmaster, meta2022human, perolat2022mastering, brown2019superhuman}, design state-of-the-art computer chips in a fraction of the time of skilled experts \citep{mirhoseini2021graph}, control nuclear fusion reactors more precisely than human-engineered solutions \citep{degrave2022magnetic, seo2024avoiding}, and replace sorting and matrix multiplication algorithms that computer scientists haven't bettered in 50 years \citep{mankowitz2023faster, fawzi2022discovering}.

Examined together, these advances suggest the RL framework captures some deep truths about general, human problem-solving. Indeed, some believe the parallels run so deep that RL is a necessary component of our meta-problem solver \citep{hutter2005universal, Sutton2018, clune2019ai, hughes2024open}, a belief summarised by Silver et al.'s hypothesis:
\begin{emphasisbox}
    \textbf{Reward is Enough} \citep{silver2021reward}. Intelligence, and its associated abilities, can be understood as subserving the maximisation of reward by an agent acting in its environment. 
\end{emphasisbox}

\section{Unfulfilled Promises}\label{chapter1: unfulfilled promises}
Given this progress and such claims, one would expect RL agents to be omnipresent;  solving problem after problem on humanity's behalf. That we are yet to meet this future is clear. But why? If an RL agent can control a Tokomak, why can't it control my central heating system? If an RL agent can beat me at poker, why can't it beat me at golf? If an RL agent can discover new sorting algorithms, why can't it discover new medicines? Answering these questions requires us to explore the consequences of RL's \textit{sample inefficiency} \citep{yu2018towards}. 

It is typical for agents to require billions of environment interactions to obtain solutions to hard problems \citep{vinyals2019grandmaster, silver2017}. Data on this scale is equivalent to many human lifetimes of learning, so we say that RL agents are \textit{sample inefficient} with respect to human learning\footnote{There are many possible explanations for this discrepancy. Perhaps the most popular is that the evolution we have been subjected to over millennia has provided us useful priors for general problem-solving \citep{lake2017building, tenenbaum2011grow}.}. Were the agent to manually collect this data from the physical world it would take them hundreds of years, so the data must necessarily come from a non-physical, \textit{simulated} world.  

\textit{Perfect simulators}\footnote{Here, ``perfect'' is used imprecisely for exposition. What we mean is that, in practice, these simulators create synthetic data that is indistinguishable from real data.} generate this data synthetically by leveraging known truths about the environment's underlying physics. For example, the physics of \textit{Space Invaders} is known because it was designed by humans, so the game can be simulated for an RL agent to play as long as is necessary to master it \citep{bellemare2013arcade}. Faraday's Law explains the behaviour of plasma inside a fusion reactor \citep{hinton1976theory}, and so can be used to simulate the plasma's response to different control strategies proposed by an agent \citep{citrin2024torax}. And the laws of linear algebra explain how operations applied to a matrix change its form, and so can be used to simulate matrix transformations under different agent-proposed routines \citep{fawzi2022discovering}. But, perfect simulators cannot be designed so easily for most real-problems because we cannot neatly summarise their physics by a system of generic equations. Occasionally, we can \citep{citrin2024torax, energyplus2001}, but even in such circumstances it can take an engineer months or years to configure the simulator to faithfully replicate reality. Indeed, it is a core assumption of this work that \textit{perfect simulators are prohibitively expensive to build for real-world problems.} 

Accepting this, we can \textit{learn} simulators from data. Here, we don't need to understand the rules of the data generating process from first principles, and can instead train a model to generate the distribution of data that reality throws up. Indeed, there has been much recent progress to this end, particularly in the learned simulation of robotic manipulation \citep{yang2023learning, eslami2018neural}, autonomous-driving \citep{hu2023gaia}, protein-folding \citep{jumper2021highly, abramson2024accurate}, and game-playing \citep{bruce2024genie, chiappa2017recurrent, ha2018, micheli2022transformers}. And one could imagine, for example, equipping a central heating system with sensors, collecting data over some period, and training a model to simulate the relationship between boiler control, air temperature and energy spend. However, such simulators are \textit{imperfect}, lacking the accuracy of perfect simulators for two reasons. First, for a complex high-dimensional problem, the space of possible behaviours is large, and therefore difficult to model from a finite dataset \citep{bellman1965}. So the model is bound to make incorrect predictions in scenarios not explained by the dataset \citep{murphy2012machine, goodfellow2016deep}. Second, the model contains finite parameters that are optimised stochastically, meaning that, even with complete data, its predictions can only ever be approximately correct \citep{goodfellow2016deep}. As a result, there will necessarily be a misalignment between the learned worlds in which we'd like to train our agent, and the real world in which our agent will be deployed. 

Such a misalignment may not necessarily worry us at first. If the learned simulator is approximately correct, presumably an RL agent can learn approximately correct behaviour? Unfortunately, even small misalignments between simulation and reality can derail conventional RL techniques \citep{zhao2020sim}, which have a well-known bias for the idiosyncrasies of their training environment \citep{zhang2018study, cobbe2019quantifying, justesen2018illuminating}. And unlearning such biases with real-world experience can take millions of further interactions, or is sometimes impossible \citep{lyle23b, lyle2022understanding}. Indeed, these agents inspect their training environments so closely that they can exploit weaknesses to hack high rewards \citep{clark2016, skalse2022defining, krakovna2020}, and expose unknown bugs in the source code \citep{rani2023deep, tufano2022using}. As a result, naively training conventional RL agents in learned simulators will not elicit real-world superhuman performance like that reported in the works that rely on perfect simulation.

\section{A Path Forward} \label{chapter1: path forward}
So, RL is a useful model of human-like sequential decision-making, but existing methods are sample-inefficient, requiring vast pre-training data. Behind each successful application of RL is a perfect simulator providing this data, but expecting these for most real-world problems is unrealistic. The tractable alternative is to learn a simulator using data collected from the environment, but these are at best approximately correct, and at worst pathologically incorrect when queried outside of the training distribution. And traditional RL agents are ill-equipped to quickly handle differences between a simulated, erroneous training environment, and a real, correct deployment environment. If our goal is to make progress toward our meta problem solver, where in this system should we intervene?

In this thesis we choose to intervene at the final stage. Our goal is to build upon agent designs that quickly adapt to differences between training and deployment environments. This situates us directly in the nascent field of \textit{zero-shot reinforcement learning} \citep{kirk2023survey}, where the goal is to develop agents that can adapt to new scenarios with \textit{zero practice shots}\footnote{A formal description of the zero-shot RL setting is left for \colouredS \ref{background: zero-shot RL}}. Excellent recent work has brought us closer to this goal. Of particular relevance is the use of \textit{forward backward representations} \citep{touati2021learning} and \textit{universal successor features} \citep{borsa2018} for zero-shot RL. Recent work has shown these methods can be trained to solve \textit{any} downstream task in an environment theoretically, with only a handful of data needed to specify the task at test time \citep{touati2022does}. Empirically, these methods can return solutions to many toy problems that are just as performant as conventional RL methods that get to see the task during training. 

Much of the work presented hereafter involves equipping similar methods with the tools to solve not just toy problems, but \textit{real-world problems}. We believe the latter exhibit (at least) three important constraints that shape the solution space:
\begin{emphasisbox}
\begin{enumerate}
    \item \textbf{The data quality constraint:} Real-world datasets are small and homogeneous.
    \item \textbf{The observability constraint:} Real-world states, dynamics and rewards are rarely fully observable.
    \item \textbf{The data availability constraint:} Real-world problems do not always provide training data \textit{a priori}.
\end{enumerate}
\end{emphasisbox}
Constraint 1 is a consequence of real-world datasets having necessarily been produced by an incumbent agent or controller that is optimising for task performance, rather than dataset size, heterogeneity or expressiveness \citep{dulac2019challenges}. Constraint 2 arises because agents observe the world through finite sensors; because the physics of the environment changes between data collection and policy execution; or because rewards are not communicated to the agent immediately after executing an action \citep{dulac2019challenges}. The justification for Constraint 3 is twofold. The first is as previously discussed; many real-world problems cannot be neatly summarised by a set of generic equations, and so cannot be perfectly simulated. The second is that many real-world problems of interest may not be controlled by an existing system from which we can extract historical data logs\footnote{As a concrete example, consider again the task of training a golf-playing agent. For this we require either a) a perfect simulator of golf's physics or b) datasets from which golf's physics can be modelled by a learned simulator. The former is impractically expensive to build and run at best, and intractable at worst. The latter do not exist because we do not collect them from the human agents who play the game.}. 

It is a core belief of this work that for an RL technique to be deployed toward solving a real-world problem it must solve tasks zero-shot, whilst subjected to at least one of these constraints. More precisely, we seek to defend the following: 
\begin{hypbox}{Thesis}{ 
    It is possible to build RL agents that solve unseen tasks zero-shot, whilst satisfying the \textit{data quality, observability} or \textit{data availability} constraints.
    }
\end{hypbox}

\section{Thesis Structure and Contributions}\label{section: thesis structure and contr.}
We advance our thesis in the subsequent chapters. We begin with a review of relevant background material in Chapter \ref{chapter: background}. This includes a formal description of the RL problem and an introduction to popular solution methods in \colouredS \ref{background: reinforcement learning}. Accepting the difficultly of building perfect simulators, this thesis will deal primarily with \textit{learned simulators}, which is the concern of \textit{offline RL}—introduced in \colouredS \ref{background: offline RL}. Methods that generalise to unseen scenarios zero-shot are necessarily distinct from the conventional techniques introduced in \colouredS \ref{background: reinforcement learning}, so they are provided their own review in \colouredS \ref{background: zero-shot RL}.

In Chapter \ref{chapter: from low quality data}, we propose \textit{conservative} zero-shot RL methods to address the failings of existing works when trained on realistic, real-world datasets. We first expose a critical failure mode of state-of-the-art zero-shot RL methods when trained on low quality datasets, namely, an overestimation of the value of state-action pairs not present in the dataset. Using regularisation techniques from the offline RL literature, we bias models to value these state-actions \textit{conservatively}, and show this greatly improves performance on all of the standard zero-shot RL benchmarks. This chapter is based on the following publication:
\begin{quote}
    Zero-Shot Reinforcement Learning from Low Quality Data. \textbf{Scott Jeen}, Tom Bewley, and Jonathan M. Cullen. In \textit{Advances in Neural Information Processing Systems 38, 2024} \citep{jeen2024zeroshot}.\\
    \href{https://arxiv.org/abs/2309.15178}{\UrlFont{[paper]}}
    \href{https://enjeeneer.io/projects/zero-shot-rl/}{\UrlFont{[website]}}
    \href{https://github.com/enjeeneer/zero-shot-RL}{\UrlFont{[code]}}
    \href{https://enjeeneer.io/posters/conservative-world-models.pdf}{\UrlFont{[poster]}}
\end{quote}
In Chapter \ref{chapter: under partial observability}, we augment zero-shot RL methods with \textit{memory models} to address their failings in partially observed problem settings. We expose two failure modes, which we call \textit{state misidentification} and \textit{task misidentification}. We show that memory-based zero-shot RL methods resolve state and task misidentification such that performance improves over memory-free methods in environments where the state, rewards or dynamics are partially observed. This chapter is based on the following publication:
\begin{quote}
    Zero-Shot Reinforcement Learning Under Partial Observability. \textbf{Scott Jeen}, Tom Bewley, and Jonathan M. Cullen. In \textit{Reinforcement Learning Conference 2, 2025} \citep{jeen2025zero}. \\
    \href{https://arxiv.org/abs/2506.15446}{\UrlFont{[paper]}}
    \href{https://enjeeneer.io/projects/bfms-with-memory/}{\UrlFont{[website]}}
    \href{https://github.com/enjeeneer/bfms-with-memory}{\UrlFont{[code]}}
\end{quote}
In Chapter \ref{chapter: no prior data}, we ask: is it possible to control a real-world system with \textit{no prior knowledge}. Our real-world problem of interest is emission-efficient building control, that is, the minimisation of emissions associated with energy-use in arbitrary buildings subject to constraints on occupant comfort. We introduce a new method we call PEARL: Probabilistic Emission Abating Reinforcement Learning, which can reduce emissions from buildings by up to 30\% with no prior knowledge, and only 180 minutes of pre-deployment data collection. This chapter is based on the following publication:
\begin{quote}
    Low Emission Building Control with Zero-Shot Reinforcement Learning. \textbf{Scott Jeen}, Alessandro Abate and Jonathan M. Cullen In \textit{AAAI Conference on Artificial Intelligence, 2023} \citep{jeen2023low}.\\
    \href{https://arxiv.org/abs/2206.14191}{\UrlFont{[paper]}}
    \href{https://enjeeneer.io/projects/pearl/}{\UrlFont{[website]}}
    \href{https://github.com/enjeeneer/PEARL}{\UrlFont{[code]}}
    \href{https://enjeeneer.io/posters/pearl.pdf}{\UrlFont{[poster]}}
\end{quote}
Other works were completed throughout the PhD that do not appear in this thesis, these are:
\begin{quote}
    Dynamics Generalisation with Behaviour Foundation Models. \textbf{Scott Jeen} and Jonathan M. Cullen. In \textit{RL Conference Workshop on Training Agents with Foundation Models, 2024} \& \textit{RL Conference Workshop on RL Beyond Rewards, 2024} \citep{jeen2024dynamics}. \\
    \href{https://openreview.net/pdf?id=A1u8YM7vuP}{\UrlFont{[paper]}}
    \href{https://enjeeneer.io/slides/rlc24/slides.html#/title-slide}{\UrlFont{[slides]}}
    \href{https://enjeeneer.io/slides/rlc24/slides.html#/title-slide}{\UrlFont{[poster]}}\\

    Beobench: a toolkit for unified access to building simulations for reinforcement learning. Arduin Findeis, Fiodar Kazhamiaka, \textbf{Scott Jeen} and Srinivasan Keshav. In \textit{Proceedings of the Thirteenth ACM International Conference on Future Energy Systems, 2022} \citep{findeis2022} \\
    \href{https://dl.acm.org/doi/abs/10.1145/3538637.3538866}{\UrlFont{[paper]}}
    \href{https://github.com/rdnfn/beobench}{\UrlFont{[code]}}\\

    Testing the domestic emission reduction potential of optimised building control alongside retrofit. Hannes Gauch, \textbf{Scott Jeen}, Jack Lynch and Andre Cabrera Serrenho.  \\
    
    Transforming agrifood production systems and supply chains with digital twins. Asaf Tzachor, Catherine Richards and \textbf{Scott Jeen}. In \textit{Nature Science of Food, 2022} \citep{tzachor2022transforming}.\\
    \href{https://www.nature.com/articles/s41538-022-00162-2}{\UrlFont{[paper]}}   
\end{quote}
We conclude the thesis in Chapter \ref{chapter: outlook} with a summary of our contributions and their limitations, and a speculative discussion on how relevant our proposals may remain in the future.

\chapter{Background}\label{chapter: background}
\section{Reinforcement Learning}\label{background: reinforcement learning}
The standard \textit{Reinforcement Learning} (RL) \citep{Sutton2018, kaelbling1996reinforcement, bertsekas2019reinforcement} setup considers an \textit{agent} interacting with an \textit{environment} sequentially to solve a \textit{task}. The agent views the state of the environment and selects an action that it thinks will help solve the task. The action is taken, the environment's \textit{state} is updated, and a \textit{reward} is communicated to the agent which reflects how useful this sequence was in solving the task. The agent's goal is to learn to take actions that maximise the cumulative sum of rewards it receives through time. Traditionally, it is assumed that the agent learns reward-maximising sequences of actions \textit{tabula rasa}, that is, with no prior knowledge of the environment or task\footnote{In \colouredS \ref{background: offline RL} we shall see that contemporary RL problems often relax this assumption and incorporate prior knowledge.}. In this setup the agent must \textit{explore} the set of actions available to find those that return high-reward, then once found, \textit{exploit} only the reward-maximising actions to solve the task. The agent's dilemma is to choose between these competing interests at every environment interaction.
 
\subsection{Markov Decision Processes}\label{background: mdps} 
The environment is usually modelled as a \textit{Markov Decision Process} (MDP) \citep{bellman1957, puterman2014markov} defined by $\mathcal{M} = \{\mathcal{S}, \mathcal{A}, P, R, \rho_0\}$. At each timestep $t$, the MDP exists in state $s_t \in \mathcal{S}$, with the starting state $s_0$ drawn from initial state distribution $\rho_0$. The agent selects an action $a_t \in \mathcal{A}$ according to it's state-conditioned policy $\pi: \mathcal{S} \rightarrow \Delta(\mathcal{A})$ (where $\Delta(X)$ denotes the set of possible distributions over $X$). Given state $s_t$ and action $a_t \sim \pi(\cdot | s_t)$, the MDP transitions to the next state $s_{t+1}$ according to its \textit{Markovian} transition function (or \textit{dynamics}): $P: \mathcal{S} \times \mathcal{A} \rightarrow \Delta(\mathcal{S})$ i.e. $s_{t+1} \sim P(\cdot | s_t, a_t)$.  For this transition, the agent receives reward $r_t$ from the reward function $R: \mathcal{S} \rightarrow \mathbb{R}$\footnote{For simplicity, we assume the reward depends only on the next state $s_{t+1}$, but this is not essential, and all methods discussed hereafter are compatible with reward functions that depend on the state-action pair $(s_t, a_t)$ or on the state-action-next state triple $(s_t, a_t, s_{t+1})$.}. In this thesis we say that a reward function defines a \textit{task} for the agent. We only deal with episodic MDPs in which the agent will eventually transition to a \textit{terminal state} at timestep $T$ that ends the sequence of interactions called an \textit{episode}. The tuple $(s_t, a_t, s_{t+1}, r_t)$ is called a \textit{transition} and a sequence of transitions $\tau = (s_1, a_1, r_1, \ldots, s_{H-1}, a_{H-1}, r_{H-1}, s_H)$ is called a \textit{trajectory} to \textit{horizon} $H$. The agent's goal is to find the policy that maximises the expected reward:
\begin{equation}\label{equation: optimal policy}
    \pi^* = \argmax_{\pi \in \Pi} \mathbb{E}_{s\sim \rho_0}\left[J(s)\right],
\end{equation}
where $\pi^*$ denotes the optimal policy, $\Pi$ is the set of all policies, and $J: \mathcal{S} \rightarrow \mathbb{R}$ is the \textit{return} of a state, calculated as
\begin{equation}\label{equation: return}
J(s) := \mathbb{E}_{\pi, P} \left[\sum_{t=0}^{T-1} R(s_{t+1}) \; | \; s_0 = s \right].
\end{equation}

\subsection{Partial Observability}
In many real-world settings the agent cannot access the full state $s_t$, but only some subset of the information in $s_t$. Here, the state is said to be \textit{partially observed} via observations $o_t \in \mathcal{O}$ \citep{aastrom1965, kaelbling1998}. This is modelled by adding an observation function $O: \mathcal{S} \times \mathcal{O} \rightarrow [0,1]$ to the standard MDP tuple, which defines a state-conditioned distribution over the observation space $\mathcal{O}$. Now, the agent samples actions from a policy conditioned on the most recent observation $a_t \sim \pi(\cdot | o_t)$ in the naïve case, or from a policy conditioned on a trajectory of observation-action pairs $a_t \sim \pi(\cdot | o_0, a_0, \dots, a_{t-1}, o_t)$ that estimate the underlying state in the improved case. This extension of an MDP is called a \textit{partially-observable MDP} (POMDP) and is represented by the tuple $\mathcal{P} =(\mathcal{S}, \mathcal{A}, \mathcal{O}, R, P, O, \rho_0, \gamma)$.

\subsection{Value Functions}\label{background: value functions}
\begin{figure}
    \centering
    \includegraphics[width=0.46\textwidth]{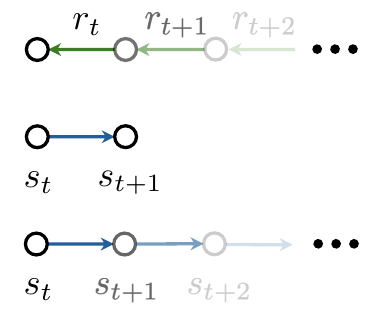}
    \caption{\textbf{Three paradigms of reinforcement learning.} (Top) \textit{Model-free} RL methods distill future {\color{OliveGreen}{rewards}} $r_t, r_{t+1}, r_{t+2}, \ldots$  into a value function, policy or both (\colouredS \ref{background: model-free RL}). (Middle) \textit{One-step model-based} RL methods predict the {\color{Blue}{state transition}} $s_t \leadsto s_{t+1}$ with a model (\colouredS \ref{background: one-step MBRL}). (Bottom) \textit{Multi-step model-based} RL methods distill future {\color{Blue}{state transitions}} $s_t, s_{t+1}, s_{t+2}, \ldots$ into a model. (\colouredS \ref{background: multi-step MBRL})}
    \label{fig: method summary}
\end{figure}

Value functions are useful for quantifying the utility of different policies. The \textit{state value function} $V^\pi(s)$ is defined as the expected discounted reward starting at $s$ and following policy $\pi$ thereafter
\begin{equation}\label{equation: state value function}
    V^{\pi}(s) := \mathbb{E}_{\pi, P} \left[\sum_{t=0}^{T-1} \gamma^t R(s_{t+1}) \; | \; s_0 = s \right],
\end{equation}
where $\gamma \in [0,1]$ is a discount factor that determines how much future reward is worth in the present. Likewise, the \textit{action-value} function $Q^\pi(s,a)$ is defined as the expected discounted reward starting at state $s$, taking action $a$, and following policy $\pi$ thereafter
\begin{equation}\label{equation: action value function}
    Q^\pi(s,a) := \mathbb{E}_{\pi, P} \left[\sum_{t=0}^{T-1} \gamma^t R(s_{t+1}) \; | \; s_0 = s, a_0 = a \right].
\end{equation}
These functions are connected by the \textit{Bellman equation} \citep{bellman1954}:
\begin{equation}
Q^\pi(s,a) = \mathbb{E}_{s^\prime \sim P(\cdot | s_t, a_t)} \left[R(s^\prime) + \gamma V^\pi (s^\prime)\right],
\end{equation}
that is, the action-value function can be decomposed into the one-step reward plus the expected value of the next state $s^\prime$ subject to the environment's dynamics and the agent's policy. Value functions can be used to rank policies $\pi_0$ and $\pi_1$; we say that $\pi_0 \geq \pi_1$ if and only if $V^{\pi_0}(s) \geq V^{\pi_1}(s) \; \forall \; s \in \mathcal{S}$. The optimal policy $\pi^*$ has associated optimal value functions $V^*(s)$ and $Q^*(s,a)$, which satisfy the \textit{Bellman optimality equation}
\begin{equation}
    Q^*(s,a) = \mathbb{E}_{s^\prime \sim P(\cdot | s_t, a_t)}\left[R(s^\prime) + \gamma V^*(s^\prime)\right].
\end{equation}
Provided access to the optimal action-value function, the optimal policy can be recovered as:
\begin{equation}
\pi^*(a|s) =
\begin{cases}
    1 & \text{if } a = \argmax_{a'} Q^*(s, a') \\
    0 & \text{otherwise.}
\end{cases}
\end{equation}
In subsequent sections we introduce three classes of methods for estimating $\pi^*$: 1) methods that do not model the environment's dynamics (\colouredS \ref{background: model-free RL}), 2) methods that model the environment's one-step dynamics (\colouredS \ref{background: one-step MBRL}), and 3) methods that model the environment's multi-step dynamics (\colouredS \ref{background: multi-step MBRL}), summarised in Figure \ref{fig: method summary}. Hereafter, we will operate in the regime of Deep Reinforcement Learning \citep{arulkumaran2017deep, li2017deep}, where value functions, policies or dynamics models are, unless otherwise stated, approximated with deep neural networks \citep{lecun2015deep, goodfellow2016deep}. Deep RL is the most recent incarnation of works that use function approximation in RL, typically to deal with large or continuous state and action spaces \citep{sutton1999policy, baird1995residual, precup2001off}. We will often discuss the loss function $\mathcal{L}$ against which a model is optimised. We do not discuss this optimisation process in detail; in practice this involves optimisation algorithms like Adaptive Moment Estimation (ADAM) \citep{kingma2014adam} and Stochastic Gradient Descent (SGD) \citep{robbins1951stochastic, leon1998online} updating model parameters in proportion to gradients obtained by backpropogation \citep{rumelhart1986}.   

\subsection{Model-free Reinforcement Learning}\label{background: model-free RL}

RL methods that search for the optimal policy without explicitly modelling the environment's dynamics are said to be \textit{model-free}. They expedite this search either by approximating the optimal policy directly, learning the optimal value function (and deriving the policy indirectly via Equation \ref{equation: optimal policy}), or a combination of both. In this section we will introduce the preferred methods in these three classes.

\textbf{Policy gradient methods.} $\;$ Methods that estimate the optimal policy directly from experience rely on the policy gradient theorem \citep{sutton1999policy}. For a policy $\pi_\theta(a|s)$ parameterised by $\theta$, the policy gradient theorem says that the gradient of the expected return with respect to the parameters is given by 
\begin{equation}\label{equation: policy gradient theorem}
\nabla_\theta J(\theta) = \mathbb{E}_{\pi_{\theta}} \left[\nabla_{\theta} \log \pi_\theta (a_t | s_t) \; G_t \right],
\end{equation}
where $G_t = \sum_{k=0}^{T - t} \gamma^k r_{t+k}$ is the \textit{discounted return} from timestep $t$ to the terminal timestep $T$. Provided a state-action-return triple $(s_t, a_t, G_t)$, the parameters of the policy are updated via
\begin{equation}
    \theta \leftarrow \theta + \alpha \cdot \nabla_\theta J(\theta),
\end{equation}
where $\alpha \in \mathbb{R}$ is a learning rate. The canonical policy gradient algorithm, REINFORCE \citep{williams1992}, estimates $G_t$ with Monte Carlo rollouts of a fixed policy in the environment; more recent methods like Trust Region Policy Optimisation (TRPO) \citep{schulman2015} and Proximal Policy Optimisation (PPO) \citep{schulman2017} instead use surrogate objectives involving value functions to stabilise learning.

An advantage of policy gradient methods is their ability to handle large or continuous action spaces, which is required for many real-world problems. The cost of this is high data inefficiency.

\textbf{Value-Based Methods.} $\;$ $Q$-learning is the canonical value-based method \citep{watkins1989, dayan1992q}. The key idea is that the agent maintains a running estimate $\hat{Q}(s,a)$ of the optimal action-value function $Q^*(s,a)$. Given a transition $(s, a, r, s^\prime)$, 

$\hat{Q}$ is updated via
\begin{equation}\label{equation: q-learning update}
    \hat{Q}(s, a) \leftarrow \hat{Q}(s, a) + \alpha \cdot \delta, 
\end{equation}
where $\delta = r + \gamma \max_{a^\prime \in \mathcal{A}} \hat{Q}(s^\prime, a^\prime) - \hat{Q}(s, a)$ is the \textit{temporal difference error} in the estimate of $\hat{Q}(s,a)$, and $\alpha \in \mathbb{R}$ is a learning rate. Updating value functions in proportion to their temporal difference error, or TD learning \citep{samuel1959, sutton1984temporal, sutton1988}, is central to nearly all modern RL algorithms that utilise value functions.

A seminal moment in the field came when \citep{mnih2015} combined $Q$-learning with deep neural networks to produce Deep $Q$-Networks (DQN). Now, the running estimate $\hat{Q}(s,a)$ of the optimal value function is approximated by a neural network $Q_\phi(s,a)$ with parameters $\phi$ and Equation \ref{equation: q-learning update} is converted to a mean squared error loss $\mathcal{L}$ on $\phi$:
\begin{equation}\label{equation: DQN loss}
\mathcal{L}_{\text{DQN}}(\phi) = \mathbb{E}_{(s,a,r,s^\prime) \sim \mathcal{D}} \left[\left(Q_\phi(s, a) - (r + \gamma \max_{a^\prime \in \mathcal{A}} Q_{\phi^-}(s^\prime, a^\prime))\right)^2 \right],
\end{equation}
where $\mathcal{D}$ is a replay buffer that stores previously observed transitions, and $\phi^-$ are parameters of a \textit{target network} updated periodically to improve stability. DQN was the first RL method to exhibit superhuman performance on the Arcade Learning Environment \citep{bellemare2013arcade}, and so catalysed the field of Deep RL. Many works have since built atop the framework of Deep $Q$-learning \citep{van2016deep, gu2016continuous, hester2018deep, hessel2018rainbow, dabney2018distributional}.

Conveniently, methods that learn value functions can arrive at the optimal policy using an order of magnitude less data than methods that attempt to extract the optimal policy directly. There is not a single explanation for this, but a useful characteristic of $Q$ learning methods are there ability to learn from \textit{off-policy} data, that is, data derived from a policy different from the our current policy or the optimal policy. This is a core idea in Offline Reinforcement Learning, a key field in this thesis, which we explore in depth in \colouredS \ref{background: offline RL}.

The primary drawback of value-based methods is the need to maintain an accurate value function for all $(s,a) \in \mathcal{S} \times \mathcal{A}$ which usually limits their applicability to problems with small, discrete action spaces. 

Value-based methods address the data inefficiency of policy gradient methods by using off-policy data to learn value functions. This comes at the cost of only being able to maintain predictions for a small, discrete set of actions. Ideally, we'd like to combine the data-efficiency of value-based methods, with the action-space flexibility of policy-gradient methods. The next section discusses this combination

\textbf{Actor critic methods.} $\;$ Actor Critic methods attempt to leverage the best aspects of policy gradient and value-based approaches \citep{konda1999actor, peters2008natural, grondman2012survey}. The usual setup involves an \textit{actor}, which is a policy $\pi_\theta(a|s)$ with parameters $\theta$ as in policy gradient methods, and a \textit{critic}, which is a value function $Q_{\phi}(s,a)$ with parameters $\phi$ as in DQN. The actor is updated to maximise the values predicted by $Q_{\phi}(s,a)$. 

Past works fall in two categories: those with stochastic policies \citep{ziebart2008maximum}, and those with deterministic policies \citep{silver2014}. The canonical method in the first category is Soft Actor Critic (SAC) \citep{Haarnoja2018}; and the canonical method in the second category is Deep Deterministic Policy Gradient (DDPG) \citep{lillicrap2015} and its derivatives Twin Delayed Deep Deterministic policy gradient (TD3) \citep{fujimoto2019} and TD7 \citep{fujimoto2024sale}. No method has proven to consistently outperform the others. In this thesis the line of work that is continued in \colouredS \ref{chapter: from low quality data} and \colouredS \ref{chapter: under partial observability} leverages the deterministic policy actor-critic framework, specifically TD3. As a result, we introduce they key ideas behind TD3 below.

TD3 addresses two failings of previous actor-critic methods. The first is the tendency for $Q$ values to be overestimated because of the $\max$ operator inside the TD error of Equation \ref{equation: q-learning update}, which is further exacerbated by function approximation \citep{thrun2014issues}. The second is the tendency for deterministic policies to overfit to peaks in the value function and so not explore the action-space efficiently. TD3 addresses the first failing with a clipped variant of Double $Q$-learning \citep{hasselt2010double}, where two critics $Q_{\phi_1}$ and $Q_{\phi_2}$ are updated towards the \textit{minimum} of their predictions. TD3 addresses the second failing by adding Gaussian noise to actions used in the critic update which has the effect of assigning similar value to nearby actions in the state-action space. As a result, the loss function for critic $i$ with parameters $\phi_i$ is a subtle evolution of the DQN loss (Equation \ref{equation: DQN loss})
\begin{equation}\label{equation: TD3 critic loss}
\mathcal{L}_{\text{TD3}}(\phi_i) = \mathbb{E}_{(s,a,r,s^\prime) \sim \mathcal{D}} \left[\left(Q_{\phi_i}(s, a) - (r + \gamma \min_{i=1,2} Q_{\phi^-_i}(s^\prime, (\pi_\theta(s^\prime) + \epsilon))\right)^2 \right],
\end{equation}
where $\epsilon \sim \text{clip}(\mathcal{N}(0, \sigma), -c, c)$ is 0-mean Gaussian noise with variance $\sigma$, clipped in $[-c,c]$. The policy loss is derived from the predictions from the first critic $Q_{\phi_1}$
\begin{equation}\label{equation: TD3 actor loss}
    \mathcal{L}_{\text{TD3}}(\theta) = - \mathbb{E}_{s \sim \mathcal{D}} \left[ Q_{\phi_1}(s, \pi_{\theta}(s))\right].
\end{equation}
\subsection{One-step Model-based Reinforcement Learning}\label{background: one-step MBRL}
A key characteristic of the RL problem (\colouredS \ref{background: reinforcement learning}) is that the system dynamics are unknown to the agent \textit{a priori}. The methods discussed in \colouredS \ref{background: model-free RL} deal with this by implicitly encoding the dynamics in the value function or policy. But is there benefit to modelling the system dynamics explicitly? That is the question asked in \textit{Model-based Reinforcement Learning} (MBRL) \citep{polydoros2017survey, moerland2023model}. 

Research on MBRL was catalysed by the \textit{Dyna} algorithm \citep{sutton1991} which built the framework that most modern MBRL algorithms follow. Here, the \textit{model} is some function $f_{\theta}$ with parameters $\theta$ that approximates the one-step system dynamics and (optionally) the one-step reward, that is, it directly approximates $P(s_{t+1} | s_t, a_t)$ and (optionally) $R(s_{t+1})$\footnote{A common assumption is that the ground truth reward function is available for calculating the rewards associated with model predictions. In this case learning an approximation of the reward function is unnecessary.}. The model is used to generate synthetic experience for updating the policy in a process called \textit{planning}. The benefit of planning is that it is conducted without interacting with the environment, which can greatly reduce the amount of real experience required to learn a policy. The remainder of this section will discuss popular modelling and planning methods.

\textbf{Dynamics models.} $\;$ The most data-efficient MBRL method to-date is PILCO \citep{deisenroth2011pilco} which models the system dynamics using Gaussian Processes \citep{engel2005reinforcement, kuss2003gaussian, williams2006}, and solves the cartpole swing-up task with 20 seconds of experience. This spurred others to experiment with Gaussian Processes for MBRL \citep{deisenroth2013gaussian, grande2014sample, levine2011nonlinear, kamthe2018, saemundsson2018meta}, but unfortunately inference using a dataset of size $n$ has complexity $\mathcal{O}(n^3)$ which becomes unwieldy with more than a few thousand samples \citep{hensman2013}. Deep neural networks scale more favourably and have been used as a substitute \citep{nagabandi2018, clavera18a, lambert2019low}. The most straightforward approach uses a model $f_\theta$ to predict the state $s_{t+1}$ that follows $(s_t, a_t)$ via maximum likelihood estimation (MLE). Given a dataset of transitions $\mathcal{D} = \{(s_t, a_t, s_{t+1})^i\}_{i=1}^{|\mathcal{D}|}$, the standard MLE loss is
\begin{equation}\label{equation: mbrl MLE}
    \mathcal{L}_{\text{model}}(\theta) = -\mathbb{E}_{(s_t, a_t, s_{t+1}) \sim \mathcal{D}} \left[ \log f_{\theta}(s_{t+1} | s_t, a_t) \right].
\end{equation}
Unlike Gaussian Processes, a network trained with the standard MLE loss does not quantify uncertainty in its predictions. More recent works like Probabilistic Ensembles with Trajectory Sampling (PETS) address this by training an ensemble of neural networks to predict a multivariate Gaussian distribution over the next state i.e. $f_\theta({\mathcal{N}(\mu_\theta(s_{t+1}), \sigma_\theta(s_{t+1})) \; | \; s_t, a_t)}$. Each member of the ensemble has a different weight initialisation and is trained on a different subset of $\mathcal{D}$ \citep{lakshminarayanan2017, clavera18a, janner2019trust}, meaning their aggregated predictions provide some estimate of epistemic uncertainty, and the multivariate Gaussian provides some estimate of aleatoric uncertainty. PETS reduced the amount of data required to complete standard MuJoCo benchmark tasks \citep{todorov2012} by more than 10x over state-of-the-art model-free methods like PPO. 

\textbf{World models.} $\;$ The methods discussed above operate on sensor-based states (e.g. joint torques and velocities), but in many real cases it is more convenient to operate on pixel-based states derived from cameras. If we were to naively apply the models discussed thus far on pixel-based states they would be required to predict the change in pixel intensity for all pixels in the image, which becomes impractical for any realistic image resolution. Instead, it is common to encode the image-based state into a compact, low-dimensional latent space, then train a model to predict the \textit{latent dynamics} in this low-dimensional space \citep{watter2015embed, hafner2019learning, banijamali2018robust}. Methods following this routine are often called \textit{world models} \citep{ha2018, schmidhuber1990making, schmidhuber1990line, schmidhuber2015learning}, with the preeminent work being the Dreamer family \citep{hafner2019, hafner2020mastering, hafner2023mastering}. The simplest setup involves a representation network $p_\theta(z_t | z_{t-1}, a_{t-1}, o_t)$ that predicts latent state $z$ given the previous latent state, action, and image observation $o_t$, and a latent dynamics model $q_\theta(z_t | z_{t-1}, a_{t-1})$ that predicts the next latent state given the current latent state and action. The representation network is a convolutional neural network \citep{lecun1989backpropagation} and the latent dynamics model is a recurrent neural network \citep{hafner2019learning, chung2014empirical}. DreamerV2 \citep{hafner2020mastering} was the first method to exhibit superhuman performance on the Arcade Learning Environment by training a policy using data generated solely by a latent dynamics model. If the world model cannot be paired with a ground-truth reward function, then an approximate \textit{reward model} must be learned from data. This can be achieved with reward-labelled states \citep{sekar2020}, via human feedback \citep{christiano2017deep}, via an observer's preference for one trajectory over another \citep{sadigh2017active, ibarz2018reward}, or using inverse RL \citep{ng2000algorithms}. 

\textbf{Planning.} $\;$ How is a model of the system dynamics and rewards used to select the right action? The canonical planning algorithm is Model Predictive Control (MPC) \citep{morari2017, morari1999model, garcia1989model, allgower2012nonlinear, camacho2013}. Provided a dynamics model $f_\theta$, reward function $R$, and initial state $s_t$, MPC searches for the sequence of actions $A^*_H = (a_t, \ldots, a_{t+H-1})$ that maximise reward over horizon $H$
\begin{equation}\label{equation: analytic mpc}
A^*_H = \argmax_{A^*_H} \sum_{k=0}^{H-1} \mathbb{E}_{f_\theta} \left[ R(s_{t+k}) \; | \; s_{t+k+1} = f_{\theta}(s_k, a_k) \right].
\end{equation}
Approximate solutions tend to be used in lieu of an analytical solution. These include \textit{random-shooting} \citep{rao2009survey}, which involves populating a matrix of random action sequences, rolling them out to horizon $H$, and executing the first action from the sequence with highest predicted reward. The Cross Entropy Method (CEM) \citep{botev2013}, and Model Predictive Path Integral (MPPI) \citep{williams2015} follow a similar routine, but perform several iterations of action selection matrix population, where each new iteration skews the samples towards action sequences with higher predicted reward.

A second popular planning algorithm is Monte Carlo Tree Search (MCTS) \citep{coulom2006}, as used by game-playing algorithms like AlphaGo \citep{silver2016mastering}, AlphaGo Zero \citep{silver2017}, AlphaZero \citep{silver2018general} and MuZero \citep{schrittwieser2020}. MCTS creates a search tree where the nodes are states and edges are actions sampled from a policy network. The tree of predicted future states is populated according to dynamics $f_\theta$ and rewards are calculated according to $R_\phi$. The reward statistics are aggregated across the tree and an action is sampled from a distribution weighted with respect to expected future reward. This procedure has been shown to be remarkably effective when the action-space is discrete, as is the case in the games mastered by the algorithms described above. However, in this thesis, and in most real-world problems, we are concerned exclusively with continuous action spaces where planning with MCTS is infeasible.

Equipped with a world model that accurately simulates the environment dynamics, a reward model that specifies a task of interest, and a planning algorithm or policy for selecting actions that maximise the reward model \colouredS \ref{background: model-free RL}, an agent can be trained entirely in simulation to generalise to \textit{one task} in the real-world zero-shot \citep{sekar2020, rigter2023reward}. Problems arise, however, if we want the agent to solve multiple tasks zero-shot. This would require learning multiple reward models, then training multiple distinct policies in parallel. To what extent does that approach scale? In the following section we will discuss methods that attempt to bypass this problem, learning policies in simulation that can generalise to \textit{many} tasks. 

\subsection{Multi-step Model-based Reinforcement Learning}\label{background: multi-step MBRL}
So far we have discussed different methods for summarising an MDP to inform policy selection. In \colouredS \ref{background: value functions} we discussed \textit{value functions} which are methods for summarising the long-run utility of states, actions and policies. Value functions are \textit{far-sighted} insofar as they model the reward an agent following some policy can expect to receive far into the future\footnote{Indeed, in the case of non-episodic MDPs, value functions summarise expected reward over infinite time-horizons, with the degree to which the future is considered mediated by the discount factor $\gamma$.}. In \colouredS \ref{background: one-step MBRL} we discussed \textit{dynamics models} which are methods for summarising the short-run evolution of the state when taking some action, from which its utility can be derived via a reward model. Dynamics models are \textit{near-sighted} insofar as they model the reward an agent following some policy can expect to receive in the short-term (where ``shortness'' here is mediated by how accurate the model remains as the planning horizon is lengthened). In this section we shall discuss a third class of methods which we think of as \textit{far-sighted dynamics models}. Informally, instead of answering the question of how likely the agent is to reach some \textit{next} state, they answer how likely the agent is to reach some \textit{future} state that may not be accessible for many timesteps. As we shall see, methods capable of providing such answers have useful, unique properties. There is no agreed-upon name for this class of methods in the literature so we call them \textit{multi-step model-based reinforcement learning} methods.

\textbf{Successor representations.} $\;$ Work on multi-step model-based RL methods began with the \textit{successor representation} (SR) \citep{dayan1993}. For an MDP with discrete state and action spaces, the SR $M^\pi(s_0, a_0)$ of state-action pair $(s_0, a_0)$ when following 
policy $\pi$ is defined as the expected discounted sum of future occurrences of each state in the state space:
\begin{equation}\label{equation: SR}
M^\pi(s_0, a_0, s) := \mathbb{E}_{\pi, P} \left[ \sum_{t=0}^T \gamma^t \mathbf{1} (s_{t+1} = s) \; | \; (s_0, a_0) \right] \; \forall \; s \; \in \mathcal{S}.    
\end{equation}
The SR can answer the following informal question: ``how often am I likely to pass through state $s$ when following policy $\pi$''. This formulation has a valuable characteristic: it can be used to derive the value function $Q_R^\pi$ for policy $\pi$ with respect to reward function $R$ using only linear operations
\begin{equation}\label{equation: Q from SR}
Q^\pi_R(s_0, a_0) = \sum_{s \in \mathcal{S}}M^\pi(s_0, a_0, s) R(s). 
\end{equation}
\textbf{Successor features and successor measures.} $\;$ SRs operate on MDPs with discrete state and action spaces. Recent work has generalised the core ideas to MDPs with continuous state and action spaces, with this first to do so being Barreto et al.'s \textit{successor features} (SFs) \citep{barreto2017}. The SFs $\psi^\pi$ of state-action pair $(s_0, a_0)$ are defined as the expected discounted sum of future state \textit{features} subject to the system dynamics and policy:

\begin{equation}\label{equation: successor features}
\psi^\pi(s_0, a_0) := \mathbb{E}_{\pi, P} \left[\sum_{t=0}^T \gamma^t \varphi(s_{t+1}) \; | \; (s_0, a_0) \right],
\end{equation} 

where $\varphi: \mathcal{S} \rightarrow \mathbb{R}^d$ is a feature-map that embeds states in a $d$-dimensional space. Note that if the state space is discrete, and if the features are a one-hot encoding of the state into $\mathbb{R}^{|\mathcal{S}|}$, then we recover the definition of the SR (Equation \ref{equation: SR}). Instead of the reward being a function of the state, it is then assumed that reward is a linear function of the features i.e. $R(s) = \varphi(s)^\top \mathbf{w}$ where $\mathbf{w} \in \mathbb{R}^d$ are the reward weights. As a result, like with SRs in Equation \ref{equation: Q from SR}, the value function for policy $\pi$ and reward $R$ can be quickly recovered from the successor features:
\begin{align}\label{equation: Q from SFs}
    Q^\pi_R(s_0, a_0) &= \mathbb{E}_{\pi, P} \left[\sum_{t=0}^T \gamma^t R(s_{t+1}) \right] \notag \\
    &= \mathbb{E}_{\pi, P} \left[\sum_{t=0}^T \gamma^t \varphi(s_{t+1})^\top \right] \mathbf{w}_R \notag  \\
    &= \psi^\pi(s_0, a_0)^\top \mathbf{w}_R.
\end{align}
An alternative approach is based on \textit{successor measures} \citep{blier2021learning}. Here, instead of computing the discounted sum of future features, it is the discounted probability of reaching some subset of the state space in the future that is computed:
\begin{equation}\label{equation: successor measure}
M^\pi(s_0, a_0, X) := \sum_{t=0}^T \gamma^t \text{Pr}_{\pi, P}(s_{t+1} \in X \; | \; s_0, a_0) \; \forall \; X \; \subset \mathcal{S},
\end{equation}
where $\text{Pr}_{\pi, P}(\cdot | s_0, a_0)$ is the probability of some event occurring subject to policy $\pi$, dynamics $P$ and initial state-action pair $(s_0, a_0)$. Both SFs and SMs are closely related to Sutton et al.'s \textit{general value functions} \citep{sutton2011horde} which view the reward expectation computed by traditional value functions (Equations \ref{equation: state value function} and \ref{equation: action value function}) as a special case of a more general \textit{cumulant}, that is, any property of an MDP's long-term behaviour that may be of interest for finding the optimal policy. In the case of SFs the cumulants are the state features, and in the case of SMs the cumulant is the state visitation probability.

\textbf{Universal successor representations.} $\;$ SRs, SFs, and SMs are all defined with respect to some policy $\pi$. In this section we shall introduce the final, most general multi-step model-based RL methods which disentangle the policy from SFs and SMs. In the case of SFs this is achieved with \textit{universal successor features} \citep{borsa2018}; and in the case of SMs this is achieved with \textit{forward-backward representations} \citep{touati2021learning}. The unifying idea between these methods is to condition the long-run dynamics predictions on some parameterisation of the policy to allow the expected behaviour of different policies to be compared. This idea is similar to Schaul et al.'s \textit{universal value function approximators} \citep{schaul2015} which condition the value function on a goal state.  We shall discuss each method in turn.

Consider a set of policies $\pi_z$ parameterised by $z \in \mathbb{R}^d$. Universal successor features (USFs) $\psi: \mathcal{S} \times \mathcal{A} \times \mathbb{R}^d \rightarrow \mathbb{R}^d$, compute the expected future features of some state-action pair conditioned on the policy induced by $z$
\begin{equation}\label{equation: universal SFs}
    \psi(s_0, a_0, z) =  \mathbb{E}_{\pi_z, P} \left[ \sum_{t=0}^T \gamma^t \varphi(s_{t+1}) \; | \; s_0, a_0 \right]  \quad \; \forall \; (s_0, a_0) \in \mathcal{S} \times \mathcal{A}, z \in \mathbb{R}^d.
\end{equation}
The policy is trained to maximise the value function induced by setting $\mathbf{w}_R = z$
\begin{equation}\label{equation: universal SFs policy}
    \pi_{z}(s) := \argmax_a \psi(s,a,z)^\top z \qquad \forall \; (s,a) \in \mathcal{S} \times \mathcal{A}, z \in \mathbb{R}^d.
\end{equation}
If Equation \ref{equation: universal SFs policy} holds, then the policy $\pi_z$ is approximately optimal for the task defined by reward weights $\mathbf{w}_R = z$. Forward-backward representations apply a similar idea in the context of successor measures. Here, a forward model $F: \mathcal{S} \times \mathcal{A} \times \mathbb{R}^d \rightarrow \mathbb{R}^d$ and a backward model $B: \mathcal{S} \rightarrow \mathbb{R}^d$ are trained to jointly approximate the successor measure:
\begin{equation}\label{equation: forward-backward framework}
    M^{\pi_z}(s_0,a_0, X)\approx \int_X F(s_0,a_0,z)^\top B(s, a)\rho(\text{d}s) \quad \forall \; (s_0, a_0) \in \mathcal{S} \times \mathcal{A}, X \subset \mathcal{S}, z \in \mathbb{R}^d,
\end{equation}
where $\rho$ is the distribution of states viewed by the agent. The forward embedding of state-action pair $(s_0, a_0)$ can be thought of as summarising their expected future trajectories subject to a policy parameterised by $z$. Because this summarisation is compressed into representation space $\mathbb{R}^d$, this is functionally identical to a USF with feature embedding space $\mathbb{R}^d$. The backward embedding $B$ of a state $s_+$ can be thought of as summarising the possible routes to $s_+$. If the dot product of these two embeddings $F^\top B$ is large, then $s_+$ is reachable from $(s_0, a_0)$. With the forward model being functionally equivalent to universal successor features, the policy is trained identically
\begin{equation}
    \pi_{z}(s) := \argmax_a F(s,a,z)^\top z \qquad \forall \; (s,a) \in \mathcal{S} \times \mathcal{A}, z \in \mathbb{R}^d.
\end{equation}
\begin{center}
    * * *
\end{center}
We began this chapter with an introduction of the traditional reinforcement learning setting: an agent interacting with an environment to learn intelligent behaviour \textit{tabula rasa}, with no prior knowledge. Another name for this setting is \textit{online} reinforcement learning. The label \textit{online} reflects the fact that, in some sense, the agent is always on the clock, interleaving new experiences, learning, planning and goal accomplishment. This framing deliberately mirrors a human life, a continuum of experience in which we too must interleave learning, planning and survival. However, can we provide them with some prior knowledge that might make online learning more sample efficient? This is the concern of \textit{offline} reinforcement learning, the subject of the next section. 

\section{Offline Reinforcement Learning}\label{background: offline RL}
Offline reinforcement learning aims to build agents that solve tasks in environments without needing to interact with them \citep{ernst2005tree, riedmiller2005neural, lange2012batch, levine2020}. To achieve this the agent is provided a historical dataset of experience, collected by some other agent or controller, from which it must derive its own policy. The dataset is said to be \textit{offline} because it is made available to the agent prior to \textit{online} evaluation in the environment. There are practical motivations for this sort of pre-training. For most real-world problems the unconstrained trial-and-error learning implied by the online reinforcement learning setting is infeasible, costly or dangerous \citep{dulac2019challenges}. Such problems are found in domains like healthcare, where trial-and-error learning could risk the patient's health; industrial control, where trial-and-error learning could damage expensive infrastructure; or autonomous driving, where trial-and-error learning could endanger road users.

\begin{figure}
    \centering
    \includegraphics[width=\textwidth]{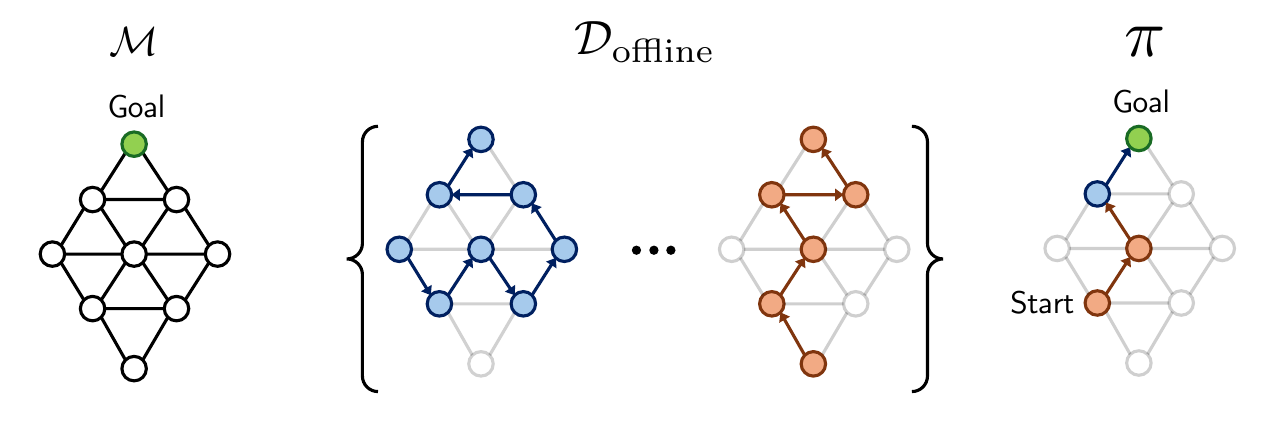}
    \caption{\textbf{Trajectory stitching in offline RL.} \textit{(Left)} A graph MDP $\mathcal{M}$ where the task is to find the shortest path to the goal state. \textit{(Middle)} A dataset of offline trajectories $\mathcal{D}_{\text{offline}}$ that may not contain the optimal trajectory for the task. \textit{(Right)} The policy $\pi$ learns to combine sub-trajectories to find the shortest path to the goal.}
    \label{fig: offline RL trajectory stitching}
\end{figure}

As with the online setting, the environment is modelled as an MDP (\colouredS\ref{background: mdps}), and the agent's goal is to find a policy $\pi$ that maximises the expected return (Equation \ref{equation: return}). For training, the agent has access to a static dataset of transitions $\mathcal{D}_{\text{offline}} = \{(s_t, a_t, r_t, s_{t+1})\}_{t=1}^{|\mathcal{D}_{\text{offline}}|}$ collected on its behalf by one or more unknown behaviour policies $\pi_{\beta}$. There are no specific assumptions made about $\pi_\beta$, but in general it is either a) optimal for the test task, or b) sub-optimal for the test task. In the case of a), the offline RL problem collapses to an imitation learning problem \citep{schaal1996learning, schaal1999imitation} where the agent only needs to copy the policy implicit in the dataset to maximise expected reward at test time. In the case of b), the agent must \textit{stitch together} sub-optimal trajectories to find a policy that is \textit{different} and \textit{better} than that which created the dataset—Figure \ref{fig: offline RL trajectory stitching}. It is case b) that interests the field because datasets demonstrating sub-optimal control are plentiful for most real-world problems. Only once the agent has used the dataset to learn a policy is it exposed to the environment for evaluation.

Because the learned policy $\pi$ must be different and better than the behaviour policy $\pi_\beta$, there will necessarily be a \textit{distribution shift} between the states visited by $\pi$ and $\pi_\beta$. This is true of the online setting too, but is less of a problem. There, the agent can find itself in parts of the state-space that it hasn't yet observed, form ``what-if'' hypotheses about where actions may lead, then falsify them by executing the actions and observing the behaviour. Clearly, the offline setting doesn't allow this luxury, so na\"ive application of online RL methods leads to irrecoverable prediction errors at states and actions not in $\mathcal{D}_{\text{offline}}$ \citep{kumar19, levine2020}. 

\begin{figure}
    \centering
    \includegraphics[width=\textwidth]{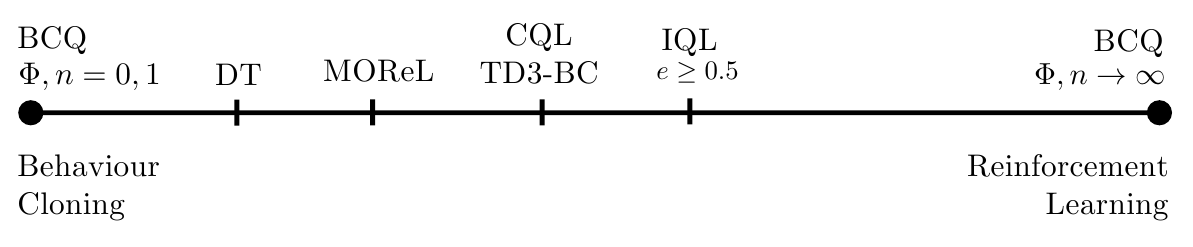}
    \caption{\textbf{Behaviour cloning $\leftrightarrow$ reinforcement learning continuum for Offline RL methods.} At the left end lie methods that attempt to mimic $\pi_{\beta}$, the policy that produced the data. The further right one moves the less the methods attempt to mimic $\pi_{\beta}$ and the closer they are to full RL methods.}
    \label{fig: offline rl continuum}
\end{figure}

Past works have developed a set of methods for \textit{constraining} the agent's behaviour to be close to the training dataset, and so minimise the distribution shift. The degree to which the agent's behaviour matches the dataset is the critical design decision for offline RL algorithms. We can plot different methods as a function of the deviation they allow—Figure \ref{fig: offline rl continuum}. At one end sit methods that attempt to exactly copy the behaviour policy (also known as \textit{behaviour cloning} \citep{baird1995residual}); at the other end sit methods that maximally deviate from the behaviour policy (which is equivalent to full reinforcement learning). The goal for any method is to move away from the behaviour policy to enable different and better behaviour, but not far enough to lose the grounding provided by the data. 
The following sections will cover the key works. We will begin with a review of \textit{policy constraints} (\colouredS\ref{background: policy constraints}), then \textit{value function constraints} (\colouredS\ref{background: value function constraints}), and finish with \textit{model constraints} (\colouredS\ref{background: model constraints}).

\subsection{Policy Constraints}\label{background: policy constraints}
Policy constraint methods control how much the learned policy deviates from the behaviour policy. Batch-Constrained Deep $Q$-learning (BCQ) \citep{fujimoto2019off} is the canonical policy-constraint offline RL method. Here, an approximation of the behaviour policy $\hat{\pi}_\beta$ is made with a Conditional Variational Autoencoder \citep{sohn2015learning}, and actions are sampled from it when training a value function. To update the value function at state-action pairs not present in the dataset some of the actions are updated with a \textit{perturbation model} $\zeta$. The $Q$ function is trained across a set of $n$ perturbed actions $\{a_i + \zeta(s, a_i, \Phi)\}_{i=1}^n$ with $a_i \sim \hat{\pi}_\beta$, and $\Phi$ defining a range $[-\Phi, \Phi]$ that constrains the perturbation. A nice property of BCQ is that the practitioner can control where $\pi$ sits on the continuum in Figure \ref{fig: offline rl continuum} using $\Phi$ and $n$. Setting $\Phi=0$ and $n=1$ dictates strict behaviour cloning of $\pi_\beta$. Increasing $\Phi$ and $n$ encourages deviation from $\pi_\beta$, and as $\Phi, n \rightarrow \infty$ full reinforcement learning is recovered.

Other works achieve a similar result without having to directly approximate $\pi_\beta$ \citep{wu2019behavior, fakoor2021continuous, ghasemipour2021emaq, peng2023weighted, matsushima2020deployment, kumar2019stabilizing, wu2022supported, yang2022behavior, fujimoto2024sale}. Perhaps the most successful of these is TD3-BC \citep{fujimoto2021minimalist}, which adds a {\color{bupu@purple}{behaviour cloning term}} to the usual TD3 loss (Equation \ref{equation: TD3 critic loss}, \colouredS \ref{background: model-free RL}):
\begin{equation}\label{equation: TD3-BC loss}
    \mathcal{L}_{\text{TD3-BC}}(\theta) = - \mathbb{E}_{(s,a) \sim \mathcal{D_{\text{offline}}}} \left[ \lambda Q_{\phi}(s, \pi_{\theta}(s)) - {\color{bupu@purple}{(\pi_\theta(s) - a)^2}}\right],
\end{equation}
where $\lambda$ controls the relative strength of the behaviour cloning term. This is related to work that uses importance sampling to align $\pi$ and $\pi_\beta$ \citep{precup2001off, sutton2016emphatic, liu2019off, nachum2019dualdice, gelada2019off}, but importance sampling has proven less successful empirically than methods akin to TD3-BC.

A parallel research direction re-frames policy-constrained offline RL as a sequence-learning problem \citep{chen2021decision, janner2021offline, lee2022, reed2022generalist, zheng2022online, chebotar2023q, furuta2021generalized, siebenborn2022crucial, krause22transformer, xu22transformer}. The first to do so was the Decision Transformer (DT) \citep{chen2021decision}, whose key idea was to train the agent to imitate high-value trajectories contained in the dataset. The DT trajectory $\tau^{\text{DT}}$ replaces the one-step reward in our definition of $\tau$ (\colouredS \ref{background: mdps}) with the undiscounted return-to-go $G_t = \sum_{k=0}^{T - t} r_{t+k}$. Given a state space $\mathcal{S} \in \mathbb{R}^N$ and an action space $\mathcal{A} \in \mathbb{R}^M$, each dimension forms an element of the trajectory of length $H$ such that:
\begin{equation}\label{equation: full trajectory sequence}
    \tau^{\text{DT}} = \left( \ldots, s_t^1, s_t^2, \ldots, s_t^N, a_t^1, a_t^2, \ldots, a_t^M, G_t, \ldots \right) \; \; t = 1, \ldots, H. 
\end{equation}
A Transformer \citep{vaswani2017attention} is trained to autoregressively predict the next token in the sequence as per the familiar language modelling setup \cite{brown2020language, achiam2023gpt, team2023gemini}. When the policy is rolled out at test-time, $G_0$ is set to the maximum cumulative reward an agent could receive in the MDP. Subsequent return-to-gos are found by subtracting the observed one step reward from the previous return-to-go $G_t = G_{t-1} - r_t$. The hope is that by conditioning each action on an approximate maximum return, the Transformer will retrieve the action that maximised the return from that state in the dataset.

\subsection{Value Function Constraints}\label{background: value function constraints} 
Value function constraint methods regularise the $Q$ function at state-action pairs not present in the dataset. Recall, DQN's value function loss from \colouredS \ref{background: model-free RL}:
\begin{equation}\notag
    \mathcal{L}_{\text{DQN}}(\phi) = \mathbb{E}_{(s,a,r,s^\prime) \sim \mathcal{D}} \left[\left(Q_\phi(s, a) - (r + \gamma \max_{a^\prime \in \mathcal{A}} Q_{\phi^-}(s^\prime, a^\prime))\right)^2 \right].
\end{equation}
Note that the TD target $(r + \gamma \max_{a^\prime \in \mathcal{A}} Q_{\phi^-}(s^\prime, a^\prime))$ contains a maximisation over the action space $\mathcal{A}$. There is no guarantee that this action is present in the dataset which can lead to an overestimation of $Q(s,a) \; \forall \; (s,a) \notin \mathcal{D}_{\text{offline}}$ and a consequent biasing of the policy toward unseen actions \citep{kumar19}. The canonical fix is Conservative $Q$-learning (CQL) \citep{kumar2020}, which adds a {\color{solarized@blue}{new term}} to the DQN loss:
\begin{equation}\label{equation: CQL loss}
    \mathcal{L}_{\text{CQL}}(\phi) = {\color{solarized@blue}{\alpha \cdot \left(\mathbb{E}_{s \sim \mathcal{D}_{\text{offline}}}[\max_{a \in \mathcal{A}}Q_\phi(s,a)] - \mathbb{E}_{(s,a) \sim \mathcal{D}_{\text{offline}}}\left[Q_\phi(s,a)\right]\right)}} + \mathcal{L}_{\text{DQN}},
\end{equation}
where $\alpha$ is a scaling parameter. The CQL loss has the dual effect of minimising peaks in the current $Q$ iterate, and maximising $Q$ for state-actions pairs in the dataset.

CQL learns a \textit{pessimistic} estimate of the true value function, but more recent works have shown it to be more pessimistic than necessary \citep{nair2020awac, yang2022rorl, lyu2022, garg2023extreme}. IQL allows the practitioner to ``tune'' how pessimistically the agent views out-of-distribution states and actions \citep{kostrikov2021offline}. IQL's $Q$ function loss is
\begin{equation}
    \mathcal{L}_{\text{IQL}}(\theta) = \mathbb{E}_{(s,a,r,s^\prime)\sim \mathcal{D}_{\text{offline}}}\left[\left(r + \gamma V_\phi(s^\prime) - Q_\theta(s,a)\right)^2\right],
\end{equation}
where the state value function $V_\phi$ is learned with
\begin{equation}
    \mathcal{L}_{\text{IQL}}(\phi) = \mathbb{E}_{(s,a)\sim \mathcal{D}_{\text{offline}}}\left[L_2^e\left(Q_{\theta^-}(s, a) - V_\phi(s)\right)\right],
\end{equation}
where $L_2^e(u) = (e - \mathbb{I}(u < 0))u^2)$ is the asymmetric least squares loss (also called expectile regression) on $u$,  mediated by $e \in [0,1]$ with $\mathbb{I}$ representing the indicator function. Setting $e = 0.5$ returns  the the expected value of the distribution, $e \geq 0.5$ decreases pessimism and $e < 0.5$ increases pessimism. Value-constraint methods like CQL and IQL have consistently outperformed policy-constraint methods on the standard offline RL benchmarks like D4RL \citep{fu2020}.

\subsection{Model Constraints}\label{background: model constraints} Model constraint methods penalise the predictions of the learned dynamics model at state-action pairs not in the dataset. The canonical method is Model-based Offline Reinforcement Learning (MOReL) \citep{kidambi2020}, which learns a \textit{pessimistic} ensemble of Gaussian dynamics models $\hat{f}_\theta$, and an amended reward function $\Hat{R}$. The key idea is to detect \textit{unknown} state-actions (i.e. those not present in $\mathcal{D}_{\text{offline}}$) with a function $U(s,a)$, then force rollouts that pass through them to transition to an absorbing state $\texttt{HALT}$ with high negative reward. The pessimistic dynamics model is given by:
\begin{equation}
\hat{f}_\theta(s^\prime | s, a) =
    \begin{cases}
        \delta(s^\prime = \texttt{HALT}) \quad \text{if} \; \; U(s,a) = \text{TRUE} \; \; \text{or} \; \; s = \texttt{HALT} \\
        \hat{f}_\theta(s' | s,a) \quad \quad \; \text{otherwise},
    \end{cases}
\end{equation}
where $\delta(s^\prime = \texttt{HALT})$ is the Dirac delta function which forces the transition to absorbing state \texttt{HALT}. The pessimistic reward function is given by:
\begin{equation}
\Hat{R}(s) = 
    \begin{cases}
        -\kappa \quad \quad \text{if} \; \; s = \texttt{HALT}  \\
        R(s) \quad \;   \text{otherwise}.
    \end{cases}
\end{equation}
where $-\kappa$ is an arbitrarily large negative reward. The dynamics model is an ensemble of $K$ models $\{f_{\theta_1}, \ldots, f_{\theta_K}\}$. If the difference in the predictions made by models in the ensemble exceeds some threshold $\lambda$ for some state-action pair $(s,a)$, then $(s,a)$ is classified as unknown i.e. $U(s,a) = \max_{i,j}||f_{\theta_i}(s,a) - f_{\theta_j}(s,a)||_2 \geq \lambda$. Several other works are built around this idea \citep{argenson2020model, yu2020, matsushima2020deployment, rafailov2021offline}. A different approach is to use an analogue to $U(s,a)$ to identify state-actions for CQL-style value function regularisation \citep{yu2021combo, rigter2022}, which is useful in continuous action spaces where CQL's $\max$ operator is tricky to estimate.
\begin{center}
    * * *
\end{center}
We have now introduced the two central paradigms of reinforcement learning. In \textit{online} RL, we provide the agent direct, \textit{online} access to the environment in which we evaluate its performance. It uses this access to perform trial-and-error learning in its search for the optimal policy. In \textit{offline} RL, the agent only receives indirect, \textit{offline} access to the environment via a static dataset of pre-collected transitions. It uses this dataset to stitch together sub-optimal trajectories and return a policy that is different and better than the one that produced the dataset. Though these settings disagree on how the agent should access the environment, they agree on an important part of the problem specification: both the \textit{system dynamics} $P$ and \textit{task} $R$ do not change between training and testing. 

Unlike the methods discussed so far, humans have ways of dealing with situations that don't conform to this assumption. Early steps have been made to replicate this ability in RL agents, steps we shall explore together in the next section.

% For example, early schooling teaches children how to write with a pencil. This process usually starts with them learning how to move their hand to form each letter in the alphabet, before they string some of these letters together to form their name, then more words, sentences and paragraphs. Once this skill is learned they can perform it with writing implements of different weight, length, and colour, and they use for different means like writing, drawing and painting. In the parlance of RL, the new writing implements have \textit{unseen dynamics} (in that the same muscle movements from the child result in different behaviour to that \textit{seen} in training), and the new means are \textit{unseen tasks} (in that the goal is different to that \textit{seen} in training). None of the methods discussed so far can adapt to unseen dynamics or tasks. Fortunately, however, many authors have identified this problem and taken steps toward solving it. The field concerns itself with \textit{zero-shot generalisation} in RL, and we will explore it together in the following section.

\section{Zero-Shot Generalisation in Reinforcement Learning}\label{background: zero-shot RL}
The traditional RL problem assumes the training and testing environments are identical. In this section we will relax this assumption, and discuss situations where there is some change in the environment at test time that the agent must deal with. We consider changes to either the reward function $R$ or transition function $P$. We shall explore works that try to design agents to deal with such changes within the first deployment episode. In other words, these works want their agents to \textit{generalise} to the new scenarios with \textit{zero practice shots}\footnote{It is this restriction that distinguishes zero-shot generalisation from the related fields of Transfer Learning \citep{taylor2009transfer, lazaric2012} and Meta Reinforcement Learning \citep{wang2016learning, finn2017model}. Though typical Transfer and Meta Reinforcement Learning setups also assume some change in the environment between training and testing, they assume the agent is allowed multiple episodes in the test context to adjust to the change.}, which is where the sub-field derives its name: \textit{zero-shot generalisation} \citep{palatucci2009zero, socher2013, kirk2023survey}.

Solving most real-world problems requires zero-shot generalisation. As we discussed in \colouredS \ref{background: offline RL}, unconstrained trial-and-error learning in the real-world is infeasible, so some amount of pre-training is inevitable. However, it is equally infeasible to think we can prepare an agent for every possible scenario it will face downstream. As a result, there will necessarily be some gap between training and deployment \textit{contexts} \citep{hallak2015contextual} that the agent must bridge to successfully solve problems. In more practical terms, this gap arises from mismatches between the real-world and pre-training worlds built with simulators or historical datasets. When we pre-train agents in simulation, it is extremely difficult to configure them to perfectly match the real-world, which is the well-established sim-to-real problem \citep{zhao2020sim}. When we pre-train with historical datasets, it is equally difficult to know if the dynamics of the real-world have changed in the interim, or to guarantee the task we want the agent to solve hasn't switched away from the one solved in the dataset. None of the methods discussed so far have mechanisms for handling these discrepancies, but many other works do, and it is those that will be the focus of this section. 

We shall discuss problem settings where only the reward function \textit{or} the system dynamics change between training and testing contexts. When we vary only the reward function between training and testing we call this \textit{task generalisation}. When we vary only the dynamics function between training and testing we call this \textit{dynamics generalisation}. To illustrate the difference consider an RL agent controlling a smart thermostat to maintain temperature setpoint $T_{\text{set}}$. If the agent were trained with $T_{\text{set}} = 22^oC$, then tested with $T_{\text{set}} = 18^oC$ it must perform task generalisation to accommodate the change in setpoint. In contrast, if the agent were trained in residential building $A$ with $T_{\text{set}} = 22^oC$, then tested in office building $B$ with $T_{\text{set}} = 22^oC$ it must perform dynamics generalisation to accommodate the change in building physics.

Methods that attempt either type of generalisation fall into two categories: 1) those that \textit{increase the similarity} between the train and test contexts, and 2) those that \textit{handle differences} between the train and test contexts \citep{kirk2023survey}. Approach 1) hopes that if $C_{\text{train}}$ is large enough it will subsume $C_{\text{test}}$, that is, the test context may appear as just another variation in the train context. Approach 2) anticipates differences between $C_{\text{train}}$ and $C_{\text{test}}$ and creates solutions for handling them explicitly. We shall use these categories to frame the remaining discussion, starting first with methods for task generalisation in \colouredS \ref{background: task generalisation} followed by methods for dynamics generalisation in \colouredS \ref{background: dynamics generalisation}.

\subsection{Task Generalisation}\label{background: task generalisation}
\textbf{Increasing the similarity between train and test contexts.} $\;$ These methods attempt to train the agent on as many tasks as possible in the hope that one may prove sufficiently similar to the test task. This is most easily operationalised in the \textit{goal-conditioned} RL setting \citep{kaelbling1993learning, schaul2015} where the agent must learn a policy $\pi(s, g)$ that minimises the number of steps taken to navigate to a goal state $g \in \mathcal{S}$. Rewards are $1$ if $s = g$, and $0$ otherwise. Given a dataset of reward-free trajectories $\mathcal{D} = \{(s_0, a_0, \ldots, a_{T-1}, s_T)^i\}_{i=1}^{|\mathcal{D}|}$, we can label them with rewards under the assumption that the terminal state $s_T$ was the intended goal state $g$, and then use them to update any actor-critic algorithm. This method is called \textit{hindsight experience replay} (HER) \citep{andrychowicz2017hindsight}, and it has the effect of creating a training context of goal-reaching tasks corresponding to all terminal states in the dataset. The wider the distribution of terminal states, the more goal-reaching tasks the agent will be trained to solve. 

A more general formulation is to instead assume that the goal state is \textit{any other} state in the dataset, rather than the terminal state in each trajectory. In this setup, one trains a \textit{universal value function approximator} \citep{schaul2015} (\colouredS \ref{background: multi-step MBRL}) $Q(s,a,g)$ or $V(s,g)$ where the goals are states sampled from $\mathcal{D}$ according to some pre-specified distribution \citep{park2023hiql}. Methods that train $V(s,g)$ are also called \textit{action-free} RL methods \citep{zheng2023semi}, which are particularly helpful for training agents from vidoes that lack action data \citep{ghosh2023reinforcement, baker2022video, schmeckpeper2020reinforcement}. 

The primary drawback of these methods is they provide no principled mechanism for improving performance on non-goal-reaching (i.e. dense reward) tasks. For that we look to methods that handle the difference between the train and test contexts.

\textbf{Handling the differences between train and test contexts.} $\;$  \textit{Skill discovery methods} are one way of performing generalisation to new tasks specified by dense reward functions \citep{da2012learning, oh2017zero, sohn2018hierarchical}. Here, agents learn a set of \textit{skills} which are combined by a learned hierarchical controller \citep{sutton1999between, mcgovern2001automatic, barto2003recent, silver2012compositional, stolle2002learning} to solve the test task. In \textit{supervised} skill discovery, the agent is provided a set of training reward functions that correspond to human-legible skills, like walking, standing or running for robotic locomotion. In \textit{unsupervised} skill discovery \citep{jaderberg2016reinforcement, gregor2016variational, machado2017laplacian, eysenbach2018}, the agent uses intrinsic rewards \citep{chentanez2004intrinsically, singh2010intrinsically, barto2013intrinsic} to learn a distinct and separable set of skills. 

\textit{In-context RL} methods train sequence models over \textit{contexts} of historical states, actions and rewards \citep{duan2016rl, lu2024structured, grigsby2023amago, laskin2022context}. The key idea is to condition the policy on these contexts, which act as examples of how one might complete the task. The canonical method, RL$^2$ \citep{duan2016rl}, conditions an RNN on a single trajectory that covers several episodes, which it uses as examples to improve its own policy. This idea has received enthusiastic recent support because language models have shown an impressive ability to perform in-context learning \citep{brown2020language, dong2022survey}. The primary limitation of \textit{skill discovery} and \textit{in-context RL} is that they usually require many episodes of experience to perform task generalisation, which violates the one-episode experience constraint of zero-shot generalisation.
\begin{center}
    * * *
\end{center}
Touati et al. \citep{touati2022does} showed that the multi-step model-based RL methods discussed in \colouredS \ref{background: multi-step MBRL} can be trained to zero-shot generalise to \textit{both} goal-reaching tasks and dense reward tasks, thus unifying the approaches discussed in this section. Recall USFs (Equation \ref{equation: universal SFs}) and their associated task-conditioned policy (Equation \ref{equation: universal SFs policy}):
\begin{equation}\label{equation: universal SFs reprise}
    \begin{cases}
    \psi(s_0, a_0, z) =  \mathbb{E}_{\pi_z, P} \left[ \sum_{t=0}^T \gamma^t \varphi(s_{t+1}) \; | \; s_0, a_0 \right]  \quad \; \forall \; (s_0, a_0) \in \mathcal{S} \times \mathcal{A}, z \in \mathbb{R}^d \\
    \pi_{z}(s) := \argmax_a \psi(s,a,z)^\top z \qquad \qquad \qquad \qquad  \forall \; (s,a) \in \mathcal{S} \times \mathcal{A}, z \in \mathbb{R}^d,
    \end{cases}
\end{equation}
and the decomposition of the reward function as $R(s) = \varphi(s)^\top z$. Touati et al. train both models with a dataset of reward-free transitions $\mathcal{D} = \{(s_t, a_t, s_{t+1})^i\}_{i=1}^{|\mathcal{D}|}$ and tasks $z$ sampled from a prior distribution over tasks $\mathcal{Z}$. If Equation \ref{equation: universal SFs reprise} holds, then the policy will be approximately optimal for all tasks in the linear span of $z$, with $z$ permitting both goal-reaching and dense reward tasks. The breadth of training tasks can be increased by increasing the representation dimension $d$. At test time, a small number of reward-labelled states $\mathcal{D}_{\text{labelled}} = \{(s, R_{\text{test}}(s))\}$ is used to infer the test task vector $z_{\text{test}}$ by linear regression: $z_{\text{test}} = \arg\min_z \mathbb{E}_{s \sim \mathcal{D}_{\text{labelled}}} \left[ (R_{\text{test}}(s) - \varphi(s)^\top z)^2 \right]$, which is then used to condition the policy. A similar protocol is used for inferring the task vector with FB representations. Touati et al. \citep{touati2022does} pre-train FB representations and USFs on large, diverse offline datasets of transitions collected by exploratory algorithms on the DeepMind Control Suite. Remarkably, they show that FB representations can return zero-shot policies for unseen tasks that are $\approx$ 80\% as performant as state-of-the-art RL methods trained to convergence on the task. As a result, these methods have recently been called \textit{foundation policies} (FPs) \citep{park2024foundation}.

\subsection{Dynamics Generalisation}\label{background: dynamics generalisation}
\textbf{Increasing the similarity between train and test contexts.} $\;$ The easiest way to increase the similarity in dynamics between the train and test contexts is with \textit{domain randomisation} (DR) \citep{tobin2017domain, sadeghi2016cad2rl, peng2018sim}. Here, we assume access to a parameterised simulation of the downstream task. For example, a robotic arm  with free parameters $\theta \in \Theta$ defining the arm's mass, length, friction coefficients etc. Each episode the agent is trained on a new simulator instance with $\theta \sim \text{Unif}(\Theta)$. This incentivises a policy that is robust to changes in the dynamics defined by $\Theta$.  Another option is to sample parameters from a distribution that is biased toward those that improve some downstream performance metric like robustness, sample efficiency, or reward. This approach is called \textit{unsupervised environment design} \citep{wang2019paired, dennis2020emergent, jiang2021replay, jiang2021prioritized, parker2022evolving}, and was used to train a humanoid hand entirely in simulation to solves Rubik's cubes with unseen initial configurations \citep{akkaya2019solving}. The primary drawback of these approaches is that they require access to a simulator that already closely approximates physical reality. For many real-world problems no simulator exists, and so such methods are unusable.

A closely related idea is \textit{data augmentation}. Here, instead of randomising the parameters of the simulation, it is a dataset of experience that is randomised\footnote{In practice, the dataset can itself be simulated, or collected from the test context as in Offline RL (\colouredS\ref{background: offline RL})}. Given a dataset of transitions $\mathcal{D} = \{(s, a, r, s^\prime)^i\}_{i=1}^{|\mathcal{D}|}$, the most general approach is to create an operator $\Lambda$ that augments each transition:
\begin{equation}\label{equation: data augmentation}
\Lambda: (s, a, r, s^\prime) \mapsto (\lambda \odot s, a, r, \lambda \odot s^\prime)
\end{equation}
with $\lambda \sim \text{Unif}([\lambda_{\text{low}}, \lambda_{\text{high}}]^{|\mathcal{S}|})$ \citep{laskin2020reinforcement}. Other works have proposed different state-based augmentation operators \citep{raileanu2020automatic, ball2021augmented}, or proposed analogous augmentations for pixel-based observations \citep{yarats2021mastering, hansen2021stabilizing, hansen2021generalization}. The difficulty here comes in selecting distribution bounds $\lambda_{\text{low}}$ and $\lambda_{\text{high}}$. They must be large enough to create qualitatively different dynamics, but not too large so that these dynamics are no longer physically realisable. No works have yet established a general method for setting these bounds, with each using task-dependent heuristics.

Another solution is to collect a large, diverse dataset of experience covering many contexts, and \textit{meta-learning} a policy across it \citep{duan2016rl, chen2018hardware, finn2017model, nagabandi2018learning, nagabandi2018deep}. This is the approach taken by GATO \citep{reed2022generalist}, a single, monolithic policy capable of solving robotics tasks, playing Atari and being a conversational chatbot. GATO is a decision transformer (\colouredS \ref{background: policy constraints}) trained across a dataset of billions of tokens from hundreds of environments. An interesting thesis of this line of work is that learnings from one environment can help an agent in a seemingly unrelated environment. Examples include pre-training on Wikipedia articles to improve performance on Atari \citep{reid2022can}, or pre-training on open-source code to improve common sense reasoning \citep{madaan2022language}. Of course, the primary drawback of these approaches is the difficulty of curating datasets that are both large and high-quality. 

\textbf{Handling the differences between train and test contexts.} $\;$ The typical way to manage differences in train and test dynamics is to condition the policy on some representation of the context during training, then infer the unseen context quickly during testing \citep{lee2020context}. A successful first-effort was Yu et al.'s UP-OSI \citep{yu2017preparing}. Here, a policy is conditioned on a context provided by an encoder network which is trained to infer the context from a history of transitions. As long as the context can be inferred from a history length that is shorter than the episode length, UP-OSI can adapt to unseen dynamics within the first deployment episode. Rapid Motor Adaptation (RMA) \citep{kumar2021rma} combines the same idea with domain randomisation to cover a broader range of robotics test contexts, including broken joints and legs. 

These methods infer the dynamics context once from a fixed set of transitions. Another approach is to continuously infer the dynamics context from the agent's recent history with \textit{memory models} \citep{bakker2001reinforcement, ramani2019short, meng2021memory, hernandez2000hierarchical, schmidhuber2019reinforcement, wierstra2007policy, zhang2016learning, beck2020amrl}. Provided a history of observation-action pairs $\tau_{k:n} = x_{k}, \ldots, x_n$, with $x = \epsilon(o, a)$ being some encoding of an observation-action pair, a \textit{memory model} is some function $f$ that outputs a new hidden state $h_n$ given a past hidden state $h_{k-1}$ and trajectory $\tau_{k:n}$:
\begin{equation}\label{equation: generalised memory model background}
    h_n = f(\tau_{k:n}, h_{k-1}).
\end{equation}
During training, $k = n - L$ in general, where $L$ is a hyperparameter called the \textit{context length}. During inference, $k = n$ because the asymptotic time complexity is $\mathcal{O}(1)$ which is helpful for fast data collection, or high-frequency motor control. Equipped with memory, and trained over a distribution of dynamics contexts, agents can generalise well to unseen dynamics functions in the single-task setting \citep{ni2021recurrent, zhao2019investigating}.  

Another line of work use hard-coded adaption techniques that do not need to infer a representation of the context. Seo et al.'s TW-MCL learns an ensemble of one-step dynamics models where each model predicts the next state for a fixed dynamics context \citep{seo2020trajectory}. At test-time, they select the dynamics model with lowest prediction error on the observed transitions and use it for MPC planning. A similar approach is employed by MOLe \citep{nagabandi2018deep}, but here they are continually updating their ensemble of dynamics models to produce aggregate predictions that are accurate for the test context.

\subsection{Without Prior Data}

\noindent The final and most demanding setting we shall discuss is one where the agent must solve a real-world task \textit{without any prior data}. Here, the agent must do something akin to traditional system identification \citep{aastrom1971system, ljung1998system} \textit{i.e.} collect a small, but expressive dataset from which to quickly learn a model or policy. The most impressive work in this setting is PILCO (discussed in \colouredS \ref{background: one-step MBRL}), which models the system dynamics using Gaussian Processes and updates its policy with analytical gradients \citep{deisenroth2011pilco}. Remarkably, PILCO solved the cartpole swing-up task with 20 seconds of real-world experience. Other works include Kendall et al.'s model-free system (\colouredS \ref{background: model-free RL}) which learned lane-following (autonomous driving), with only 30 minutes of self-collected experience \citep{kendall2019learning}; Lazic et al.'s agent that reduced energy consumption in data-centres by $\approx$ 10\% using three hours of agent-collected data \citep{lazic2018data}; and a series of works that learn robotic walking policies with 20 minutes \citep{smith2022walk}, 90 minutes \citep{haarnoja2018learning}, and 120 minutes \citep{ha2020learning} of experience, by configuring their critics with layer normalisation \citep{ba2016layer}, dropout \citep{srivastava2014dropout}, and high update-to-data ratios \citep{ball2023efficient}. 
 
These results push the limits of data-efficiency in modern RL. However, we believe they signal the relative simplicity of the tasks, rather than the broader utility of the methods. The cartpole swing-up task has small state and action spaces, and requires motor control that is straightforward relative to other robotics tasks. Lane-following requires only that the controller keeps the car straight, paired with gentle pressure on the accelerator, which is simpler than, say, navigating a complex intersection. And controlling data-centre cooling equipment requires only that the agent cool as little as possible whilst satisfying generous temperature bounds. Nonetheless, these findings beg the question: are there any other real-world settings where data is scarce, and where relatively straightforward control could prove beneficial? In Chapter \ref{chapter: no prior data}, we argue that emission-efficient building control is one such domain, and propose new methods that tackle it without prior data.

\chapter{From Low Quality Data}\label{chapter: from low quality data}

We begin our search for methods that solve real-world problems \textit{zero-shot} by addressing the \textit{data quality constraint} (\colouredS \ref{chapter1: path forward}). In \colouredS \ref{background: multi-step MBRL} we introduced FB representations, and USFs, and in \colouredS \ref{background: task generalisation} we discussed how these methods can be used to perform zero-shot RL, as defined in \citep{touati2022does}. We relayed the surprising empirical finding that these methods can return policies for arbitrary tasks, for which they received no reward supervision during training, that are $\approx$ 80\% as performant as state-of-the-art single task RL methods trained solely on each task. 

Unfortunately, these results are predicated on access to a large heterogeneous dataset of transitions for pre-training. In theory, such datasets could be curated by highly-exploratory agents during an upfront data collection phase \citep{jaderberg2016reinforcement, burda2018, eysenbach2018, pathak2017curiosity, pathak2019self, jin2020, liu2021}. However, in practice, deploying such agents in real systems can be time-consuming, costly or dangerous. To avoid these downsides, it would be convenient to skip the data collection phase and pre-train on historical datasets. Whilst these are common in the real world, they are usually produced by controllers that are not optimising for data heterogeneity \citep{dulac2019challenges} (\colouredS \ref{background: offline RL}), making them smaller and less diverse than current zero-shot RL methods expect.     

\begin{figure*}[t]
    \centering
    \includegraphics[width=\textwidth]{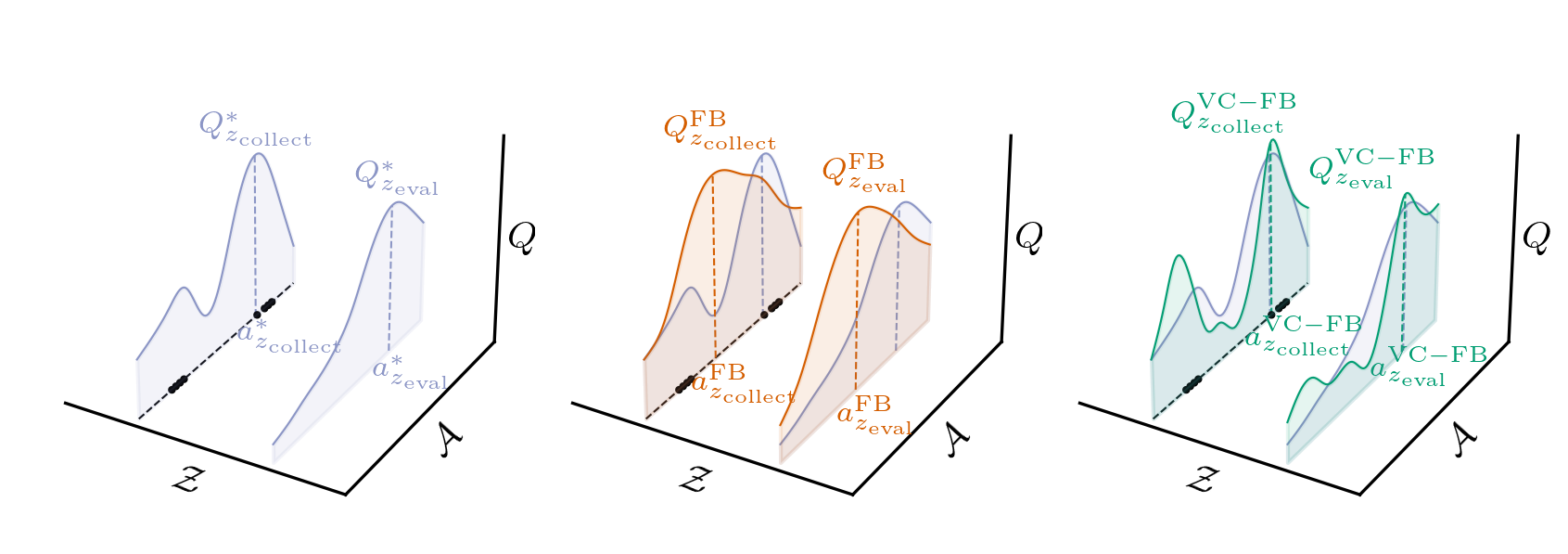}
    \caption{\textbf{\textit{Conservative} zero-shot RL.}. (\textit{Left}) Zero-shot RL methods must train on a dataset collected by a behaviour policy optimising against task $z_{\mathrm{collect}}$, yet generalise to new tasks $z_{\mathrm{eval}}$. Both tasks have associated optimal value functions $Q_{z_{\mathrm{collect}}}^*$ and $Q_{z_{\mathrm{eval}}}^*$ for a given marginal state. (\textit{Middle}) Existing methods, in this case forward-backward representations ({\color{fb-orange}{FB}}), overestimate the value of actions not in the dataset for all tasks. (\textit{Right}) Value-conservative forward-backward representations ({\color{vc-fb-green}{VC-FB}}) suppress the value of actions not in the dataset for all tasks. Black dots represent state-action samples present in the dataset.}
    \label{fig: chapter3/overview}
\end{figure*}

Can we still perform zero-shot RL using these datasets? This is the primary question this Chapter seeks to answer, and one we address in four parts. First, we investigate the performance of existing methods when trained on such datasets, finding their performance suffers because of out-of-distribution state-action value overestimation, a well-observed phenomenon in single-task offline RL. Second, we develop ideas from \textit{conservatism} in single-task offline RL for use in the zero-shot RL setting, introducing a straightforward regularizer of OOD \textit{values} or \textit{measures} that can be used by any zero-shot RL algorithm (Figure \ref{fig: chapter3/overview}). Third, we conduct experiments across varied domains, tasks and datasets, showing our \textit{conservative} zero-shot RL proposals outperform their non-conservative counterparts, and surpass the performance of methods that get to see the task in advance. Finally, we establish that our proposals do not hinder performance on large heterogeneous datasets, meaning adopting them presents little downside.

\section{Preliminaries}\label{chapter3: preliminaries}
In this section we re-introduce material that was covered in Chapter \ref{chapter: background} for the readers benefit, and to ease the exposition.

\textbf{Markov decision processes} $\;$ A \textit{reward-free} Markov Decision Process (MDP) is defined by $\mathcal{M} = \{\mathcal{S}, \mathcal{A}, P, \gamma \}$ where $\mathcal{S}$ is the state space, $\mathcal{A}$ is the action space, $P: \mathcal{S} \times \mathcal{A} \rightarrow \Delta(\mathcal{S})$ is the transition function, where $\Delta(X)$ denotes the set of possible distributions over $X$, and $\gamma \in [0, 1)$ is the discount factor \citep{Sutton2018}. Given $(s_0, a_0) \in \mathcal{S} \times \mathcal{A}$ and a policy $\pi: \mathcal{S} \rightarrow \Delta(\mathcal{A})$, we denote $\Pr(\cdot | s_0, a_0, \pi)$ and $\mathbb{E}[\cdot | s_0, a_0, \pi]$ the probabilities and expectations under state-action sequences $(s_t, a_t)_{t\geq0}$ starting at $(s_0, a_0)$ and following policy $\pi$, with $s_t \sim \mathcal{P}(\cdot |s_{t-1}, a_{t-1})$ and $a_t \sim \pi(\cdot |s_t)$. Given a reward function $R: \mathcal{S} \to \mathbb{R}_{\geq 0}$, the $Q$ function of $\pi$ for $r$ is $Q_r^\pi := \sum_{t\geq0} \gamma^t \mathbb{E}[r(s_{t+1})|s_0, a_0, \pi]$.

% \footnote{In more general formulations, rewards can be negative and dependent on ($s_t, a_t, s_{t+1}$) triplets, but we consider this special case here.}

\textbf{Problem setting (zero-shot RL) \citep{touati2022does}} $\;$ For pre-training, the agent has access to a static offline dataset of reward-free transitions $\mathcal{D} = \{(s_i, a_i, s_{i+1})\}_{i=1}^{|\mathcal{D}|}$ generated by an unknown behaviour policy, and cannot interact with the environment. At test time, a reward function $R_{\text{eval}}$ specifying a \textit{task} is revealed and the agent must return a policy for the task without any further planning or learning. Ideally, the policy should maximise the expected discounted return on the task $\mathbb{E}[\sum_{t\geq0} \gamma^t r_{\text{eval}}(s_{t+1}) | s_0, a_0, \pi]$. The reward function is specified either via a small dataset of reward-labelled states $\mathcal{D}_{\text{labelled}} = \{(s_i, R_{\text{eval}}(s_i))\}_{i=1}^{k}$ with $k \leq 10,000$ or as an explicit function $s \mapsto R_{\text{eval}}(s)$ (like $1$ at a goal state and $0$ elsewhere). 

State-of-the-art zero-shot RL methods leverage either successor measures \citep{blier2021learning} or successor features \citep{barreto2017}, with the former instantiated by \textit{forward backward representations} \citep{touati2021learning} and the latter by \textit{universal successor features} \citep{borsa2018}. The remainder of this section introduces these ideas.

\textbf{Successor measures} $\;$ The \textit{successor measure} $M^{\pi}(s_0, a_0, \cdot)$ over $\mathcal{S}$ is the cumulative discounted time spent in each future state $s_{t+1}$ after starting in state $s_0$, taking action $a_0$, and following policy $\pi$ thereafter:
\begin{equation}
    M^\pi(s_0, a_0, X) := \textstyle{\sum_{t\geq0}} \gamma^t \Pr (s_{t+1} \in X | s_0, a_0, \pi) \; \forall \; X \subset \mathcal{S}.
\end{equation}
The $Q$ function of policy $\pi$ for task $r$ is the integral of $r$ with respect to $M^\pi$:
\begin{equation}
    \label{eq:Q_from_M}
    Q^\pi_r(s_0,a_0):=\textstyle\int_{s_+\in\mathcal{S}}r(s_+)M^\pi(s_0,a_0, s_+).
\end{equation}

\textbf{The forward-backward framework} $\;$ FB representations \citep{touati2021learning} approximate the successor measures of near-optimal policies for any task. Let $\rho$ be an arbitrary state distribution, and $\mathbb{R}^d$ be a representation space. FB representations are composed of a \textit{forward} model $F:\mathcal{S}\times\mathcal{A}\times\mathbb{R}^d\to\mathbb{R}^d$, a \textit{backward} model $B:\mathcal{S}\to \mathbb{R}^d$, and set of polices $(\pi_z)_{z \in \mathbb{R}^d}$. They are trained such that
\begin{equation}
    \label{eq:M_approximation}
    M^{\pi_z}(s_0,a_0, X)\approx \int_X F(s_0,a_0,z)^\top B(s)\rho(\text{d}s) \; \forall \; s_0 \in \mathcal{S}, a_0 \in \mathcal{A}, X \subset \mathcal{S}, z \in \mathbb{R}^d,
\end{equation}
and
\begin{equation} \label{equation: policy_from_z}
    \pi_{z}(s) \approx \argmax_a F(s,a,z)^\top z \; \; \forall \; \; (s,a) \in \mathcal{S} \times \mathcal{A}, z \in \mathbb{R}^d.
\end{equation}
Intuitively, Equation \ref{eq:M_approximation} says that the approximated successor measure under $\pi_z$ from $(s_0,a_0)$ to $s$ is high if their respective forward and backward embeddings are similar i.e. have large dot product. By comparing Equation \ref{eq:Q_from_M} and Equation \ref{eq:M_approximation}, we see that an FB representation can be used to approximate the $Q$ function of $\pi_z$ with respect to any reward function $r$ as:
\begin{equation}
\begin{aligned} \label{eq:Q_from_FB}
    Q^{\pi_z}_r(s_0,a_0) &\approx \textstyle\int_{s\in\mathcal{S}}r(s)F(s_0,a_0,z)^\top B(s)\rho(\text{d}s) \\
    &= F(s_0,a_0,z)^\top\mathbb{E}_{s\sim\rho}[R(s)B(s)\ ].
\end{aligned}
\end{equation}

Training of $F$ and $B$ is done with TD learning \citep{samuel1959, sutton1988} using transition data sampled from $\mathcal{D}$:
\begin{multline}{\label{equation: fb loss}}
\mathcal{L}_{\text{FB}} = \mathbb{E}_{(s_t, a_t, s_{t+1}, s_+) \sim \mathcal{D}, z \sim \mathcal{Z}} [(F(s_t, a_t, z)^\top B(s_+) -  \gamma \bar{F}(s_{t+1}, \pi_z(s_{t+1}), z)^\top \bar{B}(s_+))^2 \\
- 2 F(s_t, a_t, z)^\top B(s_{t+1})],
\end{multline}
where $s_+$ is sampled independently of $(s_t,a_t,s_{t+1})$, $\bar{F}$ and $\bar{B}$ are lagging target networks, and $\mathcal{Z}$ is a task sampling distribution. The policy is trained in an actor-critic formulation \citep{lillicrap2015}. See \cite{touati2021learning} for a full derivation of the TD update, and Appendix \ref{appendix: z sampling implementation} for practical implementation details including the specific choice of task sampling distribution $\mathcal{Z}$.

By relating Equations \ref{equation: policy_from_z} and \ref{eq:Q_from_FB}, we find $z=\mathbb{E}_{s\sim\rho}[R(s)B(s)]$ for some reward function $r$. At test time, we can use this property to perform zero-shot RL. Using $\mathcal{D}_{\text{labelled}}$, we estimate the task as $z_{\text{eval}} \approx \mathbb{E}_{s \sim \mathcal{D_{\text{labelled}}}}[R_{\text{eval}}(s)B(s)]$ and pass it as an argument to $\pi_z$. If $z_{\text{eval}}$ lies within the task sampling distribution $\mathcal{Z}$ used during pre-training, then $\pi_z(s)\approx \argmax_{a}Q^{\pi_z}_{R_{\text{eval}}}(s,a)$, and hence this policy is approximately optimal for $r_{\text{eval}}$.

\textbf{(Universal) successor features} $\;$ \textit{Successor features} assume access to a basic feature map $\varphi: \mathcal{S} \mapsto \mathbb{R}^d$ that embeds states into a representation space, and are defined as the expected discounted sum of future features $\psi^\pi(s_0, a_0) := \mathbb{E}[\sum_{t \geq 0} \gamma^t \varphi(s_{t+1}) | s_0, a_0, \pi]$ \citep{barreto2017}. They are made \textit{universal} by conditioning their predictions on a family of policies $\pi_z$
\begin{equation}
\psi(s_0, a_0, z) = \mathbb{E}\left[\sum_{t \geq 0} \gamma^t \varphi(s_{t+1}) | s_0, a_0, \pi_z\right] \; \; \forall \; s_0 \in \mathcal{S}, a_0 \in \mathcal{A}, z \in \mathbb{R}^d,
\end{equation}
with
\begin{equation}
\pi_z(s) \approx \argmax_a \psi(s,a,z)^\top z, \; \forall \; (s_0, a_0) \in \mathcal{S} \times \mathcal{A}, z \in \mathbb{R}^d. 
\end{equation}
Like FB, USFs are trained using TD learning on
\begin{equation} \label{equation: usf loss 1}
    \mathcal{L}_{\text{SF}} = \mathbb{E}_{(s_t, a_t, s_{t+1}) \sim \mathcal{D}, z \sim \mathcal{Z}} [( \psi(s_t, a_t, z)^\top z - \varphi(s_{t+1})^\top z - \gamma \bar{\psi}(s_{t+1}, \pi_z(s_{t+1}), z)^\top z )^2],
\end{equation}

where $\bar{\psi}$ is a lagging target network, and $\mathcal{Z}$ is the same $z$ sampling distribution used for FB. We refer the reader to \cite{borsa2018} for a derivation of the TD update and full learning procedure. Test time policy inference is performed similarly to FB. Using $\mathcal{D}_{\text{labelled}}$, the task is inferred by performing a linear regression of $R_{\text{eval}}$ onto the features: $z_{\text{eval}} := \argminA_z \mathbb{E}_{s \sim \mathcal{D}_{\text{labelled}}}[(R_{\text{eval}}(s) - \varphi(s)^\top z)^2]$ before it is passed as an argument to the policy.

\section{Zero-Shot RL from Low Quality Data}\label{chapter3: conservative FB}
\begin{figure*}[t]
    \centering
    \includegraphics[width=0.8\textwidth]{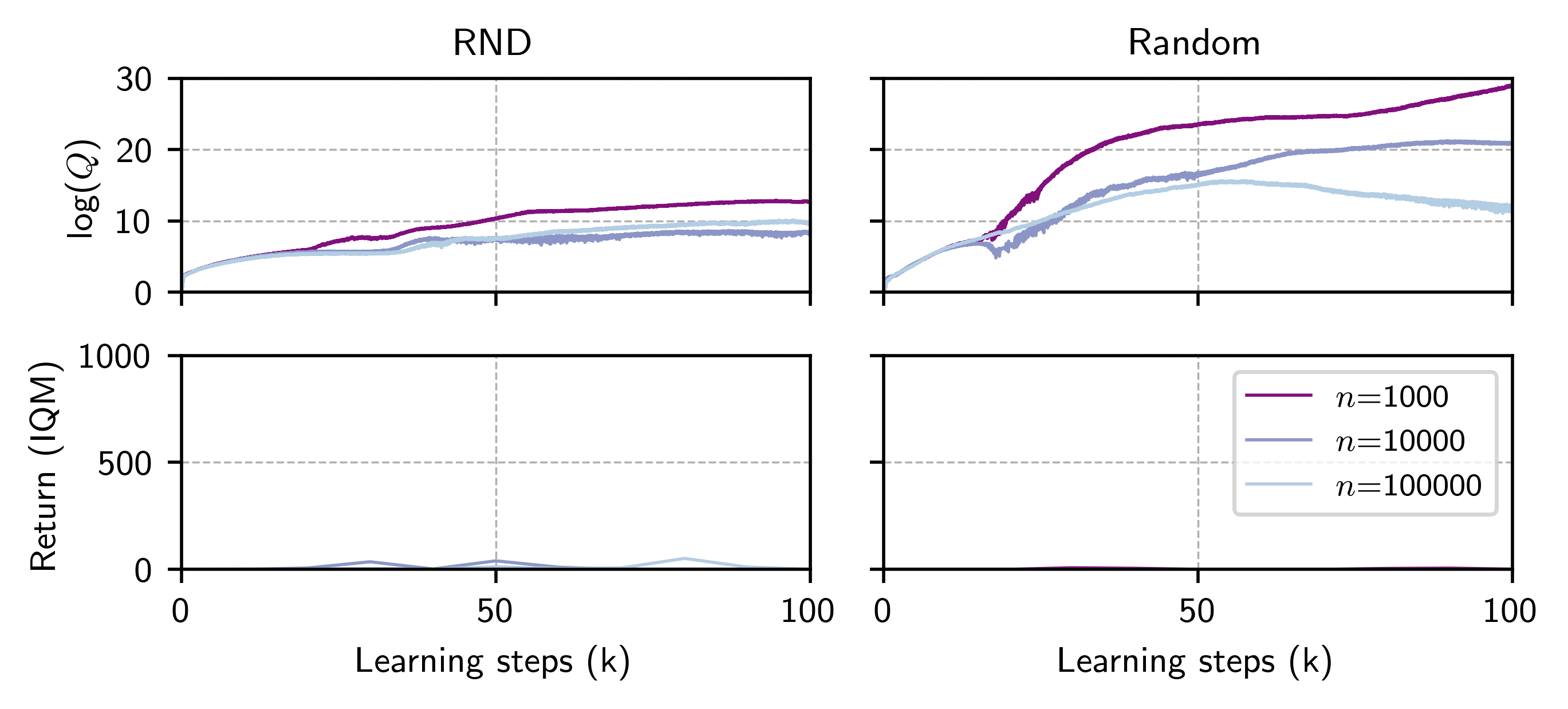}
    \caption{\textbf{FB value overestimation with respect to dataset size $n$ and quality.} Log $Q$ values and IQM of rollout performance on all Point-mass Maze tasks for datasets \textsc{Rnd} and \textsc{Random}. $Q$ values predicted during training increase as both the size and ``quality" of the dataset decrease. This contradicts the low return of all resultant policies (note: a return of 1000 is the maximum achievable for this task). Informally, we say the \textsc{Rnd} dataset is ``high" quality, and the \textsc{Random} dataset is ``low" quality--see Appendix \ref{appendix: exorl datasets} for more details.}
    \label{fig: chapter3/overestimate}
\end{figure*}

In this section we introduce methods for improving the performance of zero-shot RL methods on low quality datasets. In \colouredS \ref{chapter3: existing method failure mode}, we explore the failure mode of existing methods on such datasets. Then, in \colouredS \ref{chapter3: conservative regularisation}, we propose straightforward amendments to these methods that address the failure mode. Finally, in \colouredS \ref{chapter3: didactic}, we illustrate the usefulness of our proposals with a controlled example. We develop our methods within the FB framework because of its superior empirical performance \citep{touati2022does}, but our proposals are also compatible with USF. We push their derivation to Appendix \ref{appendix: conservative successor features implementation} for brevity.

\subsection{Failure Mode of Existing Methods}\label{chapter3: existing method failure mode}
To investigate the failure mode of existing methods we examine the FB loss (Equation \ref{equation: fb loss}) more closely. The TD target includes an action produced by the current policy $a_{t+1} \sim \pi_z(s_{t+1})$. Equation \ref{equation: policy_from_z} shows this is the policy's current best estimate of the $Q$-maximising action in state $s$ for task $z$. For finite datasets, this maximisation does not constrain the policy to actions observed in the dataset,
and so it can become biased towards out-of-distribution (OOD) actions thought to be of high value. In such instances $F$ and $B$ are updated towards targets for which the dataset provides no support. This \textit{distribution shift} is a well-observed phenomenon in single-task offline RL \citep{kumar19, levine2020, kumar2020}, and is exacerbated by small, low-diversity datasets as we explore in Figure \ref{fig: chapter3/overestimate}.

\subsection{Mitigating the Distribution Shift}\label{chapter3: conservative regularisation}
In the single-task setting, the distribution shift is addressed by applying constraints to either the policy, value function or model (\colouredS \ref{background: offline RL}). Here we re-purpose single-task value function and model regularisation for use in the zero-shot RL setting. To avoid further complicating zero-shot RL methods, we only consider regularisation techniques that do not introduce new parametric functions. We discuss the implications of this decision in \colouredS \ref{chapter3: discussion}.

Conservative $Q$-learning (CQL) \citep{kumar19, kumar2020} regularises the $Q$ function by querying OOD state-action pairs and suppressing their value. This is achieved by adding {\color{bupu@purple}{new term}} to the usual $Q$ loss function
% This means minimising the $Q$ values for OOD actions and maximising $Q$ values for in- distribution actions
\begin{equation}\label{equation: CQL update}
    \mathcal{L}_{\text{CQL}} = {\color{bupu@purple}{\alpha \cdot \mathbb{E}_{s \sim \mathcal{D}, a \sim \mu(a|s)} [Q(s,a)] - \mathbb{E}_{(s, a) \sim \mathcal{D}}[Q(s, a)]}}
    {\color{bupu@purple}{- \mathcal{H}(\mu)}} + \mathcal{L}_{\text{Q}},
\end{equation}
where $\alpha$ is a scaling parameter, $\mu(a|s)$ is a policy distribution selected to find the maximum value of the current $Q$ function iterate, $\mathcal{H}(\mu)$ is the entropy of $\mu$ used for regularisation, and $\mathcal{L}_{\text{Q}}$ is the normal TD loss on $Q$. Equation \ref{equation: CQL update} has the dual effect of minimising the peaks in $Q$ under $\mu$ whilst maximising $Q$ for state-action pairs in the dataset.

We can replicate a similar form of regularisation in the FB framework, substituting $F(s, a, z)^{\top}z$ for $Q$ in Equation \ref{equation: CQL update} and adding the normal FB loss (Equation \ref{equation: fb loss})
\begin{multline}
\label{equation: VC-FB loss}
    \mathcal{L}_{\text{VC-FB}} = {\color{vc-fb-green}{\alpha \cdot ( \mathbb{E}_{s \sim \mathcal{D}, a \sim \mu(a|s), z \sim \mathcal{Z}} [F(s, a, z)^{\top} z]}}
    {\color{vc-fb-green}{- \mathbb{E}_{(s, a) \sim \mathcal{D}, z \sim \mathcal{Z}} [F(s, a, z)^{\top}z] - \mathcal{H}(\mu))}}\\
    + \mathcal{L}_{\text{FB}}.
\end{multline}
The key difference between Equations \ref{equation: CQL update} and \ref{equation: VC-FB loss} is that the former suppresses the value of OOD actions for one task, whereas the latter does so for all task vectors drawn from $\mathcal{Z}$. We call models learnt with this loss \textit{value-conservative forward-backward representations} ({\color{vc-fb-green}{VC-FB}}).

Because FB derives $Q$ functions from successor measures (Equation \ref{eq:Q_from_FB}), and because (by assumption) rewards are non-negative, suppressing the predicted measures for OOD actions provides an alternative route to suppressing their $Q$ values. As we did with VC-FB, we can substitute FB's successor measure approximation $F(s,a,z)^\top B(s_+)$ into Equation \ref{equation: CQL update}, which yields:
\begin{multline}\label{equation: FB-CM loss}
    \mathcal{L}_{\text{MC-FB}} = {\color{mc-fb-blue}{\alpha \cdot  (\mathbb{E}_{s \sim \mathcal{D}, a \sim \mu(a|s), z \sim \mathcal{Z}, s_+ \sim \mathcal{D}} [F(s, a, z)^{\top} B(s_+)]}} \\ 
    {\color{mc-fb-blue}{- {\mathbb{E}_{(s, a) \sim \mathcal{D}, z \sim \mathcal{Z}, s_+ \sim \mathcal{D}} [F(s, a, z)^{\top} B(s_+)] - \mathcal{H}(\mu))}}} + \mathcal{L}_{\text{FB}}.
\end{multline}
Equation \ref{equation: FB-CM loss} has the effect of suppressing the expected visitation count to goal state $s_+$ when taking an OOD action for all task vectors drawn from $\mathcal{Z}$, which says, informally, if we don't know where OOD actions take us in the MDP, we assume they have low probability of taking us to any future states for all tasks. This is analogous to works that regularise model predictions in the single-task offline RL setting \citep{kidambi2020, yu2021combo, rigter2022}. As such, we call this variant a \textit{measure-conservative forward-backward representation} ({\color{mc-fb-blue}{MC-FB}}). Since it is not obvious \textit{a priori} whether the {\color{vc-fb-green}{VC-FB}} or {\color{mc-fb-blue}{MC-FB}} form of conservatism would be more effective in practice, we evaluate both in \colouredS \ref{chapter3: experiments}.

Implementing these proposals requires two new model components: 1) a conservative penalty scaling factor $\alpha$ and 2) a way of obtaining policy distribution $\mu(a|s)$ that maximises the current value or measure iterate. For 1), we observe fixed values of $\alpha$ leading to fragile performance, so dynamically tune it at each learning--see Appendix \ref{appendix: dynamically tuning alpha implementation}. For 2), the choice of maximum entropy regularisation following \cite{kumar2020}'s CQL($\mathcal{H}$) allows $\mu$ to be approximated conveniently with a log-sum exponential across $Q$ values derived from the current policy distribution and a uniform distribution. That this is true is not obvious, so we refer the reader to the detail and derivations in Section 3.2, Appendix A, and Appendix E of \cite{kumar2020}, as well as our adjustments to \cite{kumar2020}'s theory in Appendix \ref{appendix: max value estimator implementation}. Code snippets demonstrating the required changes to a vanilla FB implementation are provided in Appendix \ref{appendix: code snippets}. We emphasise these additions represent only a small increase in the number of lines required to implement existing methods.

\begin{figure*}[t]
    \centering
    \includegraphics[width=0.9\textwidth]{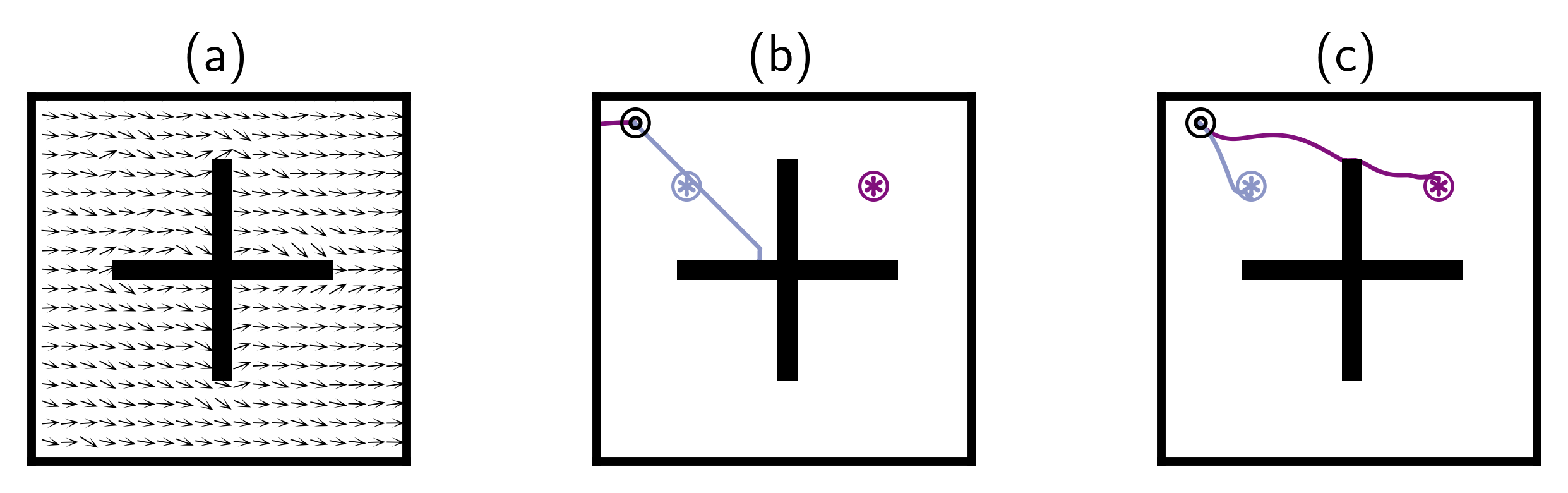}
    \caption{\textbf{Ignoring out-of-distribution actions.} The agents are tasked with learning separate policies for reaching ${\color{bupu@blue} \circledast}$ and ${\color{bupu@purple} \circledast}$. (a) \textsc{Rnd} dataset with all ``left" actions removed; quivers represent the mean action direction in each state bin. (b) Best FB rollout after 1 million learning steps. (c) Best VC-FB performance after 1 million learning steps. FB overestimates the value of OOD actions and cannot complete either task; VC-FB synthesises the requisite information from the dataset and completes both tasks.}
    \label{fig: chapter3/didactic}
\end{figure*}

\subsection{A Didactic Example} \label{chapter3: didactic}
To understand situations in which a conservative zero-shot RL methods may be useful, we introduce a modified version of Point-mass Maze from the ExORL benchmark \citep{yarats2022}. Episodes begin with a point-mass initialised in the upper left of the maze ($\circledcirc$), and the agent is tasked with selecting $x$ and $y$ tilt directions such that the mass is moved towards one of two goal locations (${\color{bupu@blue} \circledast}$ and ${\color{bupu@purple} \circledast}$). The action space is two-dimensional and bounded in $[-1,1 ]$. We take the \textsc{Rnd} dataset and remove all ``left" actions such that $a_x \in [0, 1]$ and $a_y \in [-1, 1]$, creating a dataset that has the necessary information for solving the tasks, but is inexhaustive (Figure \ref{fig: chapter3/didactic} (a)). We train FB and VC-FB on this dataset and plot the highest-reward trajectories--Figure \ref{fig: chapter3/didactic} (b) and (c). FB overestimates the value of OOD actions and cannot complete either task. Conversely, VC-FB synthesises the requisite information from the dataset and completes both tasks.

\section{Experiments} \label{chapter3: experiments}
In this section we perform an empirical study to evaluate our proposals. We seek answers to four questions: \textbf{(Q1)} Can our proposals from \colouredS \ref{chapter3: conservative FB} improve FB performance on small and/or low-quality exploratory datasets? \textbf{(Q2)} How does the performance of VC-FB and MC-FB vary with respect to task type and dataset diversity? \textbf{(Q3)} Do we sacrifice performance on full datasets for performance on small and/or low-quality datasets? \textbf{(Q4)} If our pre-training dataset only covers behaviour related to one downstream task (i.e. the dataset distribution is narrow and not exploratory), can our proposals from \colouredS \ref{chapter3: conservative FB} improve FB performance on that task?

\subsection{Setup}
\textbf{Domains.} $\;$  We respond to \textbf{Q1-Q3} using the ExORL benchmark \citep{yarats2022}. ExORL provides datasets collected by unsupervised exploratory algorithms on the DeepMind Control Suite \citep{tassa2018}. We select three of the same domains as \cite{touati2022does}: Walker, Quadruped and Point-mass Maze, but substitute Jaco for Cheetah. This provides two locomotion domains and two goal-reaching domains. Within each domain, we evaluate on all tasks provided by the DeepMind Control Suite for a total of 17 tasks across four domains. Full details are provided in Appendix \ref{appendix: exorl environements}. We respond to \textbf{Q4} using the D4RL benchmark \citep{fu2020}. We select the two MuJoCo \citep{todorov2012} environments from the Open AI gym \citep{brockman2016} that closest resemble those from ExORL: Walker2D and HalfCheetah.

\textbf{Datasets.} $\;$  For \textbf{Q1-Q3} we pre-train on three datasets of varying quality from ExORL. There is no unambiguous metric for quantifying dataset quality, so we use the reported performance of offline TD3 on Maze for each dataset as a proxy\footnote{We note that \citep{schweighofer2021dataset} propose metrics that describe dataset quality as a function of the behaviour policy's \textit{exploration} and \textit{exploitation} w.r.t. one downstream task. However, since we are interested in generalising to \textit{any} downstream task we cannot use these proposals directly, nor can we easily re-purpose them. We acknowledge that our proxy is imperfect, and that more work is required to better understand what dataset quality means in the context of zero-shot RL.}. We choose datasets collected via Random Network Distillation (\textsc{Rnd}) \citep{burda2018}, Diversity is All You Need (\textsc{Diayn}) \citep{eysenbach2018}, and \textsc{Random} policies, where agents trained on \textsc{Rnd} are the most performant, on \textsc{Diayn} are median performers, and on \textsc{Random} are the least performant. As well as selecting for quality, we also select for size by uniformly sub-sampling 100,000 transitions from each dataset. For \textbf{Q4} we choose the ``medium", ``medium-replay", and ``medium-expert" datasets from D4RL, each providing different fractions of random, medium and expert task-directed trajectories. More details on the datasets are provided in Appendix \ref{appendix: exorl datasets} and \ref{appendix: d4rl datasets}.

\subsection{Baselines}
We compare our proposals to baselines from three categories: 1) zero-shot RL methods, 2) goal-conditioned RL (GCRL) methods, and 3) single-task offline RL methods. From category 1), we use the state-of-the-art successor measure based method, FB, and the state-of-the-art successor feature based method, SF with features from Laplacian eigenfunctions (SF-LAP) \cite{touati2022does}. From category 2), we use goal-conditioned IQL (GC-IQL) \citep{park2023hiql}, a state-of-the-art GCRL method that, like our proposals, regularises the value function at OOD state-actions. We condition GC-IQL on the goal state on Maze and Jaco, and on the state in $\mathcal{D}_{\text{labelled}}$ with highest reward on Walker and Quadruped in lieu of a well-defined goal state. From category 3), we use CQL and offline TD3 trained on the same datasets relabelled with task rewards. CQL approximates what an algorithm with similar mechanistics can achieve when optimising for one task in a domain rather than all tasks. Offline TD3 exhibits the best aggregate single-task performance on the ExORL benchmark, so it should be indicative of the maximum performance we could expect to extract from a dataset. Full implementation details for all algorithms are provided in Appendix \ref{appendix: implementations}. Appendix \ref{appendix: computational resources} provides a breakdown of the computational resources used in this work.

\subsection{Evaluation Protocol} 
We evaluate the cumulative reward (hereafter called score) achieved by VC-FB, MC-FB and our baselines on each task across five seeds. We report task scores as per the best practice recommendations of \cite{agarwal2021deep}. Concretely, we run each algorithm for 1 million learning steps, evaluating task scores at checkpoints of 20,000 steps. At each checkpoint, we perform 10 rollouts, record the score of each, and find the interquartile mean (IQM). We average across seeds at each checkpoint to create the learning curves reported in Appendix \ref{learning curves appendix}. From each learning curve, we extract task scores from the learning step for which the all-task IQM is maximised across seeds. Results are reported with 95\% confidence intervals obtained via stratified bootstrapping \citep{efron1992}. Aggregation across tasks, domains and datasets is always performed by evaluating the IQM. Full implementation details are provided in Appendix \ref{appendix: FB implementation}.

\subsection{Results}\label{chapter3: results}
\begin{figure*}
    \centering
    \includegraphics[width=0.9\textwidth]{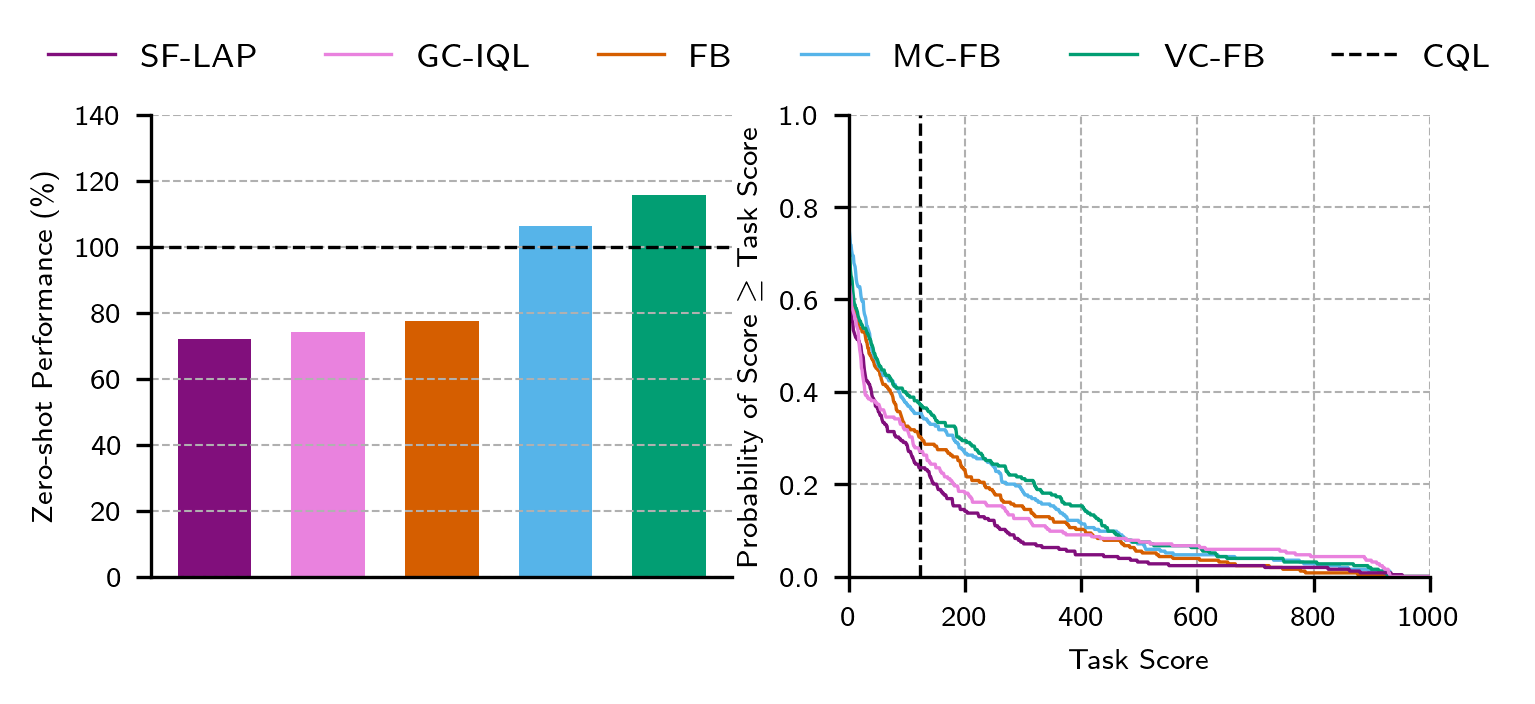}
    \caption{\textbf{Aggregate zero-shot performance on ExORL.} \textit{(Left)} IQM of task scores across datasets and domains, normalised against the performance of CQL, our baseline. \textit{(Right)} Performance profiles showing the distribution of scores across all tasks and domains. Both conservative FB variants stochastically dominate vanilla FB--see \cite{agarwal2021deep} for performance profile exposition. The black dashed line represents the IQM of CQL performance across all datasets, domains, tasks and seeds.}
    \label{fig: chapter3/all-task-performance}
\end{figure*}
\textbf{Q1} $\;$ We report the aggregate performance of our baselines and proposals on ExORL in Figure \ref{fig: chapter3/all-task-performance}. Both MC-FB and VC-FB outperform the zero-shot RL and GCRL baselines, achieving \textbf{150\%} and \textbf{137\%} of FB's IQM performance respectively.
The performance gap between FB and SF-LAP is consistent with the results in \cite{touati2022does}.  MC-FB and VC-FB outperform our single-task baseline in expectation, reaching 111\% and 120\% of CQL's IQM performance respectively \textit{despite not having access to task-specific reward labels and needing to fit policies for all tasks}. This is a surprising result, and to the best of our knowledge, the first time a multi-task offline agent has been shown to outperform a single-task analogue. CQL outperforms offline TD3 in aggregate, so we drop offline TD3 from the core analysis, but report its full results in Appendix \ref{appendix: extended results} alongside all other methods. We note FB achieves 80\% of single-task offline TD3, which roughly aligns with the 85\% performance on the full datasets reported by \cite{touati2022does}.
\begin{figure*}
    \centering
    \includegraphics[width=0.9\textwidth]{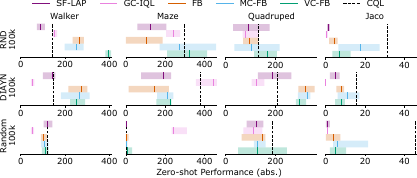}
    \caption{\textbf{Performance by dataset/domain on ExORL.} IQM scores across tasks/seeds with 95\% conf. intervals.}
    \label{fig: chapter3/dataset-domain-performance}
\end{figure*}

\textbf{Q2} $\;$ We decompose the methods' performance with respect to domain and dataset diversity in Figure \ref{fig: chapter3/dataset-domain-performance}. The largest gap in performance between the conservative FB variants and FB is on \textsc{Rnd}. VC-FB and MC-FB reach 2.5$\times$ and 1.8$\times$ of FB performance respectively, and outperform CQL on three of the four domains. On \textsc{Diayn}, the conservative variants outperform all methods and reach 1.3$\times$ CQL's score. On the \textsc{Random} dataset, all methods perform similarly poorly, except for CQL on Jaco, which outperforms all methods. However, in general, these results suggest the \textsc{Random} dataset is not informative enough to extract valuable policies--discussed further in response to Q3. There appears to be little correlation between the type of domain (Appendix \ref{appendix: exorl environements}) and the score achieved by any method. GC-IQL performs strongly on the goal-reaching domains as expected, and particularly well on Maze when trained with \textsc{Diayn} and \textsc{Random} datasets, but worse than all zero-shot methods on the locomotion tasks, irrespective of whether they are conservative or not. This is presumably because the goal-state used to condition the policy (\textit{i.e.} the state with highest reward in $\mathcal{D}_{\text{labelled}}$) is a poor proxy for the true, dense reward function.
\begin{figure*}
    \centering
    \includegraphics[width=0.65\textwidth]{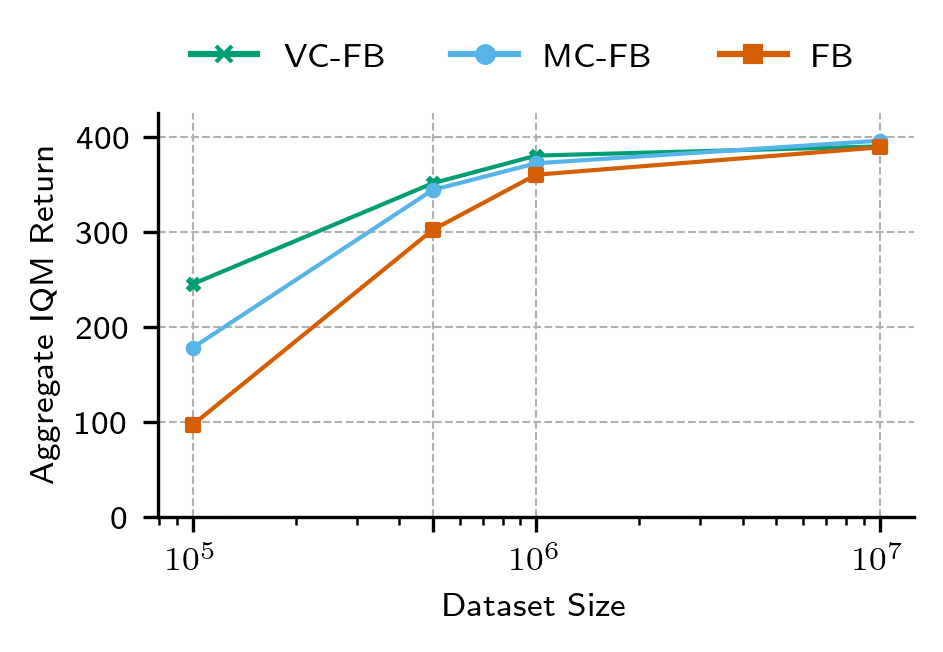}
    \caption{\textbf{Performance by dataset size.} Aggregate IQM scores across all domains and tasks as \textsc{Rnd} size is varied. The performance delta between vanilla FB and the conservative variants increases as dataset size decreases.}
    \label{fig: chapter3/dataset size performance}
\end{figure*}

\textbf{Q3} $\;$ We report the aggregated performance of all FB methods across domains when trained on the full datasets in Table \ref{table: chapter3/full-dataset-summary} (a full breakdown of results in provided in Appendix \ref{appendix: extended results}).  Both conservative FB variants slightly exceed the performance of vanilla FB in expectation. 
\begin{wrapfigure}{l}{0.5\textwidth}
    \centering
    \includegraphics[width=0.5\textwidth]{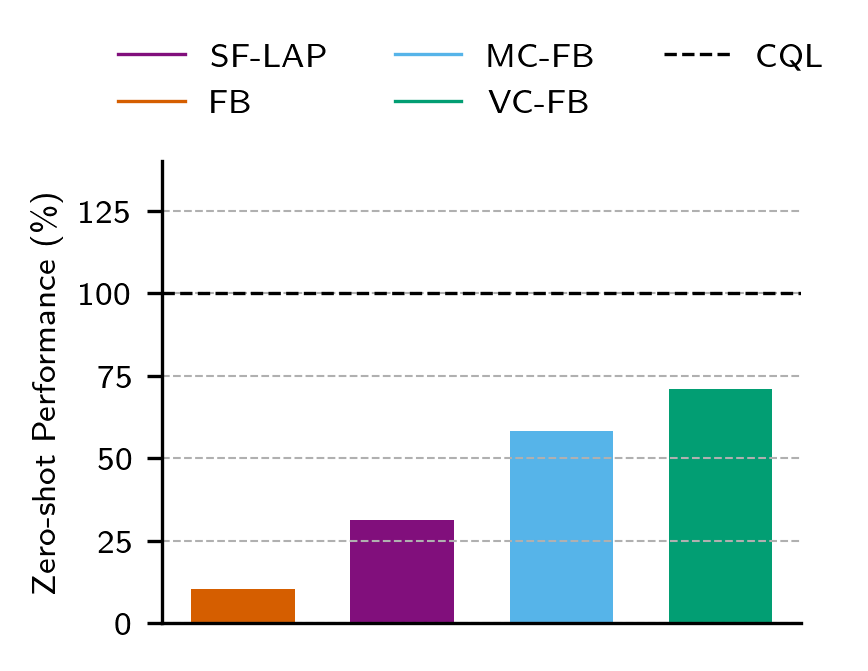}
    \caption{\textbf{Aggregate zero-shot performance on D4RL.} Aggregate IQM scores across all domains and datasets, normalised against the performance of CQL.}
    \label{fig: chapter3/d4rl-performance}
\end{wrapfigure}
The largest relative performance improvement is on the \textsc{Random} dataset--MC-FB performance is 20\% higher than FB, compared to 5\% higher on \textsc{Diayn} and 2\% higher on \textsc{Rnd}. This corroborates the hypothesis that \textsc{Random-100k} was not informative enough to extract valuable policies.

Table \ref{table: chapter3/full-dataset-summary} and Figure \ref{fig: chapter3/dataset size performance} suggest the performance gap between the conservative FB variants and vanilla FB changes as dataset size is varied. We further explore this effect in Figure \ref{fig: chapter3/dataset size performance} where we scale the $\textsc{Rnd}$ dataset size from $10^5$ through $10^7$ and plot aggregate IQM performance of FB, VC-FB and MC-FB across all domains. We find that the performance gap decreases as dataset size increases. This result is to be expected: a larger dataset size for a fixed exploration algorithm means $a_{t+1} \sim \pi_z(s_{t+1})$ in the FB TD update (Equation \ref{equation: fb loss}) is more likely to be in the dataset, the policy is less likely to become biased toward OOD actions, and conservatism is less needed. 

\textbf{Q4.} $\;$ We report the aggregate performance of all zero-shot RL methods and CQL on our D4RL domains in Figure \ref{fig: chapter3/d4rl-performance}. FB fails all domain-dataset tasks, and reaches only 10\% of CQL's aggregate performance. MC-FB and VC-FB improve on FB's considerably (by 5.6 $\times$ and 6.8 $\times$ respectively) but under-perform CQL. SF-LAP outperforms FB, but under-performs VC-FB, MC-FB and CQL. 

\begin{table}
\caption{\textbf{Aggregate performance on full ExORL datasets.} IQM scores aggregated over domains and tasks for all datasets, averaged across three seeds. Both VC-FB and MC-FB maintain the performance of FB; the largest relative performance improvement is on \textsc{Random}.}
\label{table: chapter3/full-dataset-summary}
\centering
\scalebox{0.9}{
\begin{tabular}{@{}llllll}
\toprule
\textbf{Dataset} & \textbf{Domain} & \textbf{Task} & \textbf{\color{fb-orange}{FB}} & \textbf{\color{vc-fb-green}{VC-FB}} & \textbf{\color{mc-fb-blue}{MC-FB}} \\
\midrule
  \textsc{Rnd} & all & all & $389$ & $390$ & $396$ \\
  \textsc{Diayn} & all & all & $269$ & $280$ & $283$ \\
  \textsc{Random} & all & all & $111$ & $131$ & $133$ \\
\midrule
\textsc{All} & all & all & $256$ & $267$ & $\textbf{271}$  \\
\bottomrule
\end{tabular}
}
\end{table}

\section{Discussion and Limitations}\label{chapter3: discussion}
\textbf{Performance discrepancy between conservative variants} $\;$ Why does VC-FB outperform MC-FB on both ExORL and D4RL? To understand, we inspect the regularising effect of both models more closely. VC-FB regularises OOD actions on $F(s, a, z)^\top z$, with $s \sim \mathcal{D}$, and  $z \sim \mathcal{Z}$, whilst MC-FB regularises OOD actions on $F(s, a, z)^\top B(s_+)$, with $(s, s_+) \sim \mathcal{D}$ and $z \sim \mathcal{Z}$. Note the trailing $z$ in VC-FB is replaced with $B(s_+)$ in MC-FB which ties its updates to $\mathcal{D}$ further. We hypothesised that as $|\mathcal{D}|$ reduces, $B(s_+)$ provides poorer task coverage than $z \sim \mathcal{Z}$, hence the comparable performance on full datasets and divergent performance on 100k datasets. 

To test this, we evaluate a third conservative variant called \textit{directed} (\textit{D})VC-FB which replaces all $z \sim \mathcal{Z}$ in VC-FB with $B(s_+)$ such that OOD actions are regularised on $F(s, a, B(s_+))^\top B(s_+)$ with $(s, s_+) \sim \mathcal{D}$. This ties conservative updates entirely to $\mathcal{D}$, and according to our above hypothesis, \textit{D}VC-FB should perform worse than VC-FB and MC-FB on the 100k ExORL datasets. See Appendix \ref{appendix: DVC-FB implementation} for implementation details. We evaluate this variant on all 100k ExORL datasets, domains and tasks and compare with FB, VC-FB and MC-FB in Table \ref{table: chapter/dvcfb-exposition}. See Appendix \ref{appendix: extended results} for a full breakdown. 

We find the aggregate relative performance of each method is as expected i.e. \textit{D}VC-FB $<$ MC-FB $<$ VC-FB. As a consequence we conclude that VC-FB should be preferred for small datasets with no prior knowledge of the dataset or test tasks. Of course, for a specific domain-dataset pair, $B(s_+)$ with $s_+ \sim \mathcal{D}$ may happen to cover the tasks well, and MC-FB may outperform VC-FB. We suspect this was the case for all datasets on the Jaco domain for example. Establishing whether this will be true \textit{a priori} requires either relaxing the restrictions imposed by the zero-shot RL setting, or better understanding of the distribution of tasks in $z$-space and their relationship to pre-training datasets. The latter is important future work. 

\textbf{Avoiding new parametric functions.} $\;$ State-of-the-art zero-shot RL methods are complex, and we wanted to avoid further complicating them with new parametric functions. This limited our solution-space to CQL-style regularisation techniques, but had we relaxed this constraint, other options become available. Methods like AWAC \citep{nair2020awac}, IQL \citep{kostrikov2021offline}, and $X$-QL \citep{garg2023extreme} all require an estimate of the state-value function which is not immediately accessible in the FB or USF frameworks. In theory, we could learn an action-independent USF of the form $V(s,z) = \mathbb{E}[\sum_{t \geq 0} \gamma^t \varphi(s_{t+1}) | s_0, \pi_z] \; \forall \; s_0 \in \mathcal{S}, z \in \mathbb{R}^d$ concurrently to $F$ and $B$ (or $\psi$ for USFs). If learnt with expectile regression, this function could be used to implement IQL and $\mathcal{X}$-QL style regularisation; without expectile regression it could be used to compute the advantage weighting required for AWAC. It's possible that implementing these methods could improve downstream performance and reduce computational overhead at the cost of increased training complexity. We leave this worthwhile investigation for future work. We provide detail of negative results related to downstream finetuning of FB models in Appendix \ref{appendix: negative results} to help inform future research.

\textbf{D4RL Performance.} $\;$ Unlike the ExORL results, VC-FB and MC-FB do not outperform CQL on the D4RL benchmark. We believe these narrower data distributions require a more careful selection of the conservative penalty scaling factor $\alpha$. We explore this further in Appendix \ref{learning curves appendix}, and note this is corroborated by findings in the original CQL paper \citep{kumar2020}. Methods described above, like IQL, have been shown to be more robust than CQL partly because they bypass $\alpha$ tuning. We expect that exploring the integration of these methods may improve D4RL performance.

\begin{table*}[t]
\caption{\textbf{Aggregated performance of conservative variants employing differing $z$ sampling procedures on ExORL.} \textit{D}VC-FB derives all $z$s from the backward model; VC-FB derives all $z$s from $\mathcal{Z}$; and MC-FB combines both. Performance correlates with the degree to which $z \sim \mathcal{Z}$.}
\label{table: chapter/dvcfb-exposition}
\centering
\scalebox{0.9}{
\begin{tabular}{lllllll}
\toprule
\textbf{Dataset} & \textbf{Domain} & \textbf{Task}  & \textbf{\color{fb-orange}{FB}} & \textbf{\textit{D}VC-FB}  & \textbf{\color{mc-fb-blue}{MC-FB}} & \textbf{\color{vc-fb-green}{VC-FB}} \\
\midrule
\textsc{All} (100k) & all & all & $99$ & $108$ & $136$ & $148$  \\
\bottomrule
\end{tabular}
}
\end{table*}

\section{Related Work}\label{section: chapter 3 related work}
\textbf{Zero-shot RL} $\;$ Zero-shot RL methods build upon successor representations \citep{dayan1993}, universal value function approximators \citep{schaul2015}, successor features \citep{barreto2017} and successor measures \citep{blier2021learning}. The state-of-the-art methods instantiate these ideas as either universal successor features (USFs) \citep{borsa2018} or forward-backward (FB) representations \citep{touati2021learning, touati2022does}, with recent work showing the latter can be used to perform a range of imitation learning techniques efficiently \citep{pirotta2024fast}. A representation learning method is required to learn the features for USFs, with past works using inverse curiosity modules \citep{pathak2017curiosity}, diversity methods \citep{liu2021aps, hansen2019}, Laplacian eigenfunctions \citep{wu2018laplacian}, or contrastive learning \citep{chen2020simple}. No works have yet explored the issues arising when training these methods on low quality offline datasets, and only one has investigated applying these ideas to real-world problems \cite{jeen2023low}.

Goal-conditioned RL methods train policies to reach any goal state from any other state, and so can be used to perform zero-shot RL in goal-reaching environments \citep{park2023hiql, ma2022far, yang2023essential, eysenbach2022contrastive, wang2023optimal}. However, they have no principled mechanism for conditioning policies on ``dense'' reward functions (as such tasks are not solved by simply reaching a particular state), and so are not full zero-shot RL methods. A concurrent line of work trains policies using sequence models conditioned on reward-labelled histories \citep{chen2021decision, janner2021offline, lee2022, reed2022generalist, zheng2022online, chebotar2023q, furuta2021generalized, siebenborn2022crucial, krause22transformer, xu22transformer}, but, unlike zero-shot RL methods, these works do not have a robust mechanism for generalising to different reward functions as test-time.

\textbf{Offline RL} $\;$ Offline RL algorithms require regularisation of policies, value functions, models, or a combination to manage the offline-to-online distribution shift \citep{levine2020}. Past works regularise policies with explicit constraints \citep{wu2019behavior, fakoor2021continuous, fujimoto2019, fujimoto2024sale, ghasemipour2021emaq, peng2023weighted, kumar2019stabilizing, wu2022supported, yang2022behavior}, via important sampling \citep{precup2001off, sutton2016emphatic, liu2019off, nachum2019dualdice, gelada2019off}, by leveraging uncertainty in predictions \citep{wu2021uncertainty, zanette2021provable, bai2022pessimistic, jin2021pessimism}, or by minimising OOD action queries \citep{wang2018exponentially, chen2020bail, kostrikov2021offline}, a form of imitation learning \citep{schaal1996learning, schaal1999imitation}. Other works constrain value function approximation so OOD action values are not overestimated \citep{kumar2020, kumar19, ma2021offline, ma2021conservative, lyu2022, yang2022rorl}. Offline model-based RL methods use the model to identify OOD states and penalise predicted rollouts passing through them \citep{yu2020, kidambi2020, yu2021combo, argenson2020model, matsushima2020deployment, rafailov2021offline, rigter2022}. All of these works have focused on regularising a finite number of policies; in contrast we extend this line of work to the zero-shot RL setting which is concerned with learning an infinite family of policies. 

\section{Conclusion}
In this chapter, we explored training agents to perform zero-shot RL from low quality data. We established that the existing methods suffer in this regime because they overestimate the value of out-of-distribution state-action values, a well-observed phenomena in single-task offline RL. As a resolution, we proposed a family of \textit{conservative} zero-shot RL algorithms that regularise value functions or dynamics predictions on out-of-distribution state-action pairs. In experiments across various domains, tasks and datasets, we showed our proposals outperform their non-conservative counterparts in aggregate and sometimes surpass our task-specific baseline despite lacking access to reward labels \textit{a priori}. In addition to improving performance when trained on sub-optimal datasets, we showed that performance on large, diverse datasets does not suffer as a consequence of our design decisions. We believe the proposals in this Chapter take a step toward managing the first of our three thesis constraints (\colouredS \ref{chapter1: path forward}). 

\chapter{Under Partial Observability}\label{chapter: under partial observability}
So far, we have shown that we can improve the performance of zero-shot RL methods when subjected to the \textit{data quality constraint} (\colouredS \ref{chapter1: path forward}). The data quality constraint says that real-world datasets lack the size and heterogeneity of toy datasets. We showed that, when trained on such datasets, standard zero-shot RL methods fail in predictable ways; they overestimate the values of state-action pairs not present in the dataset which poisons the policy. By regularising value functions to make conservative evaluations in such scenarios we went some way to addressing the failure mode and improving performance.

Implicit in \colouredS \ref{chapter: from low quality data}, and all prior results, is an assumed access to Markov states that provide the agent all the information it requires to solve a task. Though this is a common assumption in RL, for many interesting problems, the Markov state is only \textit{partially observed} via unreliable or incomplete observations \citep{kaelbling1998}. Observations can be unreliable because of sensor noise or issues with telemetry \citep{meng2021memory}. Observations can be incomplete because of egocentricity \citep{tirumala2024learning}, occlusions \citep{heess2015memory} or because they do not communicate a change to the environment's task or dynamics context \citep{hallak2015contextual}. This is our \textit{observability constraint} (\colouredS \ref{chapter1: path forward}).   

\begin{figure}[t]
\centering
    \includegraphics[width=\textwidth]{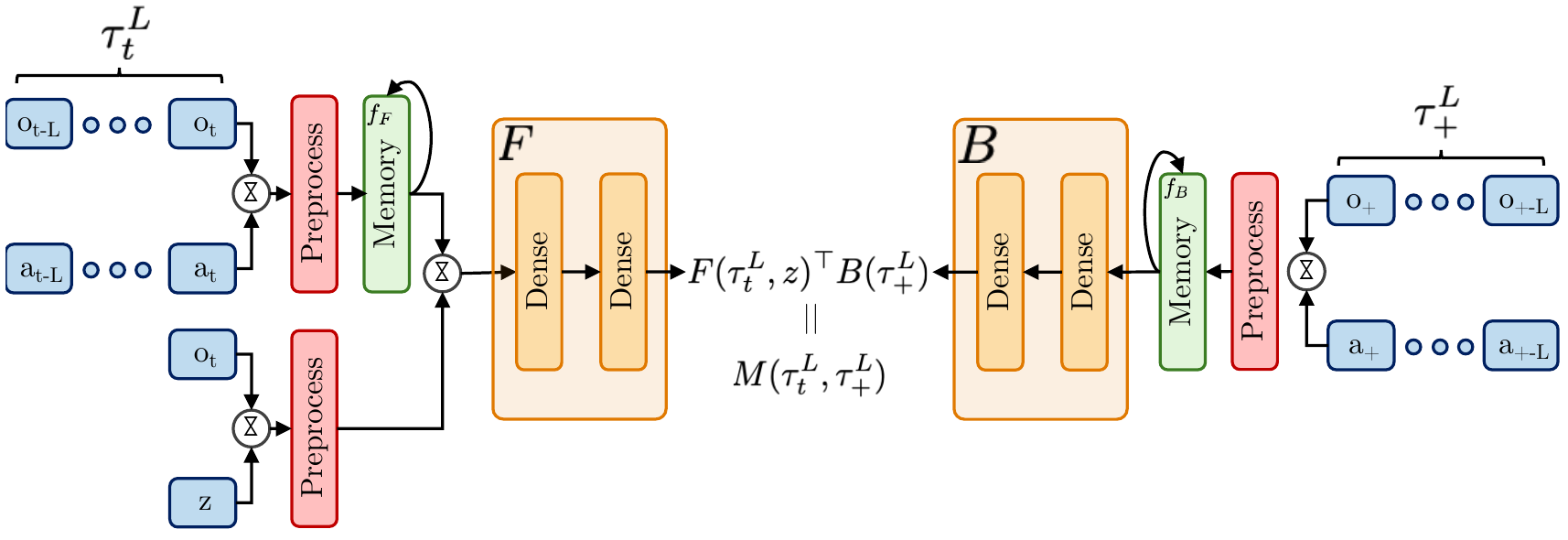}
    \caption{\textbf{Zero-shot RL methods with memory}.
    In the case of FB, the forward model $F$ and backward model $B$ condition on the output of memory models that compress trajectories of observations and actions. According to standard FB theory, their dot product predicts $M^{\pi_z}(\tau_t^L, \tau_+^L)$, the successor measure from $L$-length trajectory $\tau_t^L$ to $L$-length future trajectory $\tau_+^L$, from which a $Q$ function can be derived. Figure \ref{fig: fb architecture without memory} in Appendix \ref{appendix: implementations} illustrates memory-free FB for comparison.}
    \label{fig: chapter4/confb architecture}
\end{figure}

How do zero-shot RL methods fare when subjected to the \textit{observability constraint}? That is the primary question this Chapter seeks to answer, and one we address in three parts. First, we expose the mechanisms that cause the performance of standard zero-shot RL methods to degrade under partial observability (\colouredS \ref{subsection: failure mode of existing methods}). Second, we repurpose methods that handle partial observability in single-task RL for use in the zero-shot RL setting, that is, we add \textit{memory models} to the BFM framework (\colouredS \ref{subsection: method}, Figure \ref{fig: chapter4/confb architecture}). Third, we conduct experiments that test how well zero-shot RL methods augmented with memory models manage partially observed states (\colouredS \ref{subsection: partially observed states}) and partially observed changes in dynamics (\colouredS \ref{subsection: partially observed dynamics}). We conclude by discussing limitations and next steps.

\section{Preliminaries}\label{section: chapter4/preliminaries}
In this section we re-introduce material that was covered in Chapter \ref{chapter: background} for the readers benefit, and to ease the exposition.

\textbf{POMDPs} \; A partially observable Markov decision process (POMDP) $\mathcal{P}$ is defined by $(\mathcal{S}, \mathcal{A}, \mathcal{O}, R, P, O, \rho_0, \gamma)$, where $\mathcal{S}$ is the set of Markov states, $\mathcal{A}$ is the set of actions, $\mathcal{O}$ is the set of observations, and $\rho_0$ is the initial state distribution \citep{aastrom1965, kaelbling1998}. Let $s_t \in \mathcal{S}$ denote the Markov state at time $t$. When action $a_t \in \mathcal{A}$ is executed, the state updates via the transition function $s_{t+1} \sim P(\cdot | s_t, a_t)$, and the agent receives a scalar reward $r_{t+1} \sim R(s_{t+1})$ and observation $o_{t+1} \sim O(\cdot | s_{t+1}, a_t)$. The observation provides only partial information about the underlying Markov state. The agent samples actions from its policy $a_t \sim \pi(\cdot | \tau_{t}^L)$, where $\tau_{t}^L = (a_{t-L}, o_{t-L+1}, \ldots, a_{t-1}, o_t)$ is a \textit{trajectory} of the preceding $L$ observations and actions. We use $\mathcal{T}^L$ to denote the set of all possible trajectories of length $L$.
The policy is optimal in $\mathcal{P}$ if it  maximises the expected discounted future reward \textit{i.e.} $\pi^* = \arg\max_{\pi}\mathbb{E}[\sum_{t\geq 0}\gamma^t R(s_{t+1}) | s_0, a_0, \pi]$, where $\mathbb{E}[\cdot | s_0, a_0, \pi]$ denotes an expectation over state-action sequences $(s_t, a_t)_{t\geq0}$ starting at $(s_0, a_0)$ with $s_t \sim P(\cdot |s_{t-1}, a_{t-1})$ and $a_t \sim \pi(\cdot | \tau^{L}_t)$, and $\gamma \in [0,1]$ is a discount factor.

\textbf{Problem setting (partially observable zero-shot RL)} \; In the standard zero-shot RL problem setting, states are fully observed. For pre-training, the agent has access to a static offline dataset of reward-free transitions $\mathcal{D} = \{(s_i, a_i, s_{i+1})\}_{i=1}^{|\mathcal{D}|}$, generated by an unknown behaviour policy. At test time, a task $R_{\text{test}}$ is revealed by labelling a small number of states in $\mathcal{D}$ to create a new dataset $\mathcal{D}_{\text{labelled}} = \{(s_i, R_{\text{test}}(s_i))\}_{i=1}^{k}$ where typically $k \leq 10,000$. The agent must return a policy for this task with no further planning or learning.

In this Chapter, we consider the extended problem setting of \textit{partially observable zero-shot RL}. Here, the agent has access to an offline pre-training dataset of reward-free length-$L$ trajectories, $\mathcal{D} = \{\tau_{i}^L\}_{i=1}^{|\mathcal{D}|}$, each of which is a sequence of partial observations and actions.
% again generated by an unknown behaviour policy.
As before, a task $R_{\text{test}}$ is revealed at test time, for which the agent must return a policy. The task is specified by a small dataset of reward-labelled observation-action trajectories, where the reward is assumed to be a function of the final Markov state in the trajectory, $\mathcal{D}_{\text{labelled}} = \{(\tau_{i}^L, R_{\text{test}}(s^L_i))\}_{i=1}^{k}$.
% where $k \leq 10,000$.

\textbf{Behaviour foundation models} \; We build upon the forward-backward (FB) BFM which predicts successor measures \citep{blier2021learning}. The successor measure $M^{\pi}(s_0, a_0, \cdot)$ over $\mathcal{S}$ is the cumulative discounted time spent in each future state $s_{t+1}$ after starting in state $s_0$, taking action $a_0$, and following policy $\pi$ thereafter. Let $\rho$ be an arbitrary state distribution and $\mathbb{R}^d$ be an embedding space. FB representations are composed of a \textit{forward} model $F: S \times A \times\mathbb{R}^d\to\mathbb{R}^d$, a \textit{backward} model $B: S \to \mathbb{R}^d$, and set of polices $\pi(s, z)_{z \in \mathbb{R}^d}$. They are trained such that:
\begin{align}
    \label{eq:chapter4/M_approximation}
    M^{\pi_z}(s_0,a_0, X) &\approx \int_X F(s_0,a_0,z)^\top B(s)\rho(\text{d}s) \; \; \qquad \forall \; \; s_0 \in \mathcal{S}, a_0 \in \mathcal{A}, X \subset \mathcal{S}, z \in \mathbb{R}^d, \\
    \label{eq:chapter4/pi_approximation}
    \pi(s, z) &\approx \arg\max_a F(s,a,z)^\top z \; \qquad \qquad \qquad \qquad \forall \; \; (s,a) \in \mathcal{S} \times \mathcal{A}, z \in \mathbb{R}^d,
\end{align}
where $F(s,a,z)^\top z$ is the $Q$ function (critic) formed by the dot product of forward embeddings with a \textit{task embedding} $z$. During training, candidate task embeddings are sampled from $\mathcal{Z}$, a prior over the embedding space. During evaluation, the test task embeddings are inferred from $\mathcal{D}_{\text{labelled}}$ via:
\begin{equation}\label{equation: z inference}
z_{\text{test}} \approx \mathbb{E}_{(s, R_{\text{test}}(s)) \sim \mathcal{D}_{\text{labelled}}}[R_{\text{test}}(s)B(s)],
\end{equation}
and passed as an argument to the policy.\footnote{Equation \ref{equation: z inference} assumes $\mathcal{D}_{\text{labelled}}$ is drawn from the same distribution as $\mathcal{D}$, that is, it assumes both datasets are produced by the same, exploratory behaviour policy. Deploying a different policy to collect $\mathcal{D}_{\text{labelled}}$ is possible (\textit{e.g.} one learned via Equation \ref{eq:chapter4/pi_approximation}), but requires minor amendments to Equation \ref{equation: z inference}. We refer the reader to Appendix B.5 of \cite{touati2021learning} for further details.} 

\section{Zero-Shot RL Under Partial Observability}\label{chapter 4: method}
In this section, we adapt zero-shot RL methods for the partially observable setting. In \colouredS \ref{subsection: failure mode of existing methods}, we explore the ways in which standard zero-shot RL methods fail in this setting. Then, in \colouredS \ref{subsection: method}, we propose new methods that address these failures. We develop our methods in the context of FB, but our proposals are fully compatible with USF-based zero-shot RL methods. We report their derivation in Appendix \ref{appendix: usf with memory} for brevity.

\subsection{Failure Mode of Existing Methods}\label{subsection: failure mode of existing methods}
\begin{figure}[t]
\centering
    \includegraphics[width=\textwidth]{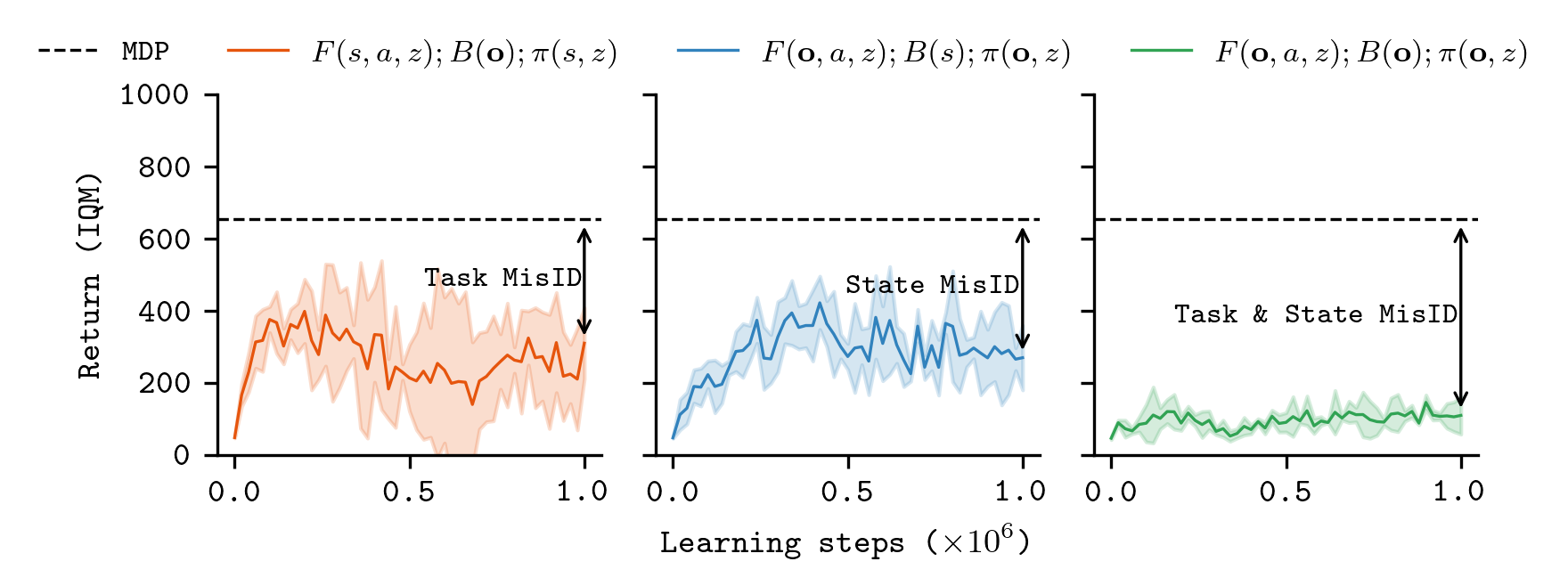}
    \vspace{-0.2cm}
    \caption{\textbf{The failure modes of zero-shot RL methods under partial observability.} FB's average (IQM) all-task return on \texttt{Walker} when observations are passed to its respective components. Observations are created by adding Gaussian noise to the underlying states. \textit{(Left)} Observations are passed as input to $B$ causing FB to misidentify the task. \textit{(Middle)} Observations are passed as input to $F$ and $\pi$ causing FB to misidentify the state. \textit{(Right)} Observations are passed as input to $F$, $\pi$ and $B$ causing FB to misidentify both the task and state.}
    \label{fig: chapter4/bfm-failure-mode}
\end{figure}

FB solves the zero-shot RL problem in two stages. First, a generalist policy is pre-trained to maximise FB's $Q$ functions for all tasks sampled from the prior $\mathcal{Z}$ (Equation \ref{eq:chapter4/M_approximation}). Second, the test task is inferred from reward-labelled states (Equation \ref{equation: z inference}) and passed to the policy. The first stage relies on an accurate approximation of $F(s,a,z)$ \textit{i.e.} the long-\texttt{run} dynamics of the environment subject to a policy attempting to solve task $z$. The second stage relies on an accurate approximation of $B(s)$ \textit{i.e.} the task associated with reaching state $s$. If the states in $F$ are replaced by observations that only partially characterise the underlying state, then the BFM will struggle to predict the long-\texttt{run} dynamics, $Q$ functions derived from $F$ will be inaccurate, and the policy will not learn optimal sequences of actions. We call this failure mode \textbf{state misidentification} (Figure \ref{fig: chapter4/bfm-failure-mode}, middle). Likewise, if the states in $B$ are replaced by partial observations, and the reward function depends on the underlying state (\colouredS \ref{section: chapter4/preliminaries}), then the BFM cannot reliably find the task $z$ associated with the set of states that maximise the reward function. We call this failure mode \textbf{task misidentification} (Figure \ref{fig: chapter4/bfm-failure-mode}, left). The failure modes occur together when both models receive partial observations (Figure \ref{fig: chapter4/bfm-failure-mode}, right).

\subsection{Addressing Partial Observability with Memory Models}\label{subsection: method}
In principle, all forms of partial observability can be resolved with \textit{memory models} that compress trajectories into a hidden state that approximates the underlying Markov state (see \colouredS \ref{section: related work} of \citet{ni2021recurrent}). A memory model is a function $f$ that outputs a new hidden state $h_t$ given a past hidden state $h_{t-L-1}$ and trajectory $\tau_{t}^L$:
\begin{equation}\label{equation: generalised memory model}
    h_t = f(\tau_{t}^L, h_{t-L-1}).
\end{equation}
Note that by setting $L=0$, we recover the standard one-step formulation of a recurrent neural network (RNN) \citep{elman1990finding, hochreiter1997, cho2014learning}. RNNs are common choice in past works \citep{wierstra2007policy, zhang2016learning, schmidhuber2019reinforcement}, but more recent works explore structured state space sequence models (S4) \citep{deng2023facing, lu2024structured} and transformers \citep{parisotto2020stabilizing, grigsby2023amago, grigsby2024amago}. In \textit{model-based} partially observable RL, state misidentification is resolved with memory-based dynamics models, and task misidentification is resolved with a memory-based reward models \citep{hafner2019, hafner2019learning, hafner2020mastering, hafner2023mastering}. In \textit{model-free} partially observable RL, the agent does not disentangle the state from the task, so task and state misidentification are resolved together by memory-based critics and policies \citep{ni2021recurrent, meng2021memory}.  
\vspace{-0.1cm}
\subsection{Zero-Shot RL Methods with Memory}
We now adapt methods from single-task partially observable RL for zero-shot RL. Standard FB operates on states (Equation \ref{eq:M_approximation}) that are inaccessible under partial observability, so we amend its formulation to operate on trajectories from which the underlying Markov state can be inferred with a memory model. The successor measure $M^\pi(\tau_0^L, \cdot)$ over $\mathcal{T}^L$ is the cumulative discounted time spent in each future trajectory $\tau_{t+1}^L$ starting from trajectory $\tau_0^L$, and following policy $\pi$ thereafter.\footnote{Note that the forward model and backward model can in principle have different context lengths. This is helpful if, for example, we know that the reward, as inferred via the backward model, depends on a shorter history length than would be required to infer the full Markov state via the forward model.} The architectures of the forward model $F$, backward model $B$, and policy $\pi$ are unchanged, but now condition on the hidden states of memory models, rather than on states and actions. They are trained such that
\begin{align}
    \label{eq:M_approximation with memory}
    M^{\pi_z}(\tau_0^L, X) &\approx \int_X F(f_F(\tau_0^L), z)^\top B(f_B(\tau^L))\rho(\text{d}\tau^L) \; \forall \; \tau_0^L \in \mathcal{T}^L, X \subset \mathcal{T}^L, z \in \mathbb{R}^d, \notag \\
    \pi(f_{\pi}(\tau^L), z) &\approx \arg\max_a F(f_F(\tau^L),z)^\top z \; \qquad \qquad \qquad \qquad  \; \; \; \;  \forall \;  \tau^L \in \mathcal{T}^L, z \in \mathbb{R}^d.
\end{align}
where $f_F, f_B, f_\pi$ are separate memory models for $F$, $B$, and $\pi$ respectively. Observation and action sequences are zero-padded for all $t - L -1 < 0$; the first hidden state in a sequence is always initialized to zero; and hidden states are dropped as arguments in Equation \ref{eq:M_approximation with memory} for brevity (\textit{c.f.} Equation \ref{equation: generalised memory model}). At test time, task embeddings are found via Equation \ref{equation: z inference}, but with reward-labelled trajectories rather than reward-labelled states:
\begin{equation}
    z_{\text{test}} \approx \mathbb{E}_{(\tau^L, R(s)) \sim \mathcal{D}_{\text{labelled}}}[R_{\text{test}}(s)B(f_B(\tau^L)].
\end{equation}
We refer to the resulting model as \textit{FB with memory} (FB-M).
The full architecture is summarised in Figure \ref{fig: chapter4/confb architecture}, and implementation details are provided in Appendix \ref{appendix: implementations}. Also note that our general proposal is BFM-agnostic; we derive the USF-based BFM formulation in Appendix \ref{appendix: usf with memory}. 

\section{Experiments}\label{section: experiments}
\subsection{Setup}\label{subsection: setup}
We evaluate our proposals in two partially observed settings: 1) partially observed states (\textit{i.e.} standard POMDPs), and partially observed changes in dynamics (\textit{i.e.} generalisation \citep{packer2018assessing}). The standard benchmarks for each of these settings only require the agent to solve one task, and so do not allow us to evaluate zero-shot RL capabilities out-of-the-box. As a result, we choose to amend the standard zero-shot RL benchmark, ExORL \citep{yarats2022}, such that it tests zero-shot RL with partially observed states and dynamics changes. 

\textbf{Partially observed states} \; 
We amend two of \cite{meng2021memory}'s partially observed state environments for the zero-shot RL setting: 1) \texttt{noisy} states, where isotropic zero-mean Gaussian noise with variance $\sigma_{\text{noise}}$ is added to the Markov state, and 2) \texttt{flickering} states, where states are dropped (zeroed) with probability $p_{\text{flick.}}$. We set $\sigma_{\text{noise}} = 0.2$ and $p_{\text{flick.}} = 0.2$ following a hyperparameter study (Appendix \ref{appendix: pomdps}). We evaluate on all tasks in the \texttt{Walker}, \texttt{Cheetah} and \texttt{Quadruped} environments.

\textbf{Partially observed changes in dynamics} \;
We amend \cite{packer2018assessing}'s dynamics generalisation tests for the zero-shot RL setting. Environment dynamics are modulated by scaling the mass and damping coefficients in the MuJoCu backend \citep{todorov2012}. The agents are trained on datasets collected from environment instances with coefficients scaled to $\{0.5\times, 1.5\times\}$ their usual values, then evaluated on environment instances with coefficients scaled by $\{1.0\times, 2.0\times\}$. Scaling by $1.0\times$ tests the agent's ability to generalise via \textit{interpolation} within the range seen during training, and scaling by $2.0\times$ tests the agent's ability to generalise via \textit{extrapolation} \citep{packer2018assessing}.

\textbf{Baselines} \; We use two state-of-the-art zero-shot RL methods as baselines: FB \citep{touati2021learning} and HILP \citep{park2024foundation}. We additionally implement a na\"ive baseline called FB-stack whose input is a \textit{stack} of the 4 most recent observations and actions, following \cite{mnih2015}'s canonical protocol. Finally, we use FB trained on the underlying MDP as an oracle policy to give us an upper-bound on expected performance. All methods inherit the default hyperparameters from previous works \citep{touati2022does, park2024foundation, jeen2024zeroshot}.  

\textbf{Datasets} \; We train all methods on datasets collected with an RND behaviour policy \citep{borsa2018} as these are the datasets that elicit best performance on ExORL. The RND datasets used in the partially observed states experiments are taken directly from ExORL. For the partially observed change in dynamics experiments, we collect these datasets ourselves by \texttt{run}ning RND in each of the environments for 5 million learning steps. Our implementation and training protocol exactly match ExORL's.

\textbf{Memory model} \; We use a GRU as our memory model \citep{cho2014learning}. GRUs are the most performant memory model on POPGym \citep{morad2023popgym} which tests partially observed single-task RL methods. We find these results hold for partially observed zero-shot RL too, as discussed in \colouredS \ref{discussion: memory model}.
% For the partially observed state and dynamics experiments,
We set the context length $L = 32$; see Appendix \ref{appendix: context lengths} for a hyperparameter study and further discussion. 

\textbf{Evaluation protocol} \;
We evaluate the cumulative reward achieved by all methods across $5$ seeds, with task scores reported as per the best practice recommendations of \cite{agarwal2021deep}. Concretely, we \texttt{run} each algorithm for 1 million learning steps, evaluating task scores at checkpoints of $20,000$ steps. At each checkpoint, we perform $10$ rollouts, record the score of each, and find the interquartile mean (IQM). We average across seeds at each checkpoint. We extract task scores from the learning step for which the all-task IQM is maximised across seeds. Results are reported with $95\%$ bootstrap confidence intervals in plots and standard deviations in tables. Aggregation across tasks, domains and datasets is always performed by evaluating the IQM. 

\subsection{Partially Observed States}\label{subsection: partially observed states}
\begin{figure}[t]
\centering
    \includegraphics[width=\textwidth]{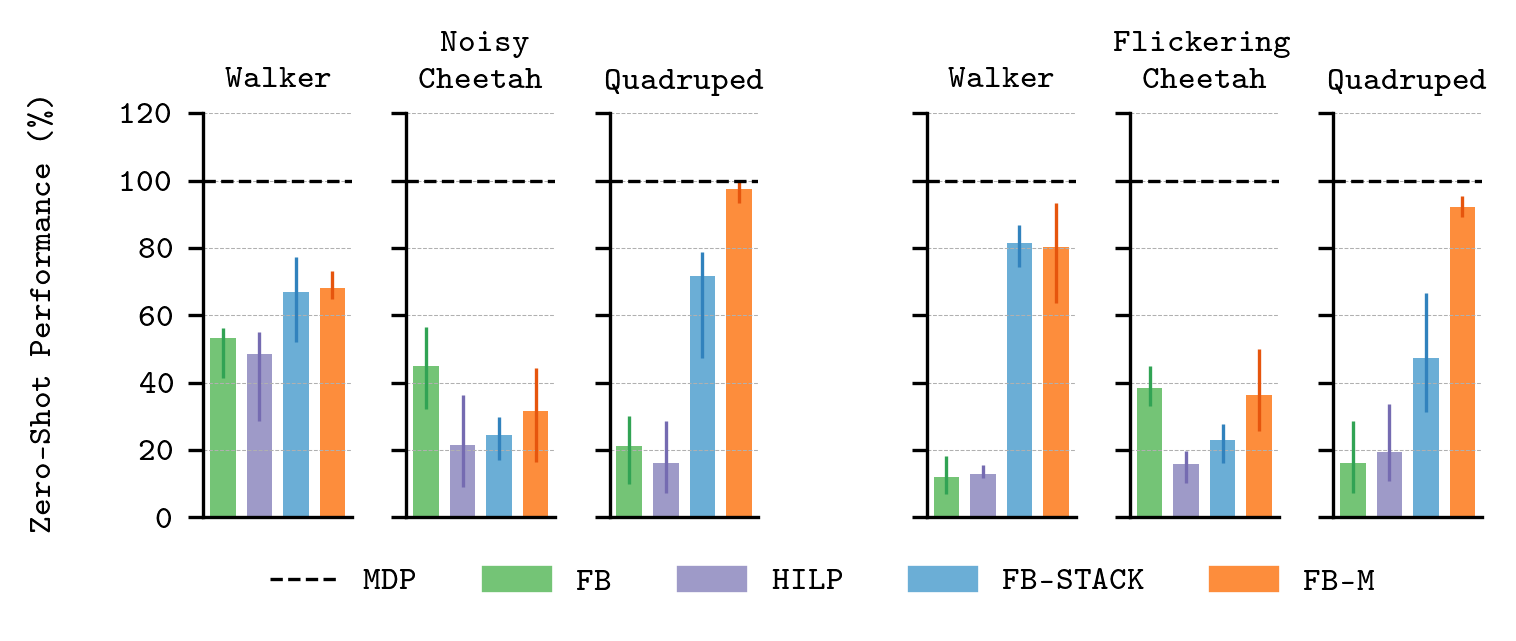}
    \caption{\textbf{Aggregate zero-shot task performance on ExORL with partially observed states.} IQM of task scores across all tasks on \texttt{noisy} and \texttt{flickering} variants of \texttt{Walker}, \texttt{Cheetah} and \texttt{Quadruped}, normalised against the performance of FB in the fully observed environment. 5 random seeds.} 
    \label{fig: chapter4/pomdp-states}
\end{figure}
Figure~\ref{fig: chapter4/pomdp-states} compares the zero-shot performance of all algorithms on our \texttt{noisy} and \texttt{flickering} variants of the standard ExORL environments. Note that these results are aggregated across all tasks in each environment, and 5 random seeds.
The performance of memory-free FB is always far below that of an oracle policy trained on the underlying MDP (dotted line), reaching less than $50\%$ of the oracle value in 5 out of 6 cases, and HILP performs similarly.
Augmenting FB by stacking recent observations mitigates the partial observability problem to some extent on \texttt{Walker} and \texttt{Quadruped}, but it performs worse than memory-free FB on \texttt{Cheetah} . Our proposed approach (FB-M) outperforms this baseline in all settings except \texttt{Walker} where it performs similarly.
The benefit of FB-M is most pronounced for the \texttt{Quadruped} environment where it achieves close to oracle performance. The full results are reported in Table \ref{table: chapter4/partially observed states} in Appendix \ref{appendix: extended results}.

\subsection{Partially Observed Changes in Dynamics}\label{subsection: partially observed dynamics}
Next, we consider the problem of partially observed dynamics changes in both the interpolation and extrapolation settings.
The results are summarised in Figure~\ref{fig: chapter4/generalisation} and reported in full in Table \ref{table: chapter4/full changed dynamics} in Appendix \ref{appendix: extended results}.
First, we find that memory-free FB performs well in the interpolation setting, matching the oracle policy in two of the three environments, but less well in the extrapolation setting where it underperforms the oracle in all environments. As in \colouredS \ref{subsection: partially observed states}, HILP is consistently the lowest scoring method in all environments. In general, stacking recent observations (as in FB-stack) harms performance, with scores lower than memory-free FB in 5 out of 6 environment/dynamics settings. FB-M performs similarly to, or better than, all algorithms in all settings. The performance difference is most pronounced in the extrapolation setting on \texttt{Cheetah} and \texttt{Quadruped} where it slightly outperforms the oracle policy. We think this is because the dataset collected under the extrapolation dynamics (with doubled mass and damping coefficients) is less expressive than the standard dynamics because the behaviour policy struggled to cover the state space. Relative differences in the non-MDP results remain valid should this be the case. 
\begin{figure}[t]
\centering
    \includegraphics[width=\textwidth]{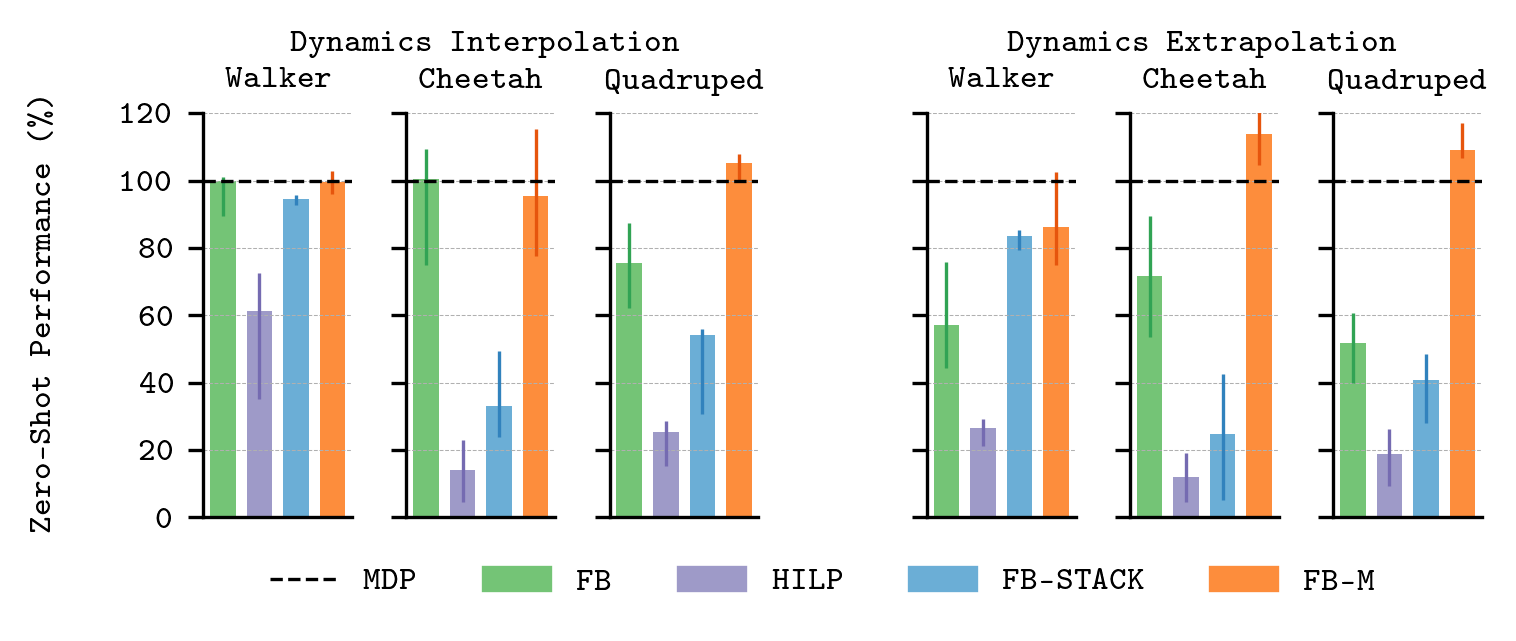}
    \caption{\textbf{Aggregate zero-shot task performance on ExORL with changed dynamics at test time.} IQM of task scores across all tasks when trained on dynamics where mass and damping coefficients are scaled to \{$0.5\times$, $1.5\times$\} their usual values and evaled on \{$1.0\times$, $2.0\times$\} their usual values, normalised against the performance of FB in the fully observed environment. To solve the test dynamics with $1.0\times$ scaling the agent must interpolate within the training set; to solve the test dynamics with $2.0\times$ scaling the agent must extrapolate from the training set.}
    \label{fig: chapter4/generalisation}
\end{figure}

\section{Discussion and Limitations}\label{section: discussion}
\subsection{Memory Model Choice} \label{discussion: memory model}
Our method uses GRUs as memory models, but much recent work has shown that transformers \citep{vaswani2017attention} and structured state-space models \citep{gu2021efficiently} outperform GRUs in natural language processing \citep{brown2020language}, computer vision \citep{dosovitskiy2020image}, and model-based RL \citep{deng2023facing}. In this section, we explore whether these findings hold for the zero-shot RL setting. We compare FB-M with GRU memory models to FB-M with transformer and diagonalised S4 (S4d) memory models \citep{gu2022parameterization}. We follow \cite{morad2023popgym} in restricting each model to a fixed hidden state size, rather than a fixed parameter count, to ensure a fair comparision. Concretely, we allow each model a hidden state size of $32^2 = 1024$ dimensions. Full implementation details are provided in Appendix \ref{appendix: implementations}. We evaluate each method in the three variants of \texttt{Walker} \texttt{flickering} used in \colouredS \ref{subsection: failure mode of existing methods} \textit{i.e.} where only the inputs to $F$ and $\pi_z$ are observations, only inputs to $B$ are observations, and where inputs to all models are observations.

\begin{figure}[ht]
\centering
    \includegraphics[width=0.9\textwidth]{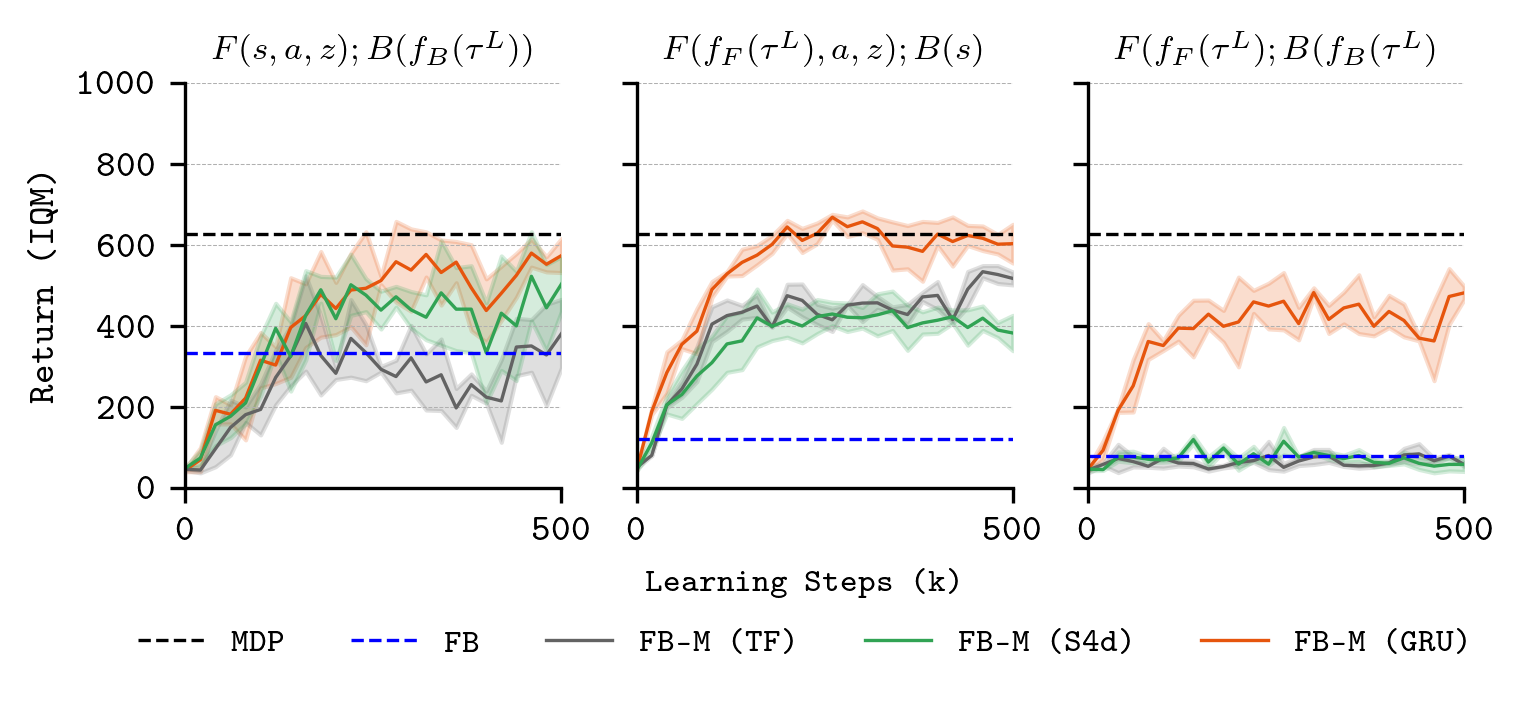}
    \vspace{-0.3cm}
    \caption{\textbf{Aggregate zero-shot task performance of FB-M with different memory models.} IQM of task scores across all tasks on \texttt{Walker flickering}. \textit{(Left)} Observations are passed only to a memory-based backward model; the forward model and policy are memory-free. \textit{(Middle)} Observations are passed only to the forward model and policy; the backward model is memory-free. \textit{(Right)} Observations are passed to all models.}
    \label{fig: chapter4/memory-models}
\end{figure}

Our results are reported in Figure \ref{fig: chapter4/memory-models}. We find that the performance of FB-M is reduced in all cases when a transformer or S4 memory model is used instead of a GRU. This corroborates \cite{morad2023popgym}'s findings that the GRU is the most performant memory model for single-task partially observed RL. Perhaps most crucially, we find that training collapses when \textit{both} $F$ and $B$ are non-GRU memory models, despite non-GRU memory models performing reasonably when added to \textit{only} $F$ or $B$, suggesting that the combined representation $M(\tau^L, \tau^L_+) \approx F(f_F(\tau^L))^\top B(f_B(\tau_+^L))$ is degenerate. Better understanding this failure mode is important future work.
\vspace{-0.35cm}
\subsection{Datasets}
As outlined in \colouredS \ref{subsection: setup}, we train all methods on datasets pre-collected with RND \citep{borsa2018} which is a highly exploratory algorithm designed for maximising data heterogeneity. However, deploying such an algorithm in any real setting may be costly, time-consuming or dangerous. As a result, our proposals are more likely to be trained on real-world datasets that are smaller and more homogeneous. It is not clear how our specific proposals will interact with such datasets. If, for example, the dataset only represents parts of the state space from which the dynamics cannot be well-inferred, because it was collected from a robot with limited freedom of movement, then we would expect our proposals to struggle. Indeed, with poor coverage of the state-action space, we would expect to see the OOD pathologies seen in the single-task offline RL setting \citep{kumar19, levine2020}. That said, the proposals of \cite{jeen2024zeroshot} for conducting zero-shot RL from less diverse datasets could be integrated into our proposals easily, and may help.

\section{Related Work}
\subsection{Zero-shot RL}

\textbf{Offline RL} \; An important part of the zero-shot RL problem is that agents must pre-train on static datasets (\colouredS \ref{section: chapter4/preliminaries}). This is the realm of offline RL \citep{lange2012batch, levine2020}, where regularisation techniques \citep{kumar2020, kidambi2020, fujimoto2021minimalist} are used to minimise the \textit{distribution shift} between the offline data and online experience \citep{kumar2019stabilizing}. In this Chapter, we only train on high-coverage datasets to isolate the problem of partial observability, so do not require such regularisation. Standard offline RL methods are trained with respect to one downstream task, so cannot generalise to new tasks at test time, as specified by the zero-shot RL problem. 

\textbf{Goal-conditioned RL} \; For goal-reaching tasks, zero-shot goal generalisation can be achieved with \textit{goal-conditioned} RL (GCRL) \citep{schaul2015, andrychowicz2017hindsight}. Here, policies are trained to reach any goal state from any other state. Past work has focused on constructing useful goal-space encodings, with contrastive \citep{eysenbach2022contrastive}, state-matching \citep{ma2022far}, and hierarchical representations \citep{park2024hiql} proving effective. However, GCRL methods do not reliably generalise to \textit{dense} reward functions that cannot be codified by a goal state,\footnote{Examples include any locomotion task \textit{e.g.} \texttt{Walker}-\texttt{\texttt{run}} in the DeepMind Control Suite.} and so cannot be said to solve the general zero-shot RL problem.

\textbf{Behaviour foundation models} \; To date, behaviour foundation models (BFMs) have shown the best zero-shot RL performance because they provide a mechanism for zero-shot generalising to \textit{both} goal-reaching and dense reward functions.\footnote{A formal justification of this statement is left for \colouredS \ref{section: chapter4/preliminaries}.} They build upon successor representations \citep{dayan1993}, universal value function approximators \citep{schaul2015}, successor features \citep{barreto2017} and successor measures \citep{blier2021learning}. State-of-the-art methods instantiate these ideas as either universal successor features (USFs) \citep{borsa2018, park2024foundation} or forward-backward (FB) representations \citep{touati2021learning, touati2022does, jeen2024zeroshot}. No works have yet explored the zero-shot RL performance of these methods under partial observability.

\subsection{Partial Observability}

\textbf{States} \; Most past works assume it is the \textit{state} that is partially observed. This is usually the result of noisy \citep{meng2021memory}, occluded \citep{heess2015memory}, aliased \citep{whitehead1990active}, egocentric \citep{tirumala2024learning} or otherwise unreliable observations. Standard solutions methods use histories of observations and actions to compute \textit{beliefs} over the true state via (approximate) Bayesian inference \citep{cassandra1994acting, kaelbling1998} or via memory-based architectures \citep{schmidhuber1990reinforcement,bakker2001reinforcement, hausknecht2015deep, ha2018}.

\textbf{Dynamics} \; Sometimes, parameters that modulate the underlying \textit{dynamics} change and are not communicated to the agent via the state. Given sets of training and testing dynamics parameters, \textit{generalisation} is a measure of the agent's average-case performance on the test set \citep{packer2018assessing, cobbe2019quantifying}. If the agent trains and tests on the same set of dynamics, \textit{robustness} is a measure of the agent's worst-case performance on this set \citep{nilim2005robust, morimoto2005robust, mankowitz2019robust}. Generalisation can be improved via regularisation \citep{farebrother2018generalization}, data augmentation \citep{tobin2017domain, raileanu2020automatic, ball2021augmented}, or dynamics context modelling \citep{seo2020trajectory, lee2020context}. Robustness can be improved with adversarial dynamics selection \citep{rajeswaran2016epopt, jiang21, rigter2023reward}.

\textbf{Rewards} \; In some cases, the utility of an action for a task may only be partially reflected in the standard one-step reward \citep{minsky1961steps, sutton1984temporal}. Such a situation arises when the reward signal is delayed \citep{arjona2019rudder} or is dependent on the entire trajectory (i.e. episodic) \citep{liu2019off}. These have traditionally been handled with sophisticated techniques that learn surrogate reward functions \citep{raposo2021synthetic, arjona2019rudder}, tune discount factors \citep{fedus2019hyperbolic}, or utilise eligibility traces \citep{xu2020meta}, among other methods.  

Each of the above methods were developed for a specific form of partial observability, but memory-based architectures are, in principle, general enough to solve all of them \citep{kaelbling1998}. Indeed, \cite{ni2021recurrent} find that a standard, but well-implemented, recurrent policy and critic can outperform methods specialised for each setting. Our proposed method (\colouredS \ref{subsection: method}) is heavily informed by this finding, and is designed to be agnostic to the specific way in which partial observability arises.

\section{Conclusion}
In this Chapter, we explored how the performance of zero-shot RL methods degrades when subjected to certain types of partial observability. We introduce memory-based zero-shot RL methods that condition $F$, $B$ and $\pi_z$ on trajectories of observation-action pairs, and show they go some way to remedying state and task misidentification. We evaluated our proposals on a suite of partially observed zero-shot RL problems, where the observations passed to the agent are noisy, dropped randomly or do not communicate a change in the underlying dynamics, and showed improved performance over memory-free baselines. We found the GRU to be the most performant memory model, and showed that transformers and s4 memory models cannot be trained stably at our scale. We believe our proposals represent a further step towards the real-world deployment of zero-shot RL methods.. We believe the proposals in this Chapter take a step toward managing the second of our three thesis constraints (\colouredS \ref{chapter1: path forward}). 

\chapter{With No Prior Data}\label{chapter: no prior data}

The methods discussed thus far have assumed access to a dataset of transitions for agent pre-training. In Chapter \ref{chapter: from low quality data} we argued that real datasets are not as expressive as typical zero-shot RL methods expect, and proposed methods for performing zero-shot RL with lower quality datasets. In Chapter \ref{chapter: under partial observability} we argued that the observations in these datasets may provide only partial information about the underlying environment state, and proposed methods for handling this misalignment efficiently. In this Chapter, we ask: is it possible to perform zero-shot RL \textit{with no prior data}.  
As discussed in \colouredS \ref{background: dynamics generalisation}, most real-world settings are likely too complex for us to expect to safely deploy an agent zero-shot with no pre-training. However, one problem setting that may be simple enough, and in which better control policies could prove useful, is emission-efficient control of heating and cooling systems in buildings. Such systems account for 31\% of global energy use, primarily in managing occupant thermal comfort and hygiene \cite{cullen2010}. They are usually regulated by rule-based controllers (RBCs) that take system temperature as input, use a temperature setpoint as an objective, and actuate equipment to minimise the error between objective and current state. Whilst usefully simple, RBCs do not maximise energy efficiency, nor can they perform \textit{demand response}: the manipulation of power consumption to better match demand with supply \citep{siano2014demand}. New techniques that demonstrate this capability across generalised settings would prove valuable climate change mitigation tools.

RL is an appealing option. Where existing advanced control techniques, like the receding-horizon architecture Model Predictive Control (MPC), require the specification of a dynamical model that can be expensive to obtain \cite{morari2017}, RL's strength is in obtaining polices \textit{tabula rasa}, and updating their parameters \textit{online} as the environment evolves. With this generality, one can imagine RL agents being placed in \textit{any} energy-intensive setting, and feasibly learning to control them more efficiently, minimising emissions.

Recent applications of RL to building control have indeed shown marked energy efficiency improvements over conventional controllers. \citet{CHEN2018195} used Q-learning to control HVAC and window actuation in a residential building with 23\% less energy than the existing RBC. Similarly, \citet{zhang2019whole} used Asynchornous Advantage Actor Critic (A3C) to reduce heating demand in a real office building by 16.7\%. However, these contributions are limited by their deployment paradigm, which is:
\begin{enumerate}
    \item \textbf{Simulate} the real-world environment's transition dynamics $p(s_{t+1} | s_t, a_t)$ using a physics-based/generative model; 
    \item \textbf{Pre-train} the agent by sampling the simulator sequentially and updating parameters until convergence; and
    \item \textbf{Deploy} the pre-trained agent in the real-world environment.
\end{enumerate}
We contend that the need to \textit{simulate} the environment \textit{a priori} limits real-world scalability. In the context of building control, creating an accurate simulator in \textit{EnergyPlus} \cite{energyplus2001}, the favoured software, can take an expert months, and is impossible without knowledge of the building topology and thermal parameters (\colouredS \ref{chapter1: unfulfilled promises}. An alternative, is to deploy agents in the real environment \textit{without} pre-training, granting them a short commissioning period of 3 hours to collect data and model the state-action space \cite{LAZIC2018}. We call this \textbf{zero-shot} building control. Zero-shot building control could scale to any building given sufficient sensor installation, but \citet{LAZIC2018}'s agent only improved cooling costs by 9\%, poorer performance than the agents trained in simulation. To maximise the emission-abating potential of RL building control, we need new systems that can elicit the performance of pre-trained agents whilst being deployed zero-shot.

In this Chapter, our primary contribution is showing that deep RL algorithms can find performant building control policies \textit{zero-shot}, without pre-training. We achieve this with a new approach called PEARL, showing it can reduce annual emissions by $\approx$ 30\% compared with an RBC whilst maintaining thermal comfort. PEARL is simple to commission, requiring no historical data or simulator access \textit{a priori}, and capable of generalising across varied building archetypes. The scaled deployment of such systems could prove a cost-effective method for tackling climate change.

\section{Related Work} \label{section: related work}
Previous research on RL for building control has focused mainly on model-free algorithms \cite{yu2021}, but as discussed in \colouredS \ref{background: model-free RL}, these methods are too data-inefficient for zero-shot deployment. The canonical works: Deep-Q Learning \cite{mnih2013}, Deep Deterministic Policy Gradient (DDPG) \cite{lillicrap2015}, and Proximal Policy Optimisation (PPO) \cite{schulman2017} each take on the order 10$^7$ samples in complex, simulated environments to obtain optimal polices. Such data inefficiency has been corroborated in the building control literature. \citet{Wei2017} train a Deep-Q agent to control the HVAC equipment of a 5-zone building with 35\% reduction in energy cost, but pre-train for 8 years in a simulator. Similarly, \citet{VALLADARES2019} require 10 years of simulated data to pre-train a Double-Q agent to control the HVAC in a university classroom with 5\% energy savings. Such data-intensive pre-training can only be achieved in bespoke building simulators that are time-consuming to create, or impossible to specify, in most cases.

One-step model-based RL algorithms demonstrate better sample efficiency (\colouredS \ref{background: one-step MBRL}). Between interactions with the environment, the agent samples its model to create additional data for updating the policy, or to predict the expected reward of a range of candidate action sequences. Either procedure reduces the agent's reliance on samples from the environment, improving data efficiency.

A popular choice of function approximator are Gaussian Processes (GP), which quantify uncertainty and work well with small datasets \cite{williams2006}. PILCO used GPs, showing state-of-the-art data efficiency on robotic tasks \cite{deisenroth2011pilco}, and \citet{jain2018} used a similar approach to curtail hotel energy-use during a simulated demand response event. However, inference using a data set of size $n$ has complexity $\mathcal{O}(n^3)$ which becomes intractable with more than a few thousand samples \cite{hensman2013}, limiting their applicability for modelling building transition functions with arbitrarily large training sets (\colouredS \ref{background: one-step MBRL}).

An alternative is to model the transition function using DNNs. In the building control setting, \citet{zhang2019} combined DNN models with MPC to reduce energy consumption in a data centre by $\approx$ 20\% with 10x fewer training steps than a model-free algorithm. \citet{jain20} then tested the algorithm in-situ, finding an 8\% energy reduction. Despite encouraging data efficiency, the performance of such agents is hampered by overfitting, or \textit{model bias}, in the low-data regime, causing poor generalisation to unobserved transitions. \citet{ding2020} attempt to mitigate model bias by deploying an ensemble of DNNs to model the transition function of a large multi-zone building showing they can achieve 8\% energy savings in 10x fewer timesteps than model-free approaches.

To the best of our knowledge, \citet{LAZIC2018} is the only work that attempts to learn a zero-shot building control policy by interacting with the real environment. Their agent fits a linear model of a datacentre's thermal dynamics using data obtained in a three-hour commissioning period, and selects actions by optimising planned trajectories using MPC. During commissioning, the agent explores the state-action space by performing a uniform random walk in each control variable, bounded to a safe operating range informed by historical data. Their choice of model expedites learning, but limits agent performance as the building's non-linear dynamics are erroneously linearised. In this study we aim to preserve the data efficiency of this approach whilst improving performance with expressive deep networks.

\section{Preliminaries}
In this section we re-introduce material that was covered in Chapter \ref{chapter: background} for the readers benefit, and to ease the exposition.

\textbf{Markov decision processes} $\;$ A Markov Decision Process (MDP) is defined by $\mathcal{M} = \{\mathcal{S}, \mathcal{A}, P, R, \rho_0$), where $\mathcal{S}$ and $\mathcal{A}$ are continuous state and action spaces, $R: \mathcal{S} \times \mathcal{A} \rightarrow \mathbb{R}$ is a reward function, and  $P: \mathcal{S} \times \mathcal{A} \rightarrow \Delta(\mathcal{S})$ is a transition function. The agent selects actions using its policy $\pi: \mathcal{S} \rightarrow \Delta(\mathcal{A})$. Its policy is optimal if it maximises the expected future reward \textit{i.e.} $\pi^* = \mathbb{E} [R(s_t, a_t) | s_0, a_0, \pi]$, where $\mathbb{E}[\cdot | s_0, a_0, \pi]$ is the expectation under state-action sequence $(s_t, a_t)_{t \geq 0}$, starting at $(s_0, a_0)$ with $s_t \sim P(\cdot | s_{t-1}, a_{t-1})$ and $a_t \sim \pi(\cdot | s_t)$.

\textbf{Problem setting (zero-shot building control).} $\;$ The agent controls a building's Heating, Ventilation and Cooling (HVAC) equipment for one year, where one timestep in simulation covers 10 minutes of real time.  It has no \textit{a priori} access to a simulation of the building, or data from the building, for pre-training, but it is allowed a three-hour \textit{commissioning period} (where $t \leq C$) in which it can collect a dataset of transitions $\mathcal{D} = \{(s_t, a_t, r_t)^i\}_{i=1}^C$ without its performance being measured. During the \textit{evaluation period} (where $C < t \leq T$) the agent must maximise scalar reward specified by reward function $R(s, a)$. The agents can access the reward function directly. The scalar reward is a linear combination of an \textit{emissions term} and a \textit{temperature term}, where the former is maximised when emissions are minimised, and the latter is maximised when the building temperature is within thermal comfort bounds. Full reward function specification is provided in Appendix \ref{appendix: energym environments}. During the evaluation period, the agent can continue to add transitions to $\mathcal{D}$, which it can use to make one set of updates to its policy, model or value function per day.

\section{PEARL: Probabilistic Emission-Abating Reinforcement Learning} \label{section: PEARL}

\begin{figure*}[t]
    \includegraphics[width=\textwidth]{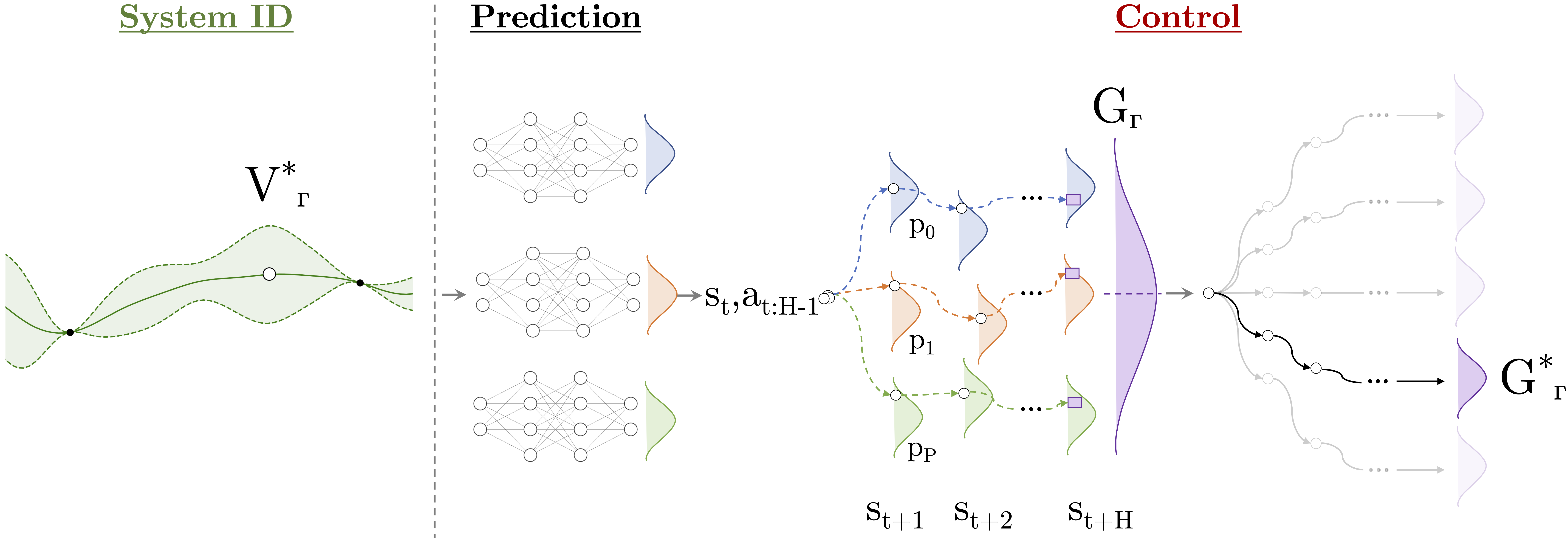}
    \caption{\textbf{PEARL: Probabilistic Emission-Abating Reinforcement Learning}. System ID: the agent takes actions to explore parts of the state-space with highest predictive variance $\circ = V_\Gamma^*$ to attempt to maximise information gain. Prediction: system dynamics are modelled with an ensemble of probabilistic deep neural networks. Control: trajectory sampling used to predict future rewards $G_\Gamma$ of one action sequence $a_{t:H-1}$, which is compared with many others to find the trajectory with optimal return $G^*_\Gamma$.}
    \label{fig:pearl}
\end{figure*}

In this section we outline our proposed approach. Our problem setting demands high data efficiency, so we build upon works from data-efficient RL. Of those, the most popular method is Probabilistic Ensembles and Trajectory Sampling (PETS) (\colouredS \ref{background: one-step MBRL}) \citep{chua2018}, which we use as the skeleton of our method. We make updates to PETS so it can make best use of the \textit{commissioning period}. We call the resulting algorithm PEARL: Probabilistic Emission-Abating Reinforcement Learning, and summarise it Figure \ref{fig:pearl} and in Algorithm 1.

\textbf{Dynamics Model} $\;$ We denote with $\tilde{P}: \mathcal{S} \times \mathcal{A} \rightarrow \mathcal{S}$ the dynamics model we train with dataset $\mathcal{D}$ to approximate the true transition function $P$. Following \citep{chua2018}, $\tilde{P}$ is an ensemble of $K$ probabilistic \cite{gal2016, higuera2018} DNNs. Where \textit{deterministic} DNNs output point predictions given an input, here the probabilistic DNNs output a multivariate Gaussian distribution over the state-space with mean $\mu$ and diagonal covariance matrix $\Sigma$; i.e: $\tilde{P}(s_{t+1} | s_t, a_t) = \mathcal{N}(\mu_\theta(s_{t+1} | s_t, a_t), \Sigma_\theta(s_{t+1} | s_t, a_t))$. Each member of the ensemble is initialised differently and trained over different subsets of the data according to the standard Maximum Likelihood Estimation (MLE) loss:
\begin{equation} \label{equation: loss}
    \mathcal{L}_{\text{MLE}} = \mathbb{E}_{(s_t, a_t, s_{t+1}) \sim \mathcal{D}} \left[ -\text{log} \tilde{P}(s_{t+1} | s_t, a_t) \right].
\end{equation}

\textbf{Planning} $\;$ The agent plans with Model Predictive Path Integral (MPPI)\footnote{We use MPPI over PETS' Cross-Entropy Method (CEM) planner \citep{botev2013} because of the superior performance reported by \citep{williams2015}} \cite{williams2015} and Trajectory Sampling \cite{chua2018}. Action sequences $a_{t:t+H-1}^u \; \forall \; u \in U$ are sampled from an initially arbitrary multivariate Gaussian distribution with diagonal covariance and parameters $(\mu^0, \sigma^0)_{t:t+H} \in \mathbb{R}^a$ which we denote with $\pi_{\text{MPPI}}$. The state-action pairs at the first planning timestep are duplicated $M$ times to create planning \textit{particles} $(s_t^m, a_t) \; \;, m \in [1, \ldots, M]$ \citep{chua2018}. The dynamics models unroll the particles autoregressively to time horizon $H$ according to the action sequences. Particles are randomly assigned to one fixed member of the dynamics ensemble each planning iteration. The return $G_{\Gamma}$ from each trajectory $\Gamma$ is predicted as the mean across particles and planning horizon according to ground-truth reward function $R$:
\begin{equation} \label{equation: expected reward}
    G_\Gamma = \frac{1}{M} \sum_{m=1}^{M} \sum_{t=1}^H R(s_t^m, a_t) \; .
\end{equation}

The top-$e$ returns $G_\Gamma^*$ are selected and the parameters of the action-selection distribution at iteration $j$ using a $G_\Gamma^*$-normalised estimate:
\begin{equation} \label{equation: mppi update}
\mu^j = \frac{\sum_{i=1}^e \Pi_i \Gamma_i^*}{\sum_{i=1}^e \Pi_i}, \; \sigma^j = \sqrt{\frac{\sum_{i=1}^e \Pi_i (\Gamma_i^* - \mu^j)^2}{\sum_{i=1}^e \Pi_i}} \; ,
\end{equation}

where $\Pi_i = e^{\tau(G_\Gamma^*, i)}$, $\tau$ is a temperature parameter that mediates weight given to the optimal trajectory, and $\Gamma_i^*$ is the $i$th top-$e$ trajectory corresponding to expected return $G_\Gamma^*$ \cite{hansen2022}. After $n$ optimisation iterations, the planner returns $\mu^{n}$, the action sequence with highest expected reward, and the first action is taken in the real environment.

\textbf{System identification.} $\;$ During the commissioning period, the task is to sample the state-action space sequentially, and update the the model, such that the error in model predictions is minimised. Inspired by Bayesian Optimisation \cite{snoek2012}, we leverage the predictive variance in our models to select trajectories that transition the agent to parts of the state-space where it is most uncertain, sometimes called Maximum Variance (MV) exploration \cite{jain2018}. We adapt the routine from the previous section to evaluate the variance $V_\Gamma$ in the predicted rewards across particle trajectories for a given action sequence
\begin{equation}
    V_\Gamma = Var_{M}\left[\sum_{t=1}^H R(s_t, a_t) \right] \; .
\end{equation}
Now the elite trajectories $\Gamma^*$ in Equation \eqref{equation: mppi update} correspond to the action sequences and that maximise predictive variance $V_\Gamma^*$. During commissioning the parameters of $\tilde{P}$ are updated after each transition is collected.

\begin{algorithm}[tb]
   \caption{PEARL}
   \label{alg: pearl}
\begin{algorithmic}[1]
\REQUIRE $\mathcal{D}$, $\tilde{P}$: empty dataset, dynamics model \\
~~~~~~~~~~~$\mu^{0}, \sigma^{0}$, $M$: $\pi_{\text{MPPI}}$ params\\
~~~~~~~~~~~$C, U, H$: commissioning steps, action sequences, planning horizon\\
\FOR{step $t=0...T$}
\FOR{iteration $n=1...N$}
\FOR{action sequence $u=1...U$}
\STATE{$a_{t:t+H-1}^u \sim \pi$} ~~~~~~~~~~~~~~~~~~~~~~~~~~~~~~~~~~~~~~~~~~~~~~~~~~~{\color{CadetBlue}$\vartriangleleft$ \emph{Sample actions}}
\STATE {$s_{t+H}^p = \tilde{P}(s_t^m, a_{t:t+H-1}^i)_\theta \; \forall \; m \in M$} ~~~~~~~~~~~~~~~~~~~~~~~~~~~~~~~~~~~~~~~{\color{CadetBlue}$\vartriangleleft$ \emph{Plan}}
\IF{$t \leq C$}
\STATE {\color{OliveGreen}$V_\Gamma = Var_{\tilde{P}} \left[\sum_{t=1}^H R(s_t, a_t) \right]$}
\STATE {\color{OliveGreen}$\mu^{n+1}, \sigma^{n+1}\leftarrow\mu^n(V_\Gamma), \sigma^n(V_\Gamma)$} ~~~~~~~~~~~~~~~~~~~~~~~~~~~~~{\color{CadetBlue} $\vartriangleleft$ \emph{Equation \ref{equation: mppi update}}}
\STATE {\color{OliveGreen} update $\tilde{P}$ given $\mathcal{D}$} ~~~~~~~~~~~~~~~~~~~~~~~~~~~~~~~~~~~~~~~~~~~~~{\color{CadetBlue}$\vartriangleleft$ \emph{Equation \ref{equation: loss}}}
\ELSE
\STATE {\color{BrickRed} $G_\Gamma = \mathbb{E}_{P} \left[\sum_{t=1}^H R(s_t, a_t) \right]$}
\STATE {\color{BrickRed} $\mu^{n+1}, \sigma^{n+1}\leftarrow\mu^n (G_\Gamma), \sigma^n(G_\Gamma)$} ~~~~~~~~~~~~~~~~~~~~~~~~~~~~~~{\color{CadetBlue} $\vartriangleleft$ \emph{Equation \ref{equation: mppi update}}}
\ENDIF
\ENDFOR
\STATE $a_{t} \leftarrow \mu_{t}^{N}$
\STATE $\mathcal{D} \leftarrow (s_t, a_t, s_{t+1})$
\ENDFOR
\IF{end of day}
\STATE {\color{BrickRed} update $\tilde{P}$ given $\mathcal{D}$} ~~~~~~~~~~~~~~~~~~~~~~~~~~~~~~~~~~~~~~~~~~~~~~~~~~~~~{\color{CadetBlue}$\vartriangleleft$ \emph{Equation \ref{equation: loss}}}
\ENDIF
\ENDFOR
\end{algorithmic}
\end{algorithm}

\section{Setup} \label{chapter5: experiments}

\textbf{Environments} $\;$ We evaluate the performance of our proposed approach using \textit{Energym}, an open-source building simulation library for benchmarking smart-grid control algorithms \cite{scharnhorst2021}. \textit{Energym} provides a Python interface for ground-truth building simulations designed in \textit{EnergyPlus} \cite{energyplus2001}, and presents buildings with varied equipment, geographies, and structural properties. We perform experiments in the following three buildings:
\begin{description}
    \item \textbf{\textit{Mixed-Use}} facility in Athens, Greece. 566m$^2$ surface area; 13 thermal zones; $\mathcal{A} \in \mathbb{R}^{12}$ and $\mathcal{S} \in \mathbb{R}^{37}$. Temperature setpoints and air handling unit (AHU) flowrates are controllable.
    \item \textbf{\textit{Office}} block in Athens, Greece. 643m$^2$ surface area; 25 thermal zones; $\mathcal{A} \in \mathbb{R}^{14}$ and $\mathcal{S} \in \mathbb{R}^{56}$. Only temperature setpoints are controllable.
    \item \textbf{\textit{Seminar Centre}} in Billund, Denmark. 1278m$^2$ surface area; 27 thermal zones; $\mathcal{A} \in \mathbb{R}^{18}$ and $\mathcal{S} \in \mathbb{R}^{59}$. Only temperature setpoints are controllable.
\end{description}
In all cases, environment states are represented by a combination of temperature, humidity and pressure sensors (among others). Full state and action spaces for each building are reported in Appendix \ref{appendix: energym environments}. Weather and grid carbon intensity match the true data in each geography for this period. 

\textbf{Baselines} $\;$ We baseline against 5 other controllers: 
\begin{description}
    \item \textbf{Soft Actor Critic} (SAC) \cite{Haarnoja2018}, a state-of-the-art model-free algorithm. 
    \item \textbf{Proximal Policy Optimisation} (PPO) \cite{schulman2017}, a popular model-free algorithm in production and used in previous works by \citet{ding2020} and \citet{zhang2019} as a baseline.
    \item \textbf{MPC with Deterministic Neural Networks} (MPC-DNN) \cite{nagabandi2018}, a simple, high-performing model-based architecture. Varying implementations have been used by previous authors, notably \citet{ding2020} and \citet{zhang2019}. We use the original implementation by \citet{nagabandi2018}.
    \item \textbf{RBC}, a generic, bang-bang controller found in most heating/cooling equipment that follows the heuristics outlined in Appendix \ref{appendix: RBC implementation}.
    \item \textbf{Oracle}, an SAC agent with hyperparameters fit to each environment using Bayesian Optimsation in Weights and Biases \cite{wandb}, and \textbf{pre-trained} in each building simulation for 10 years prior to test time. 
\end{description}
Both model-based agents plan with the same number of candidate actions over the same time horizon $H$. Full hyperparameter specifications are provided in the Appendix \ref{appendix: implementations}, PEARL's trajectory sampling and MPPI hyperparameters follow implementations by \citet{chua2018} and \citet{hansen2022} respectively.

\section{Results} \label{chapter5: discussion}
\begin{figure}[!ht]
    \centering
    \includegraphics[width=\textwidth]{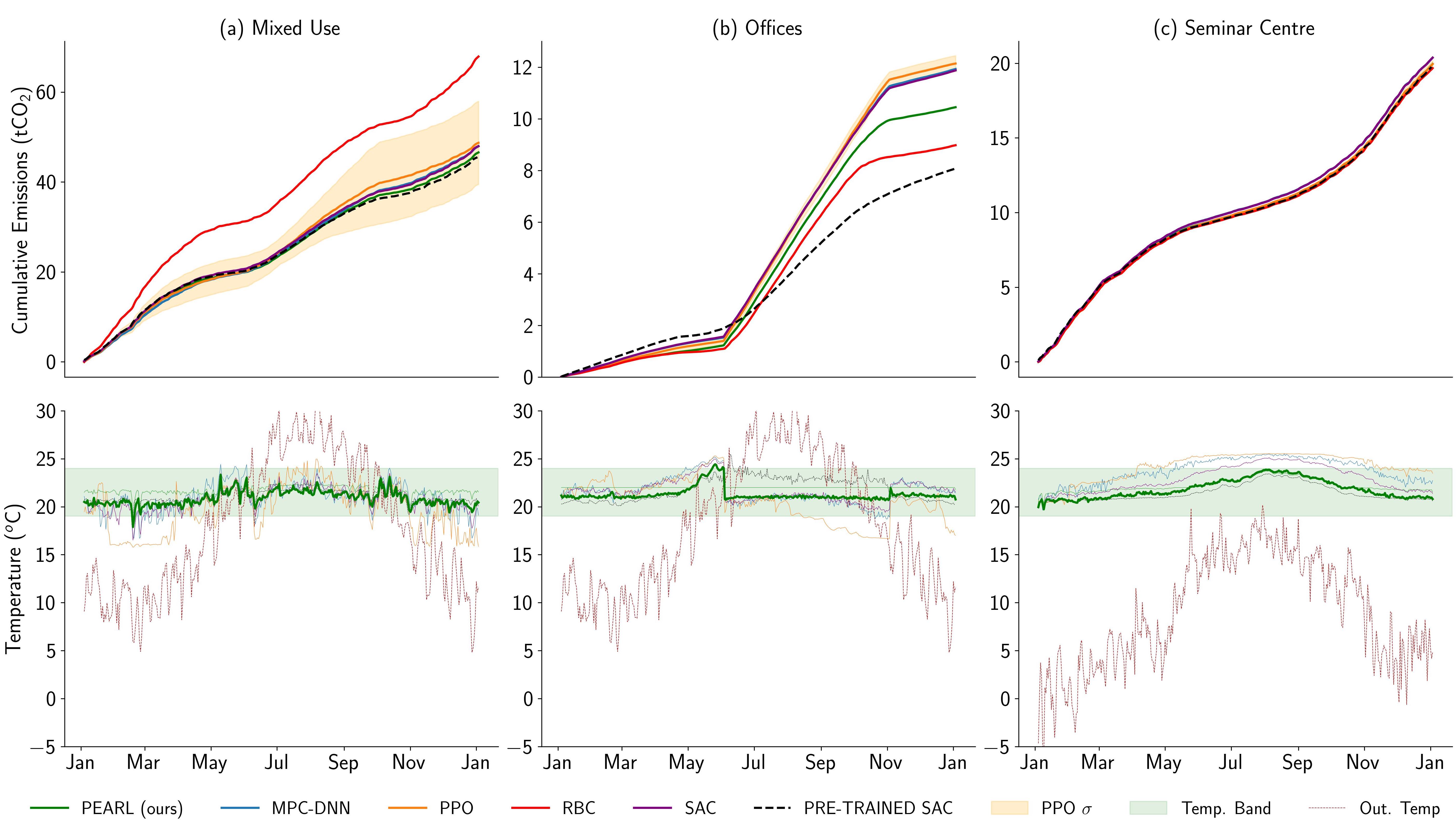}
    \caption{\textbf{Energym performance.} Top: Cumulative emissions produced by all agents across the (a) \textit{Mixed Use}, (b) \textit{Offices}, and (c) \textit{Seminar Centre} environments. Curves represent the mean of 3 runs of the experiment, shaded areas are one standard deviation (too small to see in all cases except PPO). Bottom: Mean daily building temperature produced by all agents, the green shaded area illustrates the target temperature range [19, 24].}
    \label{fig: chapter5/emissions}
\end{figure}
\begin{table}
\caption{\label{table: chapter5/results} \textbf{Energym performance}. Results for all agents across our three \textit{Energym} environments. We define the temperature infraction metric as the percentage of days where mean building temperature falls outside the target range [19, 24], and latency as the mean compute time each agent requires to select an action given its policy measured in seconds per action. Results are averaged across 3 runs and presented as mean $\pm$ standard deviation, except for the Oracle which has converged on a policy prior to deployment with multiple runs showing the same performance.}
\centering
\scalebox{0.8}{
\begin{tabular}{llllll}
\toprule
                             & Agent   & Emissions (tCO$_2$) & Temp. Infractions &  Latency & Reward \\
\midrule
\multirow{5}{*}{\rotatebox[]{90}{Mixed-Use}}   & RBC     &  $68.09 \text{\tiny{$\pm$0.00}}$ & $2.47\% \pm \text{\tiny{$\pm$0\%}}$   & — & —                   \\
                             & PPO     &   $48.80 \pm \text{\tiny{$\pm$10.35}}$  & $48.49\%\pm \text{\tiny{$\pm$6.08\%}}$ & $\textbf{0.021} \pm \text{\tiny{$\pm$0.01}}$  & $-1.47e7 \pm \text{\tiny{$\pm$2.84e6}}$ \\
                             & SAC & $48.80 \pm \text{\tiny{$\pm$0.14}}$ & $\mathbf{0\%}\pm \text{\tiny{$\pm$0\%}}$ &  $0.028 \pm \text{\tiny{$\pm$0.02}}$  & $-6.02e5 \pm \text{\tiny{$\pm$3.85e4}}$ \\
                             & MPC-DNN &    $48.03 \pm \text{\tiny{$\pm$0.46}}$ & $13.42\%\pm \text{\tiny{$\pm$0.59\%}}$ &  $0.030 \pm \text{\tiny{$\pm$0.01}}$ & $-1.12e6 \pm \text{\tiny{$\pm$3.47e4}}$ \\
                             & \textbf{PEARL (ours)} & $\mathbf{46.67} \pm \text{\tiny{$\pm$0.09}}$ & $0.55\%\pm \text{\tiny{$\pm$0.08\%}}$ & $0.870 \pm \text{\tiny{$\pm$0.15}}$ & $\mathbf{-5.76e5} \pm \text{\tiny{$\pm$2.12e3}} $\\
                             \midrule
                             & Oracle & $45.49$  & $0\%$ & $0.027$  &  $-4.77e5$ \\
\midrule
\multirow{5}{*}{\rotatebox[]{90}{Office}}   & RBC     &    $\mathbf{9.61} \pm \text{\tiny{$\pm$0.00}}$ & $1.64\% \pm \text{\tiny{$\pm$0\%}}$& —     & —             \\
                             & PPO     & $12.14 \pm \text{\tiny{$\pm$0.31}}$ & $31.51\%\pm \text{\tiny{$\pm$7.19\%}}$ & $\mathbf{0.018} \pm \text{\tiny{$\pm$0.006}}$ & $-2.26e6 \pm \text{\tiny{$\pm$1.07e6}}$ \\
                             & SAC & $11.87 \pm  \text{\tiny{$\pm$0.01}}$ & $6.58\% \pm \text{\tiny{$\pm$1.12\%}}$  & $0.025 \pm \text{\tiny{$\pm$0.01}}$ & $-2.75e5 \pm \text{\tiny{$\pm$2.03e4}}$ \\
                             & MPC-DNN & $11.93 \pm \text{\tiny{$\pm$0.022}}$ & $9.86\%\pm \text{\tiny{$\pm$0.84\%}}$ & $0.029 \pm \text{\tiny{$\pm$0.001}}$ & $-5.50e5 \pm \text{\tiny{$\pm$2.89e4}}$ \\
                             & \textbf{PEARL (ours)}   &  $10.45 \pm \text{\tiny{$\pm$0.07}}$ &  $\mathbf{1.52\%} \pm \text{\tiny{$\pm$0\%}}$ & $0.845 \pm \text{\tiny{$\pm$0.14}}$ & $\mathbf{-5.51e4} \pm \text{\tiny{$\pm$1.82e3}}$ \\
                             \midrule
                             & Oracle & $8.08$  & $2.47\%$ & $0.023$   &  $-2.63e4$ \\
\midrule
\multirow{5}{*}{\rotatebox[]{90}{Sem. Centre}} & RBC     & $\mathbf{19.74} \pm \text{\tiny{$\pm$0.00}}$ &  $\mathbf{0\%} \pm \text{\tiny{$\pm$0\%}}$&  —   & —             \\
                             & PPO     &   $20.01 \pm \text{\tiny{$\pm$0.27}}$ &  $51.23\%\pm \text{\tiny{$\pm$14.99\%}}$  &  $\mathbf{0.028} \pm \text{\tiny{$\pm$0.01}}$ & $-4.15e6 \pm \text{\tiny{$\pm$9.67e5}}$ \\
                             & SAC & $20.37 \pm \text{\tiny{$\pm$0.08}}$ & $29.32\%\pm \text{\tiny{$\pm$1.87\%}}$  & $0.033 \pm \text{\tiny{$\pm$0.02}}$  & $-1.95e6 \pm \text{\tiny{$\pm$7.11e4}}$ \\
                             & MPC-DNN &  $20.44 \pm \text{\tiny{$\pm$0.02}}$    & $49.13\%\pm \text{\tiny{$\pm$0.56\%}}$  & $0.035 \pm \text{\tiny{$\pm$0.001}}$  & $-2.45e6 \pm \text{\tiny{$\pm$3.26e4}}$ \\
                             & \textbf{PEARL (ours)}   &    $20.02 \pm \text{\tiny{$\pm$0.10}}$ & $\mathbf{0\%}\pm \text{\tiny{$\pm$0\%}}$  & $0.911 \pm \text{\tiny{$\pm$0.17}}$  & $\mathbf{-1.18e6} \pm \text{\tiny{$\pm$1.50e3}}$ \\
                             \midrule
                             & Oracle & $19.75$  & $0\%$ & $0.031$  &  $-1.14e6$ \\
\bottomrule 
\end{tabular}
}
\end{table}
Table \ref{table: chapter5/results} reports key metrics for our six controllers across our \textit{Energym} environments. Results are reported as the mean $\pm$ standard deviation for 3 runs of each experiment. The RBC performs identically across all experiments because both its policy and the environment are deterministic; varying RL agent performance is a consequence of policy and initialisation stochasticity.

\begin{figure}[t!]
    \centering
    \includegraphics[width=0.6\textwidth]{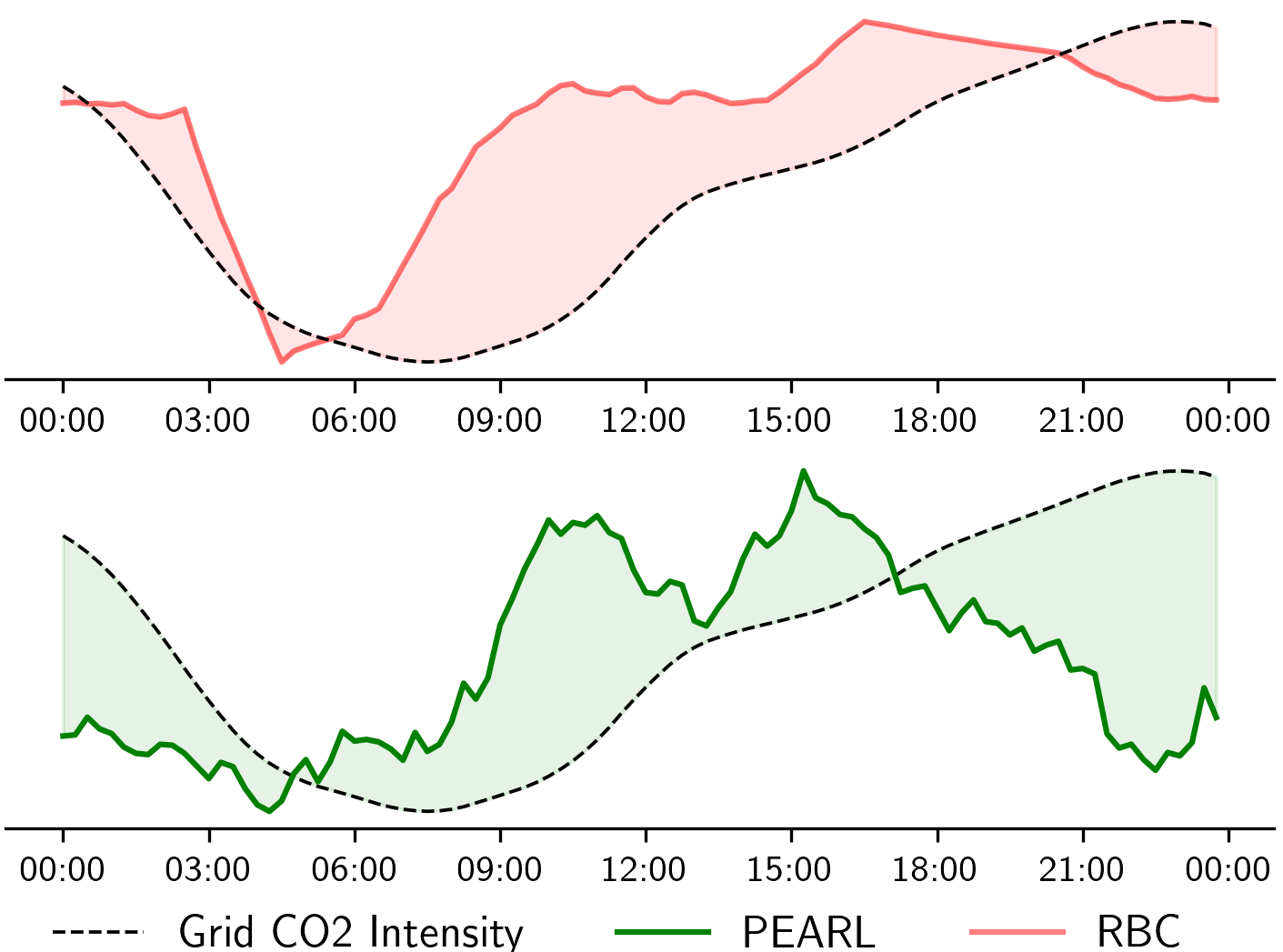}
    \caption{\textbf{Load shifting.} Power consumption for the RBC (top) and PEARL (bottom) on an exemplar day in the \textit{Office} environment, against grid carbon intensity. We wish to maximise the shaded area to minimise emissions. PEARL minimises power draw in the early morning and late evening when grid carbon intensity is highest.}
    \label{fig: chapter5/load-shifting}
\end{figure}

\textbf{Emissions.} We find PEARL produces minimum emissions in the \textit{Mixed-Use} environment, with cumulative emissions 31.46\% lower than the RBC. In the \textit{Office} block, the RBC exhibits lowest cumulative emissions, and PEARL outperforms all RL baselines. In the \textit{Seminar Centre}, the RBC minimises emissions, and PPO marginally outperforms PEARL, but does so at the cost of erroneous temperature control. Low external temperatures in the \textit{Seminar Centre} make experiments there less informative as the optimal policy is to heat most of the year, providing little room for improved control. Stepwise emission totals for each environment are illustrated in the top half of Figure \ref{fig: chapter5/emissions}. We provide an illustrative example of PEARL showing an ability to load shift in Figure \ref{fig: chapter5/load-shifting}.

\textbf{Temperature.} We find that PEARL produces minimum daily mean temperature infractions in the \textit{Office} and \textit{Seminar Centre} environments, and is slightly outperformed by SAC in the \textit{Mixed-Use} environment. The RBC is comparably performant across all environments, as would be expected. The remaining RL baselines miss the thermal bounds regularly, with PPO and MPC-DNN exhibiting temperature infraction rates as high as 51.23\% and 49.13\% respectively. In some cases the strong emissions performance of these baselines is a direct consequence of shutting off HVAC equipment and forcing uncomfortable internal temperatures. Mean daily building temperatures for each agent across the environments are plotted in the lower half of Figure \ref{fig: chapter5/emissions}.

\textbf{Latency.} PPO is the lowest-latency (mean compute time per action) controller, selecting actions in two thirds of the time required by MPC-DNN, and 41 times faster than PEARL on average, but we note the latency of all agents is far smaller than the sampling period of each environment, meaning all implementations would prove adequate for real-world deployment. Were they deployed in situ, the model-based agents could utilise the time between environment interactions fully to plan with greater numbers of action sequences which we expect would improve performance.

\textbf{Reward.} Mean annual reward captures an agent's ability to minimise emissions \textit{and} maintain thermal comfort. PEARL exhibits maximum mean reward across all environments suggesting it strikes this balance better than the other controllers. The reward curves for each agent are reported in Appendix \ref{learning curves appendix} for brevity.

\textbf{Oracle comparison.} The pre-trained oracle outperforms the baselines and PEARL in all cases as expected. However, its performance is surprisingly close to PEARL's, showing only 2.5\% and 1.3\% lower emissions in the \textit{Seminar Centre} and \textit{Mixed-Use} environments, and exhibiting similar thermal performance.

\section{Discussion}

\textbf{Why is \textit{Mixed-Use} performance an outlier?} An important observation from Table \ref{table: chapter5/results} is that PEARL only outperforms the RBC in the \textit{Mixed-Use} facility. Why would this be the case? The outcome of the \textit{Seminar Centre} results has been discussed above, but one may expect PEARL to perform as well in the \textit{Office} environment as it does in the \textit{Mixed-Use} environment. Unlike the \textit{Office} environment, in the \textit{Mixed-Use} facility the agent has access to thermostat setpoints \textit{and} continuous AHU flowrate control. This greatly increases action-space complexity and moves the control problem away from a setting where simple heuristics can be readily applied. One might infer, then, that RL building controllers should only be deployed when the action space is sufficiently complex, like when they have access to continuous control parameters.
\begin{figure}[t!]
    \centering
    \includegraphics[width=0.6\textwidth]{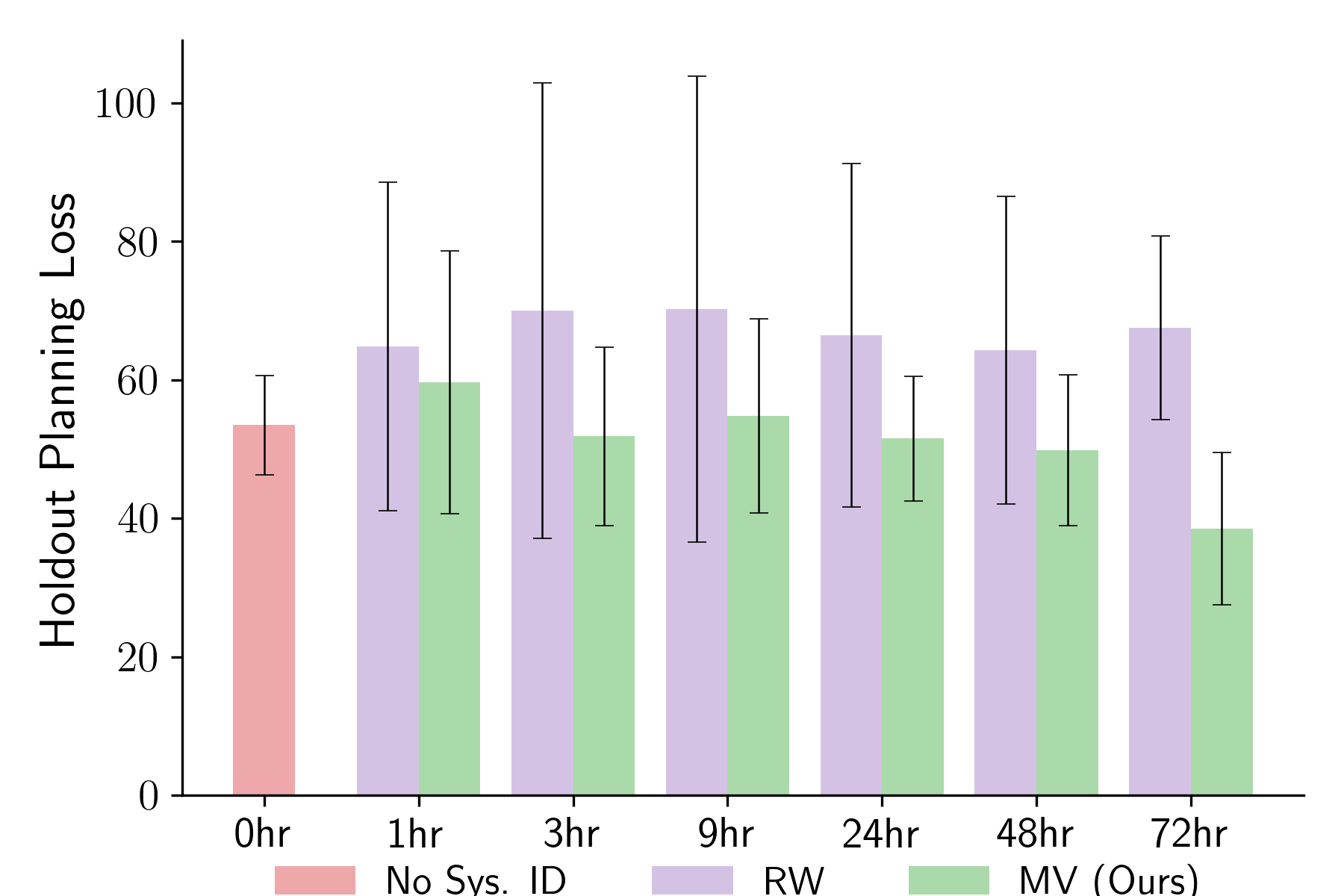}
    \caption{\textbf{System ID.} Planning MSE post-commissioning on a holdout set of 100 randomly sampled state-action trajectories, given varying system ID duration. Black bars represent one standard deviation across three runs.}
    \label{fig: chapter5/system id}
\end{figure}

\textbf{System identification.} $\;$ We test the sensitivity of PEARL's performance to system ID duration. We vary the commissioning period at seven intervals between 0 (no system ID) and 72 hours, and test predictive accuracy on a holdout set of 100 randomly sampled state-action trajectories produced by another controller in the same environment. We compare our method (MV) with \citet{LAZIC2018}'s Random Walk (RW) and plot the results in Figure \ref{fig: chapter5/system id}—see Appendix \ref{appendix: lazic} for \citet{LAZIC2018}'s exploration policy.

We find that the accuracy of models fitted via MV system ID correlates with commissioning period length, as would be expected. We observe that, in expectation, models fit using data collected via \citet{LAZIC2018}'s RW perform no better than randomly initialised networks. Using MV exploration, model accuracy is noticeably better than random only after 72 hours of system ID. However, we note from Figure \ref{fig: chapter5/emissions}, that an agent with 3 hours of system ID time remains capable of reducing emissions and maintaining thermal comfort despite poorer predictive accuracy.

\textbf{Agent decomposition.} $\;$ What components of PEARL enable performant control? We note two differences between PEARL and our model-based RL baseline MPC-DNN: 1) The use of probabilistic networks, rather than deterministic networks, and 2) The use of MPPI planning, rather than random shooting (RS). To understand their relative importance, we vary the design of PEARL to either include or exclude these components and compare performance—Table \ref{table: chapter5/agent decomposition}.
\begin{table} 
\centering
\caption{\label{table: chapter5/agent decomposition}\textbf{Agent decomposition.} Mean daily reward for four instantiations of PEARL varying the choice of network and planning algorithm. The \textit{Oracle} is reported as a baseline. Experiments conducted in the Mixed-Use environment for one year.}
\begin{tabular}{lll}
\toprule
                             & Deterministic   & Probabilistic  \\
\midrule
Random Shooting (RS) & $-7.11e5$ & $-7.30e5$   \\
MPPI & $\mathbf{-5.28e5}$  & $-5.65e5$   \\
\midrule
Oracle & $-4.77e5$ \\
\bottomrule
\end{tabular}
\end{table}
We find the performance of PEARL to be sensitive to the choice of planner and, to our surprise, insensitive to the choice of network, with agents composed of deterministic and probabilistic networks performing similarly when planner choice is controlled for. This is in contradiction to many works suggesting probabilistic modelling of dynamics function improves performance over deterministic models, particularly in complex, partially-observed state-action spaces like those exhibited in this study \cite{deisenroth2011pilco, chua2018, levine2018}.

\section{Conclusion}
In this Chapter we questioned whether it was possible to deploy an agent on a real-world task \textit{without prior data}. We considered the task of learning policies \textit{tabula rasa} that minimise emissions in buildings whilst ensuring thermal comfort, a considerably harder task than pre-training models in simulation before deployment. We have proposed PEARL (\textbf{P}robabilistic \textbf{E}mission-\textbf{A}bating \textbf{R}einforcement \textbf{L}earning), and shown it can reduce emissions from buildings by up to 31.46\% when compared with an RBC, by fitting polices \textit{online} without pre-training in simulation. When compared with existing RL baselines, our algorithm performs favourably, showing reduced emissions in all cases bar one, whilst maintaining thermal comfort more effectively. Our approach is simple to commission, requiring no historical data or simulator access \textit{a priori}, and capable of generalising across varied building archetypes. The scaled deployment of such systems could prove effective climate change mitigation tools, and represent a step toward managing the third of our three thesis constraints (\colouredS \ref{chapter1: path forward}). 

\chapter{Outlook}\label{chapter: outlook}
\textit{``This is just the beginning.''}
\begin{flushright}
    —David Silver, August 2024
\end{flushright}
\vspace{0.5in}
Modern Reinforcement Learning (RL) systems capture deep truths about general, human problem-solving. Despite their promise, the most impressive applications have come in domains that can be \textit{perfectly simulated} (\colouredS \ref{chapter1: unfulfilled promises}), allowing the system to collect as much data as is necessary to uncover useful decision-making policies. Society faces problems in which RL agents could find solutions that humans cannot, but these are often in domains which we cannot perfectly simulate. In such scenarios, we can \textit{learn} simulators from data collected from these domains, but these will only ever be approximately correct, and can be pathologically incorrect when queried outside of their training distribution. As a result, a misalignment between the environments in which we train our agents and the real-world in which we deploy our agents is inevitable.

In this work we embraced this inevitability head-on. Dealing with the misalignment is the primary concern of \textit{zero-shot reinforcement learning} research \citep{kirk2023survey, touati2022does}. Whilst impressive steps have been taken in producing general agents capable of solving tasks that are different to those seen during training \citep{borsa2018, touati2021learning}, it is our belief that new efforts are required to equip such agents with methods for solving \textit{real-world} problems. In particular, we believe real-world problems exhibit (at least) three properties that are as yet under-discussed (\colouredS \ref{section: thesis structure and contr.}). The first is the \textit{data quality constraint}. Any agent tasked with solving a real-world problem cannot expect to be trained on large heterogeneous datasets, as is the prevailing assumption. The second is the \textit{observability constraint}. Any agent tasked with solving a real-world problem cannot expect access to states that provide all the information required to solve a task. And the third is the \textit{data availability constraint}. Any agent tasked with solving a real-world problem cannot always assume \textit{a priori} access to a simulator, or a dataset from which to learn a simulator. This work sought to address each of these constraints, and consequently advance the following thesis:
\begin{hypbox}{Thesis}{ 
    It is possible to build RL agents that solve unseen tasks zero-shot, whilst satisfying the \textit{data quality, observability} or \textit{data availability} constraints.
    }
\end{hypbox}

In Chapter \ref{chapter: from low quality data}, we took steps to manage the \textit{data quality} constraint. We introduced \textit{conservative} zero-shot RL methods which address the failings of existing zero-shot RL methods when trained on realistic, real-world datasets. We exposed the critical failure mode of current methods in this setting, namely, value overestimation for state-action pairs not present in the dataset. Informed by regularisation techniques from the offline RL literature, we biased their \textit{value} and \textit{measure} predictions so they were conservatively low at those state-actions, showing this greatly improves performance on standard zero-shot RL benchmarks.

In Chapter \ref{chapter: under partial observability}, we proposed methods that handle the \textit{observability} constraint. We introduced \textit{memory-based} zero-shot RL methods which mitigate state misidentification and task misidentification--the failure mode of memory-free zero-shot RL methods trained on imperfect observations. We explored the effectiveness of zero-shot RL methods equipped with RNN-based, state space-based, and attention-based memory models, and found that GRUs perform best in environments where the states, rewards and a change in dynamics are partially observed, greatly improving the performance of memory-free methods.

In Chapter \ref{chapter: no prior data}, we explored the \textit{data availability} constraint in the context of emission-efficient building control. We introduced PEARL: Probabilistic Emission-Abating Reinforcement Learning, which reduced emissions from buildings by up to 30\% with no prior knowledge, and only 180 minutes of pre-deployment data collection. We showed that our active exploration method improved the prediction accuracy of our learned model when compared with the exploration method of the previous best zero-shot building control method.

Viewed together, these proposals make progress toward our goal of building a general agent that solves real-world problems, but many challenges remain. Our conservative zero-shot RL algorithms (\colouredS \ref{chapter: from low quality data}) improve performance when trained on low-quality datasets, but their absolute performance is far below that of vanilla zero-shot RL algorithms trained on high-quality datasets (cf. Table \ref{table: chapter3/full 100k results} versus Table H.1 in \citep{touati2022does}). Our memory-based zero-shot RL methods (\colouredS \ref{chapter: under partial observability}) can generalise to unseen tasks under unseen dynamics, but they still expect data to be collected on their behalf by many agents experiencing different dynamics with highly exploratory behaviour policies. And our zero-shot building control agent can learn a model of building physics after a 180-minute data collection period that is good enough to enable emission-efficient control, but we cannot expect this to hold for more complicated systems that require many more data to model their physics.

\section{A World We Can(not) Simulate}
Just how relevant will this work remain in the future? To provide a speculative answer, we conclude this work by returning to the assumption upon which it is predicated. In Chapter \ref{chapter: introduction}, we argued that if we want to train agents to solve real-world problems they will need to solve them with data sampled from \textit{learned} simulators, because \textit{perfect simulators} are too expensive to build for most problems. We argued that these learned simulators will only ever be approximately correct, and as a result, agents must be able to handle an inevitable misalignment between their training and deployment environments. In other words, we assumed the real-world \textit{cannot be fully simulated}. Clearly this is true today. Will it remain true indefinitely?

\begin{figure}[t]
    \centering
    \includegraphics[width=\textwidth]{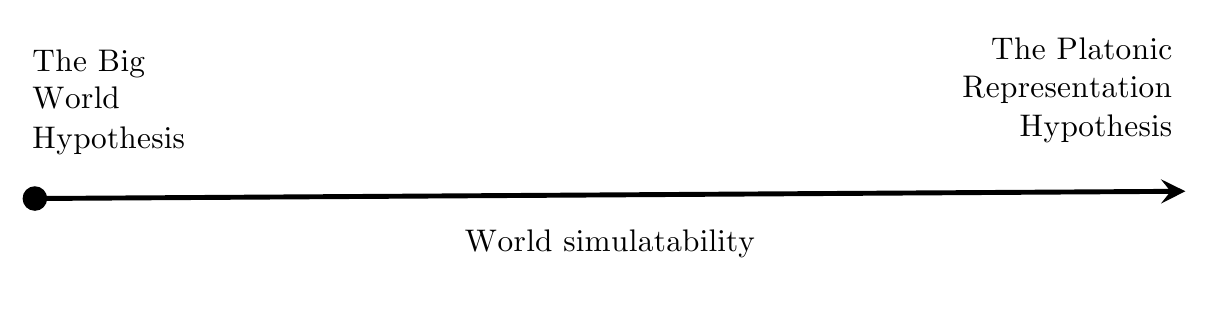}
    \caption{\textbf{How accurately could we simulate the world?} Two hypotheses have emerged. At one end lies the \textit{big world hypothesis} which argues the world will always be too large and complex to be modelled accurately \citep{javed2024the}. At the other end lies the \textit{platonic representation hypothesis} which argues that deep networks will continue to converge toward a shared statistical model of reality \citep{huh2024platonic}.}
    \label{fig: outlook/world simulatability}
\end{figure}

The community is unsure, and two opposing hypotheses have emerged, as summarised in Figure \ref{fig: outlook/world simulatability}. The first view is that the world will always be too large and complex for an agent to fully simulate. It is summarised by Javed and Sutton's \textit{big world hypothesis}:
\begin{emphasisbox}
    \textbf{The Big World Hypothesis \citep{javed2024the}.} In many decision-making problems the agent is orders of magnitude smaller than the environment. It can neither fully perceive the state of the world nor can it represent the value or optimal action for every state. Instead, it must learn to make sound decisions using its limited understanding of the environment.
\end{emphasisbox}
Most would agree that this holds for most real problems today, and Javed and Sutton suggest that it should hold moving forward in spite of exponentially growing compute. Their core argument is that as more compute becomes available for training larger simulators, the world itself becomes concurrently more complex. This may be true for (at least) two reasons. First, more compute means we can sense the world at higher fidelity\footnote{For example, today's smartphone cameras can generate more data in a week than was used to train GPT-3 \citep{brown2020language}.}. The higher the resolution of our sensors, the less easy it is to faithfully simulate the data they generate. Second, more compute means increasingly complex agents will be deployed which themselves will need to be simulated. To simulate a game of Go against a beginner, they argue, requires only that we model their policy as pseudo-random; to simulate a game of Go against AlphaZero \citep{silver2018general} requires us to fully model its complex policy. 

Much historic work implicitly sides with the big world hypothesis. If true, ideas from continual learning remain particularly relevant \citep{kirkpatrick2017overcoming, rolnick2019experience, hadsell2020embracing}. If an agent cannot compress the world's dynamics as described by a world simulator into its weights, then it will need to make decisions about what knowledge to maintain, and what to forget, through time. Equally relevant will be those related to \textit{partial models} \citep{talvitie2008simple}. If the world cannot be simulated at all states, then decisions will need to be made about what states deserve priority over others.

The second view is that there exists an underlying function producing reality, and large models will continue to approximate it with increasing accuracy. It is summarised by Huh et al.'s \textit{platonic representation hypothesis}:   
\begin{emphasisbox}
    \textbf{The Platonic Representation Hypothesis \citep{huh2024platonic}}. Neural networks, trained with different objectives on different data and modalities, are converging to a shared statistical model of reality in their representation spaces.
\end{emphasisbox}
Huh et al. observe that as the resources (parameters, data, compute) used to train modern vision and language models are scaled, their internal representations are becoming increasingly alike. They offer three justifications for this convergence. The first is that as we ask larger models to solve ever-more tasks, the space of functions that satisfy their solutions will shrink. The second is that bigger models overlap more in function space, and so are more likely to converge to the same optimal representation if it exists\footnote{Smaller models are more likely to find \textit{different} representations that converge on near-optimal representations.}. The third is that deep networks are biased toward finding simple functions that explain the data, and as they get bigger, we should expect convergence to a smaller solution space. 

Much recent work implicitly sides with the platonic representation hypothesis. If true, ideas from open-endedness are particularly relevant \citep{hughes2024open, stanley2015greatness, stanley2017open}. There, the implicit goal is to create, or evolve,  an open-ended world simulator that reflects the complexity of reality. Early efforts have been made for game-playing \citep{bruce2024genie, chiappa2017recurrent}, autonomous-driving \citep{hu2023gaia} and robotics \citep{yang2023learning, eslami2018neural}. With these, agents could then be trained over an increasingly complex curricula of tasks to learn an increasingly complex set of skills \citep{jiang2021prioritized, jiang2021replay, parker2022evolving, wang2019paired, samvelyan2023maestro}. If the platonic representation hypothesis holds, this curricula is likely to converge on the underlying function generating reality, and any agent trained in it should zero-shot transfer to the real world automatically.

A future consistent with the big world hypothesis necessitates further work on methods like those developed in Chapter \ref{chapter: no prior data}, which would remain relevant. In such a future, methods would continue to require approximate dynamics models for short-horizon planning, knowing that attempting anything more ambitious is futile (\colouredS \ref{chapter: no prior data}). A future consistent with the platonic representation hypothesis necessitates further work on methods like those developed in Chapters \ref{chapter: from low quality data} and \ref{chapter: under partial observability}. A valid current concern with these methods is that they are trying to model too much (\textit{i.e.} the successor measure from every state-action pair to every other state in the environment (\colouredS \ref{background: multi-step MBRL})). But if the plantonic representation hypothesis holds, then these methods could potentially scale to ever larger domains.

We leave it to the reader to decide which future we may inhabit. In the short run, the world will remain big, and agents will need to contend with situations they were not trained to encounter, just like humans do. For that, this thesis makes proposals that may help. In the long run, the world is likely to get smaller. How small it gets, and what that will mean for humanity, is unclear. 
\begin{spacing}{0.9}

% To use the conventional natbib style referencing
% Bibliography style previews: http://nodonn.tipido.net/bibstyle.php
% Reference styles: http://sites.stat.psu.edu/~surajit/present/bib.htm

\bibliographystyle{apalike}
\cleardoublepage
\bibliography{bibliography} % Path to your References.bib file

% If you would like to use BibLaTeX for your references, pass `custombib' as
% an option in the document class. The location of 'reference.bib' should be
% specified in the preamble.tex file in the custombib section.
% Comment out the lines related to natbib above and uncomment the following line.

%\printbibliography[heading=bibintoc, title={References}]

\end{spacing}
\clearpage
% ******* Appendices *********
\appendix

\chapter{Environments}\label{appendix: environments}
\section{ExORL}\label{appendix: exorl environements}
Across Chapters \ref{chapter: from low quality data} and \ref{chapter: under partial observability} we consider three locomotion and two goal-directed domains from the ExORL benchmark \citep{yarats2022} which is built atop the DeepMind Control Suite \citep{tassa2018}. Environments are visualised here: \url{https://www.youtube.com/watch?v=rAai4QzcYbs}. The domains are summarised in Table \ref{table: environments-appendix/exorl-summary}.\\

\noindent\textbf{Walker.} A two-legged robot required to perform locomotion starting from bent-kneed position. The state and action spaces are 24 (17 after occlusion) and 6-dimensional respectively, consisting of joint torques, velocities (removed in occluded) and positions. ExORL provides four tasks \texttt{stand, walk, run} and \texttt{flip.} The reward function for \texttt{stand} motivates straightened legs and an upright torso; \texttt{walk} and \texttt{run} are  supersets of \texttt{stand} including reward for small and large degrees of forward velocity; and \texttt{flip} motivates angular velocity of the torso after standing. Rewards are dense.\\

\noindent\textbf{Quadruped.} A four-legged robot required to perform locomotion inside a 3D maze. The state and action spaces are 78 (67 after occlusion) and 12-dimensional respectively, consisting of joint torques, velocities (removed in occluded) and positions. ExORL provides five tasks \texttt{stand, roll, roll fast, jump} and \texttt{escape.} The reward function for \texttt{stand} motivates a minimum torse height and straightened legs; \texttt{roll} and \texttt{roll fast} require the robot to flip from a position on its back with varying speed; \texttt{jump} adds a term motivating vertical displacement to stand; and \texttt{escape} requires the agent to escape from a 3D maze. Rewards are dense.\\

\noindent\textbf{Point-mass Maze.} A 2D maze with four rooms where the task is to move a point-mass to one of the rooms. The state and action spaces are 4 (2 after occlusion) and 2-dimensional respectively; the state space consists of $x, y$ positions and velocities (removed in occluded) of the mass, the action space is the $x, y$ tilt angle. ExORL provides four reaching tasks \texttt{top left, top right, bottom left} and \texttt{bottom right.} The mass is always initialised in the top left and the reward is proportional to the distance from the goal, though is sparse i.e. it only registers once the agent is reasonably close to the goal.\\

\noindent\textbf{Jaco.} A 3D robotic arm tasked with reaching an object. The state and action spaces are 55 and 6-dimensional respectively and consist of joint torques, velocities and positions. ExORL provides four reaching tasks \texttt{top left, top right, bottom left} and \texttt{bottom right.} The reward is proportional to the distance from the goal object, though is sparse i.e. it only registers once the agent is reasonably close to the goal object. \\

\noindent\textbf{Cheetah.} A running two-legged robot.  The observation and action spaces are 15 (10 after occlusion) and 6-dimensional respectively, consisting of positions of robot joints and velocities (removed in occluded). We evaluate on 4  tasks: \texttt{walk, walk backward, run} and \texttt{run backward.} Rewards are linearly proportional either a forward or backward velocity--2 m/s for walk and 10 m/s for run.

\begin{table}[b]
\caption{\textbf{ExORL domain summary.} \textit{Dimensionality} refers to the relative size of state and action spaces. \textit{Type} is the task categorisation, either locomotion (satisfy a prescribed behaviour until the episode ends) or goal-reaching (achieve a specific task to terminate the episode). \textit{Reward} is the frequency with which non-zero rewards are provided, where dense refers to non-zero rewards at every timestep and sparse refers to non-zero rewards only at positions close to the goal. {\color[HTML]{009901} Green} and {\color[HTML]{CB0000} red} colours reflect the relative difficulty of these settings.}
\label{table: environments-appendix/exorl-summary}
\centering
\begin{tabular}{llll}
\toprule
\textbf{Domain}                        & \textbf{Dimensionality}     & \textbf{Type}                        & \textbf{Reward}               \\
 \midrule
{\color[HTML]{000000} Walker}          & {\color[HTML]{009901} Low}  & {\color[HTML]{000000} Locomotion}    & {\color[HTML]{009901} Dense}  \\
Quadruped                              & {\color[HTML]{CB0000} High} & Locomotion                           & {\color[HTML]{009901} Dense}                         \\
 
{\color[HTML]{000000} Point-mass Maze} & {\color[HTML]{009901} Low}  & {\color[HTML]{000000} Goal-reaching} & {\color[HTML]{CB0000} Sparse} \\
Jaco                                   & {\color[HTML]{CB0000} High} & Goal-reaching                        & {\color[HTML]{CB0000} Sparse}                        \\
{\color[HTML]{000000} Cheetah}                                   & {\color[HTML]{009901} Low} & Locomotion                        & {\color[HTML]{009901} Dense}                       \\
\bottomrule
\end{tabular}
\end{table}

\subsection{POMDPs}\label{appendix: pomdps}
\begin{figure}[h]
\centering
    \includegraphics[width=0.8\textwidth]{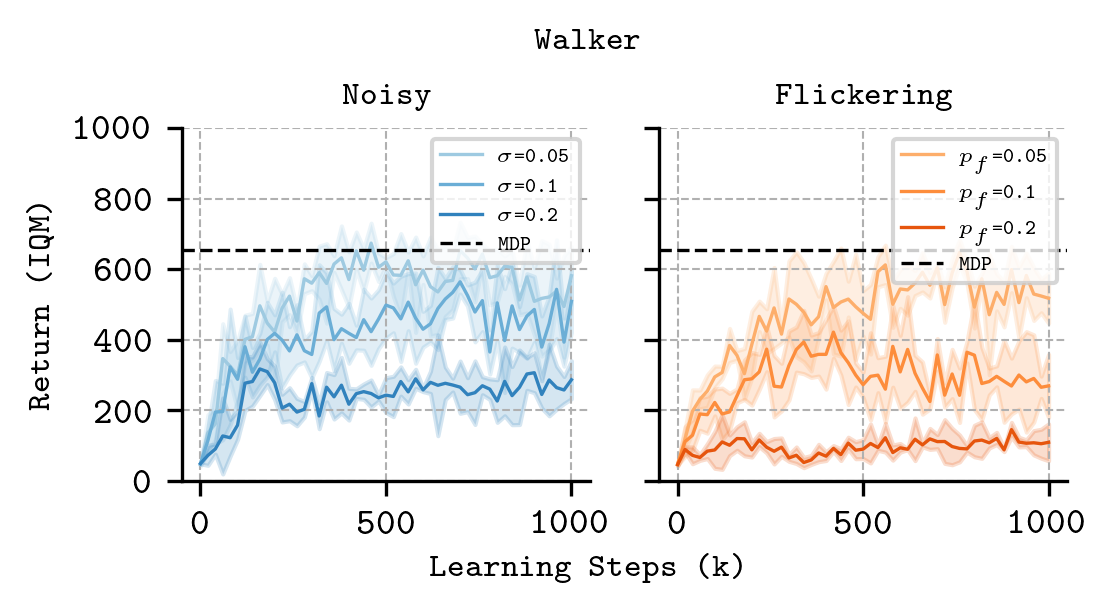}
    \caption{\textbf{POMDP hyperparameter sweep.} We evaluate the performance of standard FB on \texttt{Walker} when the states are noised according to $\sigma \in \{0.05, 0.1, 0.2\}$ and dropped according to $p_f \in \{0.05, 0.1, 0.2\}$.}
    \label{fig: appendix/pomdp hyperparameters}
\end{figure}

The \texttt{noisy} and \texttt{flickering} amendments to standard ExORL environments (\colouredS \ref{section: experiments}), Chapter \ref{chapter: under partial observability} have associated hyperparameters $\sigma$ and $p_f$. Hyperparameter $\sigma$ is the variance of the 0-mean Gaussian from which noise is sampled before being added to the state, and $p_f$ is the probability that state $s$ is dropped (zeroed) at time $t$. In Figure \ref{fig: appendix/pomdp hyperparameters} we sweep across three valued of each in $\{0.05, 0.1, 0.2\}$. From these findings we set $\sigma=0.2$ and $p_f = 0.2$ in the main experiments 

\section{D4RL}\label{appendix: d4rl environments}
In Chapter \ref{chapter: from low quality data} we consider two MuJoCo \citep{todorov2012} locomotion tasks from the D4RL benchmark \citep{fu2020}, which is built atop the v2 Open AI Gym \citep{brockman2016}. The below environment descriptions are taken from \citep{brockman2016}.\\

\noindent\textbf{Walker2D-v2}. A two-dimensional two-legged figure that consist of seven main body parts - a single torso at the top (with the two legs splitting after the torso), two thighs in the middle below the torso, two legs in the bottom below the thighs, and two feet attached to the legs on which the entire body rests. The goal is to walk in the in the forward (right) direction by applying torques on the six hinges connecting the seven body parts.\\

\noindent\textbf{HalfCheetah-v2.} A 2-dimensional robot consisting of 9 body parts and 8 joints connecting them (including two paws). The goal is to apply a torque on the joints to make the cheetah run forward (right) as fast as possible, with a positive reward allocated based on the distance moved forward and a negative reward allocated for moving backward.

\section{Energym}\label{appendix: energym environments}
The experiments in Chapter \ref{chapter: no prior data} are performed in \textit{Energym} an open-source python package providing a Gym-like interface for EnergyPlus \citep{energyplus2001} building simulations. This section provides detail on the reward function we use (\colouredS \ref{appendix: energym reward function}) and the state and action spaces of the building environments.

\subsection{Reward Function}\label{appendix: energym reward function}
The reward $R(s, a)$ is a linear combination of an \textit{emissions term} $\text{E}_t$ and a \textit{temperature term} $\text{Temp.}_t$ i.e. $R(s_t, a_t) = \text{E}_t + \text{Temp}_t$ which motivates minimal emissions whilst satisfying thermal comfort. Let the energy consumption in the environment at time $t$ be denoted with $\text{energy}_t$, and the grid carbon intensity at time $t$ be denoted with $\text{carbon}_t$, then the emissions-term reward is
\begin{equation}
	\text{E}_t = - \phi \left(\text{energy}_t \cdot \text{carbon}_t \right) \; ,
\end{equation}
where $\phi = 0.001$ is a parameter that sets the relative emphasis of emission-minimisation over thermal comfort. Let $\text{Temp}^i_t$ denote the temperature-term reward at timestep $t$ for thermal zone $i$, $T_{obs}^i$ denote the observed temperature in thermal zone $i$, and $T_{low}$ and $T_{high}$ denote the lower and upper temperature bounds on thermal comfort respectively. Then, the temperature reward is given by
\begin{equation} \label{eq: temperature reward}
 \text{Temp}^i_t = 
\begin{cases}
0: & T_{low} \leq T_{obs}^i \leq T_{high} \\
-\min \left[(T_{low} - T_{obs}^i[t])^2, (T_{high} - T_{obs}^i)^2 \right]: & \text{otherwise} \; , 
\end{cases}
\end{equation}
where the second term can be thought of as a \textit{penalty} that punishes the agent in proportion to deviations from the thermal comfort zone. The total temperature reward $\text{Temp}_t$ is obtained by summing the rewards across thermal zones i.e. $\sum_{i=1}^N \text{Temp}^i_t$.\\

\noindent\textbf{Grid carbon intensity.} The \textit{Mixed-Use} and \textit{Office} environments do not provide native support for optimisation of grid carbon intensity $\text{carbon}_t$. As such, we sourced grid carbon intensity data  from Electricity Maps, an open-source collection of global electrical grid data \cite{electricitymaps22}. Importantly, the data we sourced is for the same locale and year (Greece, 2017)  as the \textit{Energym} weather files to ensure any weather-grid dependencies are preserved. We append $\text{carbon}_t$ to the \textit{Energym} state spaces.

\subsection{Building Environments} \label{appendix: building environments}
This section reports a series of tables that describe the building environments used for the experiments in Chapter \ref{chapter: no prior data}. The \textit{Mixed-use} facility's state and action spaces are reported in Tables \ref{table: energym mixed state space} and \ref{table: energym mixed use action space}; the \textit{Office} building's state and action spaces are reported in Tables \ref{table: office state space} and \ref{table: energym office action space}; and the \textit{Seminar Centre}'s state and action spaces are reported in Tables \ref{table: seminar state space} and \ref{table: energym seminar action space}.

\begin{landscape}
\clearpage
\begin{longtable}[width=\textwidth]{ l l l l l }
			\caption{\textbf{\textit{Mixed-use} environment state-space}. Variables can be thought of as sensors installed in the building. \textit{Enerygym} clips their measurements if they fall outside lower or upper bounds.  \label{table: energym mixed state space}} \\
			\toprule
			\textbf{Variable} & \textbf{Type} & \textbf{Lower Bound} & \textbf{Upper Bound} & \textbf{Description} \\
			\midrule
			\endfirsthead
			
			\multicolumn{5}{l}{{\tablename\  \thetable{} -- continued from previous page}} \\
			
			\textbf{Variable}     & \textbf{Type} & \textbf{Lower Bound} & \textbf{Upper Bound} & \textbf{Description}\\
			\midrule
			\endhead
			
			\midrule
			\multicolumn{5}{r}{\textit{continued on next page}}
			\endfoot
			
			\bottomrule 
			\endlastfoot
			
			{\color[HTML]{212529} Bd\_T\_AHU1}                 & {\color[HTML]{212529} scalar} & {\color[HTML]{212529} 10}      & {\color[HTML]{212529} 30}       & {\color[HTML]{212529} AHU 1 temperature (°C).}           \\
{\color[HTML]{212529} Bd\_T\_AHU2}                 & {\color[HTML]{212529} scalar} & {\color[HTML]{212529} 10}      & {\color[HTML]{212529} 30}       & {\color[HTML]{212529} AHU 2 temperature (°C).}           \\
{\color[HTML]{212529} Ext\_Irr}                    & {\color[HTML]{212529} scalar} & {\color[HTML]{212529} 0}       & {\color[HTML]{212529} 1000}     & {\color[HTML]{212529} Direct normal radiation (W/m2).}   \\
{\color[HTML]{212529} Ext\_P}                      & {\color[HTML]{212529} scalar} & {\color[HTML]{212529} 80000} & {\color[HTML]{212529} 130000} & {\color[HTML]{212529} Outdoor air pressure (Pa).}        \\
{\color[HTML]{212529} Ext\_RH}                     & {\color[HTML]{212529} scalar} & {\color[HTML]{212529} 0}       & {\color[HTML]{212529} 100}      & {\color[HTML]{212529} Outdoor realtive humidity (\%RH).} \\
{\color[HTML]{212529} Ext\_T}                      & {\color[HTML]{212529} scalar} & {\color[HTML]{212529} -10}     & {\color[HTML]{212529} 40}       & {\color[HTML]{212529} Outdoor temperature (°C).}         \\
{\color[HTML]{212529} Fa\_Pw\_All}                 & {\color[HTML]{212529} scalar} & {\color[HTML]{212529} 0}       & {\color[HTML]{212529} 50000}  & {\color[HTML]{212529} Total power consumtion (W).}       \\
{\color[HTML]{212529} Z02\_Fl\_Fan}                & {\color[HTML]{212529} scalar} & {\color[HTML]{212529} 0}       & {\color[HTML]{212529} 1}        & {\color[HTML]{212529} Zone 2 fan flow rate.}             \\
\end{longtable}

\begin{longtable}{ l  l  l  l  l }
			\caption{\label{table: energym mixed use action space} \textbf{\textit{Mixed-use} environment action-space.} \textit{Energym} clips actions to the lower and upper bounds.} \\
			\toprule
			\textbf{Action}     & \textbf{Type} & \textbf{Lower Bound} & \textbf{Upper Bound} & \textbf{Description}\\
			\midrule
			\endfirsthead
			
			\multicolumn{5}{l}{{\tablename\  \thetable{} -- continued from previous page}} \\
			
			\textbf{Action}     & \textbf{Type} & \textbf{Lower Bound} & \textbf{Upper Bound} & \textbf{Description}\\
			\midrule
			\endhead
			
			\midrule
			\multicolumn{5}{r}{\textit{continued on next page}}
			\endfoot
			
			\bottomrule
			\endlastfoot
			
            {\color[HTML]{212529} Z02\_T\_Thermostat\_sp} & {\color[HTML]{212529} scalar} & {\color[HTML]{212529} 16} & {\color[HTML]{212529} 26} & {\color[HTML]{212529} Zone 2 thermostat setpoint (°C).}  \\
            {\color[HTML]{212529} Z03\_T\_Thermostat\_sp} & {\color[HTML]{212529} scalar} & {\color[HTML]{212529} 16} & {\color[HTML]{212529} 26} & {\color[HTML]{212529} Zone 3 thermostat setpoint (°C).}  \\
            {\color[HTML]{212529} Z04\_T\_Thermostat\_sp} & {\color[HTML]{212529} scalar} & {\color[HTML]{212529} 16} & {\color[HTML]{212529} 26} & {\color[HTML]{212529} Zone 4 thermostat setpoint (°C).}  \\
            {\color[HTML]{212529} Z05\_T\_Thermostat\_sp} & {\color[HTML]{212529} scalar} & {\color[HTML]{212529} 16} & {\color[HTML]{212529} 26} & {\color[HTML]{212529} Zone 5 thermostat setpoint (°C).}  \\ \\
            {\color[HTML]{212529} Z08\_T\_Thermostat\_sp} & {\color[HTML]{212529} scalar} & {\color[HTML]{212529} 16} & {\color[HTML]{212529} 26} & {\color[HTML]{212529} Zone 8 thermostat setpoint (°C).}  \\
            {\color[HTML]{212529} Z09\_T\_Thermostat\_sp} & {\color[HTML]{212529} scalar} & {\color[HTML]{212529} 16} & {\color[HTML]{212529} 26} & {\color[HTML]{212529} Zone 9 thermostat setpoint (°C).}  \\
            {\color[HTML]{212529} Z10\_T\_Thermostat\_sp} & {\color[HTML]{212529} scalar} & {\color[HTML]{212529} 16} & {\color[HTML]{212529} 26} & {\color[HTML]{212529} Zone 10 thermostat setpoint (°C).} \\
\end{longtable}

	\begin{longtable}{ l  l  l  l  l }
 
			\caption{\label{table: office state space} \textbf{\textit{Offices} environment state-space.} Variables can be thought of as sensors installed in the building. \textit{Enerygym} clips their measurements if they fall outside lower or upper bounds.} \\
			\toprule
			\textbf{Variable}     & \textbf{Type} & \textbf{Lower Bound} & \textbf{Upper Bound} & \textbf{Description}\\
			\midrule
			\endfirsthead
			
			\multicolumn{5}{l}{{\tablename\  \thetable{} -- continued from previous page}} \\
			
			\textbf{Variable}     & \textbf{Type} & \textbf{Lower Bound} & \textbf{Upper Bound} & \textbf{Description}\\
			\midrule
			\endhead
			
			\midrule
			\multicolumn{5}{r}{\textit{continued on next page}}
			\endfoot
			
			\bottomrule
			\endlastfoot
			{\color[HTML]{212529} Bd\_Pw\_All}                 & {\color[HTML]{212529} scalar} & {\color[HTML]{212529} 0}   & {\color[HTML]{212529} 5000}    & {\color[HTML]{212529} Building power consumption (W).}   \\
{\color[HTML]{212529} Ext\_Irr}                    & {\color[HTML]{212529} scalar} & {\color[HTML]{212529} 0}   & {\color[HTML]{212529} 1000}    & {\color[HTML]{212529} Direct normal radiation (W/m2).}   \\
{\color[HTML]{212529} Ext\_RH}                     & {\color[HTML]{212529} scalar} & {\color[HTML]{212529} 0}   & {\color[HTML]{212529} 100}     & {\color[HTML]{212529} Outdoor relative humidity (\%RH).} \\
{\color[HTML]{212529} Ext\_T}                      & {\color[HTML]{212529} scalar} & {\color[HTML]{212529} -10} & {\color[HTML]{212529} 40}      & {\color[HTML]{212529} Outdoor temperature (°C).}         \\
{\color[HTML]{212529} Fa\_Pw\_All}                 & {\color[HTML]{212529} scalar} & {\color[HTML]{212529} 0}   & {\color[HTML]{212529} 10000} & {\color[HTML]{212529} Total power consumption (W).}      \\
{\color[HTML]{212529} Fa\_Pw\_HVAC}                & {\color[HTML]{212529} scalar} & {\color[HTML]{212529} 0}   & {\color[HTML]{212529} 10000} & {\color[HTML]{212529} HVAC power consumption (W).}       \\
{\color[HTML]{212529} Fa\_Pw\_PV}                  & {\color[HTML]{212529} scalar} & {\color[HTML]{212529} 0}   & {\color[HTML]{212529} 2000}  & {\color[HTML]{212529} PV power production (W).}          \\
{\color[HTML]{212529} Z01\_Fl\_Fan\_sp\_out}       & {\color[HTML]{212529} scalar} & {\color[HTML]{212529} 0}   & {\color[HTML]{212529} 1}       & {\color[HTML]{212529} Zone 1 fan flow setpoint.}         \\
\end{longtable}
						
			\begin{longtable}{ l  l  l  l  l }
			\caption{\label{table: energym office action space} \textbf{\textit{Offices} environment action-space.} \textit{Energym} clips actions to the lower and upper bounds.} \\
			\toprule
			\textbf{Action}     & \textbf{Type} & \textbf{Lower Bound} & \textbf{Upper Bound} & \textbf{Description}\\
			\midrule
			\endfirsthead
			
			\multicolumn{5}{l}{{\tablename\  \thetable{} -- continued from previous page}} \\
			
			\textbf{Action}     & \textbf{Type} & \textbf{Lower Bound} & \textbf{Upper Bound} & \textbf{Description}\\
			\midrule
			\endhead
			
			\midrule
			\multicolumn{5}{r}{\textit{continued on next page}}
			\endfoot
			
			\bottomrule
			\endlastfoot

{\color[HTML]{212529} Z01\_T\_Thermostat\_sp} & {\color[HTML]{212529} scalar}   & {\color[HTML]{212529} 16} & {\color[HTML]{212529} 26} & {\color[HTML]{212529} Zone 1 thermostat setpoint (°C).}      \\

{\color[HTML]{212529} Z02\_T\_Thermostat\_sp} & {\color[HTML]{212529} scalar}   & {\color[HTML]{212529} 16} & {\color[HTML]{212529} 26} & {\color[HTML]{212529} Zone 2 thermostat setpoint (°C).}      \\

{\color[HTML]{212529} Z03\_T\_Thermostat\_sp} & {\color[HTML]{212529} scalar}   & {\color[HTML]{212529} 16} & {\color[HTML]{212529} 26} & {\color[HTML]{212529} Zone 3 thermostat setpoint (°C).}      \\

{\color[HTML]{212529} Z04\_T\_Thermostat\_sp} & {\color[HTML]{212529} scalar}   & {\color[HTML]{212529} 16} & {\color[HTML]{212529} 26} & {\color[HTML]{212529} Zone 4 thermostat setpoint (°C).}      \\

{\color[HTML]{212529} Z05\_T\_Thermostat\_sp} & {\color[HTML]{212529} scalar}   & {\color[HTML]{212529} 16} & {\color[HTML]{212529} 26} & {\color[HTML]{212529} Zone 5 thermostat setpoint (°C).}      \\

{\color[HTML]{212529} Z06\_T\_Thermostat\_sp} & {\color[HTML]{212529} scalar}   & {\color[HTML]{212529} 16} & {\color[HTML]{212529} 26} & {\color[HTML]{212529} Zone 6 thermostat setpoint (°C).}      \\

{\color[HTML]{212529} Z07\_T\_Thermostat\_sp} & {\color[HTML]{212529} scalar}   & {\color[HTML]{212529} 16} & {\color[HTML]{212529} 26} & {\color[HTML]{212529} Zone 7 thermostat setpoint (°C).}      \\

{\color[HTML]{212529} Z15\_T\_Thermostat\_sp} & {\color[HTML]{212529} scalar}   & {\color[HTML]{212529} 16} & {\color[HTML]{212529} 26} & {\color[HTML]{212529} Zone 15 thermostat setpoint (°C).}     \\
\end{longtable}
						
				\begin{longtable}{ l  l  l  l  l }
			\caption{\label{table: seminar state space} \textbf{\textit{Seminar-Thermostat} environment state-space.} Variables can be thought of as sensors installed in the building. \textit{Enerygym} clips their measurements if they fall outside lower or upper bounds.} \\
			\toprule
			\textbf{Variable}     & \textbf{Type} & \textbf{Lower Bound} & \textbf{Upper Bound} & \textbf{Description}\\
			\midrule
			\endfirsthead
			
			\multicolumn{5}{l}{{\tablename\  \thetable{} -- continued from previous page}} \\
			\textbf{Variable}     & \textbf{Type} & \textbf{Lower Bound} & \textbf{Upper Bound} & \textbf{Description}\\
			\midrule
			\endhead
			
			\midrule
			\multicolumn{5}{r}{\textit{continued on next page}}
			\endfoot
			
			\bottomrule
			\endlastfoot
			{\color[HTML]{212529} Ext\_Irr}                    & {\color[HTML]{212529} scalar} & {\color[HTML]{212529} 0}     & {\color[HTML]{212529} 1000}   & {\color[HTML]{212529} Direct normal radiation (W/m2).}      \\
            {\color[HTML]{212529} Ext\_RH}                     & {\color[HTML]{212529} scalar} & {\color[HTML]{212529} 0}     & {\color[HTML]{212529} 100}    & {\color[HTML]{212529} Outdoor relative humidity (\%RH).}    \\
            {\color[HTML]{212529} Ext\_T}                      & {\color[HTML]{212529} scalar} & {\color[HTML]{212529} -15}   & {\color[HTML]{212529} 40}     & {\color[HTML]{212529} Outdoor temperature (°C).}            \\
            {\color[HTML]{212529} Ext\_P}                      & {\color[HTML]{212529} scalar} & {\color[HTML]{212529} 80000} & {\color[HTML]{212529} 130000} & {\color[HTML]{212529} Outdoor air pressure (Pa).}           \\
            {\color[HTML]{212529} Z01\_RH}                     & {\color[HTML]{212529} scalar} & {\color[HTML]{212529} 0}     & {\color[HTML]{212529} 100}    & {\color[HTML]{212529} Zone 1 relative humidity (\%RH).}     \\
            {\color[HTML]{212529} Z01\_T}                      & {\color[HTML]{212529} scalar} & {\color[HTML]{212529} 10}    & {\color[HTML]{212529} 40}     & {\color[HTML]{212529} Zone 1 temperature (°C).}             \\
            {\color[HTML]{212529} Z01\_T\_Thermostat\_sp\_out} & {\color[HTML]{212529} scalar} & {\color[HTML]{212529} 16}    & {\color[HTML]{212529} 26}     & {\color[HTML]{212529} Zone 1 thermostat setpoint (°C).}     \\
            {\color[HTML]{212529} Z02\_RH}                     & {\color[HTML]{212529} scalar} & {\color[HTML]{212529} 0}     & {\color[HTML]{212529} 100}    & {\color[HTML]{212529} Zone 2 relative humidity (\%RH).}     \\
\end{longtable}
						
			\begin{longtable}{ l  l  l  l  l }
			\caption{\label{table: energym seminar action space} \textbf{\textit{Seminar-Thermostat} environment action-space.} \textit{Energym} clips actions to the lower and upper bounds.} \\
			\toprule
			\textbf{Action}     & \textbf{Type} & \textbf{Lower Bound} & \textbf{Upper Bound} & \textbf{Description}\\
			\midrule
			\endfirsthead
			
			\multicolumn{5}{l}{{\tablename\  \thetable{} -- continued from previous page}} \\
			
			\textbf{Action}     & \textbf{Type} & \textbf{Lower Bound} & \textbf{Upper Bound} & \textbf{Description}\\
			\midrule
			\endhead
			
			\midrule
			\multicolumn{5}{r}{\textit{continued on next page}}
			\endfoot
			
			\bottomrule
			\endlastfoot
			
			{\color[HTML]{212529} Z01\_T\_Thermostat\_sp} & {\color[HTML]{212529} scalar} & {\color[HTML]{212529} 16} & {\color[HTML]{212529} 26} & {\color[HTML]{212529} Zone 1 thermostat setpoint (°C).}  \\
            {\color[HTML]{212529} Z02\_T\_Thermostat\_sp} & {\color[HTML]{212529} scalar} & {\color[HTML]{212529} 16} & {\color[HTML]{212529} 26} & {\color[HTML]{212529} Zone 2 thermostat setpoint (°C).}  \\
            {\color[HTML]{212529} Z03\_T\_Thermostat\_sp} & {\color[HTML]{212529} scalar} & {\color[HTML]{212529} 16} & {\color[HTML]{212529} 26} & {\color[HTML]{212529} Zone 3 thermostat setpoint (°C).}  \\
            {\color[HTML]{212529} Z04\_T\_Thermostat\_sp} & {\color[HTML]{212529} scalar} & {\color[HTML]{212529} 16} & {\color[HTML]{212529} 26} & {\color[HTML]{212529} Zone 4 thermostat setpoint (°C).}  \\
            {\color[HTML]{212529} Z05\_T\_Thermostat\_sp} & {\color[HTML]{212529} scalar} & {\color[HTML]{212529} 16} & {\color[HTML]{212529} 26} & {\color[HTML]{212529} Zone 5 thermostat setpoint (°C).}  \\
            {\color[HTML]{212529} Z06\_T\_Thermostat\_sp} & {\color[HTML]{212529} scalar} & {\color[HTML]{212529} 16} & {\color[HTML]{212529} 26} & {\color[HTML]{212529} Zone 6 thermostat setpoint (°C).}  \\
            {\color[HTML]{212529} Z08\_T\_Thermostat\_sp} & {\color[HTML]{212529} scalar} & {\color[HTML]{212529} 16} & {\color[HTML]{212529} 26} & {\color[HTML]{212529} Zone 8 thermostat setpoint (°C).}  \\
            {\color[HTML]{212529} Z09\_T\_Thermostat\_sp} & {\color[HTML]{212529} scalar} & {\color[HTML]{212529} 16} & {\color[HTML]{212529} 26} & {\color[HTML]{212529} Zone 9 thermostat setpoint (°C).}  \\
\end{longtable}
\end{landscape}
\chapter{Datasets}\label{appendix: datasets}
\section{ExORL}\label{appendix: exorl datasets}
\begin{figure}[hb]
    \centering
    \includegraphics{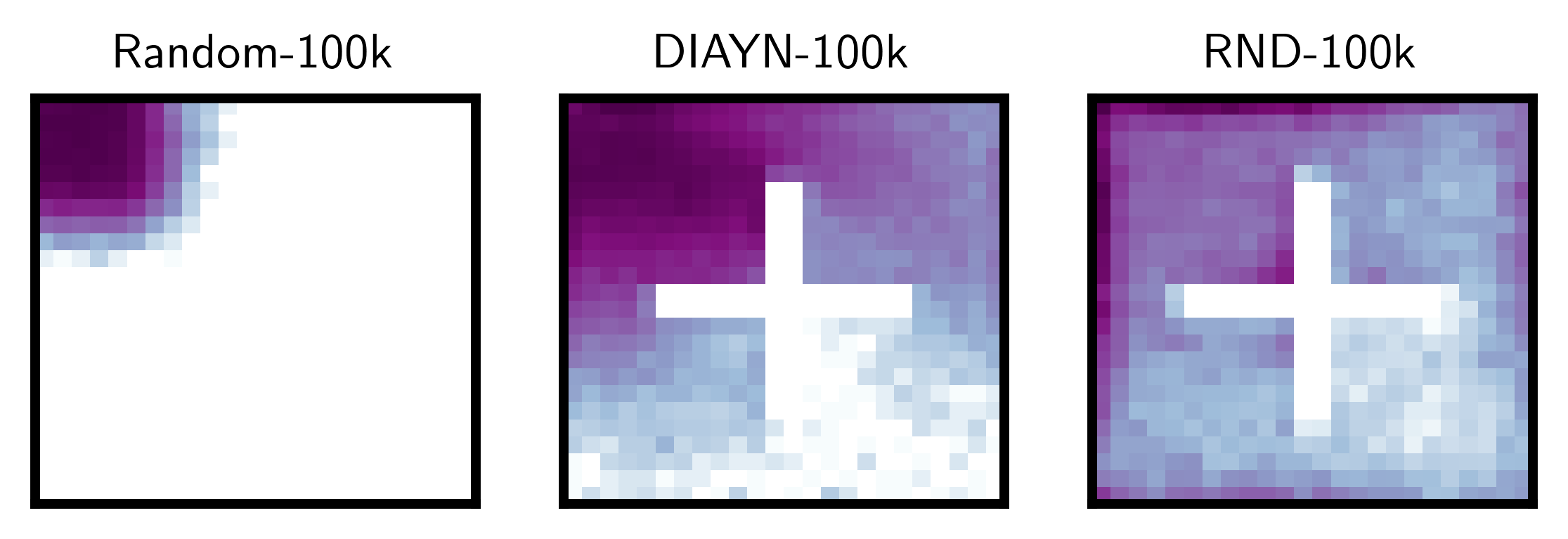}
    \caption{\textbf{Point-mass maze state coverage by dataset.} \textit{(left)} \textsc{Random}; \textit{(middle)} \textsc{Diayn}; \textit{(right)} \textsc{Rnd}.}
    \label{fig: exorl-dataset-heatmap}
\end{figure}

In Chapters \ref{chapter: from low quality data} we use train our methods on pre-collected datasets from the ExORL benchmark. Specifically, we train on 100,000 transitions uniformly sampled from three datasets collected by different unsupervised agents: \textsc{Random}, \textsc{Diayn}, and \textsc{Rnd}. The state coverage on Point-mass maze is depicted in Figure \ref{fig: exorl-dataset-heatmap}. Though harder to visualise, we found that state marginals on higher-dimensional tasks (e.g. Walker) showed a similar diversity in state coverage.\\

\noindent\textbf{\textsc{Rnd}}. An agent whose exploration is directed by the predicted error in its ensemble of dynamics models. Informally, we say \textsc{Rnd} datasets exhibit \textit{high} state diversity.\\

\noindent\textbf{\textsc{Diayn}}. An agent that attempts to sequentially learn a set of skills. Informally, we say \textsc{Diayn} datasets exhibit \textit{medium} state diversity.\\

\noindent\textbf{\textsc{Random}}. A agent that selects actions uniformly at random from the action space. Informally, we say \textsc{Random} datasets exhibit \textit{low} state diversity.\\

In Chapter \ref{chapter: under partial observability} we collect datasets from each domain under changed dynamics. We do this with our own implementation of RND \citep{borsa2018} which exactly matches the implementation from the original ExORL paper \citep{yarats2022}.

\section{D4RL}\label{appendix: d4rl datasets}
In Chapter \ref{chapter: from low quality data} we consider three goal-directed datasets from D4RL, each providing a different proportion of expert trajectories. The below dataset descriptions are taken from \citep{fu2020}.\\

\noindent\textbf{Medium.} Generated by training an SAC policy, early-stopping the training, and collecting 1M samples from this partially-trained policy.\\

\noindent\textbf{Medium-replay.} Generated by recording all samples in the replay buffer observed during training until the policy reaches the “medium” level of performance.\\

\noindent\textbf{Medium-expert.} Generated by mixing equal amounts of expert demonstrations and suboptimal data, either from a partially trained policy or by unrolling a uniform-at-random policy.\\

\chapter{Implementations}\label{appendix: implementations}
Here we detail implementations for all methods discussed in this thesis. The code to reproduce all of our experiments is open-sourced:\\

\noindent Chapter \ref{chapter: from low quality data}: \hfill \url{https://github.com/enjeeneer/zero-shot-rl} \\

\noindent Chapter \ref{chapter: under partial observability}: \hfill \url{https://github.com/enjeeneer/bfms-with-memory} \\

\noindent Chapter \ref{chapter: no prior data}: \hfill \url{https://github.com/enjeeneer/PEARL} \\

\section{Forward-Backward Representations}\label{appendix: FB implementation}
\begin{figure}[t]
\centering
    \includegraphics[width=\textwidth]{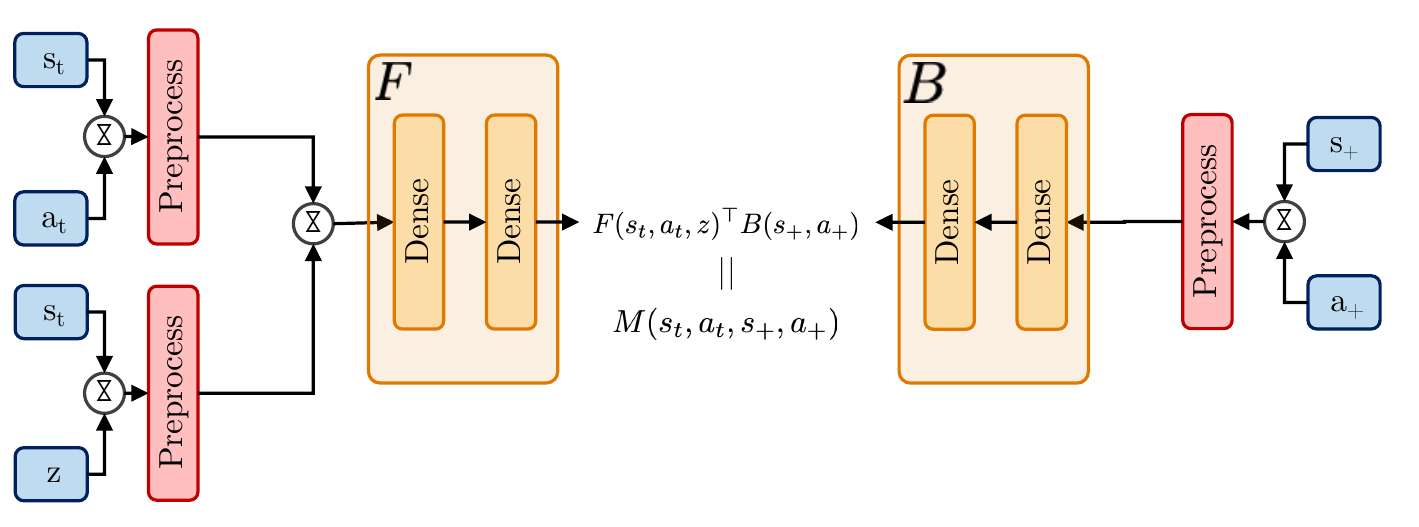}
    \caption{\textbf{Zero-shot RL methods \textit{without} memory}. FB is optimised in a standard actor critic setup \citep{konda1999actor}. The policy $\pi$ selects an action {\color{action-purple}{$a_t$}} conditioned on the current state $s_t$, and the task vector $z$. The $Q$ function formed by the USF $\psi$ evaluates the action $a_t$ given the current state $s_t$ and task $z$.}
    \label{fig: fb architecture without memory}
\end{figure}

\subsection{Architecture}
The forward-backward architecture described below follows the implementation by \cite{touati2022does} exactly, other than the batch size which we reduce from 1024 to 512. We did this to reduce the computational expense of each run without limiting performance. The hyperparameter study in Appendix J of \cite{touati2022does} shows this choice is unlikely to affect FB performance. All other hyperparameters are reported in Table \ref{table: fb hyperparameters}.\\

\noindent\textbf{Forward Representation $F(s, a, z)$}. The input to the forward representation $F$ is always preprocessed. State-action pairs $(s, a)$ and state-task pairs $(s, z)$ have their own preprocessors which are feedforward MLPs that embed their inputs into a 512-dimensional space. These embeddings are concatenated and passed through a third feedforward MLP $F$ which outputs a $d$-dimensional embedding vector. Note: the forward representation $F$ is identical to $\psi$ used by USF so their implementations are identical (see Table \ref{table: fb hyperparameters}). \\

\noindent\textbf{Backward Representation $B(s)$.} The backward representation $B$ is a feedforward MLP that takes a state as input and outputs a $d$-dimensional embedding vector.\\ 

\noindent\textbf{Actor $\pi(s, z)$.} Like the forward representation, the inputs to the policy network are similarly preprocessed. State-action pairs $(s, a)$ and state-task pairs $(s, z)$ have their own preprocessors which feedforward MLPs that embed their inputs into a 512-dimensional space. These embeddings are concatenated and passed through a third feedforward MLP which outputs a $a$-dimensional vector, where $a$ is the action-space dimensionality. A \texttt{Tanh} activation is used on the last layer to normalise their scale. As per \cite{fujimoto2019}'s recommendations, the policy is smoothed by adding Gaussian noise $\sigma$ to the actions during training. Note the actors used by FB and USFs are identical (see Table \ref{table: fb hyperparameters}).\\

\noindent\textbf{Misc.} Layer normalisation \citep{ba2016layer} and \texttt{Tanh} activations are used in the first layer of all MLPs to standardise the inputs.

\begin{table}\caption{\textbf{Hyperparameters for FB and USF.} The additional hyperparameters for Conservative FB representations are highlighted in \colorbox{solarized@lightblue}{blue} and for FB-M are in \colorbox{YellowGreen}{green}.}\label{table: fb hyperparameters}
\centering
\scalebox{1}{
\begin{tabular}{ll} 
\toprule
\textbf{Hyperparameter}                       & \textbf{Value}       \\
 \midrule
Latent dimension $d$                          & 50 (100 for maze)    \\
$F$ / $\psi$ dimensions                             & (1024, 1024)            \\
$B$ / $\varphi$ dimensions                             & (256, 256, 256)                    \\
Preprocessor dimensions & (1024, 1024) \\
Std. deviation for policy smoothing $\sigma$  & 0.2                  \\
Truncation level for policy smoothing         & 0.3                  \\
Learning steps                                & 1,000,000            \\
 
Batch size                                    & 512                  \\
Optimiser                                     & Adam    \citep{kingma2014adam}             \\
Learning rate                                 & 0.0001               \\ 
Discount $\gamma$                             & 0.98 (0.99 for maze) \\
Activations (unless otherwise stated)         & ReLU                 \\
Target network Polyak smoothing coefficient & 0.01                 \\
$z$-inference labels                          & 10,000              \\
$z$ mixing ratio                              & 0.5                   \\
\colorbox{solarized@lightblue}{Conservative budget $\tau$} & \colorbox{solarized@lightblue}{50 (45 for D4RL)} \\
\colorbox{solarized@lightblue}{OOD action samples per policy $N$}   & \colorbox{solarized@lightblue}{3}                   \\ 
\colorbox{YellowGreen}{Transformer heads} & \colorbox{YellowGreen}{4} \\ 
\colorbox{YellowGreen}{Transformer / S4d model dimension} & \colorbox{YellowGreen}{32} \\ 
\colorbox{YellowGreen}{GRU dimensions} & \colorbox{YellowGreen}{(512, 512)} \\ 
\colorbox{YellowGreen}{Context length} $L$ & \colorbox{YellowGreen}{32}  \\ 
\bottomrule
\end{tabular}
}
\end{table}

\subsection{Task Sampling Distribution $\mathcal{Z}$}\label{appendix: z sampling implementation}
FB representations require a method for sampling the task vector $z$ at each learning step. \cite{touati2022does} employ a mix of two methods, which we replicate:
\begin{enumerate}
    \item Uniform sampling of $z$ on the hypersphere surface of radius $\sqrt{d}$ around the origin of $\mathbb{R}^d$,
    \item Biased sampling of $z$ by passing states $s \sim \mathcal{D}$ through the backward representation $z = B(s)$. This also yields vectors on the hypersphere surface due to the $L2$ normalisation described above, but the distribution is non-uniform.
\end{enumerate}
We sample $z$ 50:50 from these methods at each learning step.

\subsection{Maximum Value Approximator \texorpdfstring{$\mu$}{mu}} \label{appendix: max value estimator implementation}
The conservative variants of FB require access to a policy distribution $\mu(a|s)$ that maximises the value of the current $Q$ iterate in expectation. Recall the standard CQL loss
\begin{equation}\label{equation: CQL update appendix}
    \mathcal{L}_{\text{CQL}} = \alpha \cdot \left(\mathbb{E}_{s \sim \mathcal{D}, a \sim \mu(a|s)} [Q(s,a)] - \mathbb{E}_{(s, a) \sim \mathcal{D}}[Q(s, a)] - R(\mu)\right) + \mathcal{L}_{\text{Q}},
\end{equation}
where $\alpha$ is a scaling parameter, $\mu(a|s)$ the policy distribution we seek, $R$ regularises $\mu$ and $\mathcal{L}_{\text{Q}}$ represents the normal TD loss on $Q$. \cite{kumar2020}'s most performant CQL variant (CQL($\mathcal{H}$)) utilises maximum entropy regularisation on $\mu$ i.e. $R = \mathcal{H}(\mu)$. They show that obtaining $\mu$ can be cast as a closed-form optimisation problem of the form:
\begin{equation}\label{equation: closed form mu opt.}
    \max_\mu \mathbb{E}_{x \sim \mu(x)}[f(x)] + \mathcal{H}(\mu) \; \text{s.t.} \sum_x \mu(x) = 1, \mu(x) \geq 0 \; \forall x,
\end{equation}
and has optimal solution $\mu^*(x) = \frac{1}{Z} \exp (f(x))$, where $Z$ is a normalising factor. Plugging Equation \ref{equation: closed form mu opt.} into Equation \ref{equation: CQL update appendix} we obtain:
\begin{equation}\label{equation: CQL logsumexp appendix}
    \mathcal{L}_{\text{CQL}} = \alpha \cdot \left(\mathbb{E}_{s \sim \mathcal{D}} [\log \sum_a \exp (Q(s,a))] - \mathbb{E}_{(s, a) \sim \mathcal{D}}[Q(s, a)]\right) + \mathcal{L}_{\text{Q}}.
\end{equation}
In discrete action spaces the \texttt{logsumexp} can be computed exactly; in continuous action spaces \cite{kumar2020} approximate it via importance sampling using actions sampled uniformly at random, actions from the current policy conditioned on $s_t \sim \mathcal{D}$, and from the current policy conditioned on $s_{t+1} \sim \mathcal{D}$\footnote{Conditioning on next states $s_{t+1} \sim \mathcal{D}$ is not mentioned in the paper, but is present in their official implementation.}: 
\begin{align}
\begin{aligned}\label{equation: logsumexp Q}
    \log \sum_a \exp Q(s_t, a_t) &= \log ( \frac{1}{3} \sum_a \exp Q(s_t, a_t)) + \frac{1}{3} \sum_a \exp Q(s_t, a_t))\\
    & \qquad \qquad \qquad \qquad \qquad + \frac{1}{3} \sum_a \exp (\exp Q(s_t, a_t)) , \\
    &= \log ( \frac{1}{3} \mathbb{E}_{a_t \sim \text{Unif}(\mathcal{A})} \left[\frac{\exp(Q(s_t, a_t)}{\text{Unif}(\mathcal{A})}\right] + \frac{1}{3} \mathbb{E}_{a_t \sim \pi(a_t|s_t)} \left[\frac{\exp(Q(s_t, a_t))}{\pi(a_{t}|s_t)}\right] \\
    & \qquad \frac{1}{3} \mathbb{E}_{a_{t+1} \sim \pi(a_{t+1}|s_{t+1})} \left[\frac{\exp(Q(s_t, a_t))}{\pi(a_{t+1}|s_{t+1})}\right] ), \\
    &= \log ( \frac{1}{3N} \sum_{a_i \sim \text{Unif}(\mathcal{A})}^N \left[\frac{\exp(Q(s_t, a_t))}{\text{Unif}(\mathcal{A})}\right] + \frac{1}{6N} \sum_{a_i \sim \pi(a_t|s_t)}^{2N} \left[\frac{\exp(Q(s_t, a_t))}{\pi(a_i|s_t)}\right] \\
    & \qquad \frac{1}{3N} \sum_{a_i \sim \pi(a_{t+1}|s_{t+1})}^N \left[\frac{\exp(Q(s_t, a_t))}{\pi(a_i|s_{t+1})}\right] ), \\
\end{aligned}
\end{align}
with $N$ a hyperparameter defining the number of actions to sample across the action-space. We can substitute $F(s,a,z)^\top z$ for $Q(s,a)$ in the final expression of Equation \ref{equation: logsumexp} to obtain the equivalent for VC-FB:
\begin{align}
\begin{aligned}\label{equation: logsumexp}
    \log \sum_a \exp F(s_t, a_i, z)^\top z &= \log \big( \frac{1}{3N} \sum_{a_i \sim \text{Unif}(\mathcal{A})}^N \left[\frac{\exp(F(s_t, a_i, z)^\top z)}{\text{Unif}(\mathcal{A})}\right] \\
    & \qquad \qquad + \frac{1}{6N} \sum_{a_i \sim \pi(a_t|s_t)}^{2N} \left[\frac{\exp(F(s_t, a_i, z^\top z)}{\pi(a_i|s_t)}\right] \\
    & \qquad \qquad + \frac{1}{3N} \sum_{a_i \sim \pi(a_{t+1}|s_{t+1})}^N \left[\frac{\exp(F(s_t, a_i, z)^\top z)}{\pi(a_i|s_{t+1})}\right] \big). \\
\end{aligned}
\end{align}
In Appendix \ref{learning curves appendix}, Figure \ref{fig: chapter3/action samples hyperparams} we show how the performance of VC-FB varies with the number of action samples. In general, performance improves with the number of action samples, but we limit $N = 3$ to limit computational burden. The formulation for MC-FB is identical other than each value $F(s, a, z)^T z$ being replaced with measures $F(s, a, z)^T B(s_+)$.

\subsection{Dynamically Tuning \texorpdfstring{$\alpha$}{alpha}} \label{appendix: dynamically tuning alpha implementation}
A critical hyperparameter is $\alpha$ which weights the conservative penalty with respect to other losses during each update. We initially trialled constant values of $\alpha$, but found performance to be fragile to this selection, and lacking robustness across environments. Instead, we follow \cite{kumar2020} once again, and instantiate their algorithm for dynamically tuning $\alpha$, which they call Lagrangian dual gradient-descent on $\alpha$. We introduce a conservative budget parameterised by $\tau$, and set $\alpha$ with respect to this budget:
\begin{equation}\label{equation: alpha lagrangian}
    \min_{FB} {\color{bupu@purple}{\max_{\alpha \geq 0}} \; \alpha} \cdot \left( \mathbb{E}_{s \sim \mathcal{D}, a \sim \mu(a|s) z \sim \mathcal{Z}} [F(s, a, z)^{\top} z] - \mathbb{E}_{(s, a) \sim \mathcal{D}, z \sim \mathcal{Z}} [F(s, a, z)^{\top}z] - {\color{bupu@purple}{\tau}} \right) + \mathcal{L}_{\text{FB}}.
\end{equation}
Intuitively, this implies that if the scale of overestimation $\leq \tau$ then $\alpha$ is set close to 0, and the conservative penalty does not affect the updates. If the scale of overestimation $\geq \tau$ then $\alpha$ is set proportionally to this gap, and thus the conservative penalty is proportional to the degree of overestimation above $\tau$. As above, for the MC-FB variant values $F(s, a, z)^\top z$ are replaced with measures $F(s, a, z)^\top B(s_+)$. 

\subsection{Algorithm} \label{appendix: vc-fb algorithm}
We summarise the end-to-end implementation of VC-FB as pseudo-code in Algorithm \ref{alg: VC-FB}. MC-FB representations are trained identically other than at line 10 where the conservative penalty is computed for $M$ instead of $Q$, and in line 12 where $M$s are lower bounded via Equation \ref{equation: FB-CM loss}.

\begin{algorithm}
   \caption{Pre-training value-conservative forward-backward representations}
   \label{alg: VC-FB}
\begin{algorithmic}[1]
\REQUIRE $\mathcal{D}$: dataset of trajectories \\
~~~~~~~~~~$F_{\theta_F}$, $B_{\theta_B}$, $\pi$: randomly initialised networks\\
~~~~~~~~~~$N$, $\mathcal{Z}$, $\nu$, $b$: learning steps, z-sampling distribution, polyak momentum, batch size \\
\FOR{learning step $n=1...N$}
\STATE{$\{(s_i, a_i, s_{i+1})\} \sim \mathcal{D}_{i \in |b|}$} ~~~~~~~~~~~~~~~~~~{\color{solarized@violet}{$\vartriangleleft$ \emph{Sample mini-batch of transitions}}}
\STATE{$\{z_i\}_{i \in |b|} \sim \mathcal{Z}$} ~~~~~~~~~~~~~~~~~~~~~~~~~~~~~~~~~~~~~~~{\color{solarized@violet}{$\vartriangleleft$ \emph{Sample zs (Appendix \ref{appendix: z sampling implementation})}}}
\STATE{}
\STATE{{\color{gray}{\emph{// FB Update}}}
\STATE{$\{a_{i+1}\} \sim \pi(s_{i+1}, z_i)$} ~~{\color{solarized@violet}{$\vartriangleleft$ \emph{Sample batch of actions at next states from policy}}}
\STATE{Update $FB$ given $\{(s_i, a_i, s_{i+1}, a_{i+1}, z_i)\}$} ~~~~~~~~~~~~~~~~~~~~~~~{\color{solarized@violet}{$\vartriangleleft$ \emph{Equation \ref{equation: fb loss}}}}
\STATE{}
\STATE{{\color{gray}{\emph{// Conservative Update}}}}
\STATE{$Q^{\text{max}}(s_i, a_i) \approx \log \sum_a \exp F(s_i, a_i, z_i)^\top z_i$} ~~~~~~~~~~~~~~~~~~~~~~{\color{solarized@violet}{$\vartriangleleft$ \emph{Equation \ref{equation: logsumexp}}}}
\STATE{Compute $\alpha$ given $Q^{\text{max}}$~~~~~~~~~~~~~~~~~~~~~~~~~~~~~~~~~~~~~~~~~~~~~~~~{\color{solarized@violet}{$\vartriangleleft$ \emph{Equation \ref{equation: alpha lagrangian}}}}}
\STATE{Lower bound $Q$} ~~~~~~~~~~~~~~~~~~~~~~~~~~~~~~~~~~~~~~~~~~~~~~~~~~~~{\color{solarized@violet}{$\vartriangleleft$ \emph{Equation \ref{equation: VC-FB loss}}}}}
\STATE{}
\STATE{{\color{gray}{\emph{// Actor Update}}}}
\STATE{$\{a_{i}\} \sim \pi(s_{i}, z_i)$} ~~~~~~~~~~~~~~~~~~~~~~~~~~~~~~~~~~~~{\color{solarized@violet}{$\vartriangleleft$ \emph{Sample actions from policy}}}
\STATE{Maximise $\mathbb{E} [F(s_i, a_i, z_i)^\top z_i$]} ~~~~~~~~~~~~~{\color{solarized@violet}{$\vartriangleleft$ \textit{Standard actor-critic formulation}}}
\STATE{}
\STATE{{\color{gray}{\emph{// Update target networks via polyak averaging}}}}
\STATE{$\theta_F^- \leftarrow \nu \theta_F^- + (1 - \nu)\theta_F$} ~~~~~~~~~~~~~~~~~~~~~~~~~~~~~~~~~~~~~{\color{solarized@violet}{$\vartriangleleft$ \emph{Forward target network}}}
\STATE{$\theta_B^- \leftarrow \nu \theta_B^- + (1 - \nu)\theta_B$} ~~~~~~~~~~~~~~~~~~~~~~~~~~~~~{\color{solarized@violet}{$\vartriangleleft$ \emph{Backward target network}}}
\ENDFOR{}
\end{algorithmic}
\end{algorithm}

\subsection{\textit{Directed} Value-Conservative FB Representations} \label{appendix: DVC-FB implementation}

VC-FB applies conservative updates using task vectors $z$ sampled from $\mathcal{Z}$ (which in practice is a uniform distribution over the $\sqrt{d}$-hypersphere). This will include many vectors corresponding to tasks that are never evaluated in practice in downstream applications. Intuitively, it may seem reasonable to direct conservative updates to focus on tasks that are likely to be encountered downstream. One simple way of doing this would be consider the set of all goal-reaching tasks for goal states in the training distribution, which corresponds to sampling $z=B(s_g)$ for some $s_g\sim \mathcal{D}$. This leads to the following conservative loss function:
\begin{multline}\label{equation: DVCFB loss}
    \mathcal{L}_{\text{\textit{D}VC-FB}} = \alpha \cdot \Big( \mathbb{E}_{s \sim \mathcal{D}, a \sim \mu(a|s), s_g\sim \mathcal{D}} [F(s, a, B(s_g))^{\top} B(s_g)] \\
    - \mathbb{E}_{(s, a) \sim \mathcal{D}, s_g \sim \mathcal{D}} [F(s, a, B(s_g))^{\top}B(s_g)] - \mathcal{H}(\mu)\Big) + \mathcal{L}_{\text{FB}}.
\end{multline}

We call models learnt via this loss \textit{directed}-VC-FB (\textit{D}VC-FB). While we were initially open to the possibility that \textit{D}VC-FB updates would be better targeted than those of VC-FB, and would lead to improved downstream task performance, this turns out not to be the case in our experimental settings as discussed in \colouredS \ref{chapter3: discussion}. We report scores obtained by the \textit{D}VC-FB method across all 100k datasets, domains and tasks in Appendix \ref{appendix: extended results}.

\subsection{Forward Backward Representations With Memory}\label{appendix: fb with memory}
\textbf{Memory Models $f_F(\tau^L)$, $f_B(\tau^L)$ and $f_{\pi}(\tau^L)$} \; FB-M has separate memory models for the forward model $f_F$, backward model $f_B$ and policy $f_{\pi}$ following the findings of \citep{ni2021recurrent}, but their implementations are identical. Trajectories of observation-action pairs are preprocessed by one-layer feedforward MLPs that embed their inputs into a 512-dimensional space. The memory model is a GRU whose hidden state is initialised as zeros and updated sequentially by processing each embedding in the trajectory. For the experiments in \colouredS \ref{discussion: memory model} we additionally use transformer \cite{vaswani2017attention} and s4 memory models \cite{gu2021efficiently}. Our transformer uses \textit{FlashAttention} \citep{dao2022flashattention} for faster inference, and we use diagonalised s4 (s4d) \citep{gu2022parameterization} rather than standard s4 because of its improved empirical performance on sequence modelling tasks.

\subsection{Context Lengths}\label{appendix: context lengths}
\begin{figure}[t]
\centering
    \includegraphics[width=\textwidth]{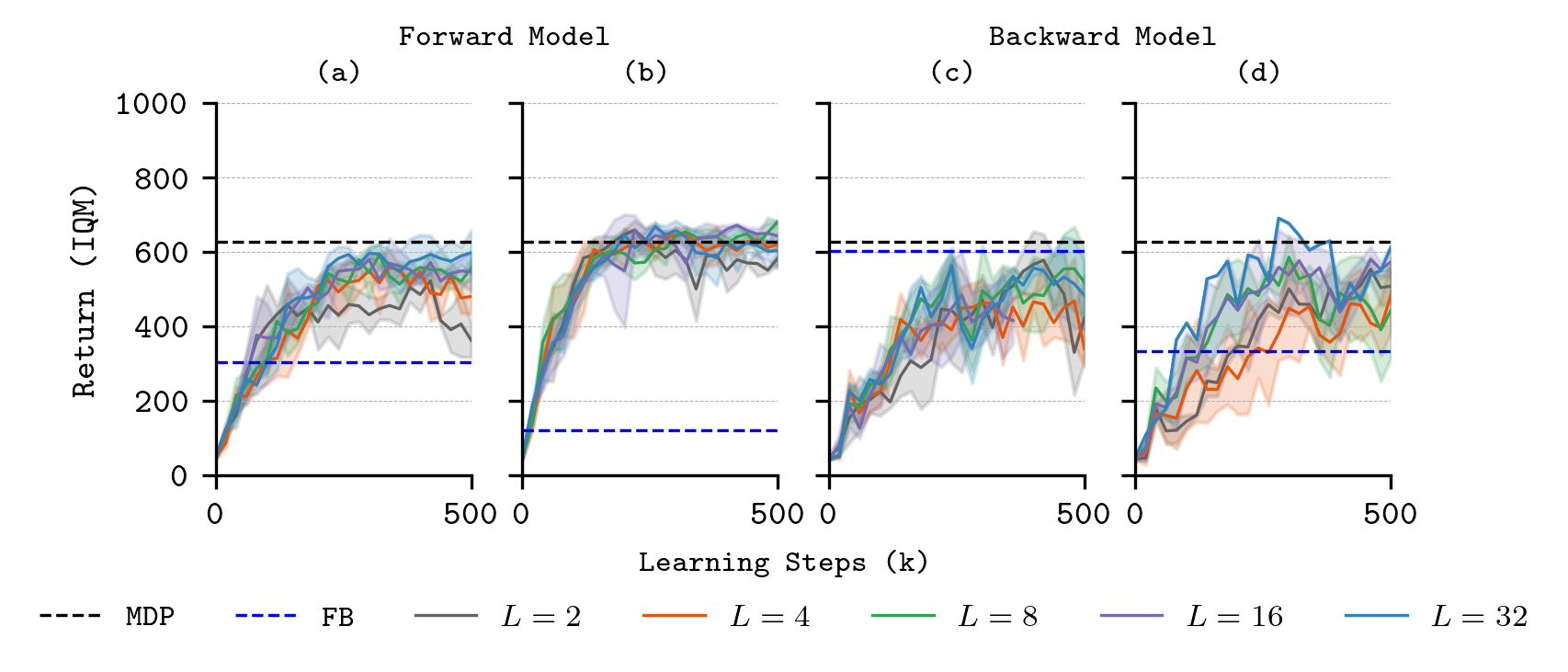}
    \caption{\textbf{Hyperparameter sweep over context length $L$.} We evaluate the performance of FB-M with GRU memory model on \texttt{Walker} \texttt{noisy} ((a) and (c) and \texttt{Walker} \texttt{flickering} ((b) and (d)). When we sweep over the forward model's context length, we pass states to the backward model and keep it memory-free; when we sweep over the backward model's context length we pass states to the forward model and policy and keep them memory-free.}
    \label{fig: appendix/context sweep}
\end{figure}

The context length $L$ of both the $F/\pi_z$ and $B$ is an important hyperparameter. When adding memory to actors or critics, it is standard practice to parallelise  training across batched trajectories of fixed $L$ (zero-padded for all $t < L$), yet condition the policy on the entire episode history during evaluation with recurrent hidden states. If $L$ is chosen to be less than the maximum episode length, as is often required with limited compute, a shift between the training and evaluation distributions is inevitable. Though this does not tend to harm performance significantly \citep{hausknecht2015deep}, the aim is generally to maximise $L$ subject to available compute.
The Markov states of different POMDPs will require different $L$, but longer $L$ increases training time and risks decreased training stability. In Figure \ref{fig: appendix/context sweep} we sweep across $L \in \{2, 4, 8, 16, 32\}$ for both $F/\pi_z$ and $B$. In general, we see small increases in performance for increased context length, and choose $L = 32$ for our main experiments.

\section{Universal Successor Features}
Our USFs with basic features $\varphi(s)$ provided by Laplacian eigenfunctions \citep{wu2018laplacian} are implemented according to \cite{touati2022does}, and with HILPs according to \citep{park2024foundation}. 

\subsection{Architecture}
\noindent\textbf{USF $\psi(s, a, z)$}. The input to the USF $\psi$ is always preprocessed. State-action pairs $(s, a)$ and state-task pairs $(s, z)$ have their own preprocessors which are feedforward MLPs that embed their inputs into a 512-dimensional space. These embeddings are concatenated and passed through a third feedforward MLP $\psi$ which outputs a $d$-dimensional embedding vector. Note this is identical to the implementation of $F$ as described in Appendix \ref{appendix: FB implementation}. All other hyperparameters are reported in Table \ref{table: fb hyperparameters}.\\

\noindent\textbf{Feature Embedding $\varphi(s)$.} The feature map $\varphi(s)$ is a feedforward MLP that takes a state as input and outputs a $d$-dimensional embedding vector. The loss function for learning the feature embedding is provided in Appendix \ref{appendix: lap loss}.\\

\noindent\textbf{Actor $\pi(s, z)$.} Like the forward representation, the inputs to the policy network are similarly preprocessed. State-action pairs $(s, a)$ and state-task pairs $(s, z)$ have their own preprocessors which are feedforward MLPs that embed their inputs into a 512-dimensional space. These embeddings are concatenated and passed through a third feedforward MLP which outputs a $a$-dimensional vector, where $a$ is the action-space dimensionality. A Tanh activation is used on the last layer to normalise their scale. As per \cite{fujimoto2019}'s recommendations, the policy is smoothed by adding Gaussian noise $\sigma$ to the actions during training. Note this is identical to the implementation of $\pi(s, z)$ as described in Appendix \ref{appendix: FB implementation}.\\

\noindent\textbf{Misc.} Layer normalisation \citep{ba2016layer} and Tanh activations are used in the first layer of all MLPs to standardise the inputs. $z$ sampling distribution $\mathcal{Z}$ is identical to FB's (Appendix \ref{appendix: z sampling implementation}).\\

\subsection{Laplacian Eigenfunctions Loss}\label{appendix: lap loss}
Laplacian eigenfunction features $\varphi(s)$ are learned as per \cite{wu2018laplacian}. They consider the symmetrized MDP graph Laplacian created by some policy $\pi$, defined as $L = \text{Id}-\frac{1}{2}(\mathcal{P}_\pi \text{diag}\rho^-1 + \text{diag}\rho^-1 (\mathcal{P}_\pi)^T)$. They learn the eigenfunctions of $L$ with the following:
\begin{equation}
\min_{\varphi}\mathbb{E}_{(s_t, s_{t+1}) \sim \mathcal{D}}\left[||\varphi(s_t) - \varphi(s_{t+1})||^2\right] + \lambda \mathbb{E}_{(s_t, s_+) \sim \mathcal{D}}\left[(\varphi(s)^\top\varphi(s_+))^2 - ||\varphi(s)||^2_2 - ||\varphi(s_+)||^2_2\right],
\end{equation}
which comes from \cite{koren2003spectral}.

\subsection{Value Conservative Universal Successor Features}\label{appendix: conservative successor features implementation}
This section develops \textit{value conservative} regularisation for use by Universal Successor Features (USF) \citep{barreto2017, borsa2018}, as alluded to in Chapter \ref{chapter: from low quality data}.

Recall from \colouredS \ref{chapter3: preliminaries} that successor features require a state-feature mapping $\varphi: \mathcal{S} \to \mathbb{R}^d$ which is usually obtained by some representation learning method \citep{barreto2017}. \textit{Universal} successor features are the expected discounted sum of these features, starting in state $s_0$, taking action $a_0$ and following the task-dependent policy $\pi_z$ thereafter
\begin{equation}
    \psi(s_0, a_0, z) := \mathbb{E}\left[ \sum_{t\geq0} \gamma^t \varphi(s_{t+1}) | s_0, a_0, \pi_z \right].
\end{equation}
USFs satisfy a Bellman equation \citep{borsa2018} and so can be trained using TD-learning on the Bellman residuals:
\begin{equation}\label{equation: usf loss}
    \mathcal{L}_{\text{SF}} = \mathbb{E}_{(s_t, a_t, s_{t+1}) \sim \mathcal{D}, z \sim \mathcal{Z}} \left( \psi(s_t, a_t, z)^\top z - \varphi(s_{t+1})^\top z - \gamma \bar{\psi}(s_{t+1}, \pi_z(s_{t+1}), z)^\top z \right)^2,
\end{equation}
where $\Bar{\psi}$ is a lagging target network updated via Polyak averaging, and $\mathcal{Z}$ is identical to that used for FB training (Appendix \ref{appendix: z sampling implementation}). As with FB representations, the policy maximises the $Q$ function defined by $\psi$:
\begin{equation}
    \pi_{z}(s) := \text{argmax}_a \psi (s,a,z)^\top z,
\end{equation}
and for continuous state and action spaces is trained in an actor critic formulation. Like FB, USF training requires next action samples $a_{t+1} \sim \pi_z(s_{t+1})$ for the TD targets. We therefore expect SFs to suffer the same failure mode discussed in \colouredS \ref{chapter3: conservative FB} (OOD state-action value overestimation) and to benefit from the same remedial measures (value conservatism). Training \textit{value-conservative successor features} ({\color{orange}VC-SF}) amounts to substituting the USF $Q$ function definition and loss for FB's in Equation \ref{equation: VC-FB loss}:
\begin{equation}
\mathcal{L}_{\text{VC-SF}} = {\color{orange}{\alpha \cdot \left( \mathbb{E}_{s \sim \mathcal{D}, a \sim \mu(a|s), z \sim \mathcal{Z}} [\psi(s, a, z)^{\top} z] - \mathbb{E}_{(s, a) \sim \mathcal{D}, z \sim \mathcal{Z}} [\psi(s, a, z)^{\top}z] \right)}} + \mathcal{L}_{\text{SF}}.
\end{equation}
Both the maximum value approximator $\mu(a|s)$ (Equation \ref{equation: logsumexp}, \colouredS \ref{appendix: max value estimator implementation}) and $\alpha$-tuning (Equation \ref{equation: alpha lagrangian}, \colouredS \ref{appendix: dynamically tuning alpha implementation}) can be extracted identically to the FB case with any occurrence of $F(s, a, z)^{\top} z$ substituted with $\psi(s, a, z)^{\top} z$. As USFs do not predict successor measures we cannot formulate measure-conservative USFs.

USFs require access to a feature map $\varphi: S \mapsto \mathbb{R}^d$ that maps states into an embedding space in which the reward is assumed to be linear \textit{i.e.} $R(s) = \varphi(s)^\top z$ with \textit{weights} $z \in \mathbb{R}^d$ representing a task \citep{barreto2017, borsa2018}. The USFs $\psi: S \times A \times \mathbb{R}^d \rightarrow \mathbb{R}^d$ are defined as the discounted sum of future features subject to a task-conditioned policy $\pi_z$, and are trained such that
\begin{align}
    \psi(s_0, a_0, z) &= \mathbb{E}\left[\sum_{t \geq 0} \gamma^t \varphi(s_{t+1}) | s_0, a_0, \pi_z \right] \qquad\; \; \; \; \; \qquad \forall \; s_0 \in S, a_0 \in A, z \in \mathbb{R}^d \\
    \pi(s, z) &\approx \arg\max_a \psi(s,a,z)^\top z, \; \; \qquad \qquad \qquad \qquad \forall \; s \in S, a \in A,  z \in \mathbb{R}^d.
\end{align}
During training candidate task weights are sampled from $\mathcal{Z}$; during evaluation, the test task weights are found by regressing labelled states onto the features: 
\begin{equation}\label{equation: usf task inference}
z_{\text{test}} \approx \argminA_z \mathbb{E}_{s \sim \mathcal{D}_{\text{test}}}[(R_{\text{test}}(s) - \varphi(s)^\top z)^2],
\end{equation}
before being passed to the policy. The features can be learned with Hilbert representations \citep{park2024foundation}, laplacian eigenfunctions, or contrastive methods \citep{touati2022does}.

\subsection{Universal Successor Features with Memory}\label{appendix: usf with memory}
We define \textit{memory-based} USFs as the discounted sum of future features extracted from the memory model's hidden state, subject to a memory-based policy $\pi_z(f_\pi(\tau^L)$:
\begin{align}
\psi(\tau_0^L, z) &= \mathbb{E}\left[\sum_{t \geq 0} \gamma^t \varphi(f_\psi(\tau_{t+1}^L)) \; | \;  \tau_0^L, \pi_z \right] \; \; \; \qquad \qquad  \forall \; \tau_0^L \in \mathcal{T}, z \in \mathbb{R}^d \\
\pi(f_\pi(\tau^L), z) &\approx \arg\max_a \psi(f_\psi(\tau^L),z)^\top z \; \; \; \qquad \qquad \qquad \; \forall \; \tau^L \in \mathcal{T},  z \in \mathbb{R}^d, 
\end{align}
where $f_\psi$ and $f_\pi$ are seperate memory models for $\psi$ and $\pi$, and the previous hidden state $h_{t-L-1}$ is dropped as an argument for brevity (\textit{c.f.} Equation \ref{equation: generalised memory model}). At test time, task embeddings are found via Equation \ref{equation: usf task inference}, but this time with reward-labelled trajectories rather than reward-labelled states:
\begin{equation}\label{equation: usf-mem task inference}
z_{\text{test}} \approx \argminA_z \mathbb{E}_{(\tau^L, R(s)) \sim \mathcal{D}_{\text{labelled}}}[(R_{\text{test}}(s) - \varphi(f_\psi(\tau^L)^\top z)^2],
\end{equation}
before being passed to the policy.

\section{GC-IQL} \label{appendix: gc-iql}
\subsection{Architecture}
We implement GC-IQL following \citep{park2023hiql}'s codebase. GC-IQL inherits all functionality from a base soft actor-critic agent \citep{Haarnoja2018}, but adds a soft conservative penalty to the goal-conditioned critic $V(s,g)$ updates. We refer the reader to the original GC-IQL paper \citep{park2023hiql} for details on the loss function used to train $V(s,g)$. Hyperparameters are reported in Table \ref{table: implementations-appendix/cql-and-td3-hyperparameters}.\\

\noindent\textbf{Critic(s).} GC-IQL trains double goal-conditioned value functions $V(s,g)$. The critics are feedforward MLPs that take a state-goal pair $(s, g)$ as input and output a value $\in \mathbb{R}^1$. During training the goals are sampled from the prior $\mathcal{G}$ described in \colouredS \ref{appendix: gc-iql goal sampling}.\\

\noindent\textbf{Actor.} The actor is a standard feedforward MLP taking the state $s$ as input and outputting an $2a$-dimensional vector, where $a$ is the action-space dimensionality. The actor predicts the mean and standard deviation of a Gaussian distribution for each action dimension; during training a value is sampled at random, during evaluation the mean is used.

\subsection{Goal Sampling Distribution $\mathcal{G}$}\label{appendix: gc-iql goal sampling}
Following \citep{park2023hiql}, goals are sampled from either random states, future states, or the current state with probabilities $0.3, 0.5$ and $0.2$ respectively. A geometric distribution $\text{Geom}(1 - \gamma)$ is used for the future state distribution, and the uniform distribution over the offline dataset is used for sampling random states.

\section{CQL} \label{appendix: cql implementation}
\subsection{Architecture}
We adopt the same implementation and hyperparameters as is used on the ExORL benchmark. CQL inherits all functionality from a base soft actor-critic agent \citep{Haarnoja2018}, but adds a conservative penalty to the critic updates (Equation \ref{equation: CQL update}). Hyperparameters are reported in Table \ref{table: implementations-appendix/cql-and-td3-hyperparameters}.\\

\noindent\textbf{Critic(s).} CQL employs double Q networks, where the target network is updated with Polyak averaging via a momentum coefficient. The critics are feedforward MLPs that take a state-action pair $(s, a)$ as input and output a value $\in \mathbb{R}^1$.\\

\noindent\textbf{Actor.} The actor is a standard feedforward MLP taking the state $s$ as input and outputting an $2a$-dimensional vector, where $a$ is the action-space dimensionality. The actor predicts the mean and standard deviation of a Gaussian distribution for each action dimension; during training a value is sampled at random, during evaluation the mean is used.

\section{TD3} \label{appendix: td3-(gru) implementation}
\subsection{Architecture}
We adopt the same implementation and hyperparameters as is used on the ExORL benchmark. Hyperparameters are reported in Table \ref{table: implementations-appendix/cql-and-td3-hyperparameters}. \\

\noindent\textbf{Critic(s).} TD3 employs double Q networks, where the target network is updated with Polyak averaging via a momentum coefficient. The critics are feedforward MLPs that take a state-action pair $(s, a)$ as input and output a value $\in \mathbb{R}^1$.\\

\noindent\textbf{Actor.} The actor is a standard feedforward MLP taking the state $s$ as input and outputting an $a$-dimensional vector, where $a$ is the action-space dimensionality. The policy is smoothed by adding Gaussian noise $\sigma$ to the actions during training.\\

\noindent\textbf{Misc.} As is usual with TD3, layer normalisation \citep{ba2016layer} is applied to the inputs of all networks.

\begin{table}[h]\caption{\textbf{Hyperparameters for CQL, Offline TD3, and GC-IQL.} Methods used in Chapter \ref{chapter: from low quality data} and \ref{chapter: under partial observability}.}\label{table: implementations-appendix/cql-and-td3-hyperparameters}
\centering
\scalebox{0.7}{
\begin{tabular}{llll} 
\toprule
\textbf{Hyperparameter}                       & \textbf{CQL} & \textbf{Offline TD3} & \textbf{GC-IQL}       \\ 
 \midrule
Critic dimensions                          & (1024, 1024) & (1024, 1024) & (1024, 1024)  \\
Actor dimensions                          & (1024, 1024) & (1024, 1024) & (1024, 1024)     \\
Learning steps                                & 1,000,000 & 1,000,000 & 1,000,000            \\
Batch size                                    & 1024 & 1024  & 1024                \\
Optimiser                                     & Adam & Adam  & Adam               \\
Learning rate                                 & 0.0001 & 0.0001 & 0.0001              \\
Discount $\gamma$                             & 0.98 (0.99 for maze) & 0.98 (0.99 for maze) & 0.98 (0.99 for maze) \\
Activations         & ReLU & ReLU  & ReLU               \\
 
Target network Polyak smoothing coefficient & 0.01 & 0.01 & 0.01                 \\ 
Sampled Actions Number & 3 & - & - \\
CQL $\alpha$ & 0.01 & - & -\\
CQL Lagrange & False & - & -\\
Std. deviation for policy smoothing $\sigma$  & - & 0.2   & -               \\
Truncation level for policy smoothing         & - & 0.3  & -         \\ 
IQL temperature & - & -  & 1         \\ 
IQL Expectile & - & -  & 0.7       \\ 
\bottomrule
\end{tabular}
}
\end{table}

\section{PEARL}\label{appendix: pearl implementation}
The hyperparameters for PEARL and MPC-DNN implementations used in Chapter \ref{chapter: no prior data} are reported in Table \ref{table: model-based hyperparameters}.

\begin{longtable}[c]{ l  l  l }
			\caption{\label{table: model-based hyperparameters} \textbf{PEARL and MPC-DNN hyperparameters.} Methods used in Chapter \ref{chapter: no prior data} only.}\\
			
			\toprule
			\textbf{Hyperparameter}                              & \textbf{PEARL} & \textbf{MPC-DNN} \\
			\midrule
			\endfirsthead
			
			\multicolumn{3}{l}{{\tablename\  \thetable{} -- continued from previous page}} \\
			
			\textbf{Hyperparameter}                              & \textbf{PEARL} & \textbf{MPC-DNN} \\
			\midrule
			\endhead
			
			\hline
			\multicolumn{3}{r}{\textit{continued on next page}}
			\endfoot
			
			\bottomrule
			\endlastfoot
			
			Horizon minutes ($H$)                                          & 300    & 300        \\
			History ($h$)                                                   & 2 & 2 \\
		Adam stepsize ($\alpha$)                & 0.0003      & 0.0003   \\
		Num. epochs                                          & 25 & 25             \\
		Minibatch size                                       & 32  & 32           \\
		No. of candidate actions ($U$)                                         & 25 & 25          \\
		No. of particles ($M$)                                         & 10 & n/a          \\
		MPPI Iterations ($N$)                                   & 5 & -- \\
		Network Fully Connected Layers                 & 5   & 5           \\
		Network Layer Dimensions                       & 200  & 200          \\
		Network Activation Function                    & Tanh  & Tanh         \\
		Ensemble Models ($K$)                & 5      & n/a       \\
		System ID Minutes ($C$)                   & 180     & n/a      \\
		Discount ($\gamma$)                                         & 1 & 1           \\
		\end{longtable}

\section{PPO}\label{appendix: PPO implementation}
The hyperparameters for PPO used in Chapter \ref{chapter: no prior data} are reported in Table \ref{table: ppo hyperparameters}. Our implementation follows the original implementation in \citep{schulman2017}.

\begin{longtable}{ l  l }
			\caption{\label{table: ppo hyperparameters} \textbf{PPO hyperparameters.} Used in Chapter \ref{chapter: no prior data} only.}\\
			
			\toprule
			\textbf{Hyperparameter}                              & \textbf{Value} \\
			\midrule
			\endfirsthead
			
			\multicolumn{2}{l}{{\tablename\  \thetable{} -- continued from previous page}} \\
			
			\textbf{Hyperparameter}     & \textbf{PPO}\\
			\midrule
			\endhead
			
			\hline
			\multicolumn{2}{r}{\textit{continued on next page}}
			\endfoot
			
			\bottomrule
			\endlastfoot
					Batch Size                                           & 32            \\
					History ($h$)                                                   & 2 \\
		Adam stepsize ($\alpha$)                & 0.0003         \\
		Epochs per update                                          & 50             \\
		Discount ($\gamma$)                                         & 0.99           \\
		GAE parameter ($\lambda$)                                    & 0.95           \\
		Clipping parameter									& 0.2            \\
		Actor Network Fully Connected Layers                 & 5              \\
		Actor Network Layer Dimensions                       & 200            \\
		Actor Network Activation Function                    & Tanh           \\
		Critic Network Fully Connected Layers                & 5           \\
		Critic Network Layer Dimensions                      & 200            \\
		Critic Network Activation Function                   & Tanh           \\
		\end{longtable}

\section{SAC}\label{appendix: SAC implementation}
The hyperparameters for SAC used in Chapter \ref{chapter: no prior data} are reported in Table \ref{table: ppo hyperparameters}. Our implementation follows the the open-source PyTorch implementation by \citep{yarats2020soft}.
\begin{longtable}{ l  l }
			\caption{\label{table: sac hyperparameters} \textbf{SAC hyperparameters.} Used in Chapter \ref{chapter: no prior data} only.}\\
			
			\toprule
			\textbf{Hyperparameter}                              & \textbf{Value} \\
			\midrule
			\endfirsthead
			
			\multicolumn{2}{l}{{\tablename\  \thetable{} -- continued from previous page}} \\
			
			\textbf{Hyperparameter}     & \textbf{SAC}\\
			\midrule
			\endhead
			
			\hline
			\multicolumn{2}{r}{\textit{continued on next page}}
			\endfoot
			
			\bottomrule
			\endlastfoot
					Batch Size                                           & 256            \\
					History ($h$)                                                   & 2 \\
		Adam stepsize ($\alpha$)                & 0.0003         \\
		Epochs per update                                          & 20             \\
		Discount ($\gamma$)                                         & 0.99           \\
		Target Smoothing Coefficient ($\tau$)                                    & 0.005           \\
		Hidden Layers (all networks)                 & 2              \\
		Layer Dimensions                       & 256            \\
		Activation Function                    & ReLU           \\
		Reward Scale                            & 2             \\
		Reward Emissions Parameter ($\phi$)       & 1              \\
		\end{longtable}
		
\begin{longtable}{ l  l  l  l }
			\caption{\label{table: sac oracle hyperparameters} \textbf{SAC Oracle hyperparameters.} Used in Chapter \ref{chapter: no prior data} only.}\\
			
			\toprule
			\textbf{Hyperparameter}                              & \textbf{Mixed-use} & \textbf{Offices} & \textbf{Seminar Centre} \\
			\midrule
			\endfirsthead
			
			\multicolumn{2}{l}{{\tablename\  \thetable{} -- continued from previous page}} \\
			
			\textbf{Hyperparameter}                              & \textbf{Mixed-use} & \textbf{Offices} & \textbf{Seminar Centre} \\
			\midrule
			\endhead
			
			\hline
			\multicolumn{4}{r}{\textit{continued on next page}}
			\endfoot
			
			\bottomrule
			\endlastfoot
					Batch Size                                           & 256   & 256 & 1024        \\
					History ($h$)                                                   & 2 & 2 & 2 \\
		Adam stepsize ($\alpha$)                & 0.0003  & 0.0015 &   0.0080    \\
		Epochs per update                                          & 1    & 1 & 1         \\
		Discount ($\gamma$)                                         & 0.9864  & 0.7539 & 0.8533          \\
		Target Smoothing Coefficient ($\tau$)                                    & 0.07815    & 0.2532 & 0.2890       \\
		Hidden Layers (all networks)                 & 2 & 2 & 2              \\
		Layer Dimensions                       & 256  & 256 & 256          \\
		Activation Function                    & ReLU   & ReLU & ReLU        \\
		Reward Scale                            & 2 & 20  &    20              \\
		\end{longtable}

\section{Rule Based Controllers}\label{appendix: RBC implementation}
In Chapter \ref{chapter: no prior data}, the RBC's objective is to maintain temperature at 22\(^\circ\)C in the building. To achieve this it selects actions as follows:
\begin{equation} \label{eq: rbc policy}
	a_t^i[t+1]= 
	\begin{cases}
		a_t^i[t] + 0.5^\circ C:& T_{obs}^i \leq 21.2^\circ C \\
		a_t^i[t] - 0.5^\circ C:& T_{obs}^i \geq 22.8^\circ C \\
		a_t^i[t]:& 21.2^\circ C \leq T_{obs}^i \leq 22.8^\circ C 
	\end{cases}
\end{equation}

\section{Random Walk Exploration} \label{appendix: lazic}
In this section we outline the random walk exploration baseline reported in Figure \ref{fig: chapter5/system id}, it follows that used by Lazic et al.'s agent for energy-efficient data-centre cooling \citep{lazic2018data}. Their agent performs uniform random walk in each control variable bounded to a safe range over the action space informed by historical data if available, or initialised conservatively and expanded if no data is available. Stepwise changes in action selection are bounded to mitigate potential hardware damage that may be caused by large swings in chosen actions.

Formally their policy as follows. Let $a^i[t]$ be the value of action $i$ at timestep $t$, with $i \in \mathcal{A}$. Let [$a_{min}^i$, $a_{max}^i$] be the safe operating range for action $i$, and let $\Delta^i$ be the maximum allowable change in action $a^i$ between timesteps. The system identification strategy, $\pi_{system ID}$, is therefore:
\begin{equation} \label{eq: system ID}
a^i[t+1] = \max\left(a_{min}^i, \min(a_{max}^i, a^i[t] + v^i)\right), \; v^i \sim \text{Uniform}\left(- \Delta ^i, \Delta^i\right) \; .
\end{equation}

\section{Computational Resources}\label{appendix: computational resources}
\noindent\textbf{Chapter \ref{chapter: from low quality data}}. $\;$ Our models are trained on NVIDIA A100 GPUs. Training a single-task offline RL method to solve one task on one GPU takes approximately 4 hours. FB and SF solve one domain (for all tasks) on one GPU in approximately 4 hours. \textit{Conservative} FB variants solve one domain (for all tasks) on one GPU in approximately 12 hours. As a result, our core experiments on the 100k datasets used approximately 110 GPU days of compute.\\

\noindent\textbf{Chapter \ref{chapter: under partial observability}}. $\;$ Again, we train our models on NVIDIA A100 GPUs. One run of FB, FB-stack and HILP on one domain (for all tasks) takes approximately 6 hours on one GPU. One \texttt{run} of the FB-M on one domain (for all tasks) on one GPU in approximately 20 hours. As a result, our core experiments on the ExORL benchmark used approximately 65 GPU days of compute. \\

\noindent\textbf{Chapter \ref{chapter: no prior data}}. $\;$ Models were trained locally on a machine with the following specifications:
\begin{itemize}
    \item RAM: 32GB 4X8GB 2666MHz DDR4
    \item CPU: Intel Core™ i7-8700 (6 Cores/12MB/12T/up to 4.6GHz/65W)
    \item GPU: 1X NVIDIA GeForce RTX 3050 Ti 
\end{itemize}

\chapter{Extended Results}\label{appendix: extended results}
In this section we report a full breakdown of our experimental results from throughout the thesis.\\

\noindent\textbf{Chapter \ref{chapter: from low quality data}.} $\;$ Table \ref{table: chapter3/full 100k results} reports results for methods trained on the 100k sub-sampled datasets, and Table \ref{table: chapter3/full dataset results} reports results for methods trained on the full datasets. We report the aggregate results for all metrics in \citep{agarwal2021deep} in Table \ref{table: chapter3/aggregate rlliable results}. D4RL results are reported in Table \ref{table: chapter3/d4rl full results}. \\

\noindent\textbf{Chapter \ref{chapter: under partial observability}.} Table \ref{table: chapter4/partially observed states} reports full results on the partially observed states experiments from \colouredS \ref{subsection: partially observed states}, and Table \ref{table: chapter4/full changed dynamics} reports full results partially observed dynamics experiments from \colouredS \ref{subsection: partially observed dynamics}.

\newpage
\begin{table}[ht]
\centering
\caption{\label{table: chapter3/full 100k results}\textbf{100k dataset experimental results on ExORL.} For each dataset-domain pair, we report the score at the step for which the all-task IQM is maximised when averaging across 5 seeds, and the constituent task scores at that step. Bracketed numbers represent the 95\% confidence interval obtained by a stratified bootstrap.}
\scalebox{0.5}{
\setlength{\tabcolsep}{5pt}
\begin{tabular}{lllllllllll}
\toprule
 Dataset & Domain  & Task  & CQL & Offline TD3 & GC-IQL & SF-LAP & FB & \textbf{VC-FB (ours)} & \textbf{\textit{D}VC-FB (ours)} & \textbf{MC-FB (ours)} \\
\midrule
\multirow[c]{17}{*}{\vspace{-9ex} {\textsc{Rnd}-100k}} & \multirow[c]{4}{*}{Walker} & Walk & $138$ \text{\tiny{($122-140$)}} & $210$ \text{\tiny{($205-231$)}} & $118$ \text{\tiny{($114-126$)}} & $58$ \text{\tiny{($33-104$)}} & $184$ \text{\tiny{($123-278$)}} & $446$ \text{\tiny{($435-460$)}} & $394$ \text{\tiny{($166-512$)}} & $247$ \text{\tiny{($164-299$)}} \\
 &  & Stand & $386$ \text{\tiny{($375-391$)}} & $362$ \text{\tiny{($335-378$)}} & $284$ \text{\tiny{($260-313$)}} & $190$ \text{\tiny{($128-233$)}} & $558$ \text{\tiny{($498-637$)}} & $624$ \text{\tiny{($604-639$)}} & $590$ \text{\tiny{($557-622$)}} & $480$ \text{\tiny{($402-517$)}} \\
 &  & Run & $71$ \text{\tiny{($64-75$)}} & $84$ \text{\tiny{($79-88$)}} & $51$ \text{\tiny{($42-56$)}} & $34$ \text{\tiny{($27-41$)}} & $101$ \text{\tiny{($90-135$)}} & $179$ \text{\tiny{($165-197$)}} & $134$ \text{\tiny{($77-191$)}} & $106$ \text{\tiny{($72-137$)}} \\
 &  & Flip & $153$ \text{\tiny{($135-174$)}} & $162$ \text{\tiny{($148-171$)}}  & $157$ \text{\tiny{($152-173$)}} & $70$ \text{\tiny{($56-84$)}} & $163$ \text{\tiny{($90-212$)}} & $325$ \text{\tiny{($292-350$)}} & $250$ \text{\tiny{($215-286$)}} & $164$ \text{\tiny{($131-192$)}} \\
\addlinespace \cline{2-11} \addlinespace
 & \multirow[c]{4}{*}{Quadruped} & Stand & $167$ \text{\tiny{($73-268$)}} & $119$ \text{\tiny{($9-338$)}}  & $43$ \text{\tiny{($14-173$)}} & $108$ \text{\tiny{($51-192$)}} & $134$ \text{\tiny{($86-176$)}} & $331$ \text{\tiny{($190-410$)}} & $269$ \text{\tiny{($152-385$)}} & $171$ \text{\tiny{($71-372$)}} \\
\bottomrule
\end{tabular}
}
\end{table}

\begin{table}[ht]
\caption{\label{table: chapter3/full dataset results}\textbf{Full dataset experimental results on ExORL.} For each dataset-domain pair, we report the score at the step for which the all-task IQM is maximised when averaging across 5 seeds, and the constituent task scores at that step. Bracketed numbers represent the 95\% confidence interval obtained by a stratified bootstrap.}
\centering
\scalebox{0.70}{
\setlength{\tabcolsep}{5pt}
\begin{tabular}{llllll}
\toprule
Dataset & Domain  & Task  & FB & \textbf{VC-FB (ours)} & \textbf{MC-FB (ours)} \\
\addlinespace \cline{1-6} \addlinespace
\multirow[c]{17}{*}{\vspace{-9ex} \textsc{Rnd}} & \multirow[c]{4}{*}{Walker} & Walk & $821$ \text{\tiny{($758-883$)}} & $864$ \text{\tiny{($850-879$)}} & $792$ \text{\tiny{($728-857$)}} \\
 &  & Stand & $928$ \text{\tiny{($925-930$)}} & $878$ \text{\tiny{($854-903$)}} & $873$ \text{\tiny{($812-934$)}} \\
 &  & Run & $281$ \text{\tiny{($242-320$)}} & $351$ \text{\tiny{($328-374$)}} & $343$ \text{\tiny{($323-366$)}} \\
 &  & Flip & $525$ \text{\tiny{($452-598$)}} & $542$ \text{\tiny{($513-571$)}} & $598$ \text{\tiny{($538-657$)}} \\
\addlinespace \cline{2-6} \addlinespace
 & \multirow[c]{5}{*}{Quadruped} & Stand & $957$ \text{\tiny{($952-963$)}} & $863$ \text{\tiny{($777-950$)}} & $949$ \text{\tiny{($939-958$)}} \\
 &  & Roll  Fast & $574$ \text{\tiny{($553-599$)}} & $512$ \text{\tiny{($471-553$)}} & $565$ \text{\tiny{($555-575$)}} \\
 &  & Roll  & $920$ \text{\tiny{($895-944$)}} & $831$ \text{\tiny{($741-920$)}} & $890$ \text{\tiny{($874-906$)}} \\
\bottomrule
\end{tabular}}
\end{table}

\clearpage

\begin{table}
\centering
\caption{\textbf{Aggregate zero-shot performance on ExORL for all evaluation statistics recommended by \cite{agarwal2021deep}.} VC-FB outperforms all methods across all evaluation statistics. $\uparrow$ means a higher score is better; $\downarrow$ means a lower score is better. Note that the optimality gap is large because we set $\gamma = 1000$ and for many dataset-domain-tasks the maximum achievable score is far from 1000.}
\label{table: chapter3/aggregate rlliable results}
\scalebox{0.8}{
\begin{tabular}{@{}lllllll}
\toprule
                   Statistic             & SF-LAP & GC-IQL & FB   &    CQL      & \textbf{MC-FB (ours)} & \textbf{VC-FB (ours)}  \\
\midrule
IQM $\uparrow$                             & $92$     & $95$ & $99$   & $128$  & $136$   & $\mathbf{148}$  \\
Mean $\uparrow$                             & $87$     & $126$ & $118$  &  $138$ & $142$   & $\mathbf{154}$  \\
Median $\uparrow$                          & $108$    & $104$ & $104$  &  $132$  & $133$   & $\mathbf{144}$  \\
Optimality Gap $\downarrow$                   & $0.92$   & $0.88$ & $0.89$ & $0.87$ & $0.86$  & $\mathbf{0.84}$ \\
\bottomrule
\end{tabular}
}
\end{table}

\begin{table}
\centering
\caption{\textbf{D4RL experimental results.} For each dataset-domain pair, we report the score at the step for which the IQM is maximised when averaging across 3 seeds. Bracketed numbers represent the 95\% confidence interval obtained by a stratified bootstrap..}
\label{table: chapter3/d4rl full results}
\scalebox{0.8}{
\begin{tabular}{lllllll}
\toprule
 Domain & Dataset  & CQL & SF-LAP & FB & \textbf{VC-FB (ours)} & \textbf{MC-FB (ours)} \\
\midrule
\multirow[m]{3}{*}{HalfCheetah} & medium & $36$ $\text{\tiny{(36-36)}}$ & $4$ $\text{\tiny{(1-7)}}$ & $3$ $\text{\tiny{(-1-8)}}$ & $39$ $\text{\tiny{(38-40)}}$ & $32$ $\text{\tiny{(28-36)}}$ \\
 & medium-expert & $43$ $\text{\tiny{(37-51)}}$ & $26$ $\text{\tiny{(20-32)}}$ & $3$ $\text{\tiny{(0-6)}}$ & $27$ $\text{\tiny{(24-28)}}$ & $24$ $\text{\tiny{(21-30)}}$ \\
 & medium-replay & $42$ $\text{\tiny{(42-42)}}$ & $29$ $\text{\tiny{(29-30)}}$ & $7$ $\text{\tiny{(2-11)}}$ & $31$ $\text{\tiny{(26-36)}}$ & $20$ $\text{\tiny{(18-22)}}$ \\
\midrule
\multirow[m]{3}{*}{Walker2d} & medium & $70$ $\text{\tiny{(68-73)}}$ & $7$ $\text{\tiny{(1-13)}}$ & $7$ $\text{\tiny{(0-14)}}$ & $42$ $\text{\tiny{(37-45)}}$ & $36$ $\text{\tiny{(33-40)}}$ \\
 & medium-expert & $102$ $\text{\tiny{(97-107)}}$ & $15$ $\text{\tiny{(0-31)}}$ & $3$ $\text{\tiny{(1-5)}}$ & $82$ $\text{\tiny{(72-92)}}$ & $74$ $\text{\tiny{(66-77)}}$ \\
 & medium-replay & $13$ $\text{\tiny{(11-14)}}$ & $11$ $\text{\tiny{(8-15)}}$ & $12$ $\text{\tiny{(7-17)}}$ & $20$ $\text{\tiny{(19-21)}}$ & $22$ $\text{\tiny{(19-24)}}$ \\
\midrule
All & All & $48$ & $15$ & $5$ & $34$ & $28$ \\
\bottomrule
\end{tabular}
}
\end{table}

\begin{table}
\centering
\caption{\textbf{Full results on partially observed states (5 seeds).} For each dataset-domain pair, we report the score at the step for which the all-task IQM is maximised when averaging across 5 seeds $\pm$ the standard deviation.}
\label{table: chapter4/partially observed states}
\scalebox{0.70}{
\begin{tabular}{lllllll|l}
\toprule
 \texttt{Environment} & \texttt{Occlusion} & \texttt{Task}  & \texttt{FB} & \texttt{HILP} & \texttt{FB-stack} & \texttt{FB-M (ours)} & \texttt{MDP} \\
\midrule \addlinespace
\multirow[c]{10}{*}{\texttt{Cheetah}} & \multirow[c]{5}{*}{\texttt{flickering}} &\texttt{overall} & $182 \pm \text{\tiny{$25$}}$ & $75 \pm \text{\tiny{$19$}}$ & $109 \pm \text{\tiny{$26$}}$ & $173 \pm \text{\tiny{$51$}}$ & $474 \pm \text{\tiny{$129$}}$ \\
 &  & \texttt{run} & $54 \pm \text{\tiny{$14$}}$ & $31 \pm \text{\tiny{$13$}}$ & $25 \pm \text{\tiny{$16$}}$ & $75 \pm \text{\tiny{$37$}}$ & $183 \pm \text{\tiny{$59$}}$ \\
 &  & \texttt{run backward} & $52 \pm \text{\tiny{$23$}}$ & $8 \pm \text{\tiny{$5$}}$ & $31 \pm \text{\tiny{$11$}}$ & $55 \pm \text{\tiny{$15$}}$ & $176 \pm \text{\tiny{$53$}}$ \\
 &  & \texttt{walk} & $279 \pm \text{\tiny{$65$}}$ & $184 \pm \text{\tiny{$66$}}$ & $186 \pm \text{\tiny{$103$}}$ & $306 \pm \text{\tiny{$175$}}$ & $726 \pm \text{\tiny{$194$}}$ \\
 &  & \texttt{walk backward} & $293 \pm \text{\tiny{$87$}}$ & $61 \pm \text{\tiny{$21$}}$ & $178 \pm \text{\tiny{$67$}}$ & $278 \pm \text{\tiny{$70$}}$ & $812 \pm \text{\tiny{$217$}}$ \\
\addlinespace\cline{2-8}\addlinespace
 & \multirow[c]{5}{*}{\texttt{noisy}} & \texttt{overall} & $213 \pm \text{\tiny{$53$}}$ & $102 \pm \text{\tiny{$59$}}$ & $116 \pm \text{\tiny{$26$}}$ & $150 \pm \text{\tiny{$59$}}$ & $474 \pm \text{\tiny{$129$}}$ \\
 &  & \texttt{run} & $65 \pm \text{\tiny{$9$}}$ & $28 \pm \text{\tiny{$30$}}$ & $34 \pm \text{\tiny{$11$}}$ & $34 \pm \text{\tiny{$17$}}$ & $183 \pm \text{\tiny{$59$}}$ \\
\bottomrule
\end{tabular}
}
\end{table}

\begin{table}
\centering
\caption{\textbf{Full results on ExORL changed dynamics experiments (5 seeds).} For each dataset-domain pair, we report the score at the step for which the all-task IQM is maximised when averaging across 5 seeds $\pm$ the standard deviation.}
\label{table: chapter4/full changed dynamics} 
\scalebox{0.70}{
\begin{tabular}{lllllll|l} 
\toprule
 \texttt{Dynamics} & \texttt{Environment} & \texttt{Task} & \texttt{FB} & \texttt{HILP} & \texttt{FB-stack} & \texttt{FB-M (ours)} & \texttt{MDP} \\
\midrule \addlinespace
\multirow[c]{15}{*}{\texttt{Interpolation}} & \multirow[c]{5}{*}{\texttt{Cheetah}} & \texttt{overall} & $476 \pm \text{\tiny{$77$}}$ & $67 \pm \text{\tiny{$37$}}$ & $156 \pm \text{\tiny{$55$}}$ & $453 \pm \text{\tiny{$120$}}$ & $474 \pm \text{\tiny{$129$}}$ \\
 &  & \texttt{run} & $167 \pm \text{\tiny{$59$}}$ & $17 \pm \text{\tiny{$11$}}$ & $59 \pm \text{\tiny{$18$}}$ & $150 \pm \text{\tiny{$68$}}$ & $183 \pm \text{\tiny{$59$}}$ \\
 &  & \texttt{run backward} & $166 \pm \text{\tiny{$21$}}$ & $6 \pm \text{\tiny{$21$}}$ & $36 \pm \text{\tiny{$38$}}$ & $192 \pm \text{\tiny{$66$}}$ & $176 \pm \text{\tiny{$53$}}$ \\
 &  & \texttt{walk} & $816 \pm \text{\tiny{$280$}}$ & $84 \pm \text{\tiny{$43$}}$ & $312 \pm \text{\tiny{$52$}}$ & $483 \pm \text{\tiny{$242$}}$ & $726 \pm \text{\tiny{$194$}}$ \\
 &  & \texttt{walk backward} & $777 \pm \text{\tiny{$71$}}$ & $160 \pm \text{\tiny{$83$}}$ & $186 \pm \text{\tiny{$226$}}$ & $956 \pm \text{\tiny{$167$}}$ & $812 \pm \text{\tiny{$217$}}$ \\
\addlinespace\cline{2-8}\addlinespace
 & \multirow[c]{5}{*}{\texttt{Quadruped}}  & \texttt{overall} & $551 \pm \text{\tiny{$82$}}$ & $186 \pm \text{\tiny{$55$}}$ & $394 \pm \text{\tiny{$76$}}$ & $767 \pm \text{\tiny{$24$}}$ & $729 \pm \text{\tiny{$6$}}$ \\
 &  & \texttt{jump} & $566 \pm \text{\tiny{$128$}}$ & $291 \pm \text{\tiny{$188$}}$ & $412 \pm \text{\tiny{$69$}}$ & $787 \pm \text{\tiny{$22$}}$ & $737 \pm \text{\tiny{$21$}}$ \\
\bottomrule
\end{tabular}
}
\end{table}

\section{Negative Results}\label{appendix: negative results}
In this section we provide detail on experiments we attempted in Chapter \ref{chapter: from low quality data}, but which did not provide results significant enough to be included in the main body.

\subsection{Downstream Finetuning}
If we relax the zero-shot requirement, could pre-trained conservative FB representations be finetuned on new tasks or domains? Base CQL models have been finetuned effectively on unseen tasks using both online and offline data \citep{kumar2022}, and we had hoped to replicate similar results with VC-FB and MC-FB. We ran offline and online finetuning experiments and provide details on their setups and results below. All experiments were conducted on the Walker domain.

\textbf{Offline finetuning.} We considered a setting where models are trained on a low quality dataset initially, before a high quality dataset becomes available downstream. We used models trained on the \textsc{Random}-100k dataset and finetuned them on both the full \textsc{Rnd} and \textsc{Rnd}-100k datasets, with models trained from scratch used as our baseline. Finetuning involved the usual training protocol as described in Algorithm \ref{alg: VC-FB}, but we limited the number of learning steps to 250k.

We found that though performance improved during finetuning, it improved no quicker than the models trained from scratch. This held for both the full \textsc{Rnd} and \textsc{Rnd}-100k datasets. We conclude that the parameter initialisation delivered after training on a low quality dataset does not obviously expedite learning when high quality data becomes available.

\begin{figure}[h]
    \centering
    \includegraphics{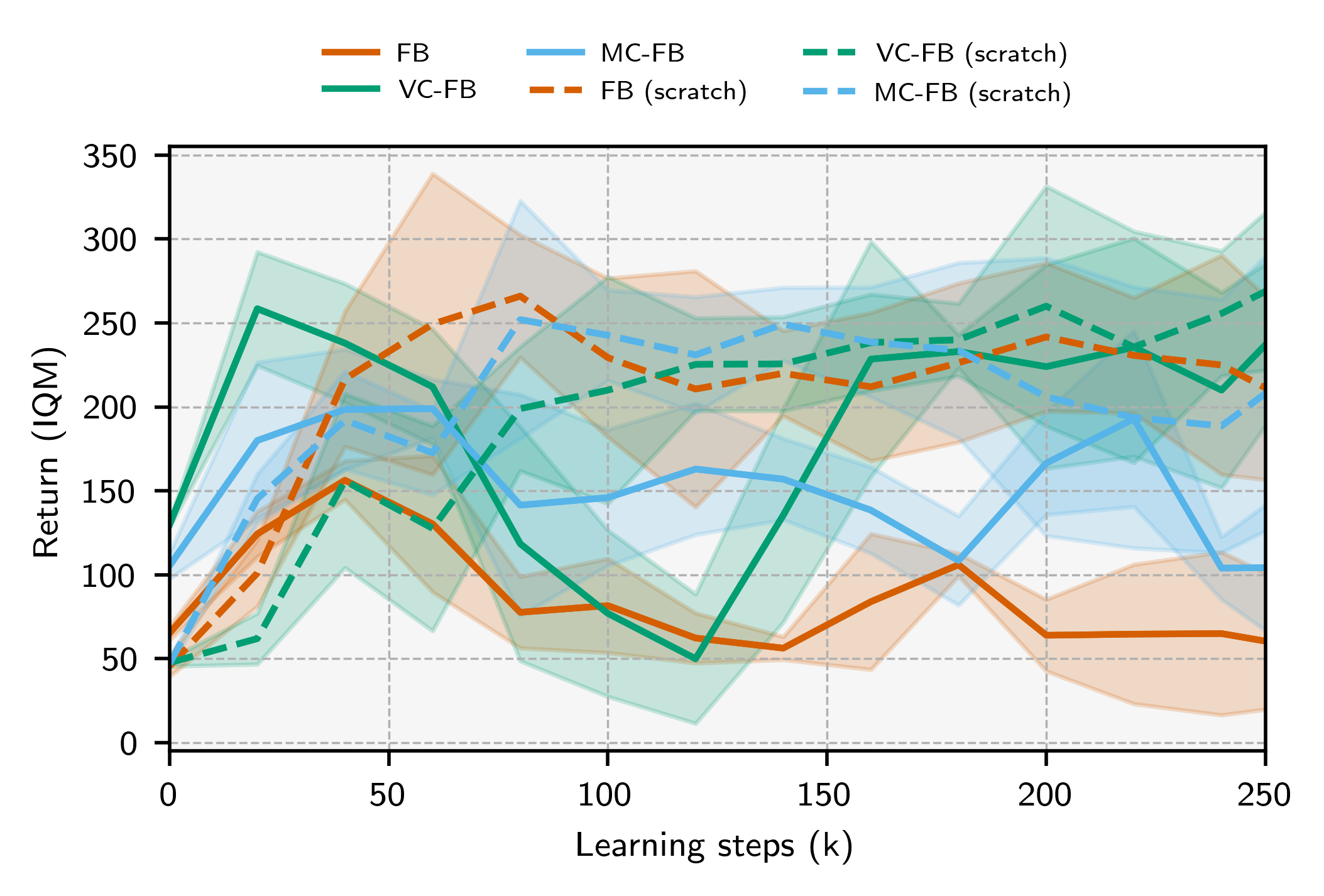}
    \caption{\textbf{Learning curves for methods finetuned on the full \textsc{Rnd} dataset.} Solid lines represent base models trained on \textsc{Random}-100k, then finetuned; dashed lines represent models trained from scratch. The finetuned models perform no better than models trained from scratch after 250k learning steps, suggesting model re-training is currently a better strategy than offline finetuning.}
    \label{fig: negative-results/offline finetuning}
\end{figure}

\textbf{Online finetuning.}
We considered the online finetuning setup where a trained representation is deployed in the target environment, required to complete a specified task, and allowed to collect a replay buffer of reward-labelled online experience. We followed a standard online RL protocol where a batch of transitions was sampled from the online replay buffer after each environment step for use in updating the model's parameters. We experimented with fixing $z$ to the target task during in the actor updates (Line 16, Algorithm \ref{alg: VC-FB}), but found it caused a quick, irrecoverable collapse in actor performance. This suggested uniform samples from $\mathcal{Z}$ provide a form of regularisation. We granted the agents 500k steps of interaction for online finetuning.

We found that performance never improved beyond the pre-trained (init) performance during finetuning. We speculated that this was similar to the well-documented failure mode of online finetuning of CQL \citep{nakamoto2023cal}, namely taking sub-optimal actions in the real env, observing unexpectedly high reward, and updating their policy toward these sub-optimal actions. But we note that FB representations do not update w.r.t observed rewards, and so conclude this cannot be the failure mode. Instead it seems likely that FB algorithms cannot use the narrow, unexploratory experience obtained from attempting to perform a specific task to improve model performance.

\begin{figure}[h]
    \centering
    \includegraphics{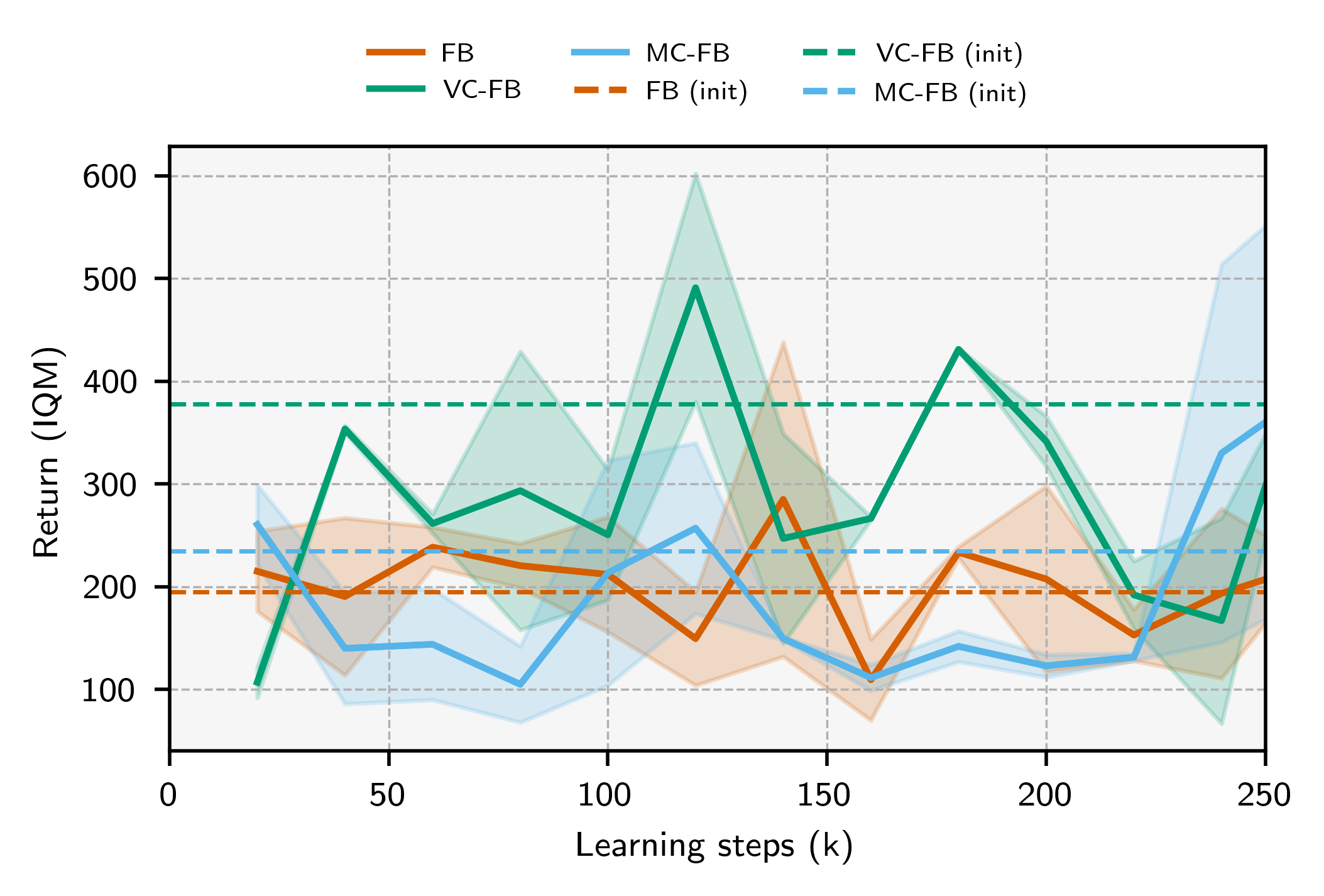}
    \caption{\textbf{Learning curves for online finetuning.} The performance at the end of pre-training (init performance) is plotted as a dashed line for each method. None of the methods consistently outperform their init performance after 250k online transitions.}
    \label{fig: negative-results/online finetuning}
\end{figure}

We believe resolving issues associated with finetuning conservative FB algorithms once the zero-shot requirement is relaxed is an important future direction and hope that details of our negative attempts to this end help facilitate future research.

\chapter{Learning Curves \& Hyperparameter Sweeps}\label{learning curves appendix}

\begin{figure}[ht]
    \centering
    \includegraphics[scale=1]{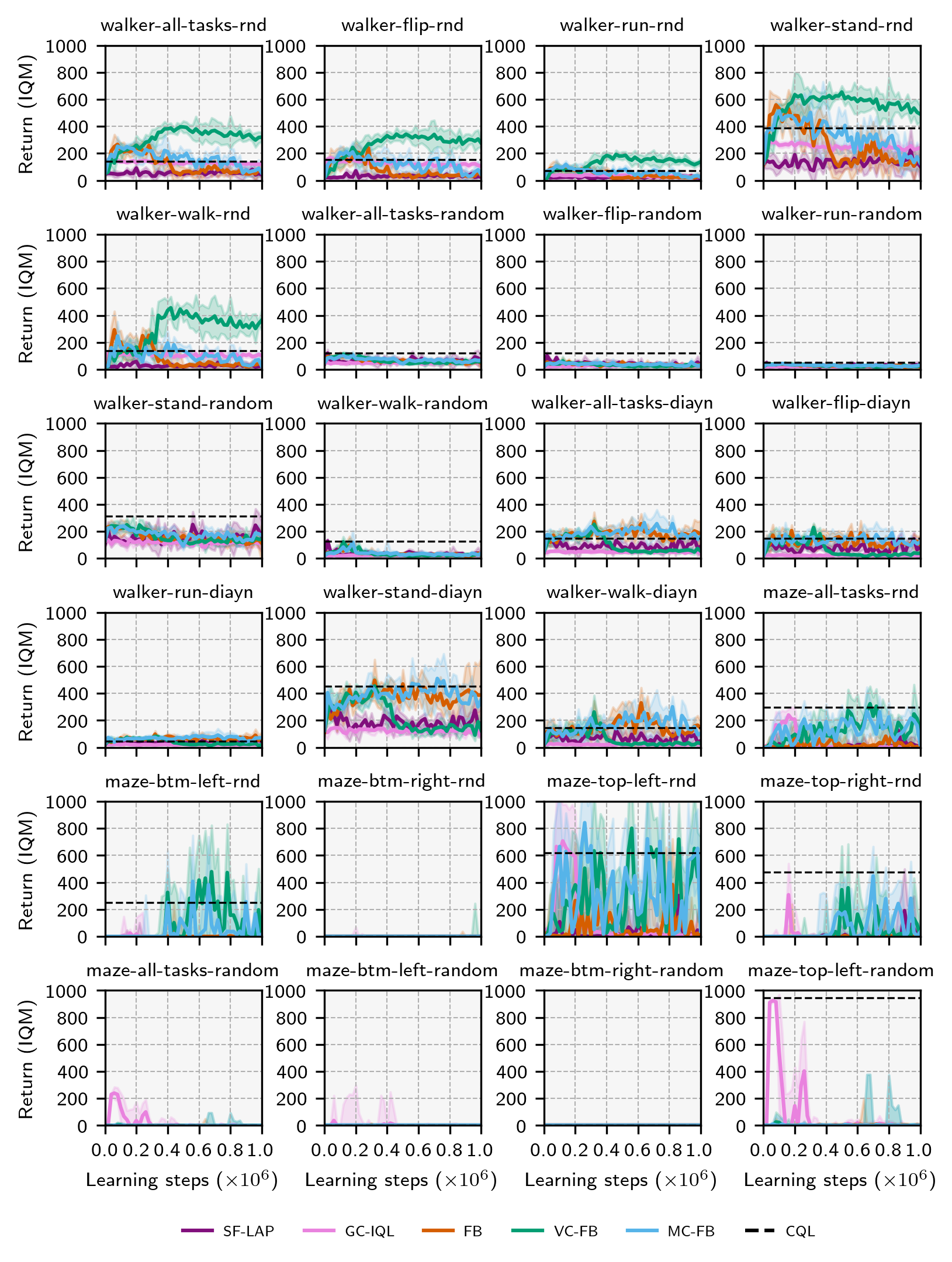}
    \caption{\textbf{Learning Curves (1/3)}. Models are evaluated every 20,000 timesteps where we perform 10 rollouts and record the IQM. Curves are the IQM of this value across 5 seeds; shaded areas are one standard deviation.}
    \label{fig: chapter3/learning curves 1}
\end{figure}

\begin{figure}
    \centering
    \includegraphics[scale=1]{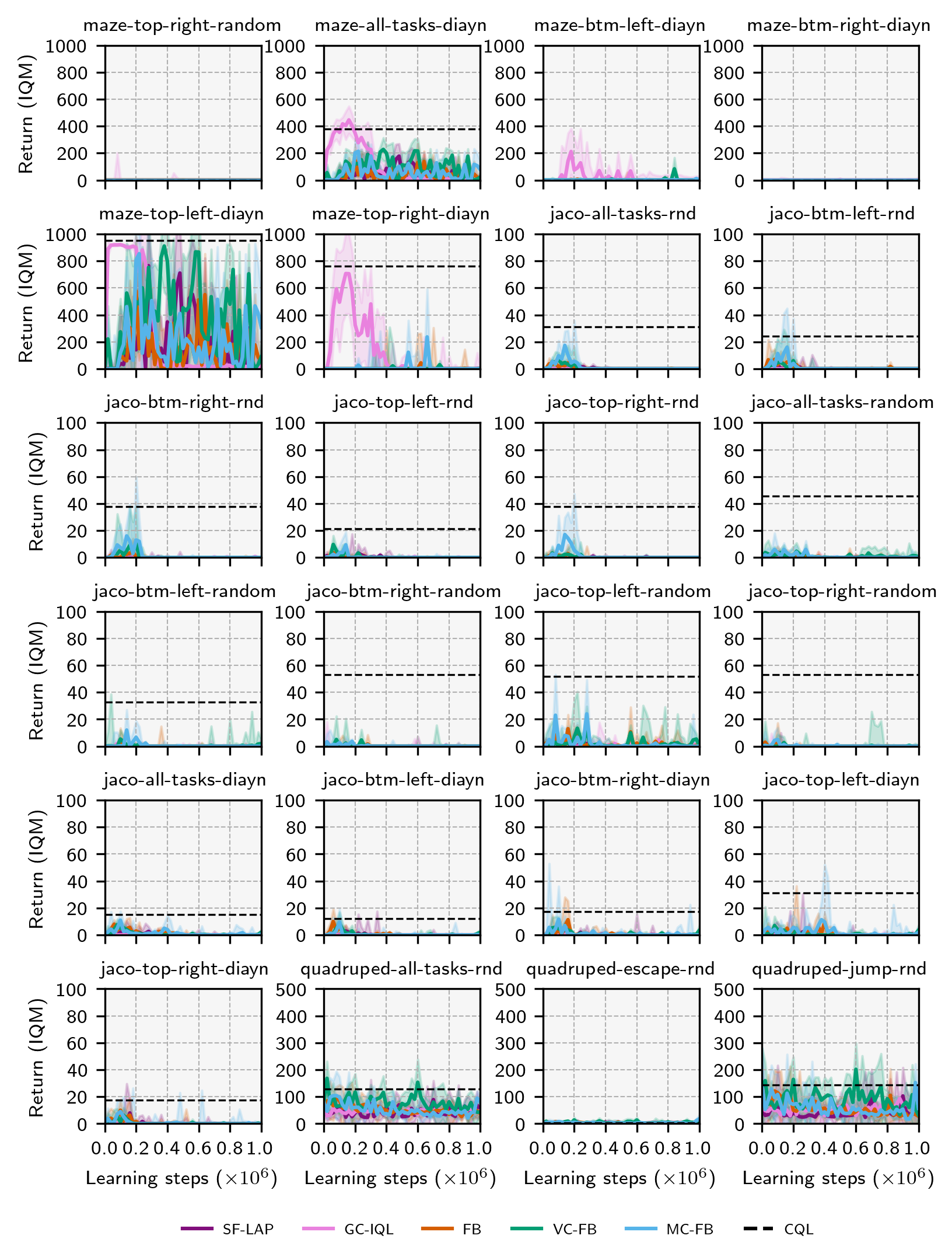}
    \caption{\textbf{Learning Curves (2/3)}. Models are evaluated every 20,000 timesteps where we perform 10 rollouts and record the IQM. Curves are the IQM of this value across 5 seeds; shaded areas are one standard deviation.}
    \label{fig: chapter3/learning curves 2}
\end{figure}

\begin{figure}
    \centering
    \includegraphics[scale=1]{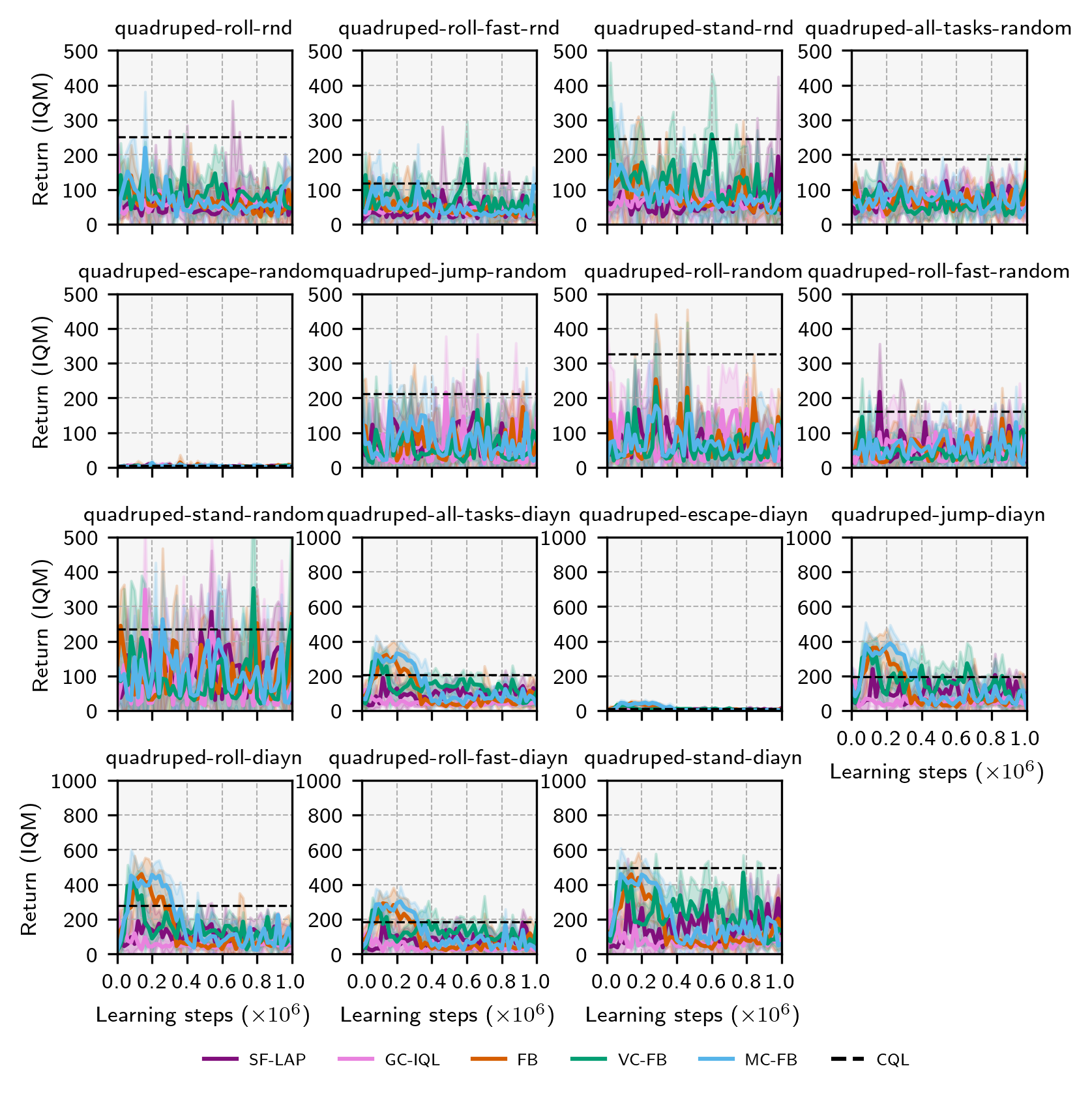}
    \caption{\textbf{Learning Curves (3/3)}. Models are evaluated every 20,000 timesteps where we perform 10 rollouts and record the IQM. Curves are the IQM of this value across 5 seeds; shaded areas are one standard deviation.}
    \label{fig: chapter3/learning curves 3}
\end{figure}

\begin{figure}[h]
    \centering
    \includegraphics[width=0.73\textwidth]{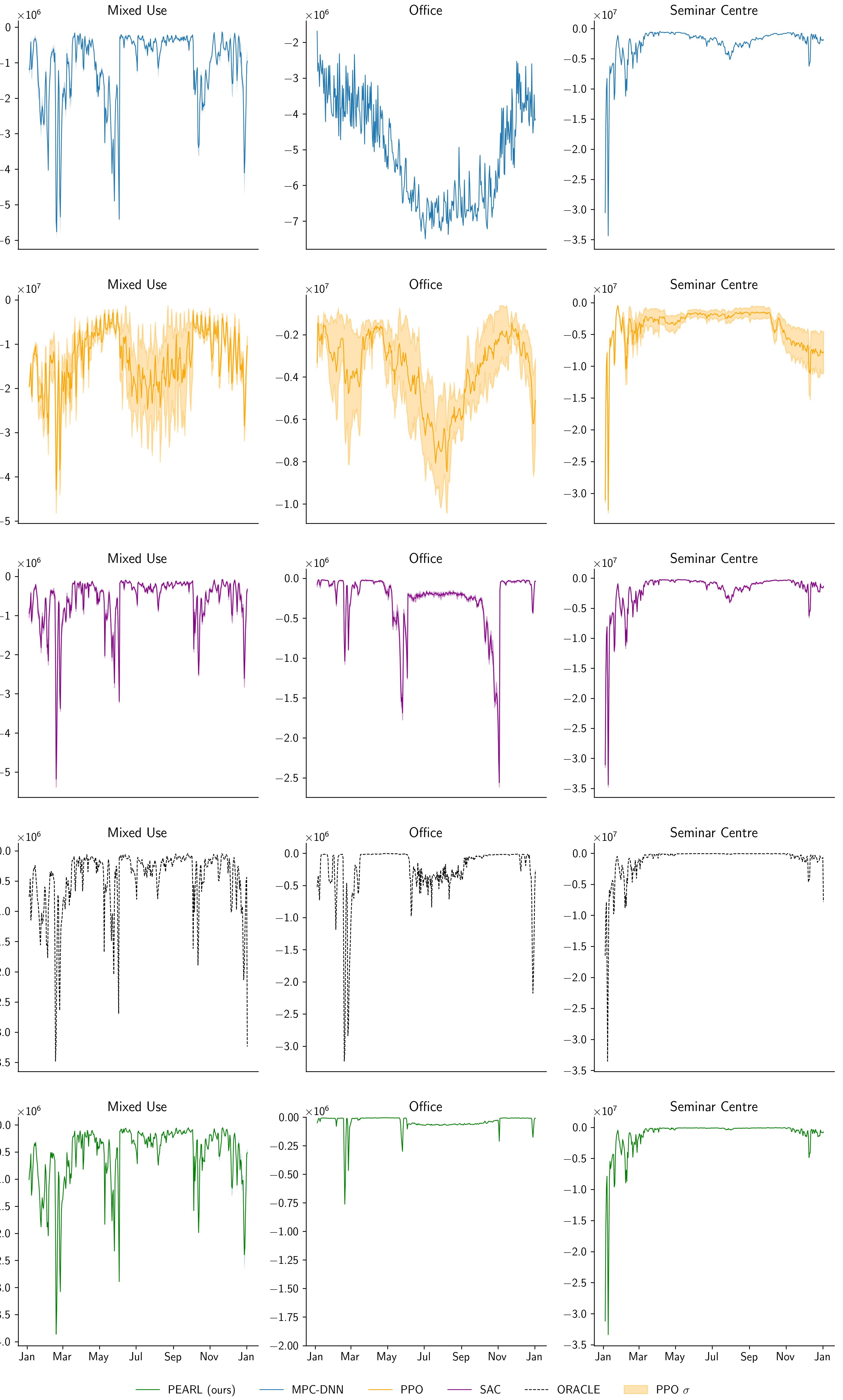}
    \caption{\textbf{Rewards.} Daily mean reward for all RL agents across each \textit{Energym} environment. Note the differing y-axes.}
    \label{fig: reward curves}
\end{figure}

\textbf{Learning curves.} $\;$ We report the learning curves for all the core experiments in Chapter \ref{chapter: from low quality data} in Figures \ref{fig: chapter3/learning curves 1}, \ref{fig: chapter3/learning curves 2}, and \ref{fig: chapter3/learning curves 3}. For all domains except Jaco, the y-axis limit is fixed at 1000 as that is the maximum score achievable in the DeepMind Control Suite. For Jaco-related figures, the y-axis limits is fixed at 100 as no method achieves a score higher than this. In Figure \ref{fig: reward curves} we report the rewards received by all methods in our core experiments from Chapter \ref{chapter: no prior data}.

\begin{figure}
    \centering
    \includegraphics[scale=0.69]{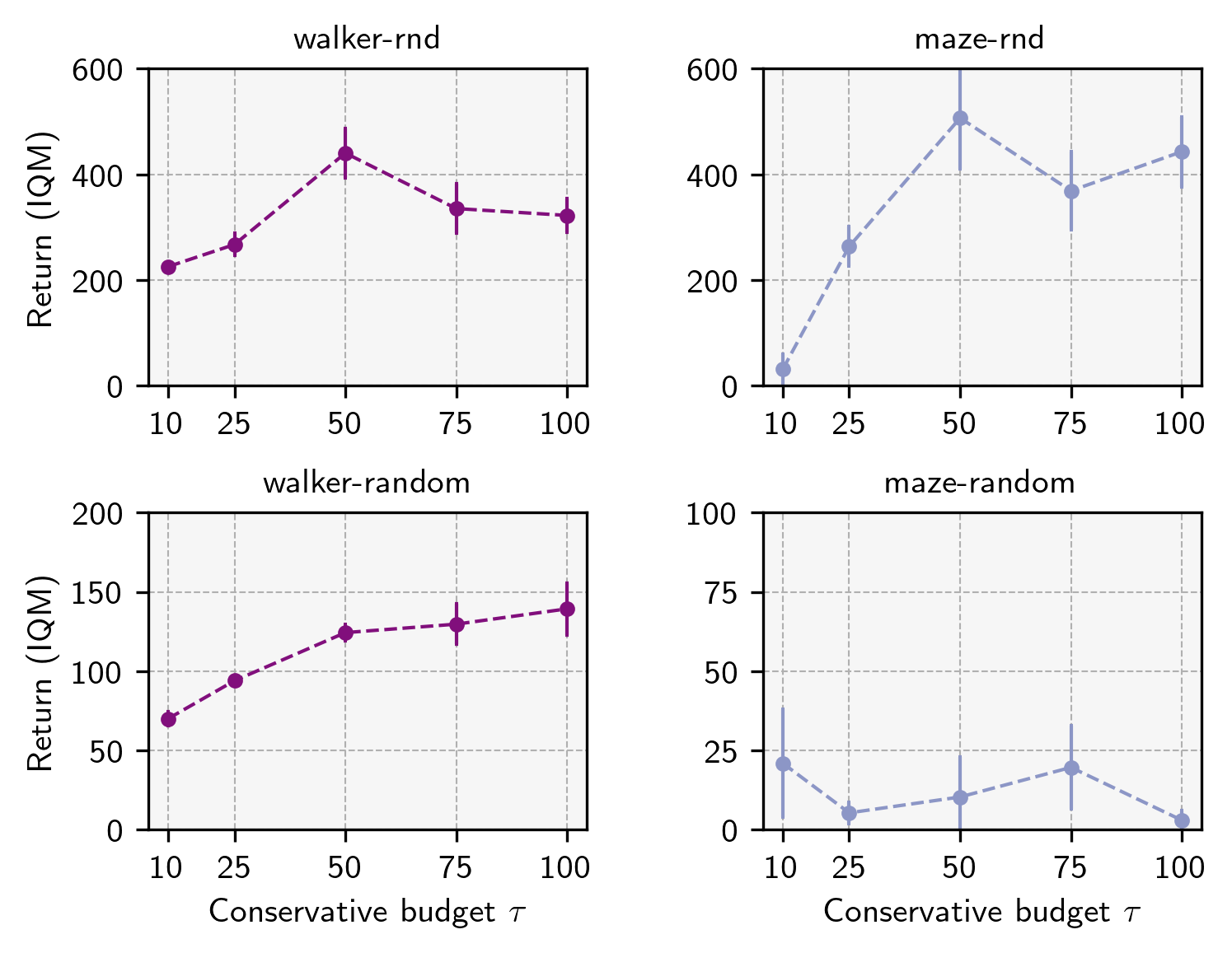}
    \caption{\textbf{VC-FB sensitivity to conservative budget $\tau$ on \textcolor{bupu@purple}{Walker} and \textcolor{bupu@blue}{Point-mass Maze}}. Top: \textsc{RND} dataset;  bottom: \textsc{Random} dataset. Maximum IQM return across the training run averaged over 3 random seeds}
    \label{fig: chapter3/vc-fb lagrange hyperparams}
\end{figure}

\begin{figure}
    \centering
    \includegraphics[scale=0.69]{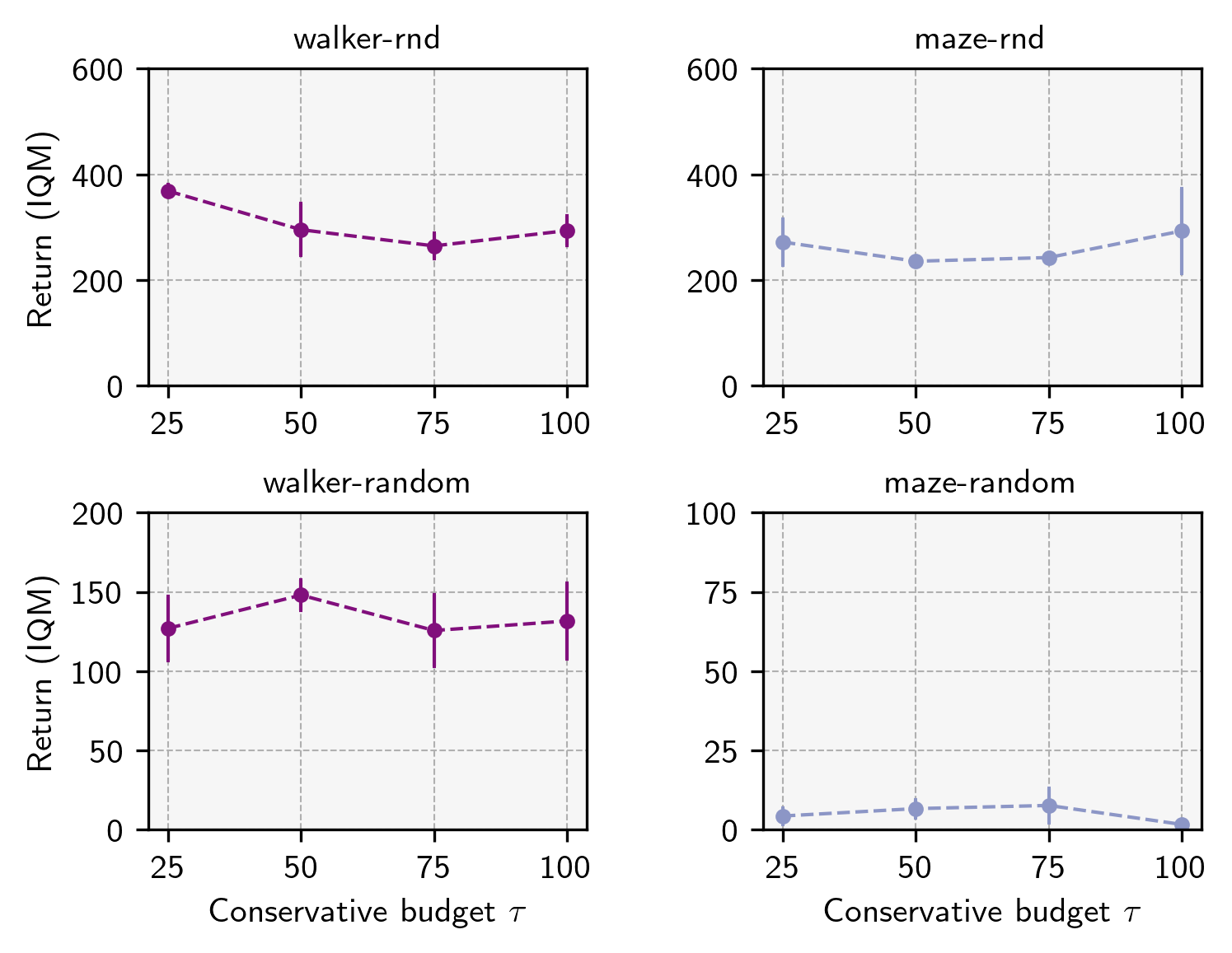}
    \caption{\textbf{MC-FB sensitivity to conservative budget $\tau$ on \textcolor{bupu@purple}{Walker} and \textcolor{bupu@blue}{Point-mass Maze}}. Top: \textsc{RND} dataset;  bottom: \textsc{Random} dataset. Maximum IQM return across the training run averaged over 3 random seeds}
    \label{fig: chapter3/mc-fb lagrange hyperparams}
\end{figure}

\begin{figure}
    \centering
    \includegraphics[scale=0.73]{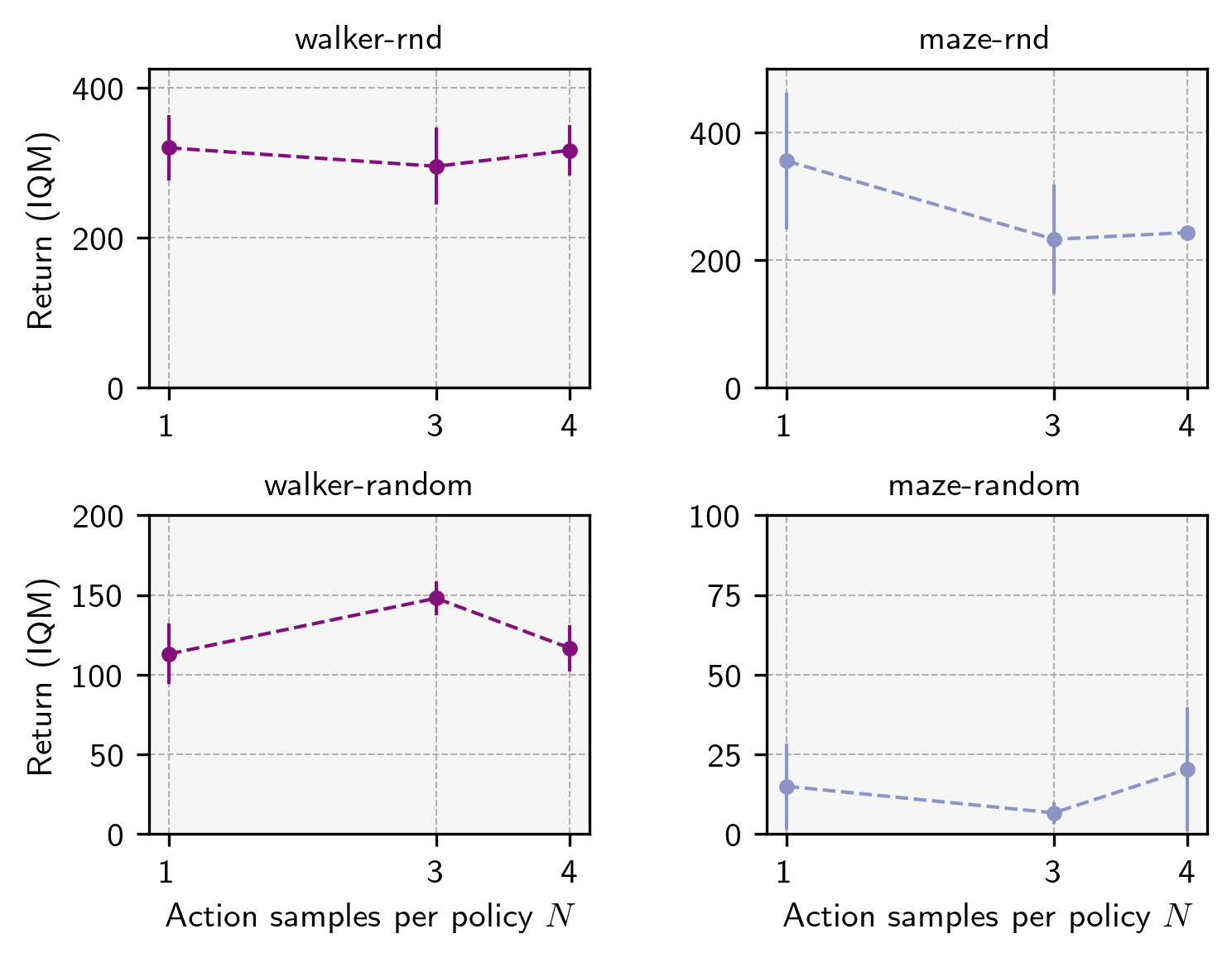}
    \caption{\textbf{MC-FB sensitivity to action samples per policy $N$ on \textcolor{bupu@purple}{Walker} and \textcolor{bupu@blue}{Point-mass Maze}}. Top: \textsc{RND} dataset;  bottom: \textsc{Random} dataset. Maximum IQM return across the training run averaged over 3 random seeds.}
    \label{fig: chapter3/action samples hyperparams}
\end{figure}

\textbf{Hyperparameter sensitivity.} $\;$ We report the sensitivity of VC-FB and MC-FB to the choice of two new hyperparameters: conservative budget $\tau$ and action samples per policy $N$ on the ExORL benchmark. Figure \ref{fig: chapter3/vc-fb lagrange hyperparams} plots the sensitivity of VC-FB to the choice of $\tau$ on Walker and Point-mass Maze domains across \textsc{Rnd} and \textsc{Random} datasets. Figure \ref{fig: chapter3/mc-fb lagrange hyperparams} plots the sensitivity of MC-FB to the choice of $\tau$ on Walker and Point-mass Maze domains across \textsc{Rnd} and \textsc{Random} datasets. Figure \ref{fig: chapter3/action samples hyperparams} plots the sensitivity of MC-FB to the choice of $N$ on Walker and Point-mass Maze domains across \textsc{Rnd} and \textsc{Random} datasets. 

\begin{figure}
    \centering
    \includegraphics[scale=0.73]{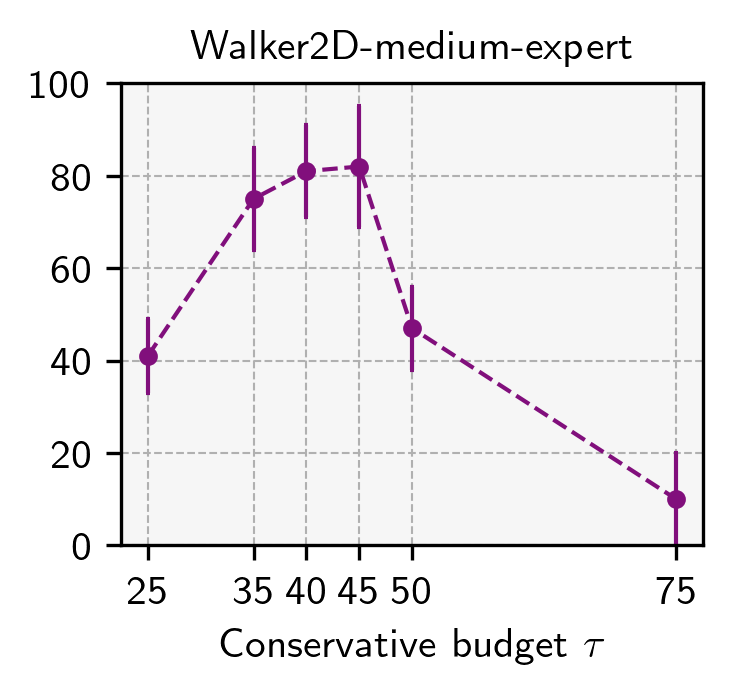}
    \caption{\textbf{VC-FB sensitivity to choice of conservative budget $\tau$ on \textcolor{bupu@blue}{Walker2D}} from the D4RL benchmark.}
    \label{fig: chapter3/d4rl hyperparams}
\end{figure}

We further explore the sensitivity of VC-FB performance on Walker2D from the D4RL benchmark w.r.t. the choice of conservative budget $\tau$. Figure \ref{fig: chapter3/d4rl hyperparams} plots this relationship when trained on the ``medium-expert" dataset from D4RL.

\clearpage

\clearpage

Mean daily reward for 3 runs of each agent in each environment are illustrated in Figure \ref{fig: reward curves}. 

\clearpage

\chapter{Code Snippets}\label{appendix: code snippets}
\section{VC-FB Update Step}
\begin{lstlisting}[language=Python]
    def update_fb(
        self,
        observations: torch.Tensor,
        actions: torch.Tensor,
        next_observations: torch.Tensor,
        discounts: torch.Tensor,
        zs: torch.Tensor,
        step: int,
    ) -> Dict[str, float]:
        """
        Calculates the loss for the forward-backward representation network.
        Loss contains two components:
            1. Forward-backward representation (core) loss: a Bellman update
               on the successor measure (equation 24, Appendix B)
            2. Conservative loss: penalises out-of-distribution actions
        Args:
            observations: observation tensor of shape [batch_size, observation_length]
            actions: action tensor of shape [batch_size, action_length]
            next_observations: next observation tensor of
                                shape [batch_size, observation_length]
            discounts: discount tensor of shape [batch_size, 1]
            zs: policy tensor of shape [batch_size, z_dimension]
            step: current training step
        Returns:
            metrics: dictionary of metrics for logging
        """

        # update step common to all FB models
        (
            core_loss,
            core_metrics,
            F1,
            F2,
            B_next,
            M1_next,
            M2_next,
            _,
            _,
            actor_std_dev,
        ) = self._update_fb_inner(
            observations=observations,
            actions=actions,
            next_observations=next_observations,
            discounts=discounts,
            zs=zs,
            step=step,
        )

        # calculate MC or VC penalty
        if self.mcfb:
            (
                conservative_penalty,
                conservative_metrics,
            ) = self._measure_conservative_penalty(
                observations=observations,
                next_observations=next_observations,
                zs=zs,
                actor_std_dev=actor_std_dev,
                F1=F1,
                F2=F2,
                B_next=B_next,
                M1_next=M1_next,
                M2_next=M2_next,
            )
        # VCFB
        else:
            (
                conservative_penalty,
                conservative_metrics,
            ) = self._value_conservative_penalty(
                observations=observations,
                next_observations=next_observations,
                zs=zs,
                actor_std_dev=actor_std_dev,
                F1=F1,
                F2=F2,
            )

        # tune alpha from conservative penalty
        alpha, alpha_metrics = self._tune_alpha(
            conservative_penalty=conservative_penalty
        )
        conservative_loss = alpha * conservative_penalty

        total_loss = core_loss + conservative_loss

        # step optimiser
        self.FB_optimiser.zero_grad(set_to_none=True)
        total_loss.backward()
        for param in self.FB.parameters():
            if param.grad is not None:
                param.grad.data.clamp_(-1, 1)
        self.FB_optimiser.step()

        return metrics
\end{lstlisting}

\section{Value-Conservative Penalty}
\begin{lstlisting}[language=Python]
def _value_conservative_penalty(
        self,
        observations: torch.Tensor,
        next_observations: torch.Tensor,
        zs: torch.Tensor,
        actor_std_dev: torch.Tensor,
        F1: torch.Tensor,
        F2: torch.Tensor,
    ) -> torch.Tensor:
        """
        Calculates the value conservative penalty for FB.
        Args:
            observations: observation tensor of shape [batch_size, observation_length]
            next_observations: next observation tensor of shape
                                                     [batch_size, observation_length]
            zs: task tensor of shape [batch_size, z_dimension]
            actor_std_dev: standard deviation of the actor
            F1: forward embedding no. 1
            F2: forward embedding no. 2
        Returns:
            conservative_penalty: the value conservative penalty
        """

        with torch.no_grad():
            # repeat observations, next_observations, zs, and Bs
            # we fold the action sample dimension into the batch dimension
            # to allow the tensors to be passed through F and B; we then
            # reshape the output back to maintain the action sample dimension
            repeated_observations_ood = observations.repeat(
                self.ood_action_samples, 1, 1
            ).reshape(self.ood_action_samples * self.batch_size, -1)
            repeated_zs_ood = zs.repeat(self.ood_action_samples, 1, 1).reshape(
                self.ood_action_samples * self.batch_size, -1
            )
            ood_actions = torch.empty(
                size=(self.ood_action_samples * self.batch_size, self.action_length),
                device=self._device,
            ).uniform_(-1, 1)

            repeated_observations_actor = observations.repeat(
                self.actor_action_samples, 1, 1
            ).reshape(self.actor_action_samples * self.batch_size, -1)
            repeated_next_observations_actor = next_observations.repeat(
                self.actor_action_samples, 1, 1
            ).reshape(self.actor_action_samples * self.batch_size, -1)
            repeated_zs_actor = zs.repeat(self.actor_action_samples, 1, 1).reshape(
                self.actor_action_samples * self.batch_size, -1
            )
            actor_current_actions, _ = self.actor(
                repeated_observations_actor,
                repeated_zs_actor,
                std=actor_std_dev,
                sample=True,
            )  # [actor_action_samples * batch_size, action_length]

            actor_next_actions, _ = self.actor(
                repeated_next_observations_actor,
                z=repeated_zs_actor,
                std=actor_std_dev,
                sample=True,
            )  # [actor_action_samples * batch_size, action_length]

        # get Fs
        ood_F1, ood_F2 = self.FB.forward_representation(
            repeated_observations_ood, ood_actions, repeated_zs_ood
        )  # [ood_action_samples * batch_size, latent_dim]

        actor_current_F1, actor_current_F2 = self.FB.forward_representation(
            repeated_observations_actor, actor_current_actions, repeated_zs_actor
        )  # [actor_action_samples * batch_size, latent_dim]
        actor_next_F1, actor_next_F2 = self.FB.forward_representation(
            repeated_next_observations_actor, actor_next_actions, repeated_zs_actor
        )  # [actor_action_samples * batch_size, latent_dim]
        repeated_F1, repeated_F2 = F1.repeat(
            self.actor_action_samples, 1, 1
        ).reshape(self.actor_action_samples * self.batch_size, -1), F2.repeat(
            self.actor_action_samples, 1, 1
        ).reshape(
            self.actor_action_samples * self.batch_size, -1
        )
        cat_F1 = torch.cat(
            [
                ood_F1,
                actor_current_F1,
                actor_next_F1,
                repeated_F1,
            ],
            dim=0,
        )
        cat_F2 = torch.cat(
            [
                ood_F2,
                actor_current_F2,
                actor_next_F2,
                repeated_F2,
            ],
            dim=0,
        )

        repeated_zs = zs.repeat(self.total_action_samples, 1, 1).reshape(
            self.total_action_samples * self.batch_size, -1
        )

        # convert to Qs
        cql_cat_Q1 = torch.einsum("sd, sd -> s", cat_F1, repeated_zs).reshape(
            self.total_action_samples, self.batch_size, -1
        )
        cql_cat_Q2 = torch.einsum("sd, sd -> s", cat_F2, repeated_zs).reshape(
            self.total_action_samples, self.batch_size, -1
        )

        cql_logsumexp = (
            torch.logsumexp(cql_cat_Q1, dim=0).mean()
            + torch.logsumexp(cql_cat_Q2, dim=0).mean()
        )

        # get existing Qs
        Q1, Q2 = [torch.einsum("sd, sd -> s", F, zs) for F in [F1, F2]]

        conservative_penalty = cql_logsumexp - (Q1 + Q2).mean()

        return conservative_penalty
\end{lstlisting}

\section{Measure-Conservative Penalty}
\begin{lstlisting}[language=Python]
def _measure_conservative_penalty(
        self,
        observations: torch.Tensor,
        next_observations: torch.Tensor,
        zs: torch.Tensor,
        actor_std_dev: torch.Tensor,
        F1: torch.Tensor,
        F2: torch.Tensor,
        B_next: torch.Tensor,
        M1_next: torch.Tensor,
        M2_next: torch.Tensor,
    ) -> torch.Tensor:
        """
        Calculates the measure conservative penalty.
        Args:
            observations: observation tensor of shape [batch_size, observation_length]
            next_observations: next observation tensor of shape
                                                        [batch_size, observation_length]
            zs: task tensor of shape [batch_size, z_dimension]
            actor_std_dev: standard deviation of the actor
            F1: forward embedding no. 1
            F2: forward embedding no. 2
            B_next: backward embedding
            M1_next: successor measure no. 1
            M2_next: successor measure no. 2
        Returns:
            conservative_penalty: the measure conservative penalty
        """

        with torch.no_grad():
            # repeat observations, next_observations, zs, and Bs
            # we fold the action sample dimension into the batch dimension
            # to allow the tensors to be passed through F and B; we then
            # reshape the output back to maintain the action sample dimension
            repeated_observations_ood = observations.repeat(
                self.ood_action_samples, 1, 1
            ).reshape(self.ood_action_samples * self.batch_size, -1)
            repeated_zs_ood = zs.repeat(self.ood_action_samples, 1, 1).reshape(
                self.ood_action_samples * self.batch_size, -1
            )
            ood_actions = torch.empty(
                size=(self.ood_action_samples * self.batch_size, self.action_length),
                device=self._device,
            ).uniform_(-1, 1)

            repeated_observations_actor = observations.repeat(
                self.actor_action_samples, 1, 1
            ).reshape(self.actor_action_samples * self.batch_size, -1)
            repeated_next_observations_actor = next_observations.repeat(
                self.actor_action_samples, 1, 1
            ).reshape(self.actor_action_samples * self.batch_size, -1)
            repeated_zs_actor = zs.repeat(self.actor_action_samples, 1, 1).reshape(
                self.actor_action_samples * self.batch_size, -1
            )
            actor_current_actions, _ = self.actor(
                repeated_observations_actor,
                repeated_zs_actor,
                std=actor_std_dev,
                sample=True,
            )  # [actor_action_samples * batch_size, action_length]

            actor_next_actions, _ = self.actor(
                repeated_next_observations_actor,
                z=repeated_zs_actor,
                std=actor_std_dev,
                sample=True,
            )  # [actor_action_samples * batch_size, action_length]

        # get Fs
        ood_F1, ood_F2 = self.FB.forward_representation(
            repeated_observations_ood, ood_actions, repeated_zs_ood
        )  # [ood_action_samples * batch_size, latent_dim]

        actor_current_F1, actor_current_F2 = self.FB.forward_representation(
            repeated_observations_actor, actor_current_actions, repeated_zs_actor
        )  # [actor_action_samples * batch_size, latent_dim]
        actor_next_F1, actor_next_F2 = self.FB.forward_representation(
            repeated_next_observations_actor, actor_next_actions, repeated_zs_actor
        )  # [actor_action_samples * batch_size, latent_dim]
        repeated_F1, repeated_F2 = F1.repeat(
            self.actor_action_samples, 1, 1
        ).reshape(self.actor_action_samples * self.batch_size, -1), F2.repeat(
            self.actor_action_samples, 1, 1
        ).reshape(
            self.actor_action_samples * self.batch_size, -1
        )
        cat_F1 = torch.cat(
            [
                ood_F1,
                actor_current_F1,
                actor_next_F1,
                repeated_F1,
            ],
            dim=0,
        )
        cat_F2 = torch.cat(
            [
                ood_F2,
                actor_current_F2,
                actor_next_F2,
                repeated_F2,
            ],
            dim=0,
        )

        cml_cat_M1 = torch.einsum("sd, td -> st", cat_F1, B_next).reshape(
            self.total_action_samples, self.batch_size, -1
        )
        cml_cat_M2 = torch.einsum("sd, td -> st", cat_F2, B_next).reshape(
            self.total_action_samples, self.batch_size, -1
        )

        cml_logsumexp = (
            torch.logsumexp(cml_cat_M1, dim=0).mean()
            + torch.logsumexp(cml_cat_M2, dim=0).mean()
        )

        conservative_penalty = cml_logsumexp - (M1_next + M2_next).mean()

        return conservative_penalty
\end{lstlisting}

\newpage
\section{\texorpdfstring{$\alpha$}{alpha} Tuning}
\begin{lstlisting}[language=Python]
def _tune_alpha(
        self,
        conservative_penalty: torch.Tensor,
    ) -> torch.Tensor:
        """
        Tunes the conservative penalty weight (alpha) w.r.t. target penalty.
        Discussed in Appendix B.1.4
        Args:
            conservative_penalty: the current conservative penalty
        Returns:
            alpha: the updated alpha
        """

        # alpha auto-tuning
        alpha = torch.clamp(self.critic_log_alpha.exp(), min=0.0, max=1e6)
        alpha_loss = (
            -0.5 * alpha * (conservative_penalty - self.target_conservative_penalty)
        )

        self.critic_alpha_optimiser.zero_grad()
        alpha_loss.backward(retain_graph=True)
        self.critic_alpha_optimiser.step()
        alpha = torch.clamp(self.critic_log_alpha.exp(), min=0.0, max=1e6).detach()

        return alpha
\end{lstlisting}

\end{document}